%% file: main.tex
\documentclass[10pt,twocolumn,letterpaper]{article}

\usepackage[pagenumbers]{cvpr} 

\input{preamble}

\definecolor{cvprblue}{rgb}{0.21,0.49,0.74}
\usepackage[pagebackref,breaklinks,colorlinks,allcolors=cvprblue]{hyperref}

\def\paperID{30618} 
\def\confName{CVPR}
\def\confYear{2026}

\title{Action Motifs: Self-Supervised Hierarchical Representation of Human Body Movements}

\author{Genki Kinoshita$^{1}$
\hspace{0.15cm}
Shu Nakamura$^{1}$
\hspace{0.15cm}
Ryo Kawahara$^{1}$
\hspace{0.15cm}
Shohei Nobuhara$^{2}$
\\
\vspace{1mm}
Yasutomo Kawanishi$^{3}$
\hspace{0.15cm}
Ko Nishino$^{1}$
\\
\vspace{1mm}
$^1$Kyoto University
\hspace{0.5em} $^2$Kyoto Institute of Technology
\hspace{0.5em} $^3$RIKEN \\
{\tt\small \url{https://vision.ist.i.kyoto-u.ac.jp/research/action-motifs/}}
}

\begin{document}
\twocolumn[{
\maketitle
\begin{center}
    \vspace{-5mm}
    \captionsetup{type=figure}
    \includegraphics[width=\linewidth]{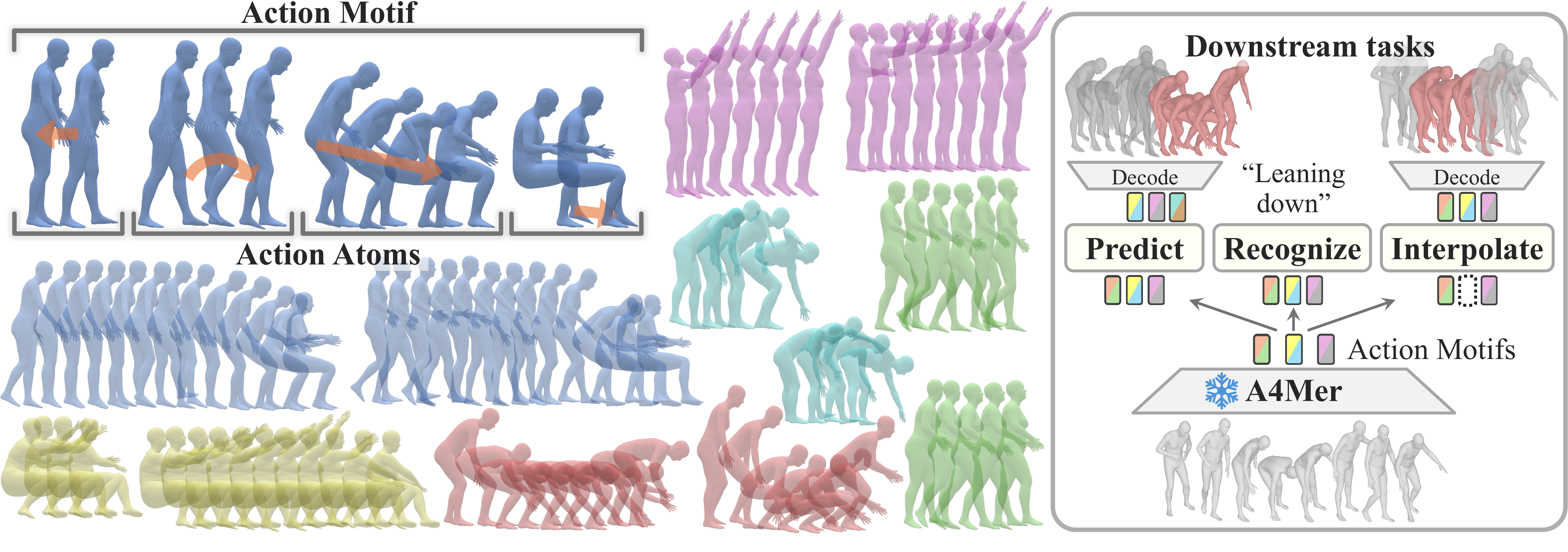}
    \vspace{-1.8\baselineskip}
    \captionof{figure}{
        We introduce A4Mer, a novel unsupervised method for learning a hierarchical representation of human body movements consisting of Action Atoms and Action Motifs that capture the atomic body movements and their temporal compositions together with their contextual variations, respectively.
        The learned Action Motif representations (depicted in the same color) encode common movements regardless of durations and individual differences.
        These Action Motif sequences enable accurate motion prediction and interpolation, and also action recognition.}
    \label{fig:Teaser}
\end{center}
}]

\begin{abstract}
Effective human behavior modeling requires a representation of the human body movement that capitalizes on its compositionality.
We propose a hierarchical representation consisting of Action Atoms that capture the atomic joint movements and Action Motifs that are formed by their temporal compositions and encode similar body movements found across different overall human actions. We derive A4Mer, a nested latent Transformer to learn this hierarchical representation from human pose data in a fully self-supervised manner. A4Mer splits a 3D pose sequence into variable-length segments and represents each segment as a single latent token (Action Atoms). Through bottom-up representation learning, temporal patterns composed of these Action Atoms, which capture meaningful temporal spans of reusable, semantic segments of body movements, naturally emerge (Action Motifs). A4Mer achieves this with a unified pretext task of masked token prediction in their respective latent spaces. We also introduce Action Motif Dataset (AMD), a large-scale dataset of multi-view human behavior videos with full SMPL annotations. We introduce a novel use of cameras by mounting them on the feet to achieve their frame-wise annotations despite frequent and heavy body occlusions. Experimental results demonstrate the effectiveness of A4Mer for extracting meaningful Action Motifs, which significantly benefit human behavior modeling tasks including action recognition, motion prediction, and motion interpolation.
\end{abstract}


\section{Introduction}
\label{sec:intro}

Humans can instantly interpret and even anticipate others' actions.
Reproducing this cognitive capability on a computer would require a representation of the body movements that capture their global coordination and local instantiations in space-time both accurately and efficiently.
The motion of raising an arm can be part of body movements for opening a door or reaching up, which are part of actions like grabbing food from a fridge or taking a book from a shelf. Each movement and action would also vary depending on the person, their condition (\eg, mood), and surrounding context.

We argue that a ``good" representation of human body movement should encode this compositionality. It should (1) express basic body movements with the same atomic motions of the body; (2) build up distinct actions as sequences of unique patterns of temporal compositions of these atomic units; and (3) abstract away contextual effects causing variations and permutations in the instantiations of the atomic movements and their temporal compositions so that they capture the semantics of the human behavior.
If such atomic motions were words in language, their recurring temporal patterns that collectively express distinct actions would be phrases---reusable components that appear across different sentences (\ie, actions) and collectively compose the prose.

Existing works learn fixed-length representations at the frame-wise (like alphabets), clip-wise (n-grams), or video-wise units (entire texts), which are too fine and redundant, misaligned with semantic boundaries, or gloss over reusable patterns of human movements, respectively.
Supervision with manual annotation for extracting such hierarchical representations is infeasible, as it involves segmenting and labeling at two interdependent levels. It fundamentally needs to be discovered as emergent patterns from human movement data without explicit supervision.

In this paper, we learn these phrase-level movements as \textit{Action Motifs} and their finer building blocks as \textit{Action Atoms}. Each Action Motif represents ``a semantically meaningful and reusable temporal composition of Action Atoms.''
Two challenging entanglements need to be resolved to extract these Action Motifs from human behavior observations. The first is segmentation and semantics.
Even if a representation space can be constructed, a different temporal segmentation alters action semantics and consequently distances in that space. Without knowing the proper segmentation, a model cannot understand the semantics and similarities between motions. Yet, without understanding semantics, it cannot determine the temporal boundaries.
The second is representation and composition.
Even if the boundaries between actions were known, extracting movement representations still requires jointly learning the representations themselves and their temporal compositions.
The compositions define what kinds of representations are needed, while the representations constrain how the compositions can be formed.

We overcome these entanglements with \textit{A4Mer}, a novel Trans\underline{former} for learning \underline{A}ction \underline{A}toms \underline{A}nd \underline{A}ction \underline{M}otifs from 3D pose sequences.
The first key idea is to let meaningful movement patterns emerge through a bottom-up hierarchical composition of representations starting from small motion units.
A4Mer begins by segmenting motion sequences into Action Atoms using simple kinematic cues and then learning their representations.
By detecting recurring patterns of Action Atoms across sequences, A4Mer learns higher-level representations.
This hierarchical process alternately enables motion semantics to be learned and motion segments to emerge, resolving their chicken-and-egg dependency.
To represent the resulting variable-length segments in a unified manner, A4Mer consolidates each segment into a single latent token through cross-attention.

The second key idea is to jointly learn the representations themselves and their temporal compositions by solving a pretext task. Given a pose sequence with segment-wise masking, A4Mer predicts the latent tokens representing the masked segments.
This adoption of a joint-embedding predictive architecture (JEPA) \cite{JEPA} naturally encourages the model to learn semantically abstracted segment-wise representations while considering the action semantics conveyed by their temporal compositions.
To promote context-independent representations, we decompose each predicted representation into global and local components, and encourage the model to focus more strongly on learning the latter.

We also introduce Action Motif Dataset (\textbf{AMD}) and use it to train A4Mer. AMD captures diverse human daily activities. AMD consists of a large number of multi-view RGB videos of people naturally performing chores in a real house, along with accurate SMPL \cite{SMPL} annotations.

To realize occlusion-robust annotation in natural indoor scenes, we exploit two simple observations: legs are most frequently occluded, and rooms have ceilings. We attach tiny foot-mounted cameras and place markers on the ceilings and the undersides of tables to enable accurate foot localization without constraining room layout or action diversity. As these markers are not observable from the ceiling-mounted cameras---key viewpoints for applications such as child and elderly monitoring---AMD retains the natural appearance of both subjects and environments.

Through extensive experiments, we show that learned Action Motifs encode common movement patterns in a consistent manner and their temporal compositions capture action semantics, leading to significant accuracy gains on downstream tasks including action recognition, motion prediction, and motion interpolation.
We believe that Action Motifs can serve as a fundamental building block for human behavior modeling, and that AMD will provide a sound platform for learning and evaluating such representations.


\section{Related Work}
\label{sec:RelatedWork}

\begin{table*}[t!]
    \scriptsize
    \centering
    \renewcommand{\arraystretch}{1.1}
    \setlength{\tabcolsep}{1.6pt}
    \caption{Comparison of human 3D pose datasets.
    3rdPV: provides third-person-view videos.
    MultiView: multi-view videos are available.
    Daily: includes indoor daily activities such as chores.
    MultiAct: a single video contains goal-directed motions composed of multiple actions.
    RealHome: recorded in real home where subjects can move freely in a wide area.
    $\dagger$: noisy when occluded.
    }
    \label{tab:Datasets}
    \vspace{-0.75\baselineskip}
    \begin{tabular}{@{}lccccccccccccc@{}}
        \toprule
        \textbf{Attribute} &
        PiGraphs \cite{PiGraphs} & PROX \cite{PROX} & RICH \cite{RICH} & HOI\text{-}M${}^3$ \cite{HOIM3} & EgoBody \cite{EgoBody} & H36M \cite{Human3.6M} &
        EgoExo4D \cite{EgoExo4D} & MotionX \cite{MotionX} &
        ParaHome \cite{ParaHome} & Nymeria \cite{Nymeria} & HiK \cite{HiK} & \cellcolor{red!15}\textbf{AMD} \\
        \midrule
        Real        & \checkmark & \checkmark & \checkmark & \checkmark & \checkmark & \checkmark &
                       \checkmark & \checkmark &
                       \checkmark & \checkmark & \checkmark & \checkmark \\
        3rdPV       & \checkmark & \checkmark & \checkmark & \checkmark & \checkmark & \checkmark &
                       \checkmark & \checkmark &
                       \na        & \na        & \na        & \checkmark \\
        MultiView   & \na        & \na        & \checkmark & \checkmark & \checkmark & \checkmark &
                       \checkmark & \checkmark &
                       \na        & \na        & \na        & \checkmark \\
        Sensor      & \na        & \na        & \na        & \na        & HMD        & MoCap      &
                       Aria       & \na        &
                       MoCap      & MoCap      & \na        & FootCam \\
        Daily       & \na        & \na        & \na        & \na        & \na        & \na        &
                       \checkmark & \na        &
                       \checkmark & \checkmark & \checkmark & \checkmark \\
        MultiAct    & \na        & \na        & \na        & \na        & \checkmark & \na        &
                       \checkmark & \na        &
                       \checkmark & \checkmark & \checkmark & \checkmark \\
        RealHome    & \na        & \na        & \na        & \na        & \na        & \na        &
                       \na        & \na        &
                       \na        & \checkmark & \checkmark & \checkmark \\
        Format      & 3DKpts$^\dagger$ & SMPL-X$^\dagger$ & SMPL-X & SMPL$^\dagger$ & SMPL-X & SMPL &
                       3DKpts$^\dagger$ & SMPL-X &
                       SMPL-X & 3DKpts & SMPL$^\dagger$ & SMPL \\
        \bottomrule
    \end{tabular}
    \vspace{-0.75\baselineskip}
\end{table*}

We review past representative works on human movement modeling and datasets.

\vspace{-8pt}
\paragraph{Self-Supervision.}
Various forms of self-supervision have been explored for learning human movement representations from 3D pose sequences including reconstruction \cite{PUMPS,PCM3,MotionConsistencyAndContinuity}, masked modeling \cite{BehaveMAE,MAMP}, 2D-to-3D lifting \cite{H2oT,MotionBERT}, denoising \cite{MacDiff}, and contrastive learning \cite{USDRL,ActCLR,SkeletonMAE,HiCLR,SCD-Net}.
Unlike non-contrastive self-supervision, we compute the loss in the latent space using JEPA \cite{JEPA}, which enables the model to capture the underlying semantics of body movements rather than their superficial pose differences.
In contrast to contrastive learning, which requires carefully designed semantics-preserving augmentations, JEPA learns through latent token prediction without hand-crafted augmentations.

While few past methods \cite{S-JEPA,GFP} employ JEPA by masking several joints, A4Mer masks latent tokens that consolidate motion segments.
Predicting abstracted representations enables A4Mer to capture segment semantics beyond mere pose similarity.

All these works learn fixed-length representations at the frame, clip, or video levels.
In contrast, A4Mer learns semantically meaningful segments, \ie, Action Motifs, whose temporal spans vary depending on the types of actions, individuals, and contexts.

\vspace{-8pt}
\paragraph{Action Label Supervision.}
Several works \cite{RevealingKeyDetailsToSeeDifferences,SA-DVAE,PURLS,DuoCLR} leverage action labels for contrastive learning.
Although these methods achieve higher accuracy in action recognition than those trained solely on pose sequences, annotating such labels is labor-intensive and does not scale to large datasets.
Besides, they are inherently constrained by the predefined set of action classes, which makes it challenging to generalize to unseen actions.
In contrast, learning directly from many untrimmed sequences containing diverse natural behaviors enables models to acquire generalized representations.
Our method follows this direction and does not require action labels for training.

\vspace{-8pt}
\paragraph{3D Human Pose Datasets.}
Most existing 3D human pose datasets either use MoCap systems \cite{AMASS,KIT,Nymeria,CIRCLE,OMOMO,ParaHome} or RGB(+D) videos \cite{PROX,PiGraphs,RICH,HOIM3,HiK}.
Although MoCap systems can record precise pose data, MoCap suits make appearances unnatural, limiting their applicability to methods that rely on visual information for human behavior modeling.
Video datasets, in contrast, capture the natural appearances throughout their movements.
They are, however, susceptible to occlusions induced by the environment and the body itself, which severely restricts the environments.
Consequently, the kinds of actions captured in such datasets tend to be limited.

Existing datasets typically contain either a single brief action or a sequence of purposeless random motions.
Few datasets \cite{HiK,ParaHome,Nymeria} capture long enough sequences of actions that compose daily activities in real houses.
Unfortunately, they do not provide third-person-view RGB videos.

Our dataset, AMD, provides multi-view RGB videos of natural sequential actions consisting of indoor daily activities in a real house.
We realize occlusion-robust automatic SMPL annotations by leveraging tiny foot-mounted cameras and markers on the ceilings and the undersides of tables.
\cref{tab:Datasets} summarizes the differences among these related datasets.


\begin{figure*}[t!]
    \centering
    \includegraphics[width=0.98\linewidth]{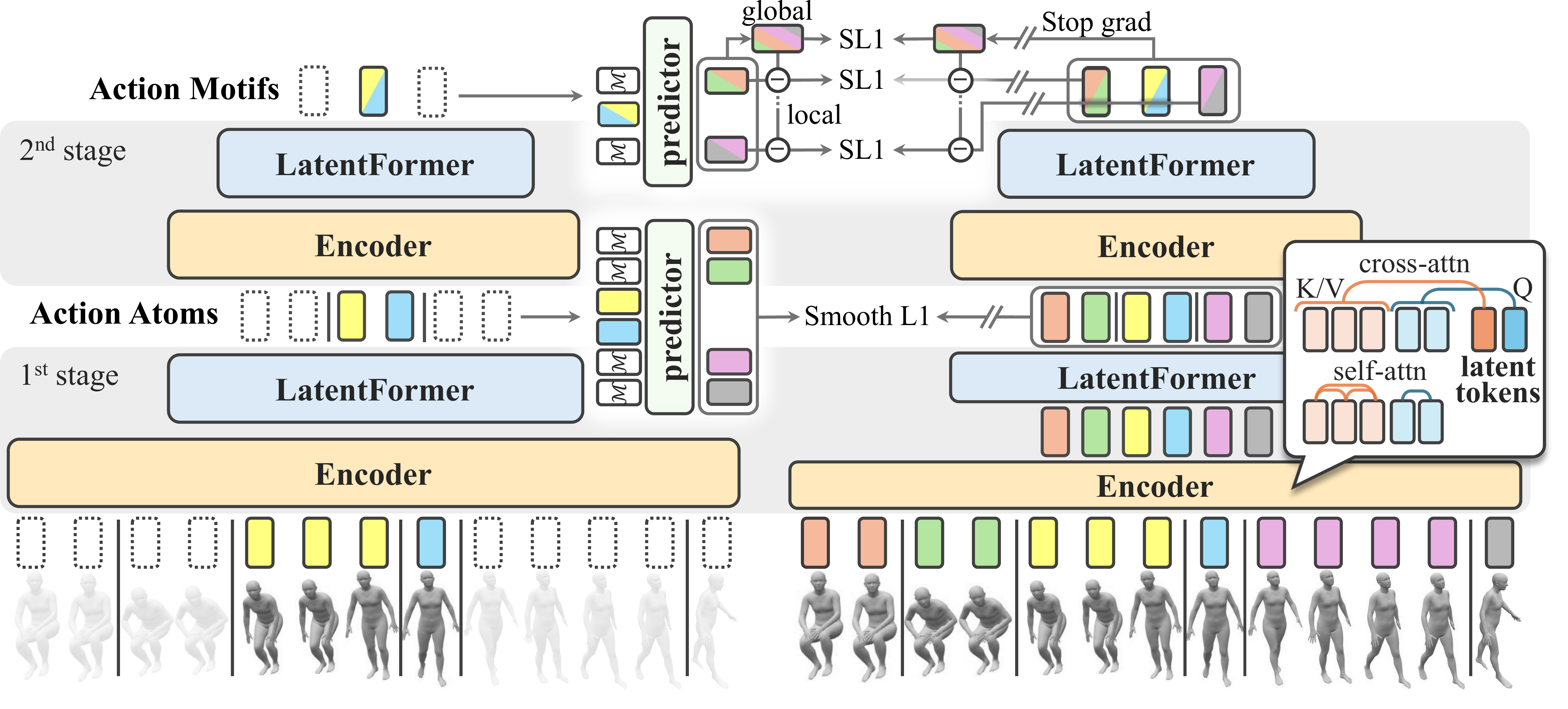}
    \vspace{-\baselineskip}
    \caption{A4Mer extracts a hierarchical representation of human body movements consisting of Action Atoms, which in turn compose Action Motifs, with Encoder and LatentFormer. Encoder consolidates each segment into a latent token via cross-attention, and LatentFormer captures their temporal relationships.
    During training, A4Mer takes 3D poses with several segments masked as input, and outputs latent tokens of the unmasked segments (left).
    For each masked segment, a learnable token $\mathcal{M}$ is inserted at the corresponding latent position, and a predictor predicts the masked latent tokens. The prediction targets are extracted from the complete, unmasked inputs (right).
    To promote learning context-independent representations, the loss on the latent tokens is decomposed into global and local components with adaptive emphasis on the local component.
}
    \label{fig:MethodOverview}
    \vspace{-\baselineskip}
\end{figure*}

\section{Learning Action Motifs}

Our goal is to learn a representation of human movement that fully exploits its compositionality so that it can serve as a basis for accurate and efficient human behavior modeling.

\subsection{Architecture for Action Motif Emergence}
As shown in \cref{fig:MethodOverview}, A4Mer employs a two-stage hierarchical architecture that composes higher-level Action Motifs from lower-level Action Atoms.
This bottom-up design allows higher-level motion segments to naturally emerge as recurrent patterns over the lower-level motions.
We first train the first-stage model, find these patterns that define the second-stage segments, and then train both stages end-to-end.

Both stages share the same architecture, consisting of Encoder and LatentFormer.
Encoder aims to produce a single representation for each motion segment. This is non-trivial since segment durations vary with the kind of motion, individual, and context.
To represent such variable-length segments in a unified manner, Encoder consolidates all input tokens belonging to a segment into a single latent token through cross-attention, inspired by BLT \cite{BLT}.

Let $X = (x_1, \ldots, x_T)$ denote a 3D pose sequence of $T$ frames.
For each segment $k\in\{1, \ldots, K\}$, Encoder initializes a latent token $z_k$ by max-pooling the input tokens within the segment $X_k=\{x_t \mid s(t) = k\}$, and performs cross-attention by using $z_k$ as the query, while $X_k$ serves as keys and values.
Here, $s(t)$ returns the segment index that frame $t$ belongs to. Note that the number of segments $K$ in a sequence varies depending on the input.

\vspace{-8pt}
\paragraph{Segment-Wise Attention.}
Encoder adopts a Transformer-decoder architecture that alternates between self-attention among the input tokens and the aforementioned cross-attention.
Performing self-attention over all input tokens in the sequence may cause the model to respond excessively to minor pose similarities between distant frames, hindering semantic understanding of the body movements.
Instead, we restrict self-attention to tokens within the same segment, allowing Encoder to focus on consolidating segment information.
By explicitly decoupling intra-segment consolidation from inter-segment relation modeling and assigning the latter to LatentFormer, which performs self-attention among the latent tokens, A4Mer effectively learns to encode the temporal relationships between motions at the semantic level.

\subsection{Segmentation}
\label{subsec:Segmentation}
\paragraph{Action Atom Segmentation.}
We define the boundaries of the first-stage segments by nonlinear changes in joint trajectories, which we regard as the initiation of the fine-grained movements, \ie, Action Atoms.
These are detected by evaluating the discrepancy between linearly extrapolated and observed joint trajectories (see \cref{subsec:ActionAtomSegmentation} for details).
Using these segments, the first-stage model is trained to learn the Action Atoms.

\vspace{-8pt}
\paragraph{Action Motif Segmentation.}
To capture the emerging Action Atom patterns shared across different actions (\ie, Action Motifs) and use these patterns as the second-stage segments, we employ the Generalized Sequential Pattern (GSP) algorithm \cite{GSP}.
The entire dataset is first fed into the trained first-stage model to obtain sequences of Action Atoms.
Action Atoms are then clustered using $k$-means and discretized into categorical codes, with the number of clusters set to 512.
Note that this clustering is performed only during training, and during inference each Action Atom is assigned to a cluster via nearest-neighbor matching.
From these categorical code sequences, we find co-occurring patterns by iteratively extending shorter frequent subsequences that occur more frequently than a minimum threshold.

A single Action Atom in a sequence may be covered by multiple detected patterns. To obtain a non-overlapping pattern assignment, we apply dynamic programming to each sequence and select a pattern set that covers the sequence with the minimum number of patterns.

\subsection{Learning by Predicting Latent Tokens}
To jointly learn the Action Atoms, Motifs and their temporal compositions, A4Mer solves the same pretext task in both stages. It receives masked inputs and predicts the latent tokens representing the masked segments.
To realize this joint-embedding predictive architecture (JEPA) \cite{JEPA}, we incorporate a predictor $g_\phi$ into A4Mer, alongside the feature extractor $f_\theta$ composed of Encoder and LatentFormer.
During training, we randomly mask a subset of segments $\mathcal{K} \subset \{1, \ldots, K\}$, the feature extractor outputs latent tokens for the visible segments $X_{\text{vis}} = \{ x_t \mid s(t) \notin \mathcal{K} \}$, and the predictor then estimates those of the masked segments.
In the original JEPA formulation, both are jointly trained with
\begin{align}
\label{eq:OriginalJEPA}
    &\min_{\theta, \phi, \mathcal{M}}\sum\nolimits_{k \in \mathcal{K}}
    \mathrm{SL1}\big( \hat{z}_k - \sg(z_k) \big)\,, \\
    Z = \{ z_k & \}^K_{k=1} = f_{\bar{\theta}}(X)\,, \quad
    \hat{Z} = g_\phi(\mathcal{M}, f_\theta(X_{\text{vis}}))\,,
\end{align}
where $\mathrm{SL1}$ indicates the smooth L1 loss, and $\mathcal{M}$ denotes learnable mask tokens for the missing segments.
To prevent representation collapse, A4Mer detaches the prediction target $Z$ with a stop-gradient operation $\sg(\cdot)$, and updates the parameters of the target feature extractor $\bar{\theta}$ as the exponential moving average of $\theta$.
Solving this pretext task encourages the model to account for temporal context and produce segment-wise representations.

When training the two stages end-to-end, Action Motif segments are randomly selected for masking and Action Atom segments belonging to them are also masked.
This hierarchical masking strategy prevents the second stage from accessing the information of its target representations leaked from the first stage.

\vspace{-8pt}
\paragraph{Decomposition into Local and Global Components.}
The objective in \cref{eq:OriginalJEPA} does not explicitly constrain whether contextual information should be baked in the representation itself or the model should interpret it from the sequences.
To encourage the model to produce context-independent and highly reusable representations, we decompose each predicted latent token into a global component that represents the overall movements in the sequence and a local component.
We then place greater emphasis on minimizing the error of the latter.
Formally, for the predicted and target sequences $\hat{Z}$ and $Z$, we compute their global means ($\hat{z}^\glob, z^\glob$) and local deviation ($\hat{z}^\local_k, z^\local_k$) as
\begin{align}
    \hat{z}^\glob &= \frac{1}{|\mathcal{K}|} \sum\nolimits_{k \in \mathcal{K}} \hat{z}_k\,, &
    z^\glob &= \frac{1}{|\mathcal{K}|} \sum\nolimits_{k \in \mathcal{K}} z_k\,, \\
    \hat{z}^\local_k &= \hat{z}_k - \hat{z}^\glob\,, &
    \quad z^\local_k &= z_k - z^\glob\,,
\end{align}
respectively.
Our objective $\loss$ combines a local loss $\loss^\local_k$ and a global loss $\loss^\glob$ with a dynamic weighting that prioritizes the local one:

\begin{align}
    \loss =
    \frac{1}{|\mathcal{K}|}\sum\nolimits_{k \in \mathcal{K}} \loss^\local_k &+ \frac{\alpha \lambda_k}{|\mathcal{K}|}\, \loss^\glob\,, \label{eq:Objective} \\
    \loss^\local_k = \mathrm{SL1}\big( \hat{z}^\local_k - z^\local_k \big)\,, \quad &
    \loss^\glob = \mathrm{SL1}\big( \hat{z}^\glob - z^\glob \big)\,, \label{eq:ObjectiveDetails}
\end{align}
where $\lambda_k = \mathrm{sg}(\mathcal{L}^\local_k/\mathcal{L}^\glob)$, $\alpha = 0.05$.
This decomposition explicitly guides the model to capture segment-wise motion semantics while suppressing sequence-level biases.

\subsection{Using Action Motifs}
\label{subsec:Application}

The Action Motifs emerging from A4Mer can be used as the core representation for fundamental human behavior modeling tasks.

\textit{Action Recognition} can be realized by zero-shot transfer with a simple weighted $k$-NN classifier \cite{Lemniscate} or transfer learning.
For zero-shot transfer, we assign a ground-truth action label to each latent token extracted from the training data. When an Action Motif segment does not exactly match the ground-truth action segment, we determine its class label by taking the majority vote of the action labels within the segment. Then, we classify each test latent token by weighted voting among the $k$-nearest training tokens.

For transfer learning, we predict the action class of each observed frame from the latent token sequence.
Specifically, we repeat each latent token according to the length of its segment and feed the resulting sequence into a 1-layer Transformer encoder trained to output per-frame action labels.

\textit{Long-term Motion Prediction} forecasts the future poses at each time step.
Since the temporal compositions of Action Motifs represent action semantics, using them enables semantic-level forecasting of long-term future motions, beyond merely producing temporally smooth poses.
This can be achieved with a next latent token prediction head attached to A4Mer (\cref{fig:Teaser}). The autoregressively predicted latent tokens are then decoded into 3D poses with a decoder trained independently of the prediction head.

For decoding, each predicted latent token is repeated according to the number of frames belonging to its segment, and the decoder outputs the pose for each frame.
The decoder takes both the predicted Action Motifs and those used as input to the prediction head, enabling action-semantic-aware decoding.
Since the segments of future frames are unavailable during inference, the prediction head jointly estimates both the next latent token and its segment length.

\textit{Motion Interpolation} recovers unobserved poses from partial observations. With Action Motifs, this can be achieved by inferring the missing latent tokens from the observed ones with an interpolation head and decoding the resulting latent sequence into pose space. In the interpolation head, learnable tokens are inserted at unobserved latent positions, and their outputs are treated as the interpolated latent tokens.

\begin{figure*}[t]
    \centering
    \begin{minipage}[t]{0.34\linewidth}
        \vspace{0pt}
        \centering
        \includegraphics[width=\linewidth]{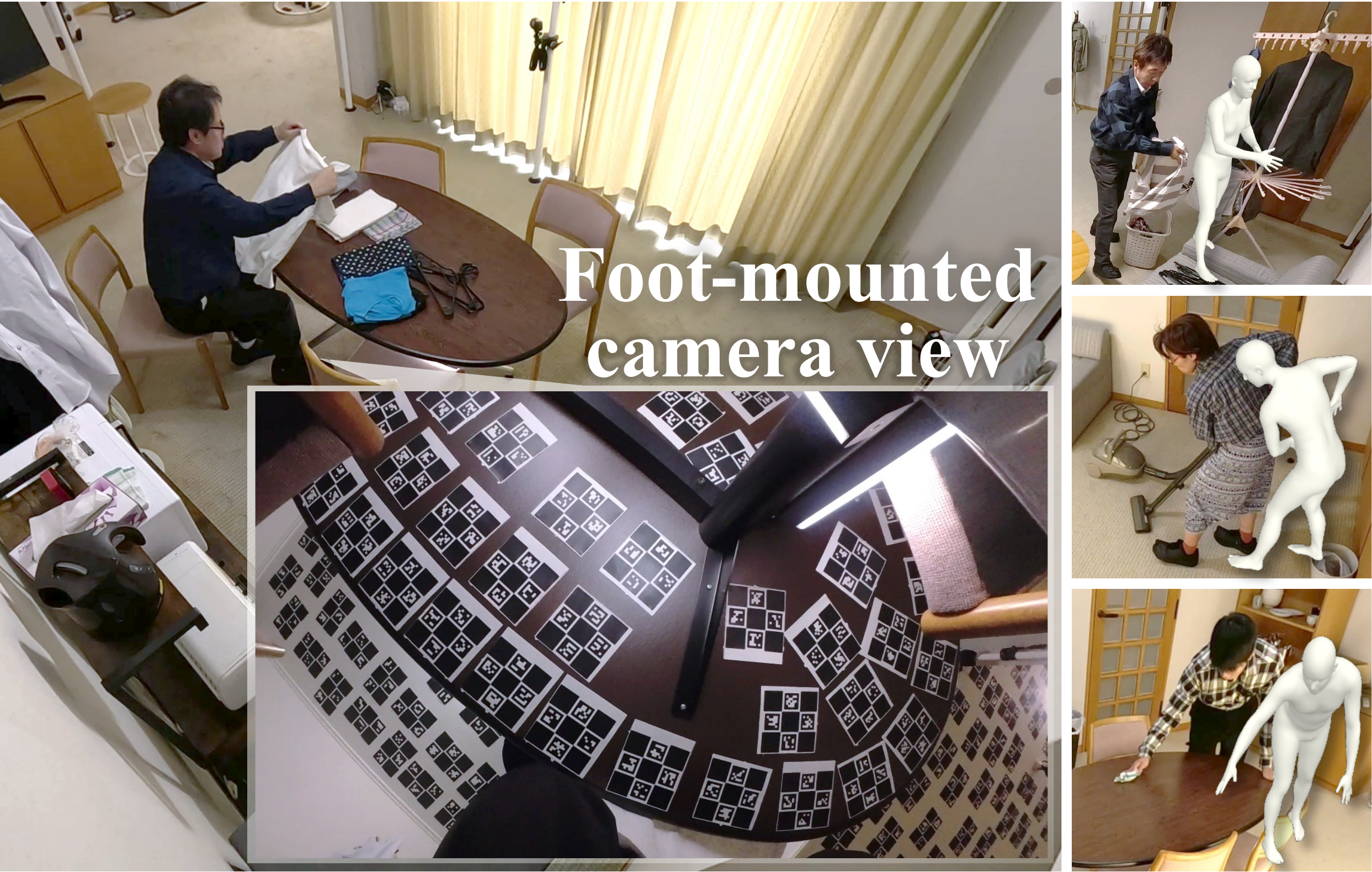}
        \vspace{-15pt}
        \caption{AMD captures diverse daily activities with accurate SMPL annotations despite frequent and heavy occlusions by leveraging foot-mounted cameras and markers.
        }
        \label{fig:FootCameraView}
        \vspace{1em}
        \includegraphics[width=\linewidth]{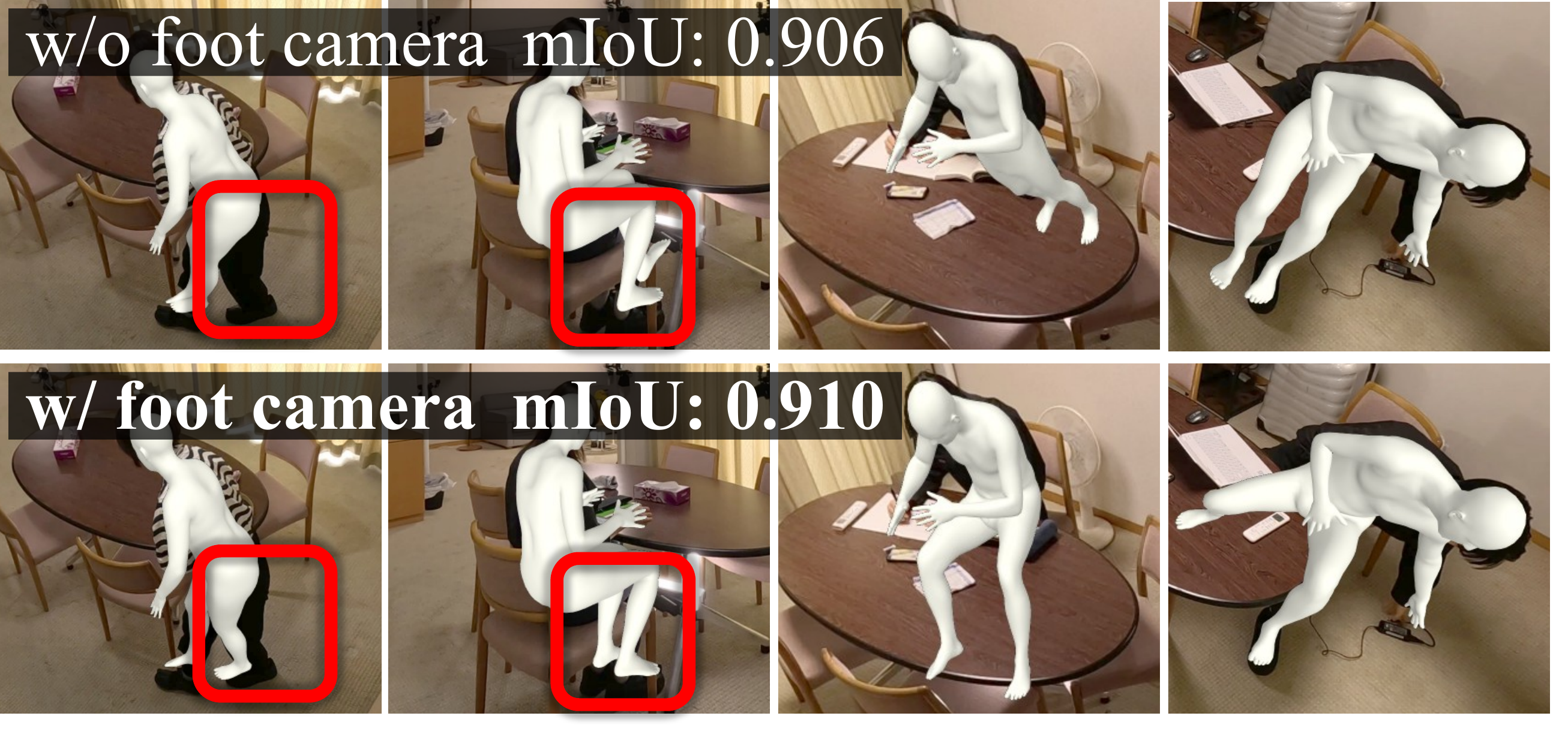}
        \vspace{-20pt}
        \caption{Using the foot camera poses for SMPL fitting leads to accurate pose annotations.}
        \vspace{-0.8\baselineskip}
        \label{fig:FootCameraConstraint}
    \end{minipage}
    \hfill
    \begin{minipage}[t]{0.64\linewidth}
        \vspace{0pt}
        \centering
        \includegraphics[width=\linewidth]{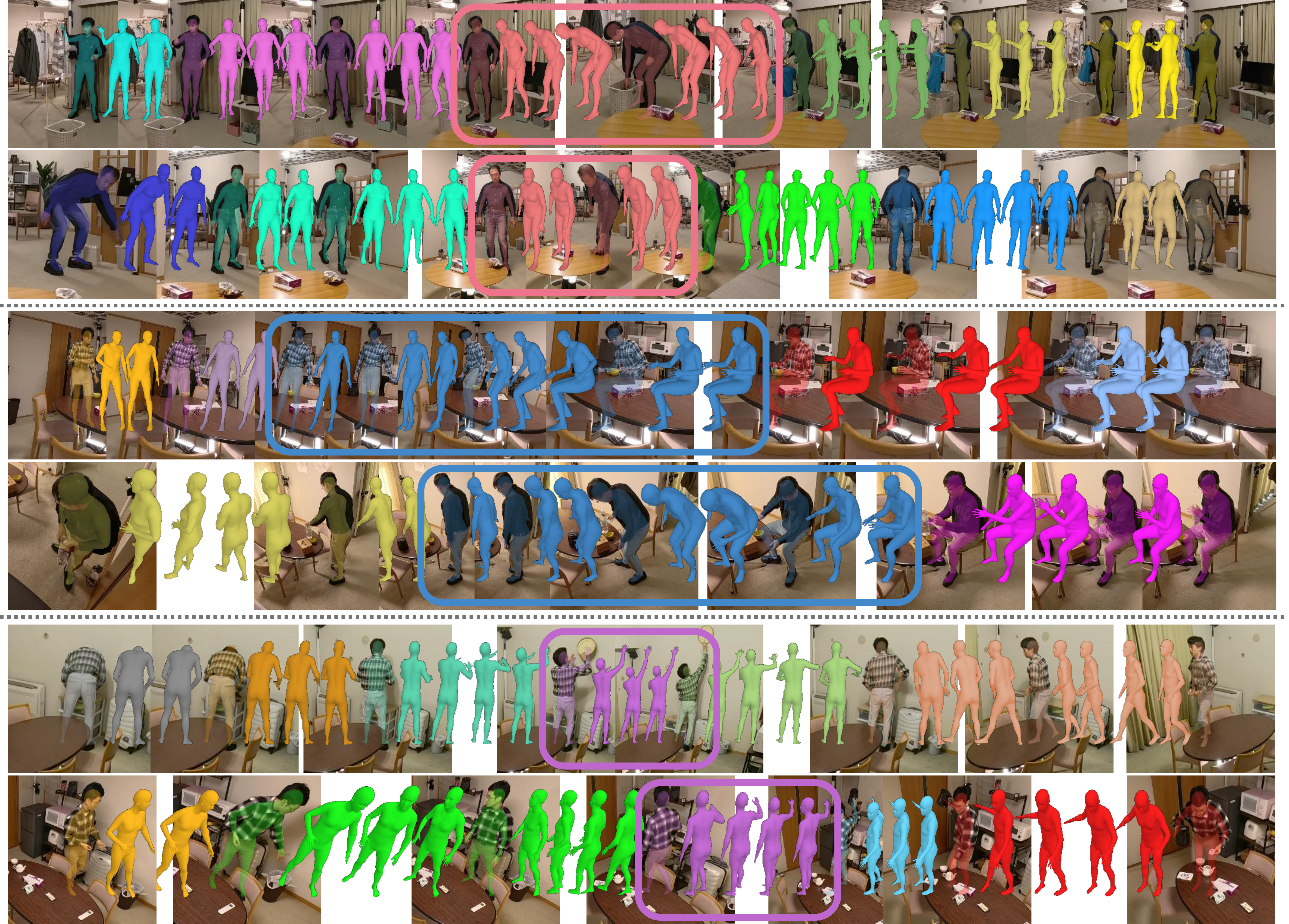}
        \vspace{-1.4\baselineskip}
        \caption{
            Action Motif sequences on AMD. SMPL color denotes cluster IDs assigned with $k$-means clustering on the Action Motif representation space. Common movements across actions have the same colors, which indicates that shared Action Motifs are successfully represented as similar latent tokens.
            }
        \vspace{-0.8\baselineskip}
        \label{fig:ActionMotifs}
    \end{minipage}
\end{figure*}

\section{Action Motif Dataset}
\label{sec:Dataset}
We introduce Action Motif Dataset (AMD), which captures people performing daily chores with multi-view cameras mounted on room corners and provides SMPL annotations by exploiting a novel use of cameras, namely mounting them on the feet (\cref{fig:FootCameraView}).
The abundance and diversity of natural human behavior captured in AMD with 3D pose annotations make it a suitable platform for learning and evaluating human movement representations.

AMD consists of videos of 50 subjects (27 males and 23 females) aged 21--69, recorded in a furnished living-dining room with 24 cameras, resulting in 14.2 hours of footage at 30~fps.
Each subject participated in 3--9 sessions (avg. 3.5), with the furniture arrangement varied across sessions.
Before each session, subjects were instructed on the activities to perform, but no specific action order was imposed to preserve natural movement and personal variation.
Each session lasted 1--17 minutes (avg. 4.8) and includes multiple activities.

\subsection{SMPL Annotation}
\label{subsec:SMPLAnnot}

To obtain the SMPL pose $\pose$, we extend SMPLify-X \cite{SMPLify-X} to enable multi-view optimization while fixing shape parameters obtained with a body scan. We minimize the objective $\smplloss$ with 2D joints estimated in all views:
\begin{equation}
    \label{eq:BasicPolicy}
    \begin{split}
        &\smplloss(\pose) = \jointw \jointl + \regw \regl + \floorw \floorl\,, \\
    \end{split}
\end{equation}
where $\jointl$ and $\regl$ are the data term and the regularization term \cite{SMPLify-X}, respectively.
We also introduce a floor intersection loss $\floorl = \sum_{\vertex \in V}\max(-v_z, 0)$ to force the SMPL model to stay above the floor by penalizing the SMPL vertices $V$ with $z$ coordinates below zero, assuming that the $x$-$y$ plane of the world coordinate system is aligned with the floor and its $z$-axis points upward.
Each $\lambda$ denotes the weight for the corresponding loss term.

\subsection{Foot Camera Constraint}
To achieve accurate SMPL annotation despite the frequent and heavy occlusions, especially of the legs, we mount a tiny camera on each subject's foot (`foot camera') and attach ChArUco markers \cite{ChArUco} to the ceilings and the undersides of tables.
These markers do not alter the scene appearance captured from the ceiling-mounted cameras.
We localize foot cameras with the PnP algorithm using the calibrated markers.
The foot localization data are incorporated into the SMPL pose fitting (\cref{eq:BasicPolicy}) as relative-pose constraints between the foot cameras and the SMPL feet.
We first derive this relative pose by capturing the subject in A-pose at the room center before each session.
We fit SMPL to this A-pose video using \cref{eq:BasicPolicy}, assuming a perfect fit, and compute the rotation and translation from each foot camera to the SMPL foot.
When fitting SMPL for the main sessions that capture daily activities, we add loss terms that preserve this relative transformation (see \cref{sec:DetailsOfAMD} for details).

To assess the effectiveness of the foot camera constraints, we compare the mean IoU between the human masks predicted by OneFormer \cite{OneFormer} and the rendered SMPL masks under two conditions: fitting with and without the loss terms related to the foot cameras.
Since the evaluation uses 10,000 frames where the entire body is visible, making mask estimation with OneFormer easier, the quantitative differences shown in \cref{fig:FootCameraConstraint} remain relatively small. However, as the qualitative results depict, the actual impact of the foot camera constraints goes beyond the evaluation metrics and leads to more accurate pose annotations.

\begin{table*}[t!]
    \centering
    \begin{minipage}[t]{0.61\textwidth}
        \centering
        \scriptsize
        \caption{
            Comparison of methods pretrained on AMD for action recognition, motion prediction, and motion interpolation.
            Action recognition is evaluated on HiK \cite{HiK} with both $k$-NN classification and a head trained on HiK. The other tasks are evaluated on AMD and zero-shot transferred to HiK.
            \HTwoOT~does not treat temporally contiguous frames as a segment.
            A4Mer achieves the best accuracy across tasks.
        }
        \label{tab:Main}
        \vspace{-0.8\baselineskip}
        \setlength{\tabcolsep}{0.5em}
        \begin{tabular}[t]{@{}l|c|c|cc|cc|cc@{}}
        \toprule
        \multirow{2}{*}{\textbf{Method}} &
        \multirow{2}{*}{\textbf{Objective}} &
        \multirow{2}{*}{\textbf{Segm}} &
        \multicolumn{2}{c|}{\textbf{Recog} (\%) $\uparrow$} &
        \multicolumn{2}{c|}{\textbf{Pred} (mm) $\downarrow$} &
        \multicolumn{2}{c}{\textbf{Interp} (mm) $\downarrow$}  \\
        & & & $k$-NN (top-1/-3) & head & AMD & HiK & AMD & HiK  \\
        \midrule
        MotionBERT \cite{MotionBERT} & 2D$\to$3D & frame & 1.77 / 0.35 & 27.9 & 237 & 199 & 141 & 124 \\
        USDRL \cite{USDRL} & contrast & frame            & \underline{31.1} / 43.0 & 30.1 & 171 & 155 & \underline{137} & 127 \\
        PUMPS \cite{PUMPS} & recon & frame               & 16.1 / 31.2 & 14.0 & 214 & 209 & 197 & 188 \\
        MacDiff \cite{MacDiff} & denoise & clip          & 22.1 / \underline{46.5} & 30.3 & 210 & \underline{132} & 186 & \textbf{110} \\
        BehaveMAE \cite{BehaveMAE} & MAE & clip          & 20.9 / 22.1 & \underline{35.6} & \underline{167} & 288 & 163 & 362 \\
        \HTwoOT~\cite{H2oT} & 2D$\to$3D & --             & 26.8 / 28.7 & 31.8 & 187 & 145 & 143 & \underline{123} \\
        \rowcolor{red!15}
        \textbf{A4Mer} & JEPA & variable                 & \textbf{31.7} / \textbf{59.0} & \textbf{38.1} & \textbf{150} & \textbf{120} & \textbf{126} & \textbf{110} \\
        \bottomrule
        \end{tabular}
        \vspace{-0.5\baselineskip}
    \end{minipage}%
    \hfill
    \begin{minipage}[t]{0.37\textwidth}
        \centering
        {
        \scriptsize
        \caption{
            Ablation studies of key components. SL indicates the average length of Action Motif segments across the entire dataset.
        }
        \label{tab:Ablation}
        \vspace{-0.8\baselineskip}
        \renewcommand{\arraystretch}{0.9}
        \setlength{\tabcolsep}{0.4em}
        \begin{tabular}[t]{@{}l|l|c|cc|c|c@{}}
        \toprule
        \multirow{2}{*}{\textbf{Key}} & \multirow{2}{*}{\textbf{Subst.}} & \multirow{2}{*}{\textbf{SL}} & \multicolumn{2}{c|}{\textbf{Recog} $\uparrow$} & \multicolumn{1}{c|}{\textbf{Pred}} & \multicolumn{1}{c}{\textbf{Interp}} \\
                                      &                                  &                              & $k$-NN & head                       & $\downarrow$                   & $\downarrow$                                \\
        \midrule
        \multirow{3}{*}{$\loss^\local_k + \lambda_k\loss^\glob$}
          & \cref{eq:OriginalJEPA}         & 24.1 & 15.1 & 51.7 & 222 & 210  \\
          & $\loss^\local_k$               & 22.3 & 16.2 & 51.0 & 254 & 218  \\
          & $\loss^\local_k + \loss^\glob$ & 14.5 & 25.8 & 50.2 & 188 & 181  \\
        \midrule
        segm attn & full attn              & 22.0 & 20.6 & 54.8 & 212 & 197  \\
        \midrule
        \multirow{2}{*}{segment}
          & frame                          & 1 & 31.5 & \underline{58.8} & 309 & 154  \\
          & clip                           & 10 & 26.3 & 55.5 & 208 & 183  \\
        \midrule
        JEPA
          & BERT\,\cite{BERT}              & 13.2 & \underline{34.8} & 58.2 & \underline{169} & \textbf{112}  \\
        \midrule
        full & ---                         & 10.6 & \textbf{38.1} & \textbf{59.0} & \textbf{150} & \underline{126}  \\
        \bottomrule
        \end{tabular}
        }
        \vspace{-0.5\baselineskip}
    \end{minipage}
\end{table*}

\section{Experimental Results}
Implementation and training details are provided in \cref{sec:ImplementationDetails}.
For comparison in the following experiments, we group existing self-supervised methods by their pretext tasks, and for each category, we select a state-of-the-art method whose implementations are publicly available.

\subsection{Datasets}
We use AMD to pretrain all models and to train task-specific heads for all tasks except action recognition, which is trained on the Humans in Kitchens (HiK) dataset \cite{HiK}.
For the other tasks, we evaluate both on AMD and by zero-shot transfer to HiK.
HiK provides multi-label action classes per frame. Other datasets with action labels and 3D poses typically contain only short or sequential unrelated actions, making them unsuitable for evaluating long-term motion understanding.
Due to HiK’s fine-grained, imbalanced, and occasionally noisy annotations, we experimentally found that multi-label training is too challenging regardless of the methods. We thus group semantically related actions into coarser single-label categories and use these custom labels for our recognition experiments (see \cref{supp:subsec:Datasets}).

To efficiently handle long sequences while preserving the fidelity of motion semantics, all datasets are downsampled to 5~fps.
All models take a 30-second pose sequence as input.

\subsection{Action Motifs on AMD}
\cref{fig:ActionMotifs} visualizes Action Motif sequences of daily activities in AMD. Different SMPL colors correspond to different $k$-means clusters in the representation space. Shared movements across sequences are assigned the same color, even when they occur in different actions. This shows that the learned Action Motif representations successfully capture the semantic similarity of movements with the latent tokens.

\subsection{Action Recognition}
\paragraph{Zero-Shot Transfer.}
We conduct zero-shot action recognition on HiK as in \cref{subsec:Application} and evaluate its classification accuracy.
As shown in \cref{tab:Main}, Action Motifs produced by A4Mer significantly outperform existing representations, demonstrating that they capture motion semantics and successfully encode semantically common movements in a consistent manner.
This strong zero-shot performance also reflects the diversity of AMD.

\paragraph{Transfer Learning.}
We attach a 1-layer Transformer encoder to each pretrained model and perform transfer learning for per-frame action recognition.
Note that $k$-NN classification measures semantic awareness of individual representations, whereas this task tests how much their temporal compositions convey action semantics.
As shown in \cref{tab:Main}, A4Mer achieves the highest accuracy among all compared methods, thanks to the learned Action Motifs whose temporal compositions capture the action semantics.

\subsection{Long-Term Motion Prediction}
\label{subsec:ResultLongTermMotionPrediction}
As detailed in \cref{subsec:Application}, we perform motion prediction by autoregressively predicting the next latent token and decoding it into a pose sequence.
We evaluate the prediction accuracy over a 3-second future horizon using MPJPE.

\begin{figure*}[t!]
    \centering
    \begin{subfigure}[b]{0.34\textwidth}
        \centering
        \includegraphics[width=\linewidth,trim=1pt 3pt 3pt 5pt,clip]{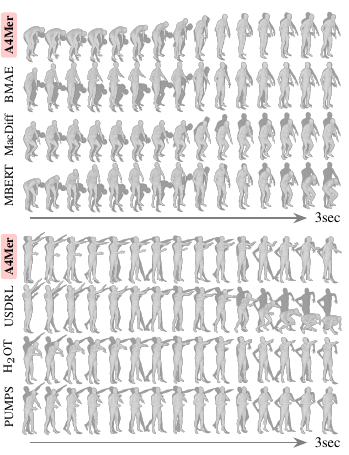}
        \caption{}
        \label{fig:Prediction}
    \end{subfigure}
    \hfill
    \begin{subfigure}[b]{0.645\textwidth}
        \centering
        \includegraphics[width=\linewidth,trim=1pt 3pt 3pt 1pt,clip]{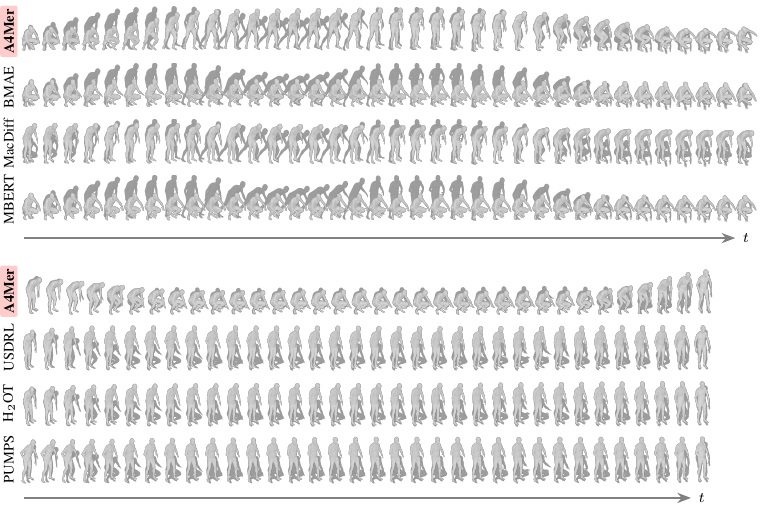}
        \caption{}
        \label{fig:Interpolation}
    \end{subfigure}
    \vspace{-0.5\baselineskip}
    \caption{(a) Predicted poses through auto-regressive latent token prediction and decoding. (b) Interpolated poses through latent token interpolation and decoding. The dark-colored SMPLs are the ground truth. BMAE and MBERT denote BehaveMAE and MotionBERT, respectively. Compared to the other representations, the Action Motifs extracted by A4Mer provide stronger semantic cues, leading to more accurate and semantically meaningful forecasts and interpolations rather than simple linear extrapolation or interpolation.}
    \vspace{-0.3\baselineskip}
\end{figure*}

\cref{fig:Prediction} shows qualitative results, and \cref{tab:Main} reports quantitative comparisons. Action Motifs used as the underlying movement representation for motion prediction outperform all existing representations, even in zero-shot evaluations on HiK.
Past representations struggle to model long-term dynamics since they do not sufficiently capture the semantics of the movements.
In contrast, Action Motifs emerge as semantically meaningful patterns, and their temporal compositions represent action semantics, enabling the prediction head to reason about future actions at a semantic level beyond motion smoothness.

\subsection{Motion Interpolation}
For motion interpolation, we randomly mask contiguous Action Motif segments spanning 4--8 seconds for each input sequence, and extract latent tokens of unmasked segments using the pretrained models. The latent tokens corresponding to the masked segments are inferred with an interpolation head. The resulting latent token sequence is then decoded into poses (see \cref{subsec:Application}). We evaluate interpolation accuracy using MPJPE over the interpolated poses. \Cref{tab:Main} shows that the Action Motifs extracted by A4Mer achieve lower MPJPE than the other representations. Moreover, \cref{fig:Interpolation} shows that, while the others often produce nearly linear interpolations, the Action Motifs enable semantically meaningful interpolation of missing motions. This suggests that the temporal compositions of Action Motifs represent distinct actions and thus provide stronger semantic constraints for motion interpolation.

\subsection{Ablation Study}
We conduct ablation studies by removing each key component of A4Mer and transferring the extracted representations to the downstream tasks.
\cref{tab:Ablation} quantitatively reports the contribution of each component.

\vspace{-8pt}
\paragraph{Global and Local Decomposition.}
We find that decomposing the objective into $\loss^\local_k$ and $\loss^\glob$ and dynamically adjusting their weights so that the former dominates leads to better performance.
Without this decomposition, all latent tokens within a sequence tend to collapse into similar representations, causing the model to extract excessively long patterns when determining Action Motif segments.
Conversely, using only the local term discards global contextual information and prevents the loss from reflecting the semantics conveyed by the temporal composition of latent tokens.

\vspace{-8pt}
\paragraph{Segment-Wise Attention}
In Encoder, restricting self-attention to tokens within the same segment encourages it to focus on intra-segment information aggregation.
As a result, LatentFormer can handle already abstracted latent tokens, facilitating semantic reasoning across segments and yielding more robust representations.

\vspace{-8pt}
\paragraph{Action Motif Segments.}
When learning representations at fixed lengths---either per frame or per clip (with the clip length set to the average Action Motif segment length obtained by our method)---the model fails to capture motion semantics, leading to degraded performance across tasks.

\vspace{-8pt}
\paragraph{JEPA.}
Computing loss in the latent space encourages representations to capture the semantic essence of movements while remaining invariant to trivial pose variations, whereas pose-space losses preserve fine-grained motion details beneficial to certain tasks.
Moreover, with the smooth L1 loss (\cref{eq:Objective}), the latent space exhibits a Manhattan-like geometry, making it amenable to k-means clustering and facilitating the second-stage segmentation.
Together, these properties yield effective representations for diverse downstream tasks.

\section{Conclusion}
We introduced Action Motifs---semantically meaningful and reusable movement patterns naturally emerging from 3D pose data---and proposed A4Mer for self-supervised learning of them from pose sequences.
Bottom-up learning through masked latent token prediction lets a hierarchical movement representation naturally emerge.
Learned Action Motifs and their temporal compositions experimentally showed semantic awareness, outperforming existing fixed-length representations in diverse downstream tasks.
We also introduced AMD, a large-scale dataset that captures various daily activities with per-frame SMPL annotations with the help of a novel use of cameras. We hope A4Mer and AMD will serve as foundations for furthering research on human behavior modeling.

\paragraph*{Acknowledgement}
This work was in part supported by
JSPS KAKENHI 
21H04893 
and JST JPMJAP2305.

\appendix
\renewcommand{\thefigure}{\Alph{figure}}
\renewcommand{\thetable}{\Alph{table}}
\renewcommand{\theequation}{\Alph{equation}}
\renewcommand{\thesection}{\Alph{section}}
\setcounter{figure}{0}
\setcounter{table}{0}
\setcounter{equation}{0}

\section{Implementation Details}
\label{sec:ImplementationDetails}
\subsection{Pose Representation}
For the input to A4Mer, we use the same pose representation as HumanML3D \cite{HumanML3D}, where body shapes are normalized and poses are expressed in a local coordinate system aligned with the gravity direction. A pose is defined as $\left(\bm{c},\, \dot{r}^{a},\, \dot{\bm{r}}^{v},\, \bm{j},\, \bm{j}^{v},\, \bm{j}^{r}\right)$, where $\bm{c}\in\{0,1\}^{4}$ denotes the binary indicators of the ground contacts for left and right heels and toes; $\dot{r}^{a}\in\real$ is the pelvis angular velocity along the gravity direction; $\dot{\bm{r}}^{v}\in\real^{2}$ is the pelvis velocity on the horizontal plane; $\bm{j}\in\real^{J\times 3}$, $\bm{j}^{v}\in\real^{J\times 3}$, and $\bm{j}^{r}\in\real^{J\times 6}$ represent joint positions, velocities, and rotations, respectively, where $J$ is the number of joints. The joint rotations $\bm{j}^{r}$ are parameterized using the 6D continuous rotation representation \cite{OnTheContinuityOfRotationRepresentations}.
For comparison with existing methods in our experiments, we use their original pose representations exactly as specified in the respective papers.

\subsection{Action Atom Segmentation}
\label{subsec:ActionAtomSegmentation}
To determine the boundaries between Action Atom segments based on nonlinear changes in joint trajectories, A4Mer starts by linearly predicting local joint positions $\bm{j}_t\in \real^{J\times 3}$ at frame $t$. With least-squares fitting to the past $w$ frames $\{\bm{j}_{t-w}, \dots, \bm{j}_{t-1}\}$, a linear prediction of the joint trajectory at $\bm{j}_t$ is computed. The prediction error is computed as the $L_1$ difference $\mathcal{E}_{j,t} \in \real^3$ for each joint $j$. To eliminate joint-specific variation in the motion range, each joint’s error is normalized with the mean $\mu_j$ and the standard deviation $\sigma_j$ computed over all frames within the same video. The normalized errors are averaged over joints and coordinates to obtain a scalar error signal $e_t$ for each frame.
To capture movement transitions from this error signal, the first-order temporal difference $\Delta e_t = e_t - e_{t-1}$ and the second-order difference $\Delta^2 e_t = \Delta e_t - \Delta e_{t-1}$ are computed. While $\Delta e_t$ is effective at capturing large movement transitions, it may fail to respond when the transition is sharp yet low in magnitude. The second difference $\Delta^2 e_t$ compensates for this by responding strongly to abrupt accelerations in the error signal.
Based on these quantities, candidate boundary frames are detected using the logical condition
\begin{equation}
    \label{eq:ActionAtomSegmentation}
    \bigl(\Delta e_t > \tau_1 \ \wedge\  \Delta^2 e_t > 0\bigr) \lor \bigl( \Delta^2 e_t > \tau_2 \bigr) \,,
\end{equation}
where $\tau_1=0.005$ and $\tau_2 = \mathrm{mean}_t(|\Delta^2 e_t|)$.
The first term selects frames where the error is not only increasing beyond a threshold but also accelerating, indicating the onset of a new movement.
The second term captures abrupt events, even when their first-order differences remain small.
To prevent excessive fragmentation of segments, once a boundary is identified, we suppress further boundary detection for the subsequent $0.5$ seconds, under the assumption that human movements typically preserve coherence for at least this duration.
Note that this boundary-detection procedure is applied to the 30 fps streams, in contrast to the 5 fps used as input to the representation learning models.

\subsection{Action Motif Segmentation}
The Generalized Sequential Pattern (GSP) algorithm \cite{GSP} is used to find frequent patterns shared across sequences of Action Atoms discretized into categorical codes by $k$-means.
GSP is grounded in the Apriori principle, \ie, any supersequence of an infrequent subsequence must also be infrequent, and discovers frequent patterns by iteratively expanding the current set of frequent patterns into longer pattern candidates.
At each iteration, candidates are generated by concatenating pairs of frequent patterns obtained from all previous iterations. Candidates whose occurrence counts exceed a threshold $o$ are then registered as new frequent patterns.
This iterative generate-and-prune procedure enables GSP to efficiently extract recurring Action Atom patterns, \ie, Action Motifs.

For an efficient matrix-based implementation, we set the maximum pattern length $p_\mathrm{max}\,(=20)$. The minimum occurrence threshold is linearly decreased from $o_\mathrm{max}\,(=15)$ to $o_\mathrm{min}\,(=5)$ as the pattern length increases.
To mitigate the risk of missing patterns due to intervening trivial or minor movements, two candidate patterns are determined as being equivalent when their Hamming distance does not exceed $h$, and their occurrences are summed when counting. The Hamming-distance threshold $h$ is linearly increased from $h_\mathrm{min}\,(=0)$ to $h_\mathrm{max}\,(=4)$ as the pattern length grows. Furthermore, contiguous runs of identical categorical codes are treated as continuations of the same action, and such runs---regardless of whether their length exceeds $p_\mathrm{max}$---are also considered valid patterns.

\subsection{Network Architectures}

\begin{table}[t]
    \centering
    \caption{Architectural configuration of A4Mer and task-specific heads.}
    \label{tab:Architecture}
    \scriptsize
    \setlength{\tabcolsep}{0.45em}
    \begin{tabular}{lcccc}
        \toprule
        Module & Submodule & Layers & Token Dim. & Attn.\! Heads \\
        \midrule
        \multirow{3}{*}{A4Mer}
            & Encoder        & 4 & 256 & 4 \\
            & LatentFormer   & 4 & 256 & 4 \\
            & predictor $g_\phi$      & 3 & 256 & 4 \\
        \midrule
        \multicolumn{1}{l}{Recognition Head} & -- & 1 & 256 & 4 \\
        \multicolumn{1}{l}{Decoding Head}    & -- & 4 & 256 & 4 \\
        \multicolumn{1}{l}{Motion Prediction Head}  & -- & 4 & 64  & 4 \\
        \multicolumn{1}{l}{Motion Interpolation Head}    & -- & 4 & 256 & 4 \\
        \bottomrule
    \end{tabular}
\end{table}

The task-specific heads and A4Mer including the predictor $g_\phi$ used for JEPA \cite{JEPA} are built upon Transformer \cite{Transformer} architectures. All self-attention layers in these networks employ the rotary positional embeddings (RoPE) \cite{RoPE}. The architectural hyperparameters of each network are summarized in \cref{tab:Architecture}.

\paragraph{A4Mer.}
Encoder consists of a stack of Transformer-decoder layers, each composed of segment-wise self-attention and cross-attention from latent tokens $z_k$ to their corresponding input segments. Here, we detail these two attention mechanisms.
We represent the attention operator as
\begin{align}
    \mathrm{Attn}&(\mathrm{Q}, \mathrm{K}, \mathrm{V})
    = \notag \\
    &\mathrm{softmax}\!\left(
        \frac{(\mathrm{Q} W_\mathrm{Q})(\mathrm{K} W_\mathrm{K})^{\top}}{\sqrt{d}}
    \right)
    (\mathrm{V} W_\mathrm{V})\,,
\end{align}
where $W_\mathrm{Q}, W_\mathrm{K}, W_\mathrm{V}$ denote learnable projection matrices for queries $\mathrm{Q}$, keys $\mathrm{K}$, and values $\mathrm{V}$, respectively, and $d$ denotes the dimensionality of the projected queries and keys.

For convenience, we restate the notation from the main paper: the input sequence $X=(x_1,\ldots,x_T)$ is partitioned into segments $X_k=\{x_t \mid s(t)=k\}$, where $s(t)$ returns the segment index that frame $t$ belongs to.
Each layer first applies self-attention within each segment $k \in \{1,\ldots,K\}$:
\begin{equation}
    \mathrm{SelfAttn}_k
    = \mathrm{Attn}(X_k, X_k, X_k)\,,
\end{equation}
which restricts the receptive field to intra-segment interactions, helping the model consolidate local movement information without responding to coincidental similarities across distant frames.

Each variable-length segment $X_k$ is consolidated into a single latent token through cross-attention:
\begin{equation}
    \mathrm{CrossAttn}_k
    = \mathrm{Attn}(z_k, X_k, X_k)\,.
\end{equation}

LatentFormer operates on the set of latent tokens $Z = (z_1,\ldots,z_K)$ and performs standard self-attention $\mathrm{Attn}(Z, Z, Z)$ to capture temporal relationships between movements at the abstracted semantic level.

\paragraph{Predictor in JEPA.}
The predictor $g_{\phi}$ takes a latent-token sequence in which the positions indexed by $\mathcal{K} \subset \{1,\ldots,K\}$ are masked out.
For each $k \in \mathcal{K}$, a learnable mask token $\mathcal{M}$ (shared across $\mathcal{K}$) is inserted. Formally, we construct $\tilde{Z} = (\tilde{z}_1,\ldots,\tilde{z}_K)$, where
\begin{equation}
    \tilde{z}_k =
        \begin{cases}
            \mathcal{M}, & k \in \mathcal{K}, \\[4pt]
            z_k, & \text{otherwise}\,.
        \end{cases}
\end{equation}
The resulting sequence $\tilde{Z}$ is then processed using self-attention, and the outputs corresponding to the mask tokens are passed through a single linear layer to predict the latent tokens at the masked positions.

\subsection{Training Settings}
\paragraph{Pretraining of A4Mer.}
We pretrain A4Mer in two steps: an initial training of the first-stage model, and a subsequent end-to-end training of the first- and second-stage models.
Both training steps are conducted for 400 epochs with a batch size of 64.
Optimization is performed using AdamW with a learning rate of $1\times10^{-3}$ and weight decay of $0.1$. We apply a linear warm-up for the first 2{,}000 iterations, followed by a cosine-annealing schedule over 100{,}000 iterations with a minimum learning rate of $1\times10^{-6}$.
For all downstream heads, we adopt the same optimization configuration as above, except that the number of training epochs is reduced to 200.
During pretraining, the parameters of the target feature extractor $f_{\bar{\theta}}$ are updated using an exponential moving average with a decay rate of $0.996$, and 50\% of latent tokens are randomly masked to create the pretext task where the model predicts masked latent tokens from visible ones.

\paragraph{Action Recognition.}
For action recognition, a 1-layer Transformer-encoder classification head is trained by minimizing the cross-entropy loss $\loss_\mathrm{recog}$ with class-balanced weighting to mitigate class imbalance:
\begin{equation}
    \loss_\mathrm{recog}
    = -\sum_{\mathrm{c}=1}^{\mathrm{C}} w_\mathrm{c}\, y_\mathrm{c} \log p_\mathrm{c} \,,
\end{equation}
where $p_\mathrm{c}$ is the predicted probability for class $\mathrm{c}\in\{1,\ldots, \mathrm{C}\}$, $y_\mathrm{c}$ is the one-hot label, and $w_\mathrm{c}$ is the inverse of the frequency of class $\mathrm{c}$ in the dataset.

\paragraph{Decoding.}
Motion prediction and motion interpolation are realized by decoding predicted or interpolated latent tokens into poses.
For this, we train a decoding head with a Transformer-encoder architecture.
The decoding head is trained using the objective $\loss_{\mathrm{dec}}$, which combines the joint-position reconstruction loss $\loss_{\mathrm{pos}}$ and the velocity reconstruction loss $\loss_{\mathrm{vel}}$:
\begin{align}
    \loss_{\mathrm{dec}} &= \lambda_\mathrm{pos}\loss_{\mathrm{pos}} + \lambda_\mathrm{vel}\loss_{\mathrm{vel}}\,, \\
    \loss_{\mathrm{pos}} &= \frac{1}{J T}\sum_{t=1}^T \sum_{j=1}^J \left\| \hat{\bm{j}}_{j,t} - \bm{j}_{j,t} \right\|\,, \\
    \loss_{\mathrm{vel}} &= \frac{1}{J T}\sum_{t=2}^{T-1} \sum_{j=1}^J \left\| \hat{\Delta\bm{j}}_{j,t} - \Delta\bm{j}_{j,t} \right\|\,, \\
    \hat{\Delta\bm{j}}_{j,t} &= \hat{\bm{j}}_{j,t} - \hat{\bm{j}}_{j,t-1}, \quad \Delta\bm{j}_{j,t} = \bm{j}_{j,t} - \bm{j}_{j,t-1} \,,
\end{align}
where $\bm{j}_{j,t}$ denotes the position of the joint $j$ at frame $t$ and $\hat{\bm{j}}_{j,t}$ is the predicted one. We set $\lambda_\mathrm{pos}$ as $1.5$ and $\lambda_\mathrm{vel}$ as $1.0$.

\paragraph{Long-term Motion Prediction.}
For long-term motion prediction, we train a prediction head for forecasting the next latent token $z_{k+1}$ from the preceding sequence $z_{1:k}$. In addition to minimizing the latent token prediction error $\loss_{\mathrm{token}}$, the prediction head also minimizes the joint-space reconstruction error $\loss_{\mathrm{dec}}$ computed by feeding the predicted latent tokens into the pretrained, frozen decoder. This auxiliary objective enables the model to focus its predictions on the meaningful dimensions within the latent space.

Moreover, since future segments are not observable at inference time, the prediction head is trained to predict not only latent tokens but also their segment lengths, and minimizes the segment length error $\loss_{\mathrm{segm}}$. Since segment lengths follow a long-tailed distribution where short segments are far more common than long ones (see \cref{fig:SegmLenDistribution}), direct regression on the raw scale might cause the loss to be dominated by rare but large values. To alleviate this imbalance, we apply a logarithmic transformation to the ground-truth segment lengths and train the model by minimizing the prediction error in log-space. This transformation compresses the dynamic range of the target values, reduces the undue influence of rare long segments, and yields a more uniform target distribution.
Note that the ``ground-truth'' segment lengths used here are not manually annotated; rather, they are determined by the pretrained A4Mer.

The final prediction loss $\loss_{\mathrm{pred}}$ is represented as
\begin{align}
    \loss_{\mathrm{pred}}
    &= \loss_{\mathrm{token}}
    + \loss_{\mathrm{dec}}
    + \loss_{\mathrm{segm}} \,, \\
    \loss_{\mathrm{token}} &= \sum_{k=1}^{K-1} \mathrm{SmoothL1}\left(z_{k+1} - \hat{z}_{k+1}\right) \,, \\
    \loss_{\mathrm{segm}} &= \sum_{k=1}^{K-1} \left|\log{l_{k+1}} - \mathrm{Softplus}\left(\hat{l}_{k+1}\right)\right| \,,
\end{align}
where $\hat{\cdot}$ denotes an estimated value by the prediction head, and $l_k$ is the segment length of the latent token $k$.

For the baseline methods, we similarly train the next-token prediction head using the representations output by each method as input. However, since these methods except \HTwoOT~\cite{H2oT} rely on fixed, predefined segment lengths, the segment length loss $\loss_{\mathrm{segm}}$ is not applied.
\HTwoOT~does not treat consecutive frames as temporal segments. Instead, it groups feature vectors based on their similarity with DPC-$k$NN \cite{DPCkNN}, producing clusters that do not correspond to fixed temporal durations. As a result, the segment length is inherently undefined for this method.
To perform motion prediction under the same formulation as other methods, we approximate the segment length by dividing the input pose-sequence length by the predefined number of clusters and use this value during both training and inference.

\paragraph{Motion Interpolation.}
For motion interpolation, we train an interpolation head that infers latent tokens of unobserved frames from the latent tokens extracted by A4Mer from partially observed pose sequences. We insert learnable tokens at the latent positions corresponding to the unobserved frames and feed the resulting sequence to the interpolation head. This head is trained by minimizing a loss $\loss_{\mathrm{latent}}$ between the predicted latent tokens and the latent tokens extracted by A4Mer from fully observed input sequences. Although the primary goal is to recover the latent tokens of the unobserved frames, partially masking the input also slightly shifts the distribution of the latent tokens corresponding to the observed frames. Therefore, we compute the loss not only for the latent tokens produced from the learnable tokens but also for the latent tokens of the observed frames, allowing the interpolation head to additionally correct the output distribution. Formally, the loss $\loss_{\mathrm{latent}}$ is formulated as:
\begin{align}
    \loss_{\mathrm{latent}} = \sum_{k=1}^{K} \mathrm{SmoothL1}\left(z_{k} - \hat{z}_{k}\right) \,.
\end{align}

In addition, as in motion prediction, we introduce an auxiliary objective in the pose space $\loss_{\mathrm{dec}}$ by passing the interpolated latent token sequences through the pretrained, frozen decoder.

\subsection{Training Time}
\cref{tab:TrainingTime} reports the training times of A4Mer and other representation learning methods. Although A4Mer involves a two-step pre-training procedure unlike the others, its overall training time remains comparable to, or even faster than theirs.

\begin{table}[t]
    \centering
    \caption{Pretraining time. Although A4Mer requires two steps for pretraining, its overall training time is comparable to or faster than other methods.}
    \label{tab:TrainingTime}
    \renewcommand{\arraystretch}{1.1}
    \setlength{\tabcolsep}{2pt}
    \begin{tabular}{@{}llc@{}}
        \toprule
        Method & Step & Time (hours) \\
        \midrule
        MotionBERT \cite{MotionBERT} & training & 32.86 \\
        USDRL \cite{USDRL} & training & 13.45 \\
        PUMPS \cite{PUMPS} & training & 2.926 \\
        MacDiff \cite{MacDiff} & training & 10.32 \\
        BehaveMAE \cite{BehaveMAE} & training & 21.83 \\
        \midrule
        \multirow{4}{*}{\textbf{A4Mer}}
            & 1st-stage training        & 3.096 \\
            & segmentation              & 0.056 \\
            & end-to-end training       & 4.400 \\
            \cmidrule{2-3}
            & total & 7.552 \\
        \bottomrule
    \end{tabular}
\end{table}

\subsection{Datasets}
\label{supp:subsec:Datasets}
\paragraph{AMD.}
All representation learning methods, as well as the decoding, motion-prediction, and motion-interpolation heads, are trained on our proposed AMD, which contains 14.2 hours of footage.
We split AMD into 80\% for training, 10\% for validation, and 10\% for testing, ensuring that no subjects overlap across splits. While the scenarios performed by participants are shared across splits, variations exist in the specific actions performed, the order of actions, and the arrangement of furniture, depending on the subject.
For evaluation of motion prediction and motion interpolation, we automatically exclude trivial sequences where the subject merely remains stationary by detecting them based on the magnitude of joint velocities.

\paragraph{Humans in Kitchens.}
We train the action recognition head on the Humans in Kitchens (HiK) dataset \cite{HiK}. Following the protocol in the original paper, recordings from \emph{Kitchen~D} are used exclusively for testing, while data from \emph{Kitchens~A,~B,} and \emph{C} are split into 90\% for training and 10\% for validation, respectively.
Since HiK contains fine-grained, imbalanced, and occasionally noisy action annotations, multi-label classification is unreasonably challenging across all methods we examined, as verified in \cref{tab:MultiLabelRecog}.
Instead, we group semantically related actions into coarser, single-label categories and adopt these custom labels for our recognition experiments.
For this, we automatically assign new task class labels through the following procedure:
\begin{enumerate}
    \item We first define a set of task classes as shown in the first column of \cref{tab:HiKClass}.
    \item We then categorize the original action labels into
    four groups: \textit{main actions}, \textit{sub-actions}, \textit{transition actions}, and
    \textit{others}.
    \begin{itemize}
        \item \textbf{Main actions:} Actions uniquely associated with a specific task class.
        \item \textbf{Sub-actions:} Actions that accompany main actions and may appear across
        multiple task classes.
        \item \textbf{Transition actions:} Movements associated with transitions between actions,
        such as \textit{sitting down} or \textit{standing up}.
        \item \textbf{Others:} Rarely occurring actions and state-like labels (\eg, \textit{sitting})
        that do not convey task-specific semantics.
    \end{itemize}

    \item We next identify all frames annotated with any main action and assign the corresponding task class label to these frames (the first and second columns of \cref{tab:HiKClass}).

    \item For each frame with task class labels, if sub-actions belonging to the same task class appear in its temporal neighborhood, we propagate the task class label to cover the entire contiguous interval containing these sub-actions (the first and third columns of \cref{tab:HiKClass}).

    \item Finally, among the remaining unlabeled frames, those containing transition actions are assigned the task class label associated with the corresponding transition (the first and fourth columns of \cref{tab:HiKClass}).
\end{enumerate}
Any frames that remain unlabeled after this procedure are excluded from both training and evaluation, resulting in a curated subset comprising 30.8\% of the original data.

For motion prediction and motion interpolation evaluation on HiK, we follow the same criterion as in AMD and exclude trivial sequences in which the subject remains nearly stationary.

\begin{table}[t]
    \centering
    \caption{
Multi-label action recognition on HiK with its original fine-grained labels.
Although the Action Motifs extracted by A4Mer attain slightly better performance than the others, the overall scores are consistently low, making this evaluation setting inadequate for assessing the utility of the learned representations. Therefore, in the main text, we instead train and evaluate models using the single-label classes that we annotated following the procedure described in \cref{supp:subsec:Datasets}.
    }
    \label{tab:MultiLabelRecog}
    \renewcommand{\arraystretch}{1.1}
    \setlength{\tabcolsep}{2pt}
    \begin{tabular}[t]{@{}lc@{}}
    \toprule
    Method & mAP $\uparrow$ \\
    \midrule
    MotionBERT \cite{MotionBERT} & 0.113 \\
    USDRL \cite{USDRL} & 0.087 \\
    PUMPS \cite{PUMPS} & 0.057 \\
    MacDiff \cite{MacDiff} & 0.072 \\
    BehaveMAE \cite{BehaveMAE} & 0.113 \\
    \HTwoOT~\cite{H2oT} & 0.096 \\
    \rowcolor{red!15}\textbf{A4Mer} & \textbf{0.121} \\
    \bottomrule
    \end{tabular}
\end{table}

\begin{table*}[t]
\centering
\scriptsize
\renewcommand{\arraystretch}{0.75}
\begin{tabular}{@{}l|l|l|l@{}}
\toprule
\multicolumn{1}{c|}{Task classes} &
\multicolumn{1}{c|}{Main actions} &
\multicolumn{1}{c|}{Sub-actions} &
\multicolumn{1}{c}{Transition actions} \\
\midrule
\multirow{3}{*}{take dish out of cupboard} & open cupboard & & \\
 & take dish out of cupboard & & \\
 & close cupboard & & \\
 \midrule
drink & drink & carry cup & \\
 \midrule
phone call & phone call & & \\
 \midrule
use smartphone & use smartphone & & \\
 \midrule
use laptop & use laptop & & \\
 \midrule
fussball & fussball & & \\
 \midrule
draw on whiteboard & draw on whiteboard & & \\
 \midrule
erase on whiteboard & erase on whiteboard & & \\
 \midrule
remove sheet from whiteboard & remove sheet from whiteboard & & \\
 \midrule
eat fruit & eat fruit & & \\
 \midrule
peal fruit & peal fruit & & \\
 \midrule
read paper & read paper & & \\
 \midrule
\multirow{3}{*}{use dishwasher} & open dishwasher & & \\
 & place in dishwasher & & \\
 & close dishwasher & & \\
 \midrule
\multirow{4}{*}{take cake out of fridge} & open fridge & & \\
 & take cake out of fridge & & \\
 & put cake in fridge & & \\
 & close fridge & & \\
 \midrule
\multirow{2}{*}{take sth. out of drawer} & open drawer & & \\
 & close drawer & & \\
 \midrule
\multirow{2}{*}{pour milk} & pour milk & carry cup & \\
 & take milk & & \\
 \midrule
washing hands & washing hands & take water from sink & \\
 \midrule
clean dish & clean dish & take water from sink & \\
 \midrule
put cup in microwave & put cup in microwave & & \\
 \midrule
take cup out of microwave & take cup out of microwave & & \\
 \midrule
\multirow{2}{*}{put teabag in cup} & put teabag in cup & & \\
 & take teabag & & \\
 \midrule
\multirow{2}{*}{put water in kettle} & put water in kettle & take water from sink & \\
 & & take kettle & \\
 \midrule
pour kettle & pour kettle & take kettle & \\
 \midrule
\multirow{2}{*}{make coffee} & place cup onto coffee machine & & \\
 & press coffee button & & \\
 \midrule
take cup from coffee machine & take cup from coffee machine & & \\
 \midrule
mark coffee & mark coffee & & \\
 \midrule
\multirow{6}{*}{prepare coffee machine} & check water in coffee machine & take water from sink & \\
 & take water tank from coffee machine & & \\
 & empty water from coffee machine & & \\
 & fill water to coffee machine & & \\
 & fill water tank & & \\
 & place water tank in coffee machine & & \\
 \midrule
\multirow{3}{*}{cut cake} & take piece of cake & & \\
 & place cake on plate & & \\
 & cut cake in pieces & & \\
 \midrule
place cake on table & place cake on table & & \\
 \midrule
eat cake & eat cake & & \\
 \midrule
put water in glass & put water in glass & take water from sink & \\
 \midrule
throw in trash & throw in trash & & \\
 \midrule
sitting down & & & sitting down \\
 \midrule
standing up & & & standing up \\
 \midrule
leaning down & & & leaning down \\
 \midrule
kneeling down & & & kneeling down \\
\bottomrule
\end{tabular}
    \vspace{4pt}
    \caption{Our definition of the task classes on Humans in Kitchens. We first categorize the original HiK action labels into \textit{main actions}, \textit{sub-actions}, \textit{transition actions} (the second to fourth columns), and \textit{others}, and then assigned our new task classes considering the context of the actions (the first column). HiK action labels not listed in this table are categorized as \textit{others}. This includes labels describging states rather than actions (\eg, ``standing'', ``leaning''), labels that is annotated to co-occur with other actions (\eg, ``start microwave'' appears while performing another action), and labels that occur only rarely (\eg, ``clean countertop'').  See \cref{supp:subsec:Datasets} for details.
    }
    \label{tab:HiKClass}
\end{table*}


\section{Additional Experimental Results}
\subsection{Segment Length Distributions}
\begin{figure}[t]
    \centering
    \includegraphics[width=\linewidth]{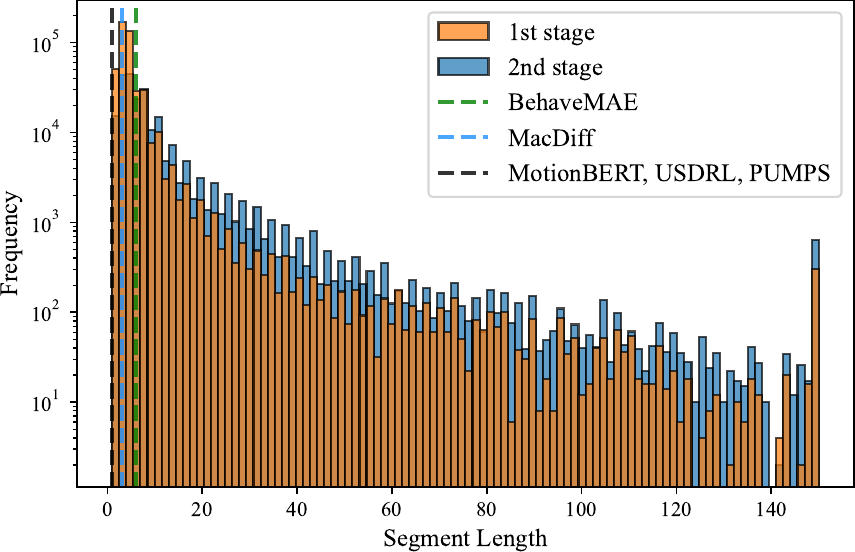}
    \caption{Distribution of segment lengths.  Compared with the other methods that adopt fixed segment lengths (dashed lines), Action Motif segments are substantially longer. Note that our implementation caps the maximum length at 150, so the rightmost bin has a relatively large count.
    }
    \label{fig:SegmLenDistribution}
\end{figure}

As shown in \cref{fig:SegmLenDistribution}, our Action Motif segments are substantially longer than those of existing fixed-length representations, indicating a significantly higher compression rate of pose information. Despite this aggressive compression, Action Motifs achieve superior performance across major downstream tasks, as reported in Tab.~2 of the main text, since their segments correspond to semantically meaningful movements.

\subsection{Sensitivity to Hyperparameters}
We analyze the impact of A4Mer hyperparameters used during pretraining (and, in some cases, during inference) on downstream task performance.

\paragraph{$\tau_1$ in Action Atom Segmentation.}
Varying $\tau_1$ in \cref{eq:ActionAtomSegmentation} during both pretraining and inference results in different Action Atom segments. As shown in \cref{tab:ActionAtomSegmentation}, although different choices of $\tau_1$ lead to minor variations in performance, our method generally outperforms existing methods, and achieves comparable performance in the remaining cases.

\paragraph{The Number of Clusters of $k$-means.}
During pretraining of A4Mer, we first discretize the Action Atoms output by the pretrained first-stage model using $k$-means clustering to discover Action Atom patterns. \cref{tab:KMeans} reports that $k=512$ offers consistently solid performance across all tasks, although it is not always the best.
Consequently, we adopt $k=512$ in the main paper.
Note that other choices of $k$ achieve performance that is comparable to or exceeds that of competing methods on most tasks.

\paragraph{$o_\text{max}$ and $o_\text{min}$ in GSP.}
As shown in \cref{tab:OmaxOmin}, varying the occurrence thresholds $o_\text{max}$ and $o_\text{min}$ in GSP leads to slight performance variations, yet A4Mer achieves competitive performance across all threshold settings and, in most cases, outperforms prior methods across downstream tasks.

\begin{table}[t]
    \centering
    \caption{Impact of varying $\tau_1$ in the Action Atom segmentation. Across downstream tasks and different values of $\tau_1$, A4Mer outperforms prior methods in most cases and achieves comparable performance otherwise.
    }
    \footnotesize
    \setlength{\tabcolsep}{0.5em}
    \begin{tabular}[t]{@{}l|cc|cc|cc@{}}
    \toprule
    \multirow{2}{*}{$\tau_1$} &
    \multicolumn{2}{c|}{\textbf{Recog} (\%) $\uparrow$} &
    \multicolumn{2}{c|}{\textbf{Pred} (mm) $\downarrow$} &
    \multicolumn{2}{c}{\textbf{Interp} (mm) $\downarrow$}  \\
    & $k$-NN (top-1/-3) & head & AMD & HiK & AMD & HiK  \\
    \midrule
    0.01        & 30.2 / 59.3 & 39.6 & 164 & 128 & 134 & 118 \\
    0.001       & 31.1 / 57.1 & 34.5 & 155 & 138 & 142 & 124 \\
    0.005       & 31.7 / 59.0 & 38.1 & 150 & 120 & 126 & 110 \\
    \midrule
    best in priors & 31.1 / 46.5 & 35.6 & 167 & 132 & 137 & 110 \\
    \bottomrule
    \end{tabular}
    \label{tab:ActionAtomSegmentation}
\end{table}

\begin{table}[t]
    \centering
    \caption{Effect of varying $k$ in the $k$-means clustering.
    Although not always optimal, $k=512$ offers consistently strong performance across all tasks.
    We therefore adopt $k=512$ in the main paper.
    Other choices of $k$ remain comparable to, and in many cases exceed, prior methods.
    }
    \footnotesize
    \setlength{\tabcolsep}{0.4em}
    \begin{tabular}[t]{@{}l|cc|cc|cc@{}}
    \toprule
    \multirow{2}{*}{$k$ for $k$-means} &
    \multicolumn{2}{c|}{\textbf{Recog} (\%) $\uparrow$} &
    \multicolumn{2}{c|}{\textbf{Pred} (mm) $\downarrow$} &
    \multicolumn{2}{c}{\textbf{Interp} (mm) $\downarrow$}  \\
    & $k$-NN (top-1/-3) & head & AMD & HiK & AMD & HiK  \\
    \midrule
    128            & 29.1 / 57.7 & 35.2 & 163 & 130 & 140 & 125 \\
    256            & 28.0 / 57.1 & 35.9 & 155 & 126 & 134 & 117 \\
    512            & 31.7 / 59.0 & 38.1 & 150 & 120 & 126 & 110 \\
    1024           & 31.5 / 58.8 & 38.6 & 156 & 121 & 123 & 112 \\
    \midrule
    best in priors & 31.1 / 46.5 & 35.6 & 167 & 132 & 137 & 110 \\
    \bottomrule
    \end{tabular}
    \label{tab:KMeans}
\end{table}

\begin{table}[t]
    \centering
    \caption{
        Effect of varying $o_\text{max}$ and $o_\text{min}$ in GSP. Although different threshold choices lead to slight variations in downstream task performance, A4Mer outperforms prior methods in most cases and remains competitive otherwise.
    }
    \footnotesize
    \setlength{\tabcolsep}{0.5em}
    \begin{tabular}[t]{@{}l|cc|cc|cc@{}}
    \toprule
    \multirow{2}{*}{$o_\text{max}$ / $o_\text{min}$} &
    \multicolumn{2}{c|}{\textbf{Recog} (\%) $\uparrow$} &
    \multicolumn{2}{c|}{\textbf{Pred} (mm) $\downarrow$} &
    \multicolumn{2}{c}{\textbf{Interp} (mm) $\downarrow$}  \\
    & $k$-NN (top-1/-3) & head & AMD & HiK & AMD & HiK  \\
    \midrule
    30 / 5         & 32.0 / 60.3 & 38.8 & 143 & 116 & 130 & 110 \\
    20 / 5         & 31.8 / 60.1 & 39.2 & 153 & 116 & 128 & 112 \\
    15 / 5         & 31.7 / 59.0 & 38.1 & 150 & 120 & 126 & 110 \\
    10 / 5         & 28.0 / 55.9 & 41.7 & 161 & 126 & 131 & 120 \\
    15 / 3         & 32.3 / 61.4 & 39.2 & 166 & 134 & 131 & 117 \\
    15 / 7         & 30.9 / 58.8 & 38.8 & 167 & 131 & 130 & 114 \\
    \midrule
    best in priors & 31.1 / 46.5 & 35.6 & 167 & 132 & 137 & 110 \\
    \bottomrule
    \end{tabular}
    \label{tab:OmaxOmin}
\end{table}

\subsection{Additional Qualitative Results}
\begin{figure*}[t!]
    \centering
    \includegraphics[width=\linewidth,trim=1pt 3pt 3pt 5pt,clip]{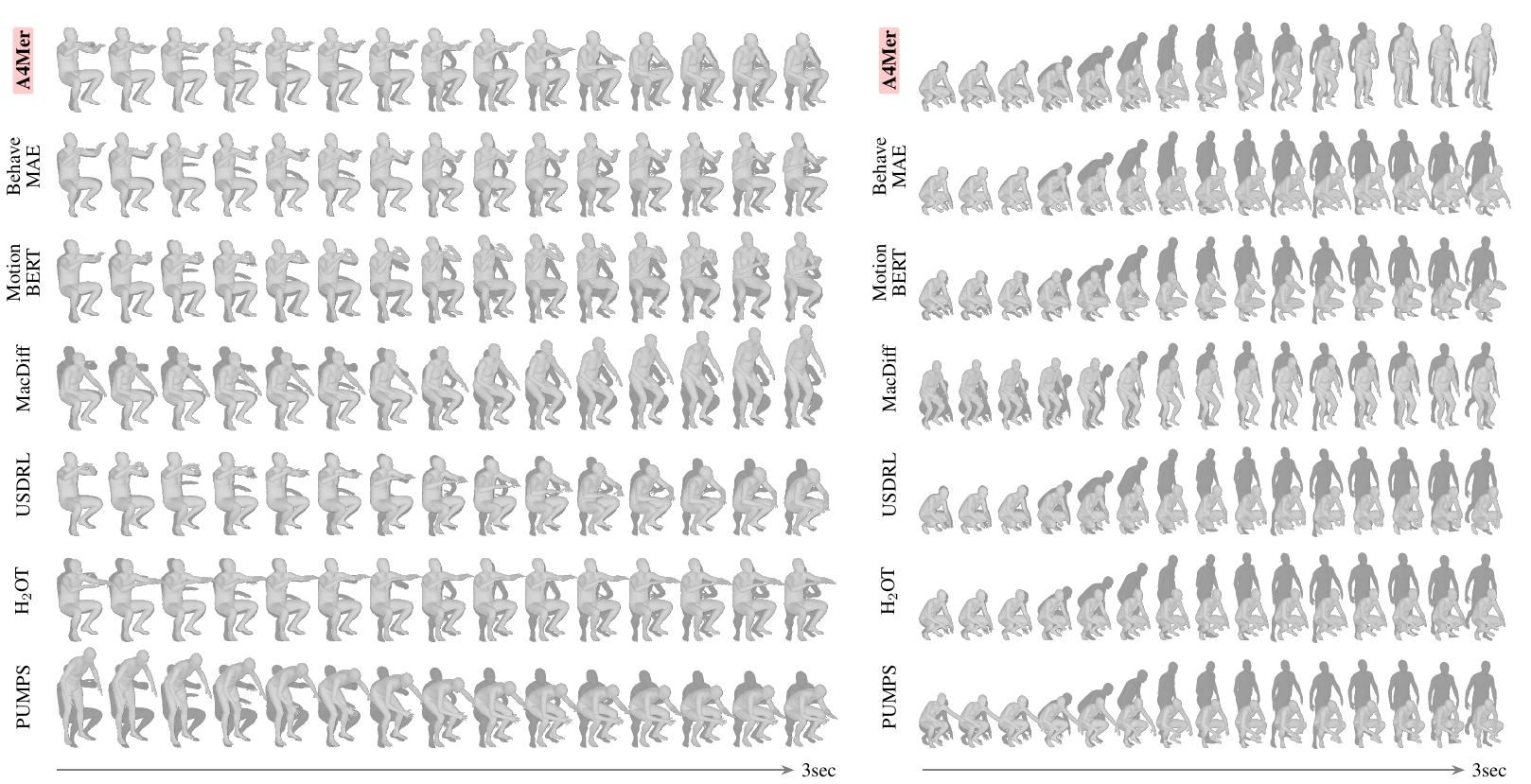}
    \caption{Predicted poses on AMD obtained via auto-regressive latent token prediction and decoding. The dark-colored SMPLs indicate the ground-truth.}
    \label{fig:SuppPrediction}
\end{figure*}

\begin{figure*}[t!]
    \centering
    \includegraphics[width=\linewidth,trim=1pt 3pt 3pt 5pt,clip]{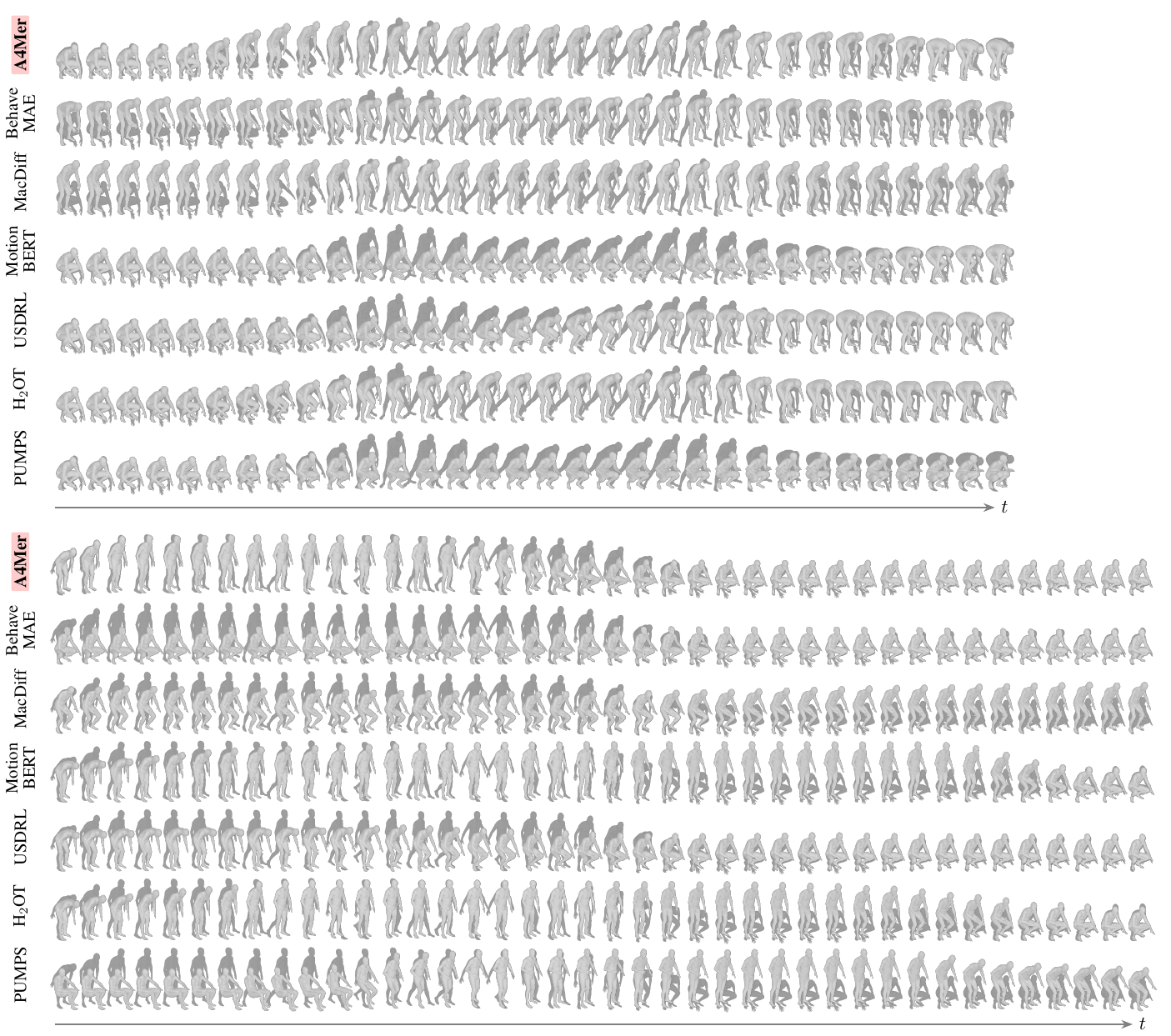}
    \caption{Interpolated poses through latent token interpolation and decoding. The SMPLs in dark color are the ground-truth.}
    \label{fig:SuppInterp}
\end{figure*}

We show additional qualitative results of motion prediction in \cref{fig:SuppPrediction} and motion interpolation in \cref{fig:SuppInterp}.

\section{Details of AMD}
\label{sec:DetailsOfAMD}
Here, we describe the details of the SMPL annotation procedure for AMD as well as the contents of the dataset.

\begin{figure}[t]
    \centering
    \includegraphics[width=0.7\linewidth]{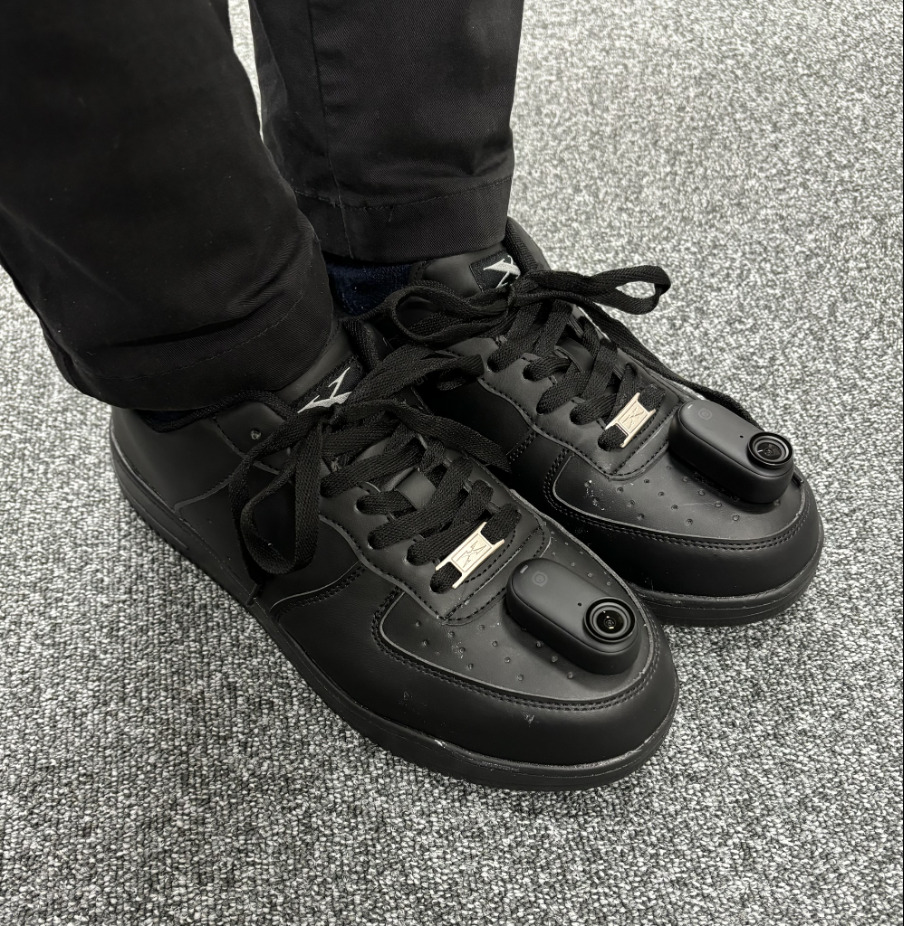}
    \caption{Foot cameras attached to subjects' feet. These cameras do not restrict the subjects’ movements and are small enough to remain invisible to the room-mounted cameras.}
    \label{fig:Go3}
\end{figure}

\subsection{Shape Fitting}
Before recording daily activities, each subject performs an A-pose, which is captured by four RGB-D cameras (RealSense L515) separate from the room RGB cameras. We fit a SMPL model to the obtained point cloud \cite{MPI1, MPI2} and use the optimized shape parameters $\shape$ as ground truth.

\subsection{Foot Camera Localization}
To robustly annotate SMPL parameters despite occlusions around the legs, which frequently occur in natural indoor environments, we attach tiny cameras (Insta360 GO3) to the feet (\cref{fig:Go3}) and place ChArUco markers \cite{ChArUco} on the ceiling and the undersides of tables.
To localize the foot cameras with the Perspective-n-Points (PnP) algorithm, we need to know the 3D positions of the markers in the global coordinate system.
Before recording subjects, we first calibrated the extrinsic parameters of the room-mounted cameras (`room cameras') using Structure-from-Motion on videos capturing a marker rig moving throughout the room.
During this calibration, the ceiling markers observed from the room cameras were also accurately localized.
To localize additional markers---including those placed on the underside of tables, which are not observed by the room cameras---we captured supplementary images from diverse viewpoints targeting these markers. The camera poses for these newly captured images were estimated via PnP using the already-localized markers as reference points. We then perform bundle adjustment to localize the previously unobserved markers, while ensuring consistency with the already localized ones.
Through this two-stage marker localization, the 3D positions of all markers in the environment were obtained.

During recording subjects, we estimated the pose of each foot camera in every frame by applying the PnP algorithm to the localized markers detected from the foot camera.
Note that the room cameras and the foot cameras were time-synchronized using QR codes displayed at the beginning of each session, in the same manner as Ego-Exo4D~\cite{EgoExo4D}.

\subsection{Postprocessing of Foot Camera Pose}
\begin{figure*}[t]
    \centering
    \includegraphics[width=\linewidth]{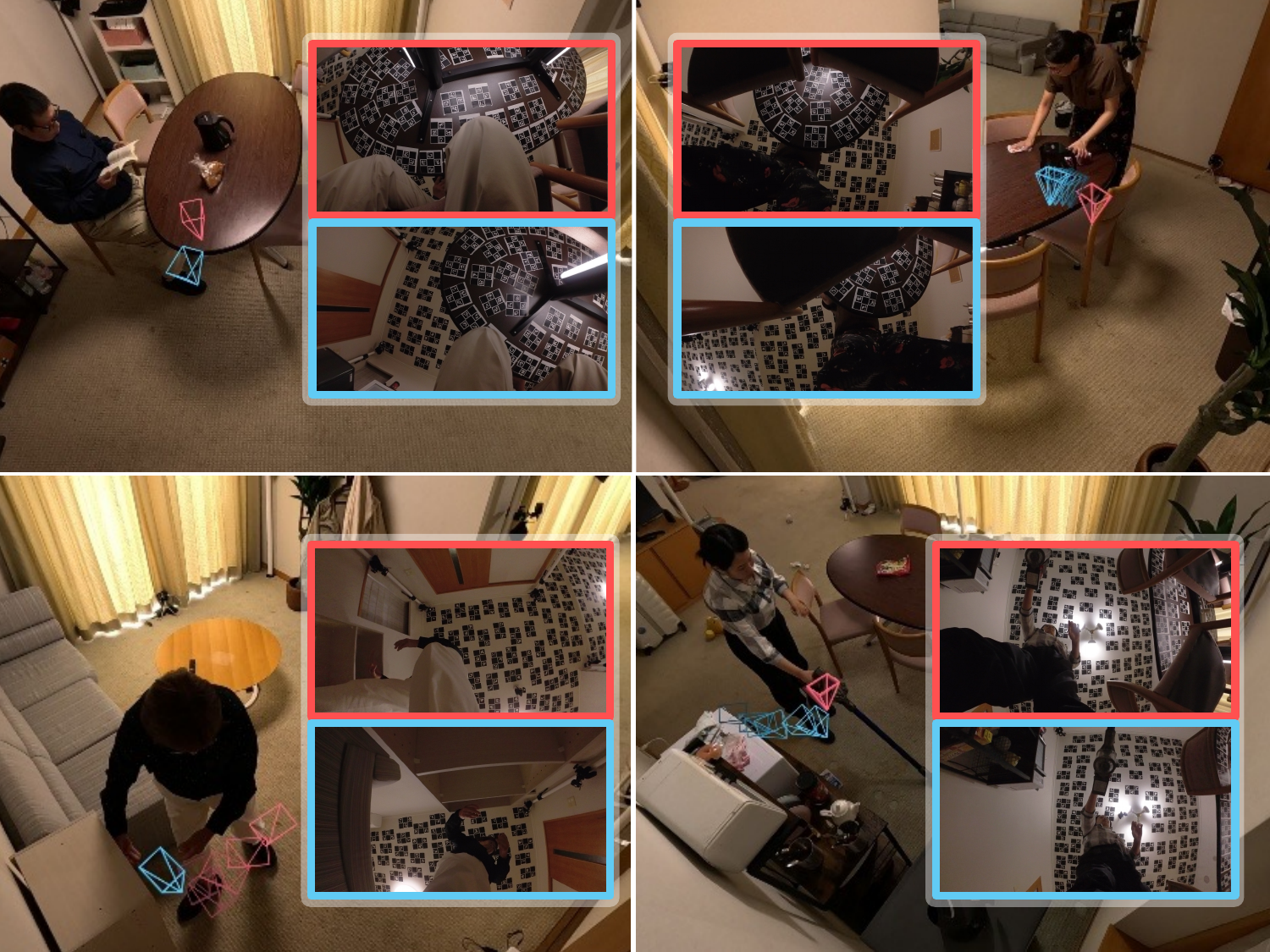}
    \caption{Localized foot camera poses and the images captured from these cameras. The wide field of view of the foot cameras enables capture of many markers and accurate pose localization.}
    \label{fig:FootCameraPose}
\end{figure*}

When subjects perform rapid or large movements, the images from the foot cameras might suffer from momentary motion blur, causing marker detection to fail. Besides, the PnP algorithm might occasionally produce inaccurate pose estimates. To reduce the number of frames in which the camera pose cannot be localized and to obtain temporally consistent trajectories, we apply the following postprocessing procedure.
First, we compare the raw PnP estimates with predictions smoothed by the Rauch–Tung–Striebel (RTS) smoother \cite{RTSSmoother} and identify frames with large discrepancies as outliers, which are subsequently discarded.
Next, we apply an exponential moving average to the foot camera rotations and translations to further stabilize the trajectories.
Finally, we interpolate the poses of unlocalized frames using Kochanek-Bartels splines \cite{KochanekBartels}.
The resulting foot camera poses are illustrated in \cref{fig:FootCameraPose}.

\subsection{Fitting Objective Derivation}
To achieve accurate SMPL fitting by leveraging the foot camera poses, we augment Eq.~(7) in the main text with additional constraint terms.
To prevent the fitted SMPL model from intersecting the foot cameras, we add a loss term $\underfootl$ to ensure that the SMPL foot vertices $V_\foot$ remain below the foot camera plane $P_\cam$ formed by the width and height axes of the camera coordinate system:
\begin{equation}
    \underfootl = \sum_{\vertex \in V_\foot} \text{max}\!\left(\vertex \cdot \normal_\cam + d_\cam, 0\right)\,,
\end{equation}
where $\normal_\cam$ and $d_\cam$ denote the normal vector and the constant of $P_\cam$, respectively.
To fully leverage the foot camera information, we also incorporate $\footdistl$ and $\footanglel$ into Eq.~(7).
$\footdistl$ constrains the SMPL foot dorsum vertices $V_\dorsum$ to lie close to the foot camera plane $P_\cam$.
$\footanglel$ penalizes the SMPL foot bone $\boneSMPL$, the vector from the toe to the heel joint, to be in the same direction as a designated axis $\bonecam$ of the foot camera coordinate frame.
This angular constraint is realized by the physical mounting of the foot camera such that one of its coordinate axes is oriented along the toe-to-heel direction of the foot. These loss terms are formulated as:
\begin{align}
    \footdistl &= \sum_{\vertex \in V_\dorsum} \left|\vertex \cdot \normal_\cam + d_\cam \right| \,, \\
    \footanglel &= 1 - \frac{\boneSMPL \cdot \bonecam}{\left\| \boneSMPL \right\|}\,.
\end{align}

In addition to the aforementioned constraints, we also add loss terms, $\Rrelfootl$ and $\trelfootl$, which are introduced in the main text, to preserve the relative pose between each foot camera and the corresponding SMPL foot.
We define a SMPL-foot coordinate system whose origin $\tSMPL$ is placed at the toe joint. The basis vectors $\RSMPL$ are constructed as follows:
(i) the first axis is obtained by projecting $\boneSMPL$ onto the plane of the foot dorsum computed from $V_\dorsum$ with cross product and normalizing it; (ii) the second axis is the normal vector of this plane; and (iii) the third axis is defined to be orthogonal to the preceding two axes.
From the fitted result of the A-pose videos, we derive a rigid transformation $(\RcamSMPL\,, \tcamSMPL)$ that maps the foot camera pose $(\Rcam\,, \tcam)$ to the corresponding SMPL foot pose $(\RSMPL\,,\tSMPL)$.
$\Rrelfootl$ and $\trelfootl$ enforce that the relative rotation between $\Rcam$ and $\RSMPL$, and the relative translation between $\tcam$ and $\tSMPL$, remain consistent with these transformations:
\begin{align}
    &\Rrelfootl = \left\| \bm{I} - \diag{\{\left(\Rcam \RcamSMPL\right)^\top \RSMPL\}} \right\|\,, \\
    &\trelfootl = \left\| \tSMPL - \left(\Rcam\tcamSMPL + \tcam\right) \right\|\,,
\end{align}
where $\bm{I} \in \real^{3 \times 3}$ denotes the identity matrix.
Combining these terms with Eq.~(7), the full optimization objective is given by
\begin{align}
    \label{eq:SMPLFitting}
    &\smplloss(\pose) = \jointw \jointl + \regw \regl + \footw \footl \,, \\
    &\footl = \underfootw \underfootl + \footdistw \footdistl + \footanglew \footanglel + \Rrelfootw \Rrelfootl + \trelfootw \trelfootl \,,
\end{align}
where each $\lambda$ denotes the weight for the corresponding loss term.

After the SMPL fitting with \cref{eq:SMPLFitting}, we further improve temporal consistency by imposing a smoothness constraint on the SMPL joints $\bm{J}$ and the pose parameters $\pose$ through the velocity loss $\loss_{\mathrm{vel}}$ and acceleration loss $\loss_{\mathrm{acc}}$.
We dynamically modulate the weights of these loss terms based on the average 2D joint confidence $c$ estimated by ViTPose \cite{ViTPose} across all views, assigning larger weights to joints with lower confidence.
Furthermore, for frames in which the foot camera positions remain nearly constant, \ie, the foot is effectively stationary, we assign larger weights to the foot joints of these two terms.
This second-stage SMPL fitting minimizes the following objective function:
\begin{align}
    &\smplloss^\mathrm{2nd}(\pose) = \sum_{t} \smplloss^t + \velw \vell^t + \accw \accl^t \,, \\
    &\vell^t\!=\!\sum_j\!\Lambda_j  \Big(\!\left\| \bm{J}^t_j - \bm{J}^{t-1}_j \right\| + \bm{\delta}^t_j \Big) \,, \\
    &\accl^t\!=\!\sum_j\!\Lambda_j \Big(\!\left\| \bm{J}^t_j\!-2\bm{J}^{t-1}_j\!+\!\bm{J}^{t-2}_j \right\|\!+\bm{\delta}^{t}_j - \bm{\delta}^{t-1}_j\Big) \,,\\
    &\bm{\delta}^t_j = d_\mathrm{geo}\left( \pose^t_j, \pose^{t+1}_j \right) \,, \\
    &\Lambda_j = \frac{1}{c_j} + \lambda_s \indicator\left\{ \left\| \tcam^t - \tcam^{t-1} \right\| < \epsilon \right\}
\end{align}
where $j$ and $t$ denote the joint and time indices, respectively, $d_\mathrm{geo}(\cdot,\cdot)$ is the geodesic distance, and $\indicator\{\cdot\}$ is the indicator function.
This optimization is performed every 100 frames.

\subsection{Evaluation of the Effectiveness of the Foot Camera Constraint Using 2D Masks}
To quantitatively evaluate the effectiveness of the foot camera constraint, we compute the IoU between the human masks predicted by OneFormer \cite{OneFormer}, which we regard as a pseudo ground truth mask, and the mask obtained by rendering the fitted SMPL mesh from the corresponding camera viewpoint, as described in Sec.~4.2 of the main text.

For a more reliable evaluation, we further apply two preprocessing steps to the pseudo ground truth mask: (i) removing regions that extend beyond the body surface (e.g., clothing and hair), and (ii) interpolating missing regions in the mask caused by occlusions.

\paragraph{Removal of Regions Extending Beyond the Body Surface.}
The human masks predicted by OneFormer include regions corresponding to clothing and hair, which often extend beyond the actual body surface. As a result, these masks tend to be larger than the masks obtained from the SMPL model, which represents only the body surface. To remove such extraneous regions from the pseudo ground truth mask, we apply a morphological erosion operation.

\paragraph{Interpolation of Missing Regions Caused by Occlusion.}
For the evaluation, we randomly sample 10,000 frames in which the full body is visible. This is determined based on the confidence scores of the body joints predicted by ViTPose \cite{ViTPose}. However, even with this sampling, some frames still contain partial occlusions by furniture or other objects, resulting in unobserved, missing regions in the human mask.

To mitigate this issue, we render the SMPL meshes obtained from the fittings with and without the foot camera constraint.
We then treat the overlapping regions of the two rendered masks as a reliable human region, since this is consistently supported by both fitting configurations. This region is added to the pseudo ground truth mask predicted by OneFormer.

This procedure not only restores human regions that were occluded, but also compensates for regions that may have been excessively removed by the aforementioned erosion process.

\subsection{Contents of AMD}
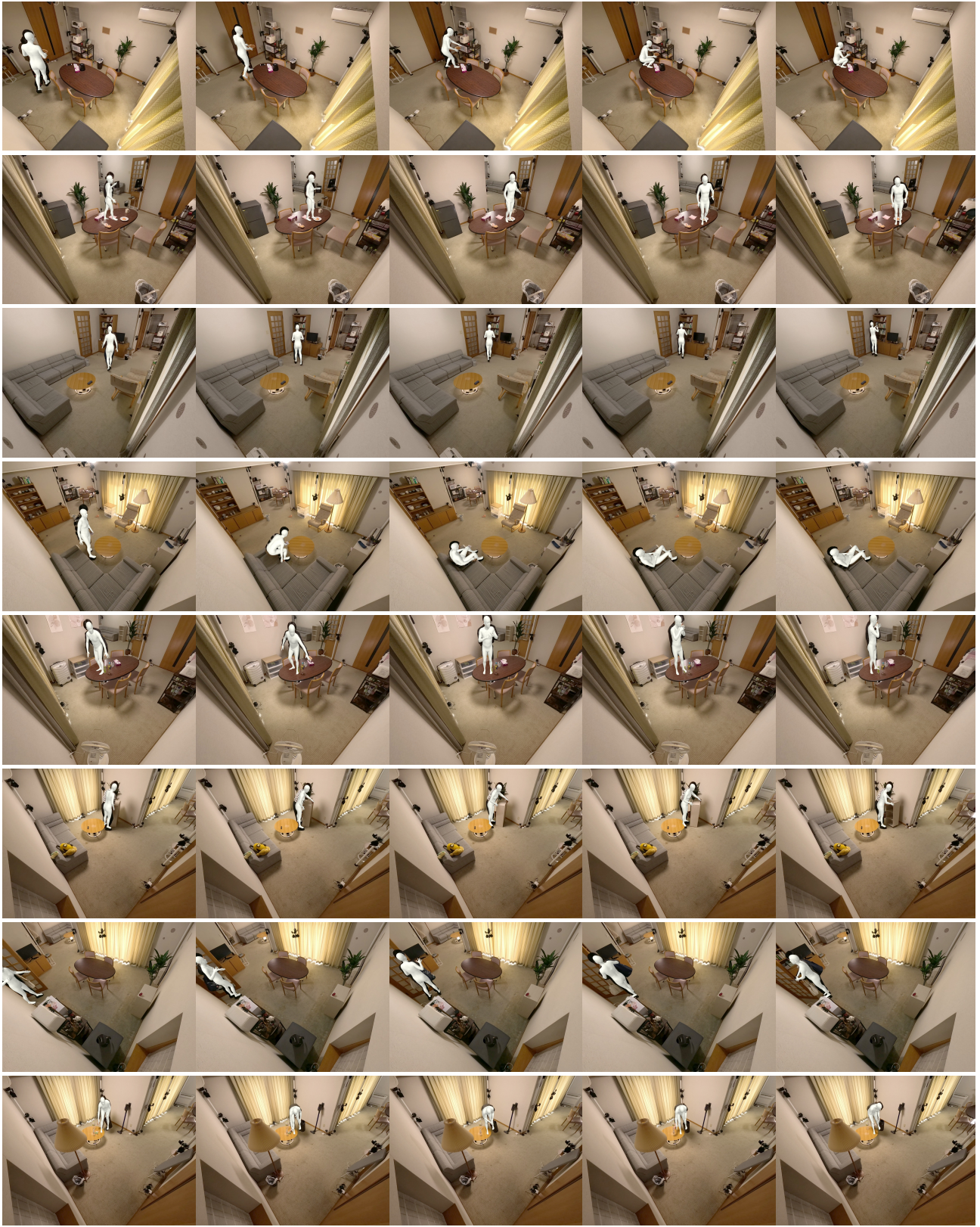
\begin{figure*}[t]
    \centering
    \input{fig/suppl_amd_samples1}
    \vspace{-0.25\baselineskip}
    \caption{Samples of AMD. Our dataset contains videos of people conducting various daily activities with accurate per-frame SMPL annotations.}
    \label{fig:AMDSamples1}
\end{figure*}

\begin{figure*}[t]
    \centering
    \input{fig/suppl_amd_samples2}
    \vspace{-0.25\baselineskip}
    \caption{More samples of AMD.}
    \label{fig:AMDSamples2}
\end{figure*}
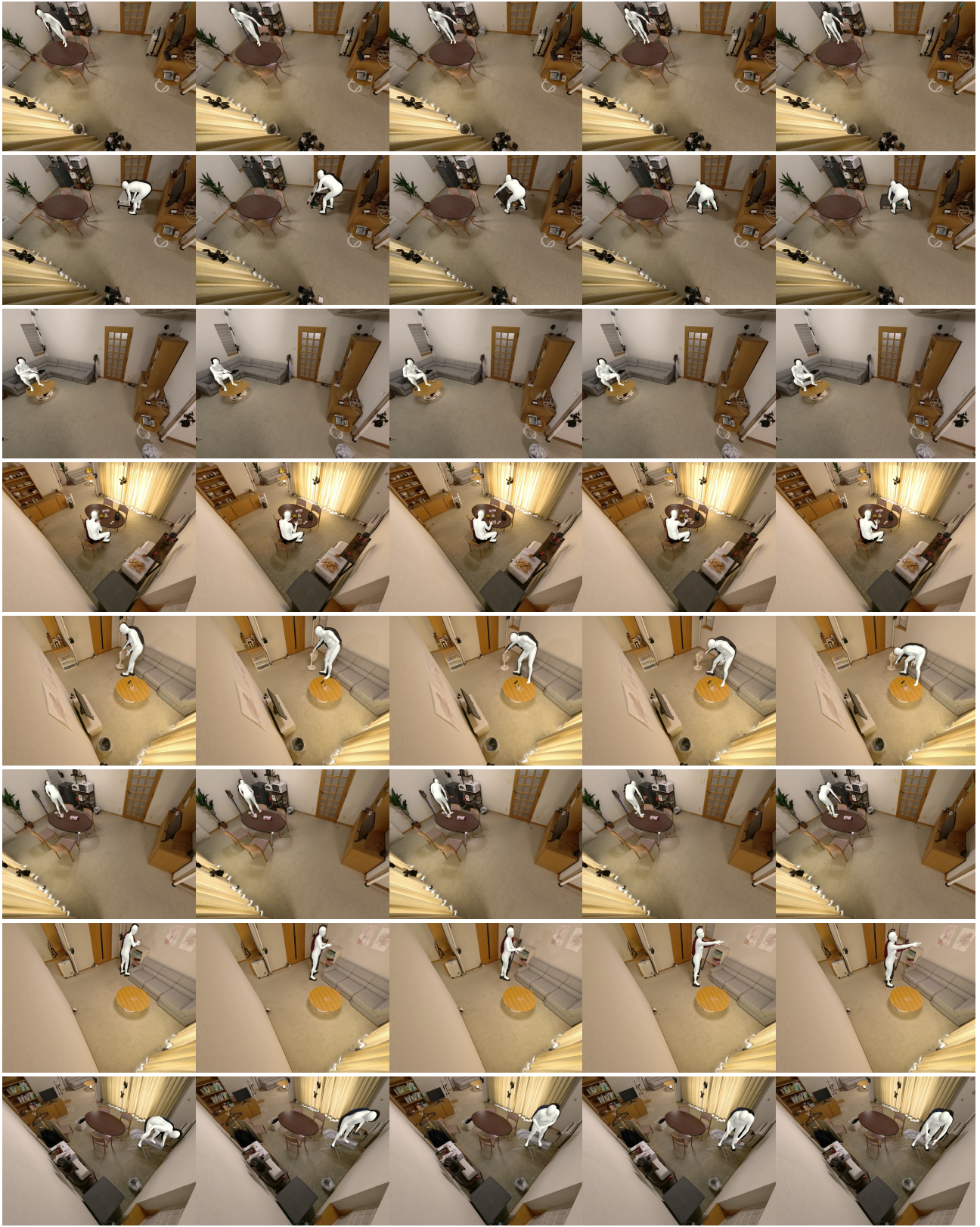

In \cref{tab:AMDActions}, we summarize the diverse daily actions contained in AMD. We show sample sequences of AMD in \cref{fig:AMDSamples1,fig:AMDSamples2}.

\include{action_tables}

{
    \small
    \bibliographystyle{ieeenat_fullname}
    \bibliography{main}
}

\end{document}

%% file: preamble.tex
\usepackage{booktabs}   
\usepackage{multirow}   
\usepackage{comment}
\usepackage[table]{xcolor}
\usepackage{bm}
\usepackage{zebra-goodies}
\usepackage{caption}
\usepackage{subcaption}
\usetikzlibrary{arrows.meta}

\newcommand{\diag}{\mathrm{diag}}
\newcommand{\loss}{\mathcal{L}}
\newcommand{\local}{l}
\newcommand{\glob}{g}
\newcommand{\sg}{\mathrm{sg}}

\newcommand{\na}{--}
\newcommand{\shape}{\bm{\beta}}
\newcommand{\pose}{\bm{\theta}}

\newcommand{\joint}{\bm{J}}
\newcommand{\smplloss}{E}
\newcommand{\jointl}{\smplloss_{\joint}}
\newcommand{\jointw}{\lambda_{\joint}}
\newcommand{\regl}{\smplloss_\mathrm{reg}}
\newcommand{\regw}{\lambda_\mathrm{reg}}

\newcommand{\floorl}{\smplloss_f}
\newcommand{\floorw}{\lambda_f}
\newcommand{\underfootl}{\smplloss_u}
\newcommand{\underfootw}{\lambda_u}
\newcommand{\footdistl}{\smplloss_d}
\newcommand{\footdistw}{\lambda_d}
\newcommand{\footanglel}{\smplloss_a}
\newcommand{\footanglew}{\lambda_a}
\newcommand{\Rrelfootl}{\smplloss_{R}}
\newcommand{\Rrelfootw}{\lambda_{R}}
\newcommand{\trelfootl}{\smplloss_{t}}
\newcommand{\trelfootw}{\lambda_{t}}
\newcommand{\footl}{\smplloss_{\mathrm{foot}}}
\newcommand{\footw}{\lambda_{\mathrm{foot}}}
\newcommand{\vell}{\smplloss_{\mathrm{vel}}}
\newcommand{\velw}{\lambda_{\mathrm{vel}}}
\newcommand{\accl}{\smplloss_{\mathrm{acc}}}
\newcommand{\accw}{\lambda_{\mathrm{acc}}}
\newcommand{\indicator}{\mbox{1}\hspace{-0.25em}\mbox{l}}

\newcommand{\SMPL}{\mathrm{smpl}}
\newcommand{\normal}{\bm{n}}
\newcommand{\vertex}{\bm{v}}
\newcommand{\foot}{\mathrm{foot}}
\newcommand{\dorsum}{\mathrm{drsm}}
\newcommand{\cam}{\mathrm{cam}}
\newcommand{\boneSMPL}{\bm{b}}
\newcommand{\bonecam}{\bm{n}_b}
\newcommand{\Rcam}{\bm{R}_\cam}
\newcommand{\tcam}{\bm{t}_\cam}
\newcommand{\RSMPL}{\bm{R}_\SMPL}
\newcommand{\tSMPL}{\bm{t}_\SMPL}
\newcommand{\RcamSMPL}{\bm{R}^\cam_\SMPL}
\newcommand{\tcamSMPL}{\bm{t}^\cam_\SMPL}
\newcommand{\real}{\mathbb{R}}

\newcommand{\HTwoOT}{H$_2$OT}

%% file: fig/suppl_amd_samples1.tex
\begin{tikzpicture}[x=1mm,y=1mm,every node/.style={inner sep=0pt, text depth=0pt}]
    \newcount\colwidth \colwidth=34  
    \newcount\rowheight \rowheight=-27 

    \newcount\vx \vx=0
    \newcount\vy \vy=0

    \newcommand\imagewidth{0.2\textwidth} 

    \def\RootDir{fig/camera_ready/supp/dataset_examples}
    \def\SceneA{use_microwave/2024-09/17/batch1/a-1/gopro11/8}
    \def\SceneB{make_coffee/2024-09/17/batch1/b-1/gopro11/11}
    \def\SceneC{return_book/2024-09/17/batch1/c-1/gopro11/3}
    \def\SceneD{read_book/2024-09/23/batch1/labboy3-3/gopro11/1}
    \def\SceneE{blow_nose/2024-09/23/batch1/labboy4-5/gopro11/11}
    \def\SceneF{move_shelf/2024-09/19/batch1/b-3/gopro11/13}
    \def\SceneG{open_door/2024-09/19/batch2/a-2/gopro11/1}
    \def\SceneH{throw_trash/2024-09/19/batch2/b-3/gopro11/13}

    \def\Pa#1{\includegraphics[width=\imagewidth]{#1}}
    \def\Pb#1{\includegraphics[width=\imagewidth]{#1}}
    \def\Pc#1{\includegraphics[width=\imagewidth]{#1}}
    \def\Pd#1{\includegraphics[width=\imagewidth]{#1}}
    \def\Pe#1{\includegraphics[width=\imagewidth]{#1}}
    \def\Pf#1{\includegraphics[width=\imagewidth]{#1}}
    \def\Pg#1{\includegraphics[width=\imagewidth]{#1}}
    \def\Ph#1{\includegraphics[width=\imagewidth]{#1}}

    \vy = 0
    \node[anchor=north] at (\vx, \vy) {\Pa{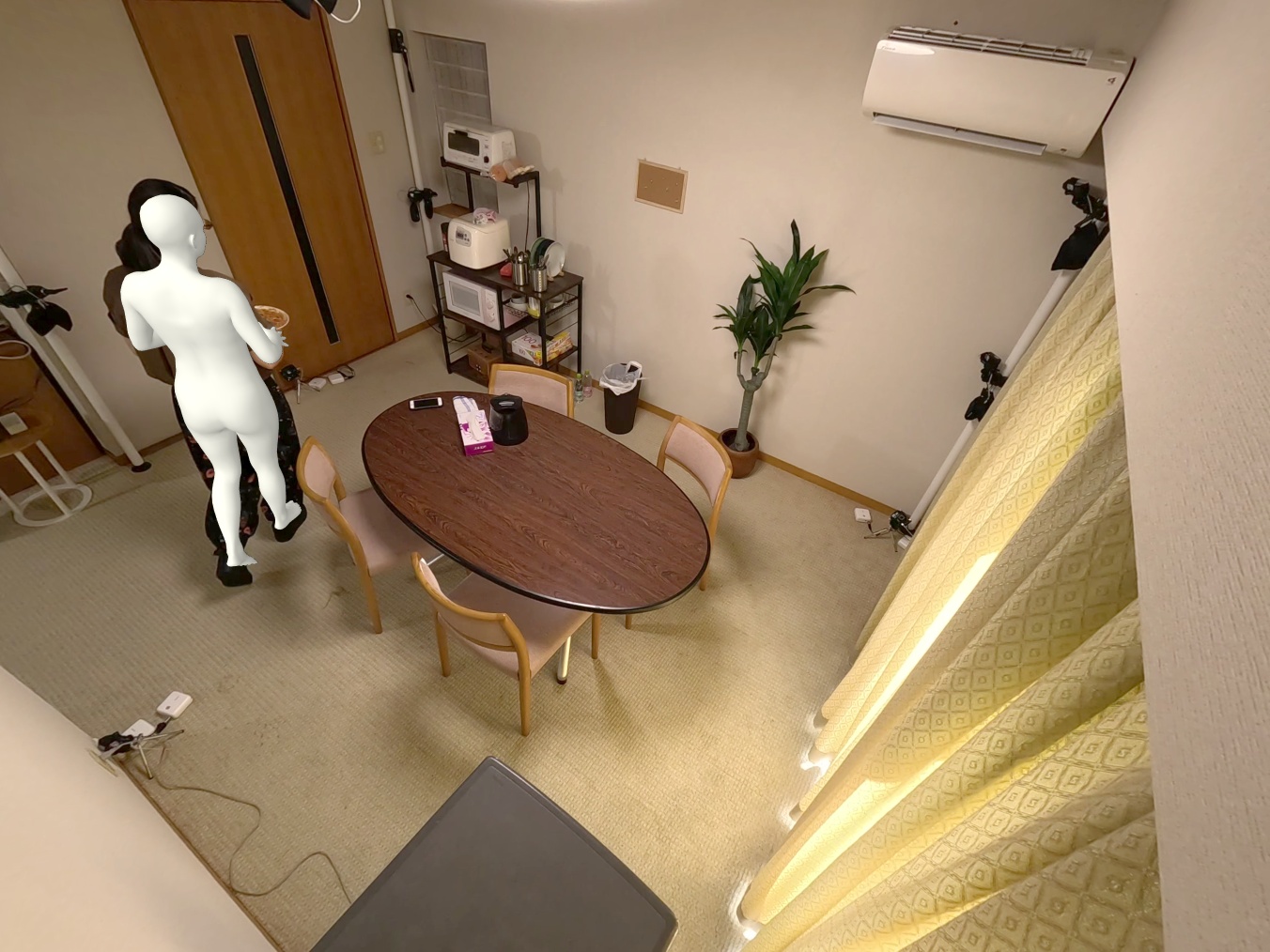}}; \advance \vx by \colwidth
    \node[anchor=north] at (\vx, \vy) {\Pa{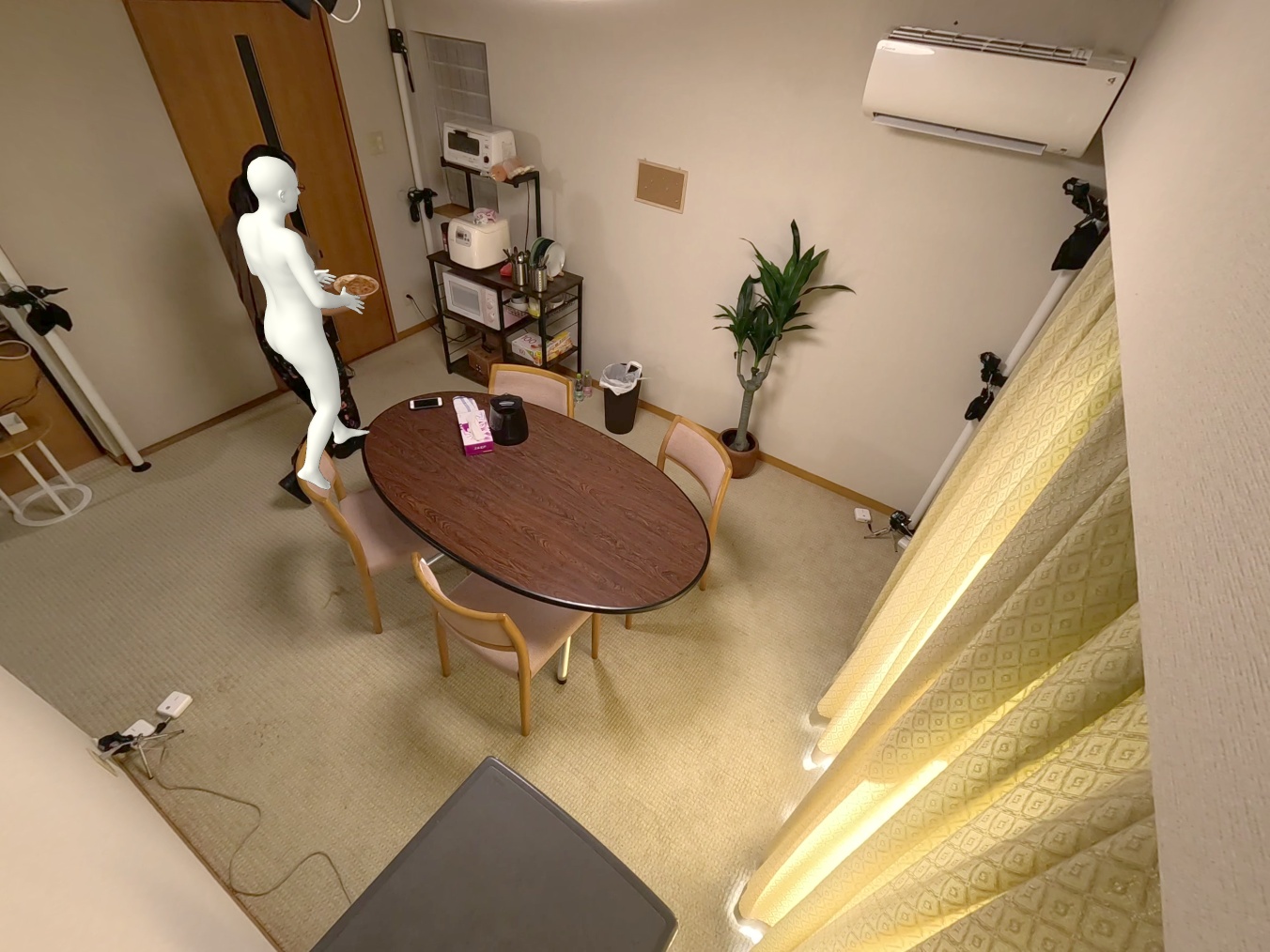}}; \advance \vx by \colwidth
    \node[anchor=north] at (\vx, \vy) {\Pa{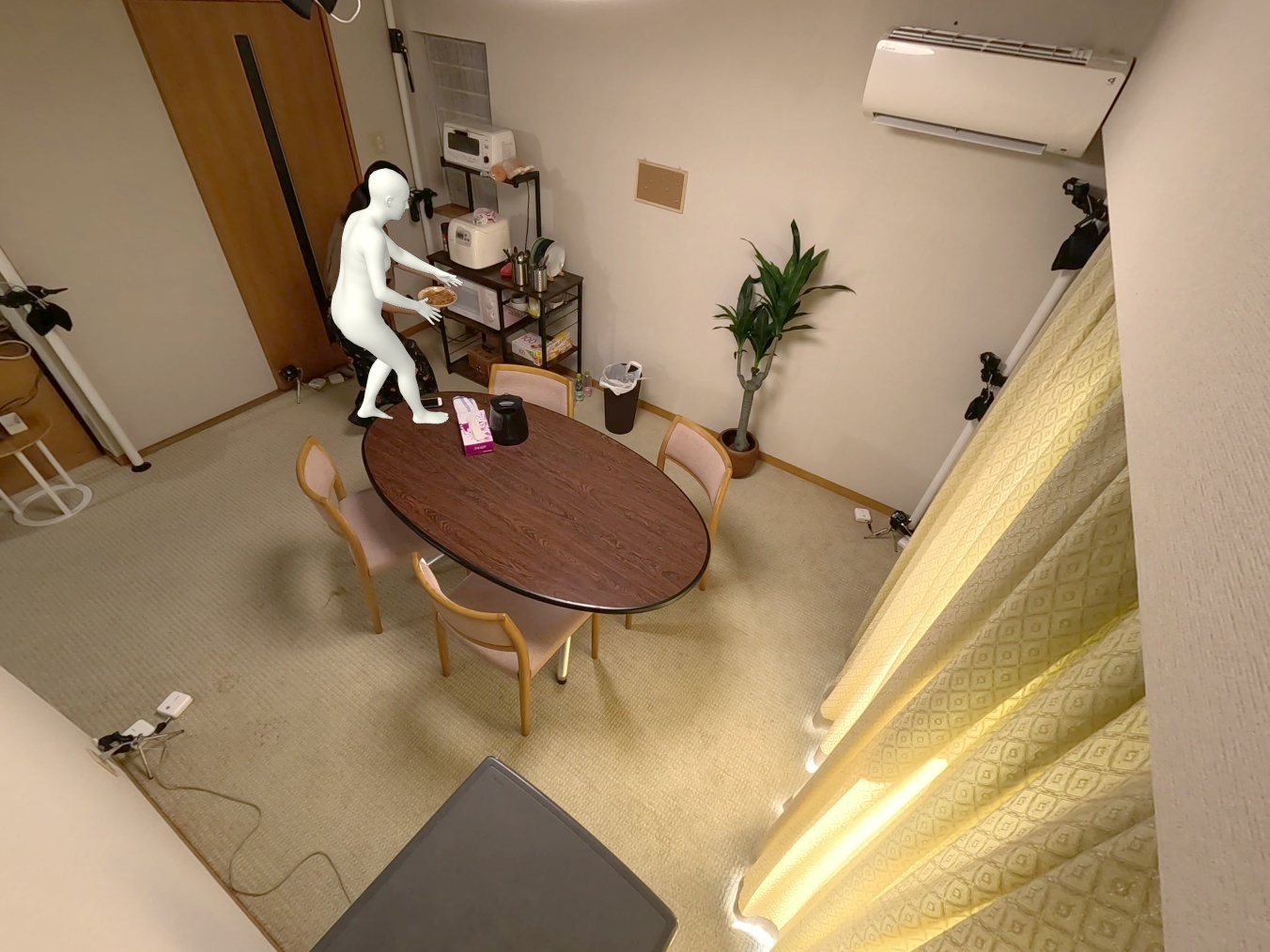}}; \advance \vx by \colwidth
    \node[anchor=north] at (\vx, \vy) {\Pa{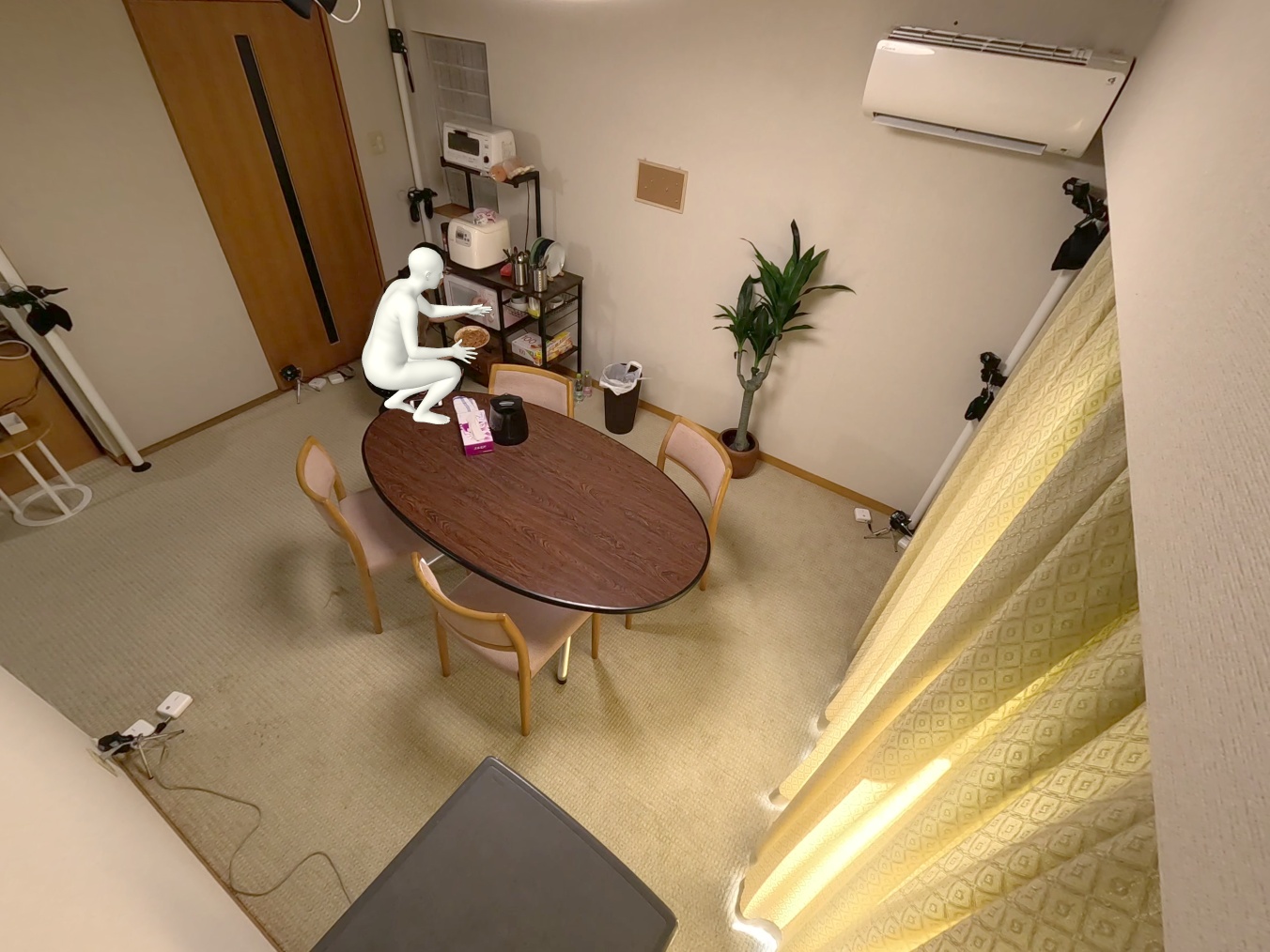}}; \advance \vx by \colwidth
    \node[anchor=north] at (\vx, \vy) {\Pa{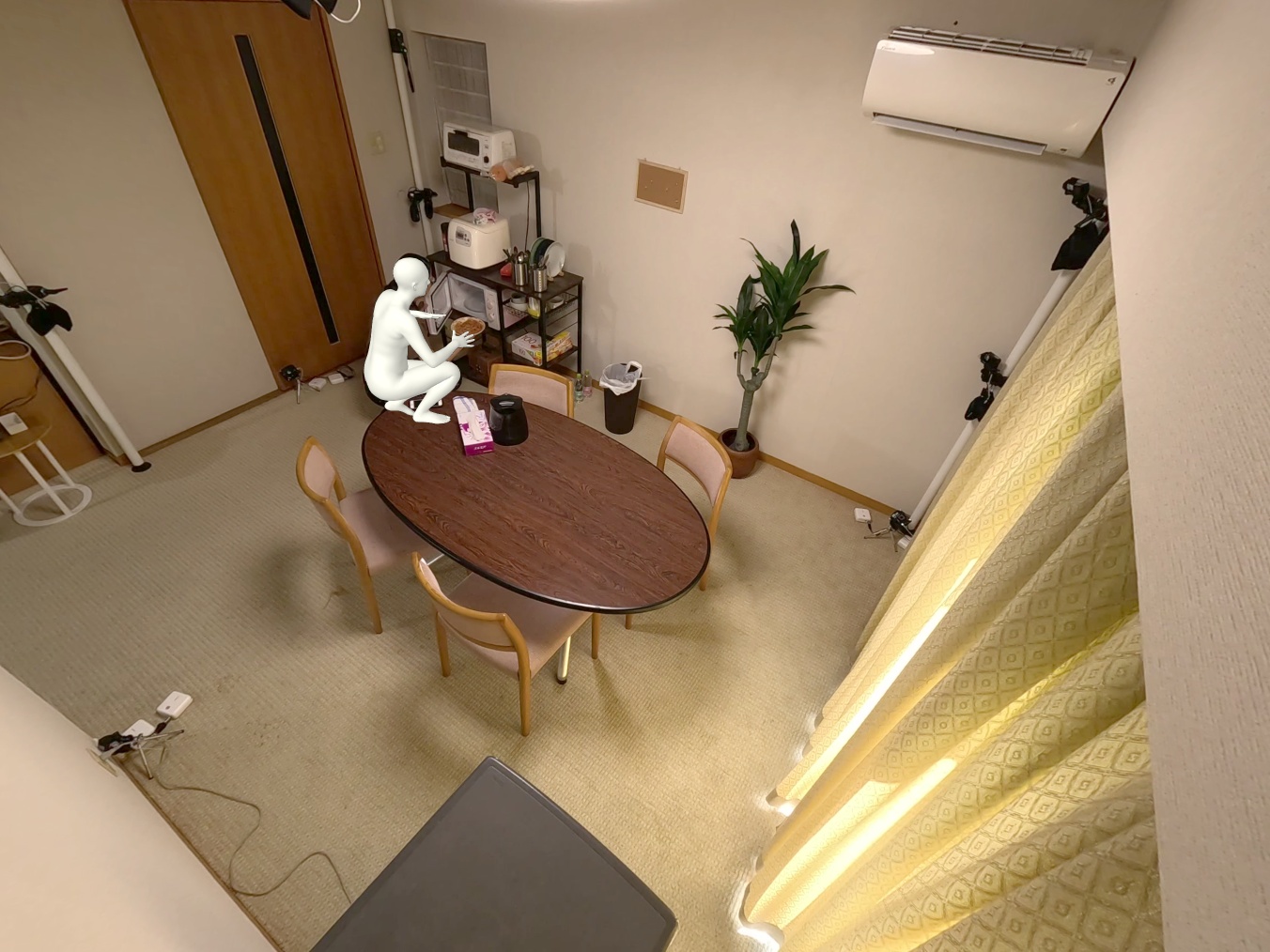}}; \advance \vx by \colwidth

    \vx = 0
    \advance \vy by \rowheight
    \node[anchor=north] at (\vx, \vy) {\Pb{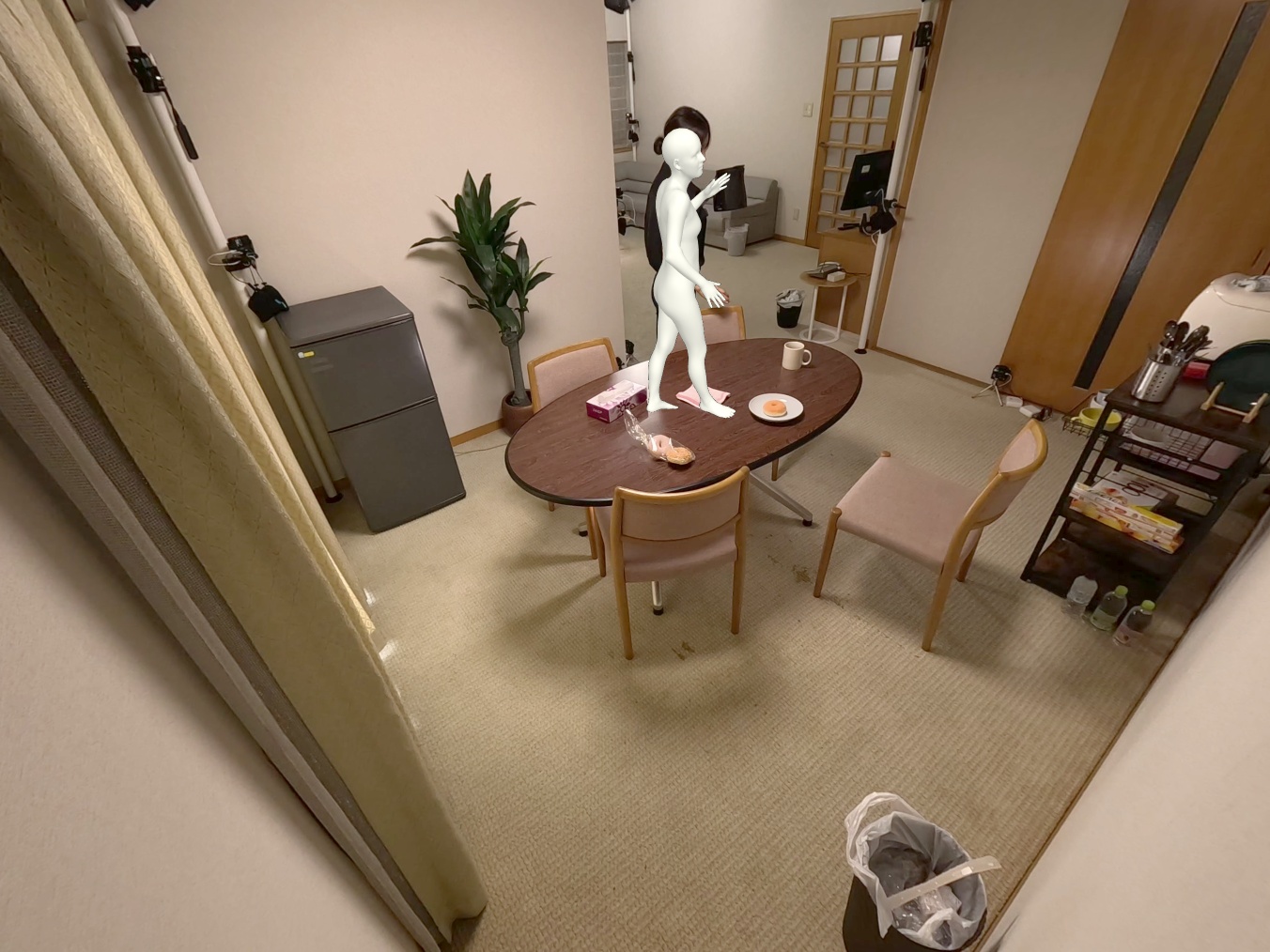}}; \advance \vx by \colwidth
    \node[anchor=north] at (\vx, \vy) {\Pb{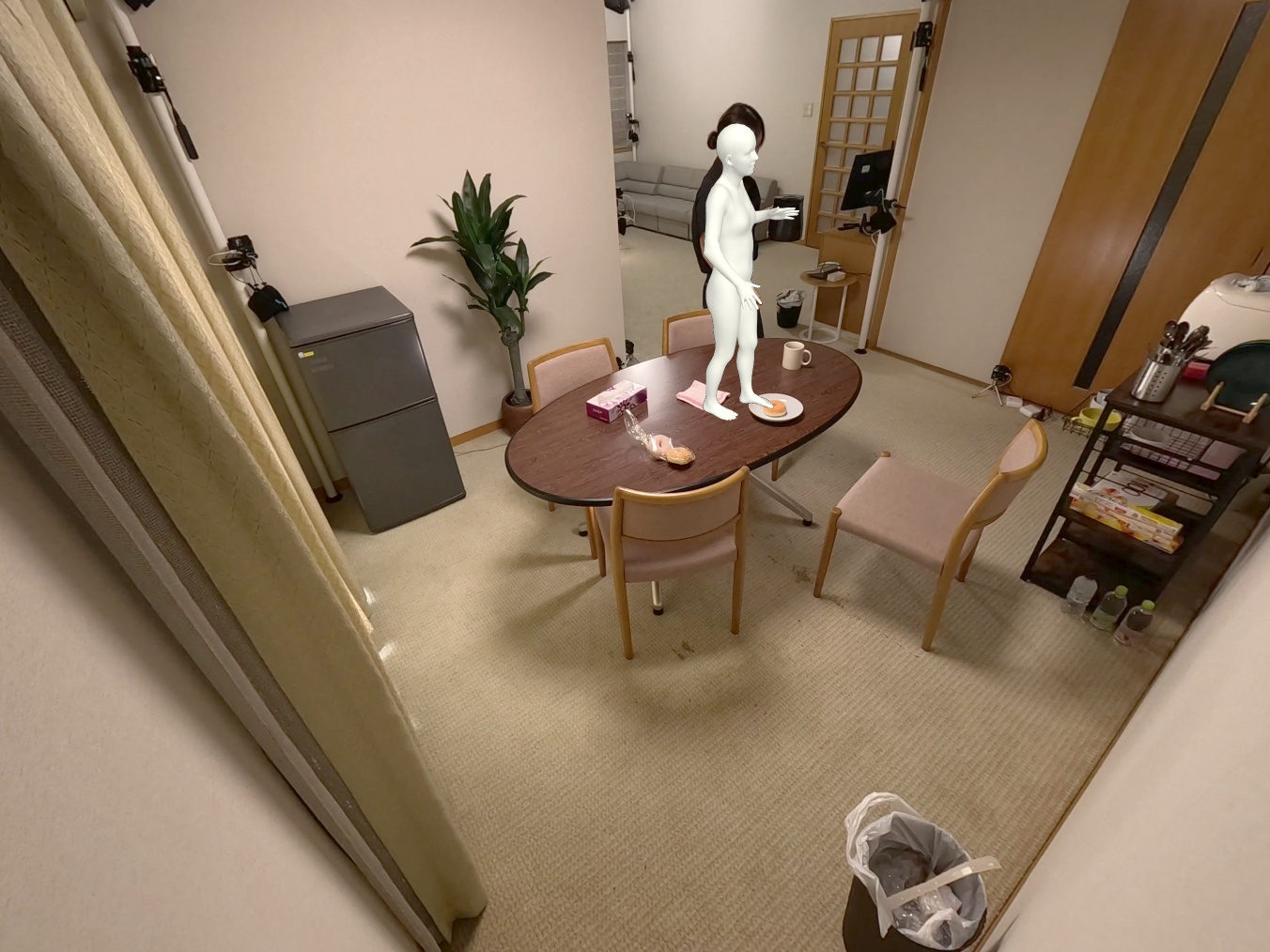}}; \advance \vx by \colwidth
    \node[anchor=north] at (\vx, \vy) {\Pb{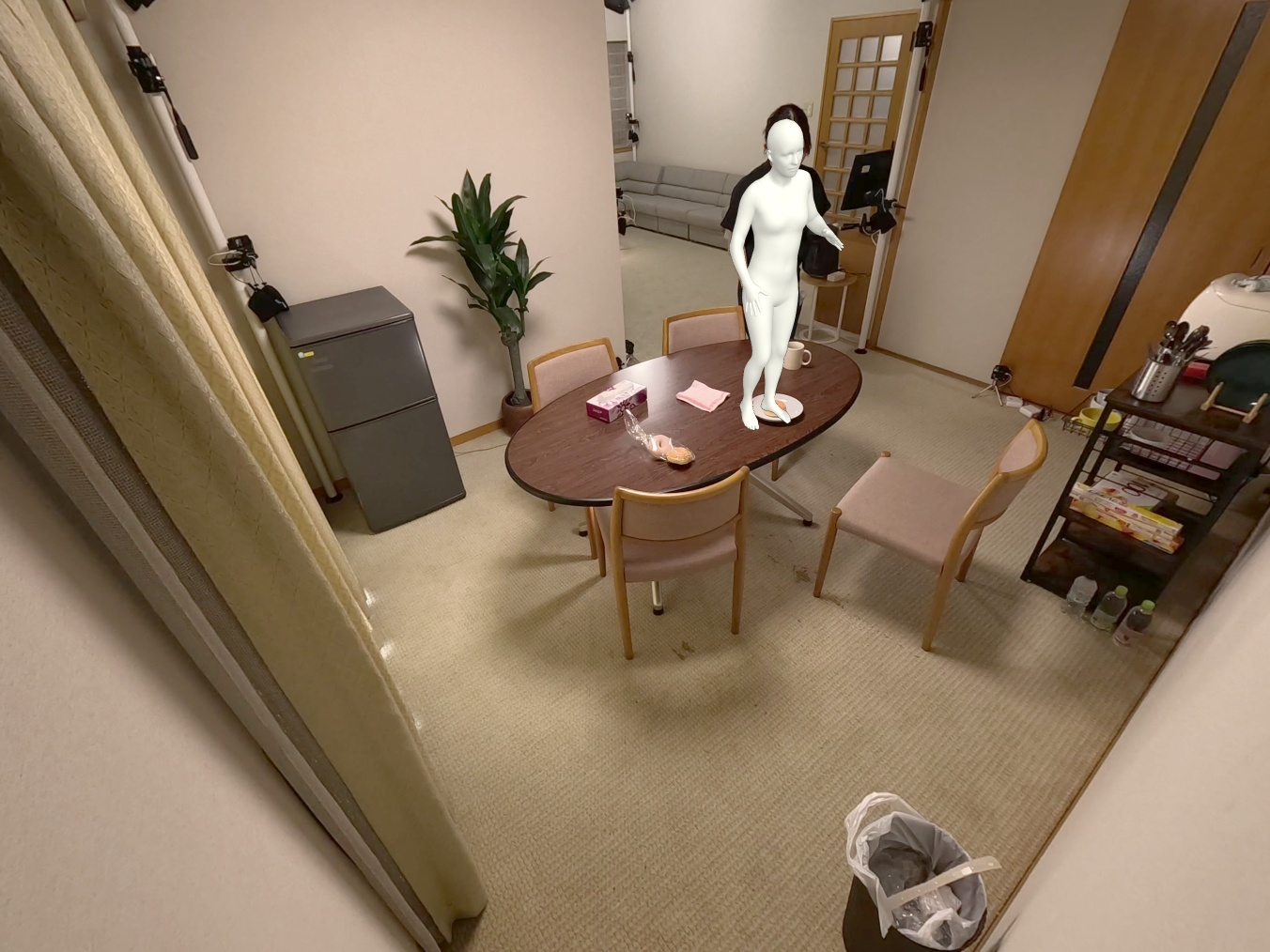}}; \advance \vx by \colwidth
    \node[anchor=north] at (\vx, \vy) {\Pb{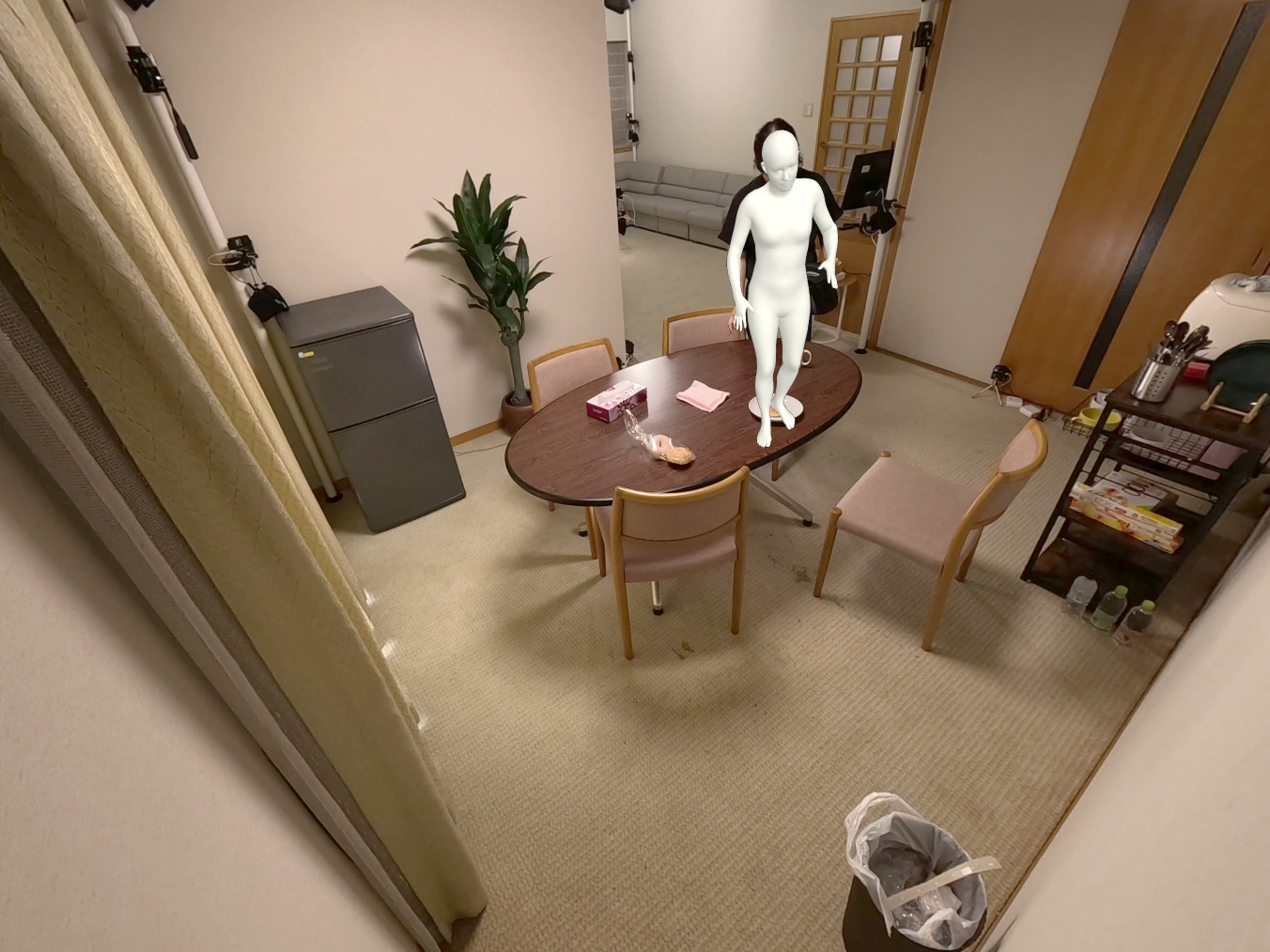}}; \advance \vx by \colwidth
    \node[anchor=north] at (\vx, \vy) {\Pb{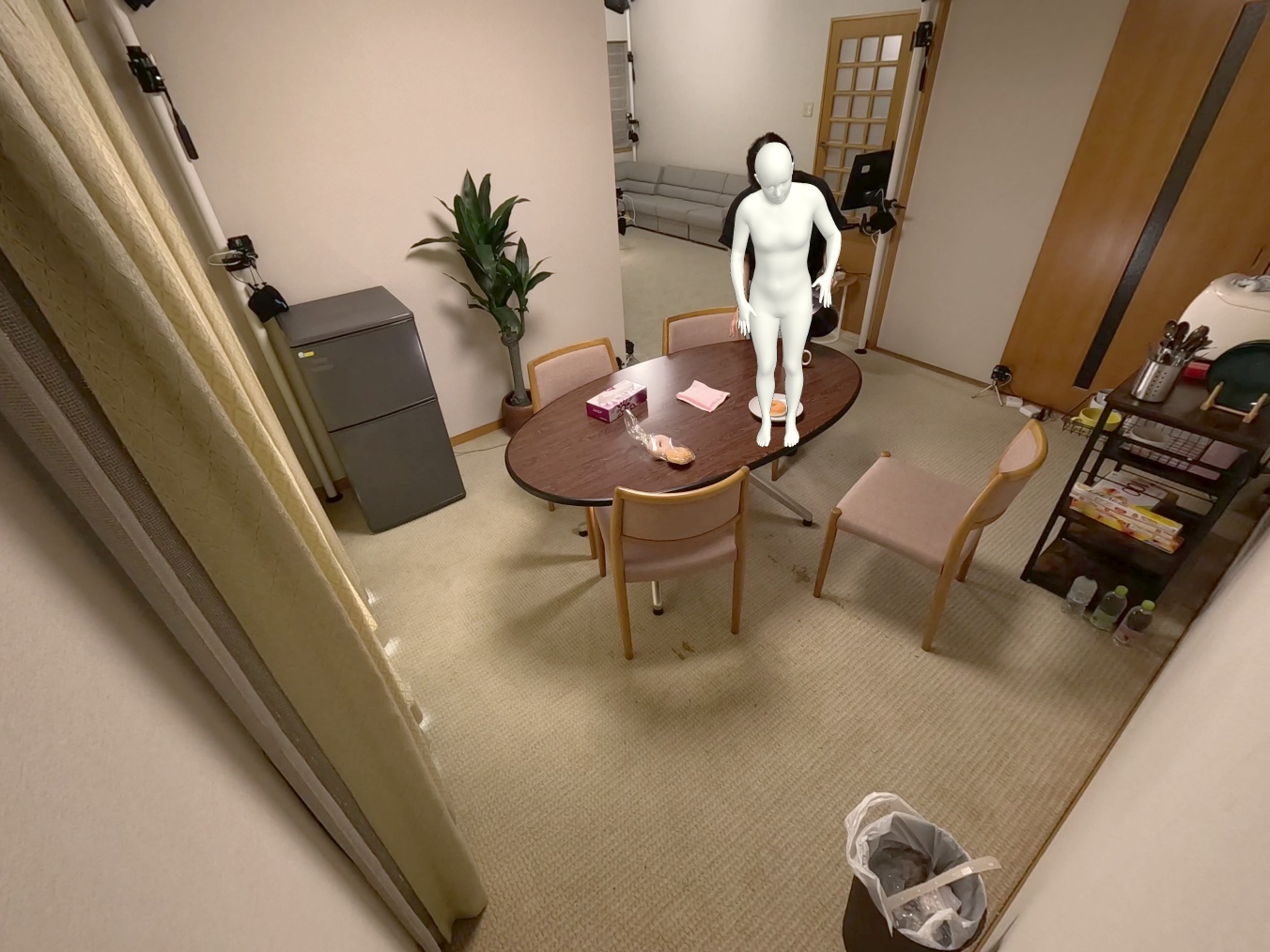}}; \advance \vx by \colwidth

    \vx = 0
    \advance \vy by \rowheight
    \node[anchor=north] at (\vx, \vy) {\Pc{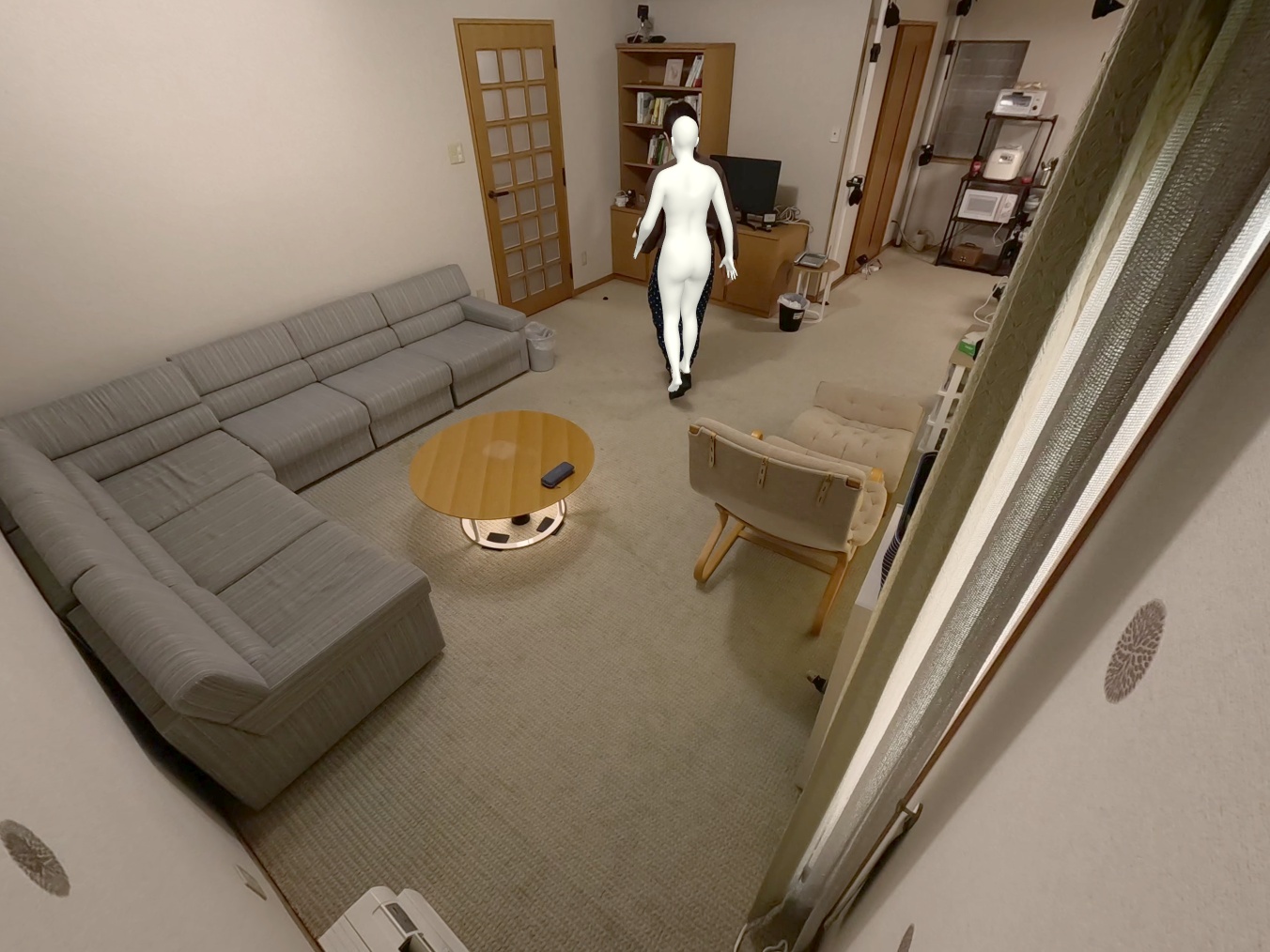}}; \advance \vx by \colwidth
    \node[anchor=north] at (\vx, \vy) {\Pc{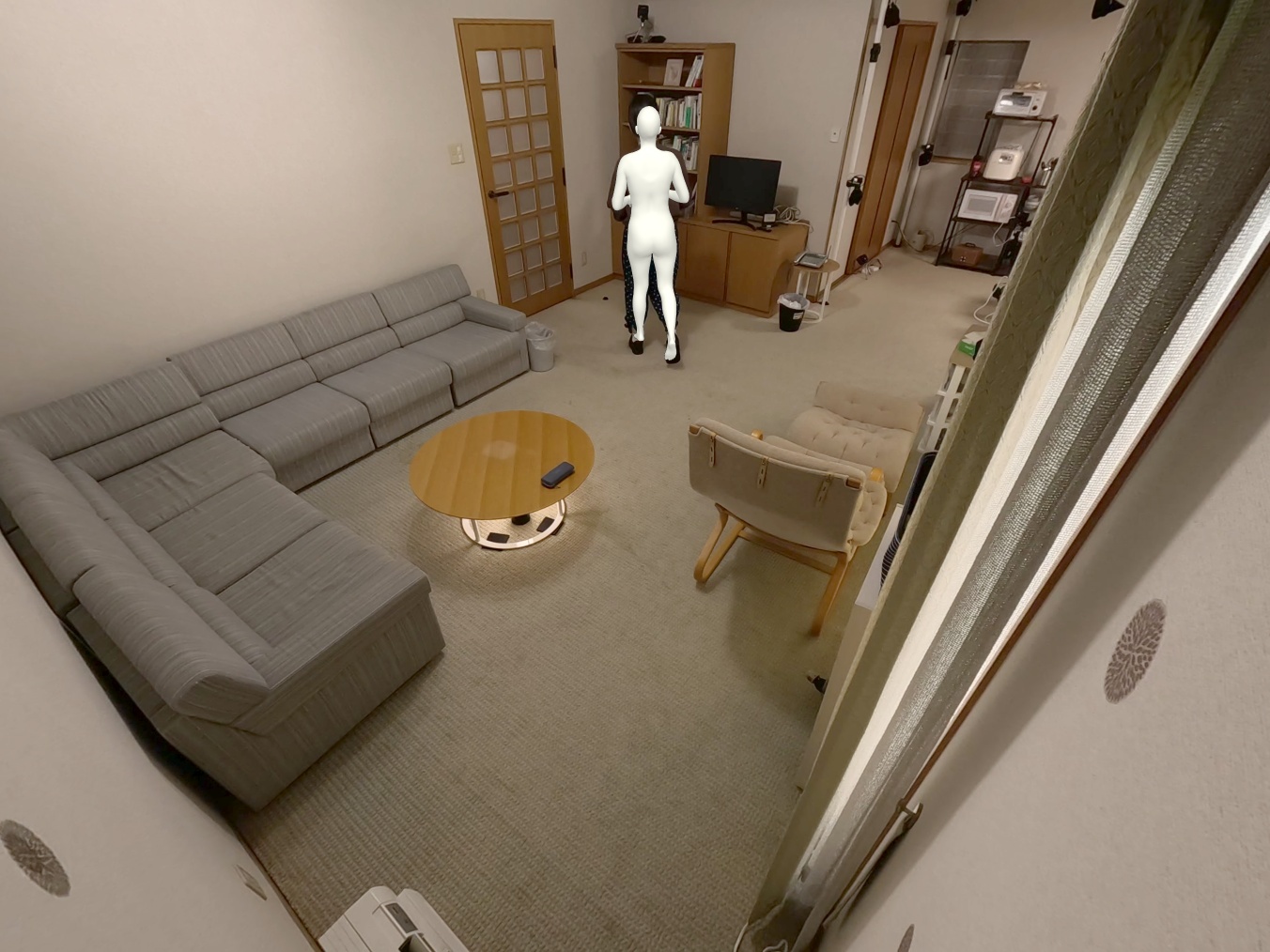}}; \advance \vx by \colwidth
    \node[anchor=north] at (\vx, \vy) {\Pc{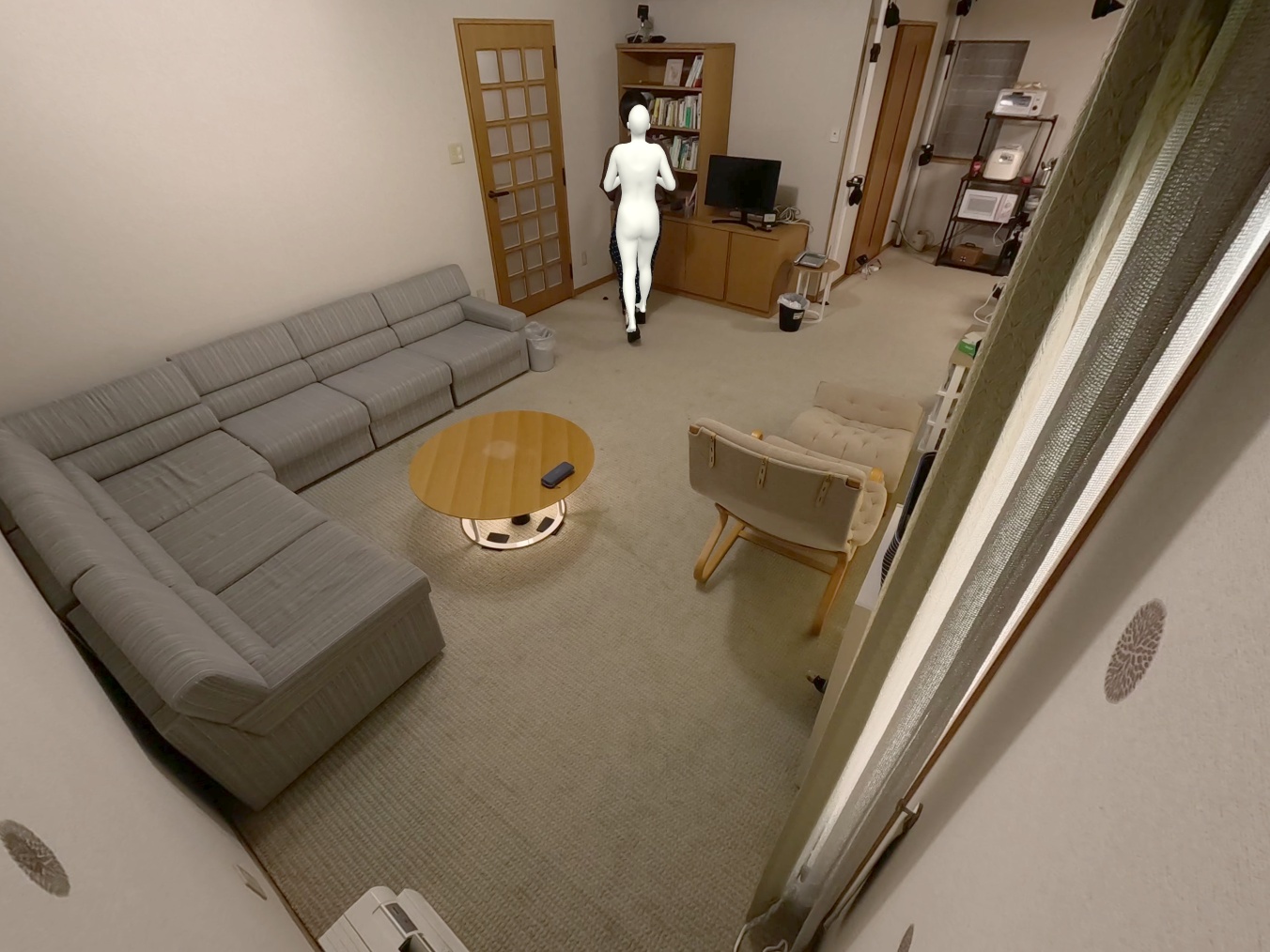}}; \advance \vx by \colwidth
    \node[anchor=north] at (\vx, \vy) {\Pc{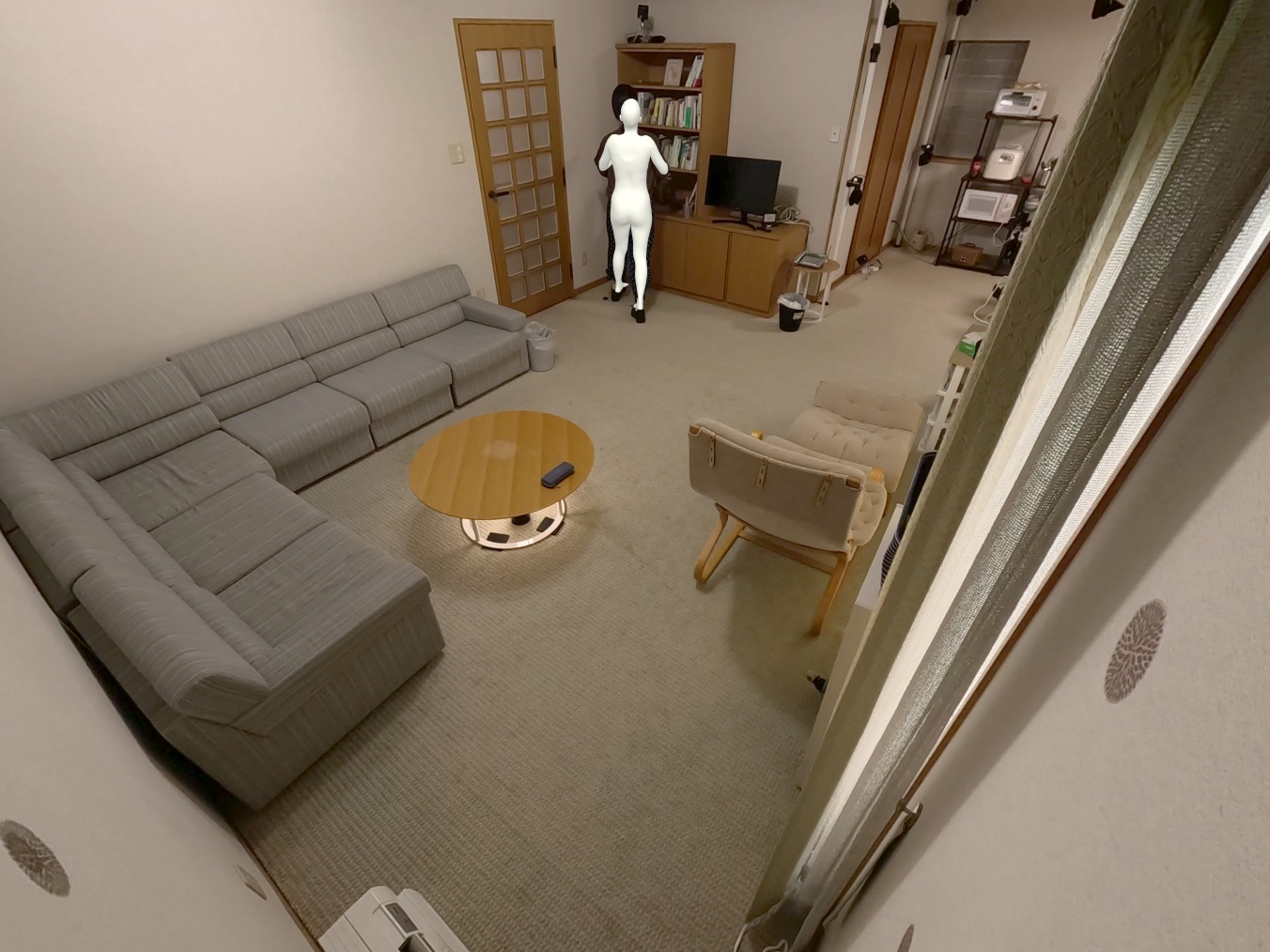}}; \advance \vx by \colwidth
    \node[anchor=north] at (\vx, \vy) {\Pc{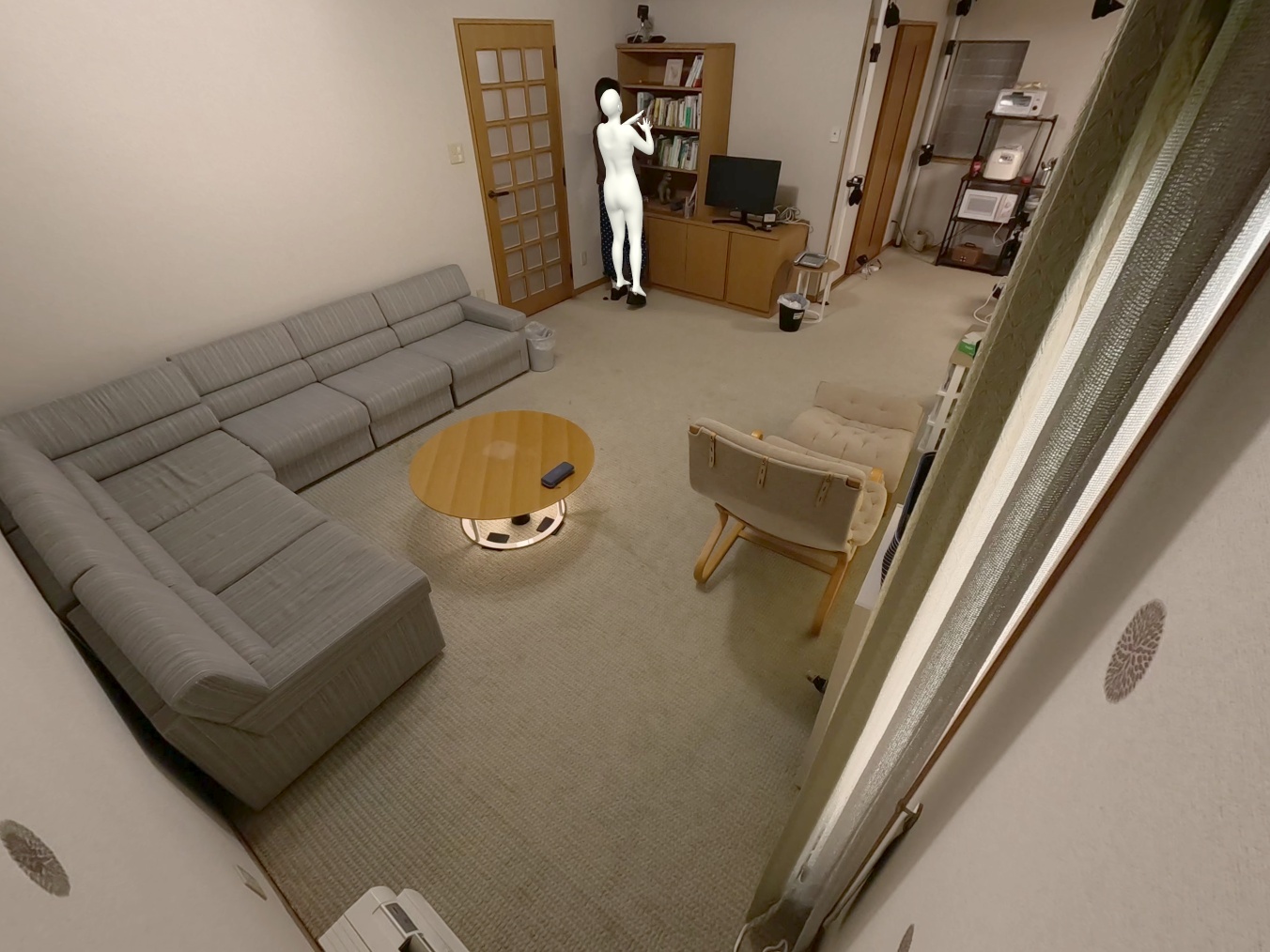}}; \advance \vx by \colwidth

    \vx = 0
    \advance \vy by \rowheight
    \node[anchor=north] at (\vx, \vy) {\Pd{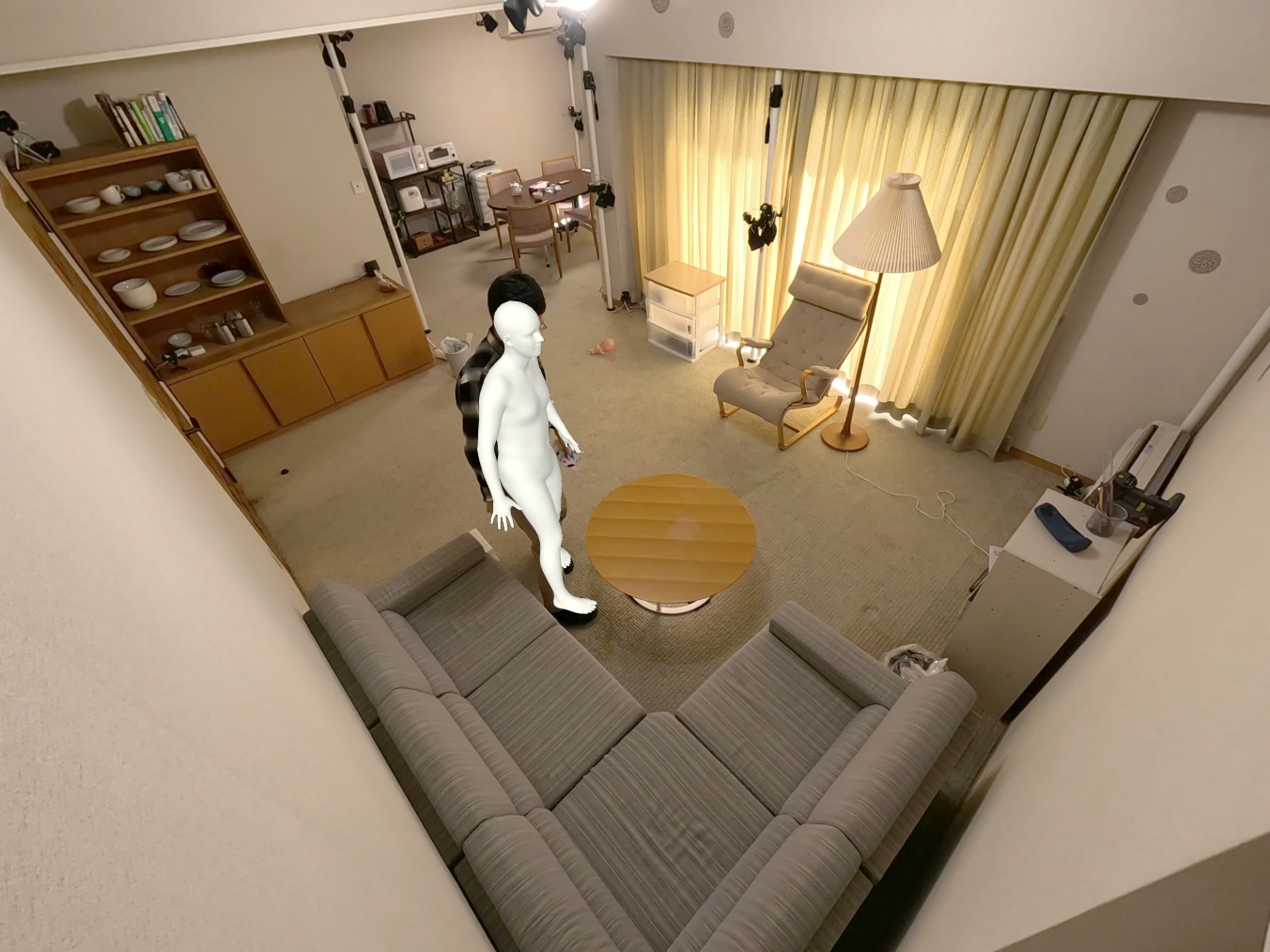}}; \advance \vx by \colwidth
    \node[anchor=north] at (\vx, \vy) {\Pd{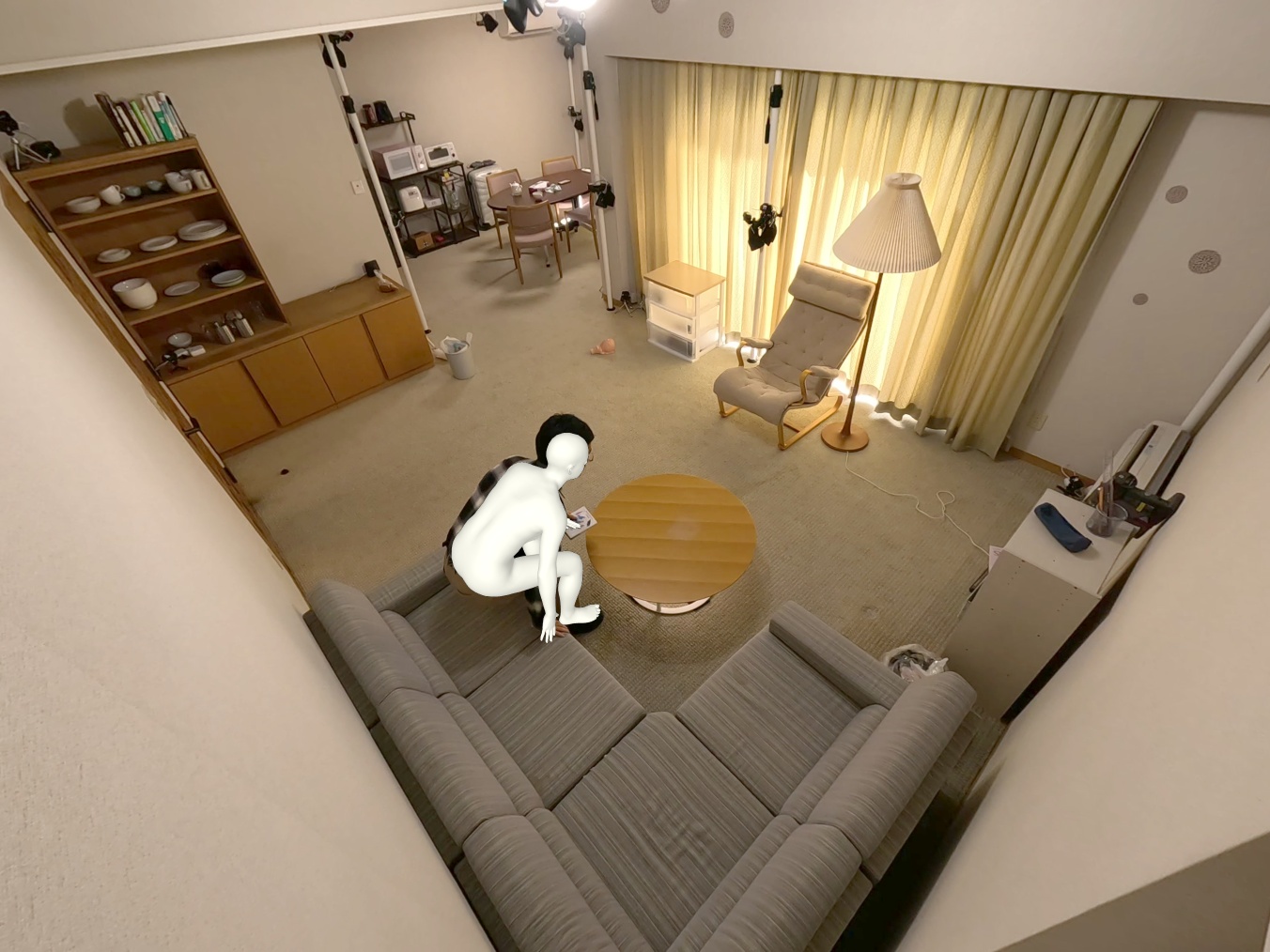}}; \advance \vx by \colwidth
    \node[anchor=north] at (\vx, \vy) {\Pd{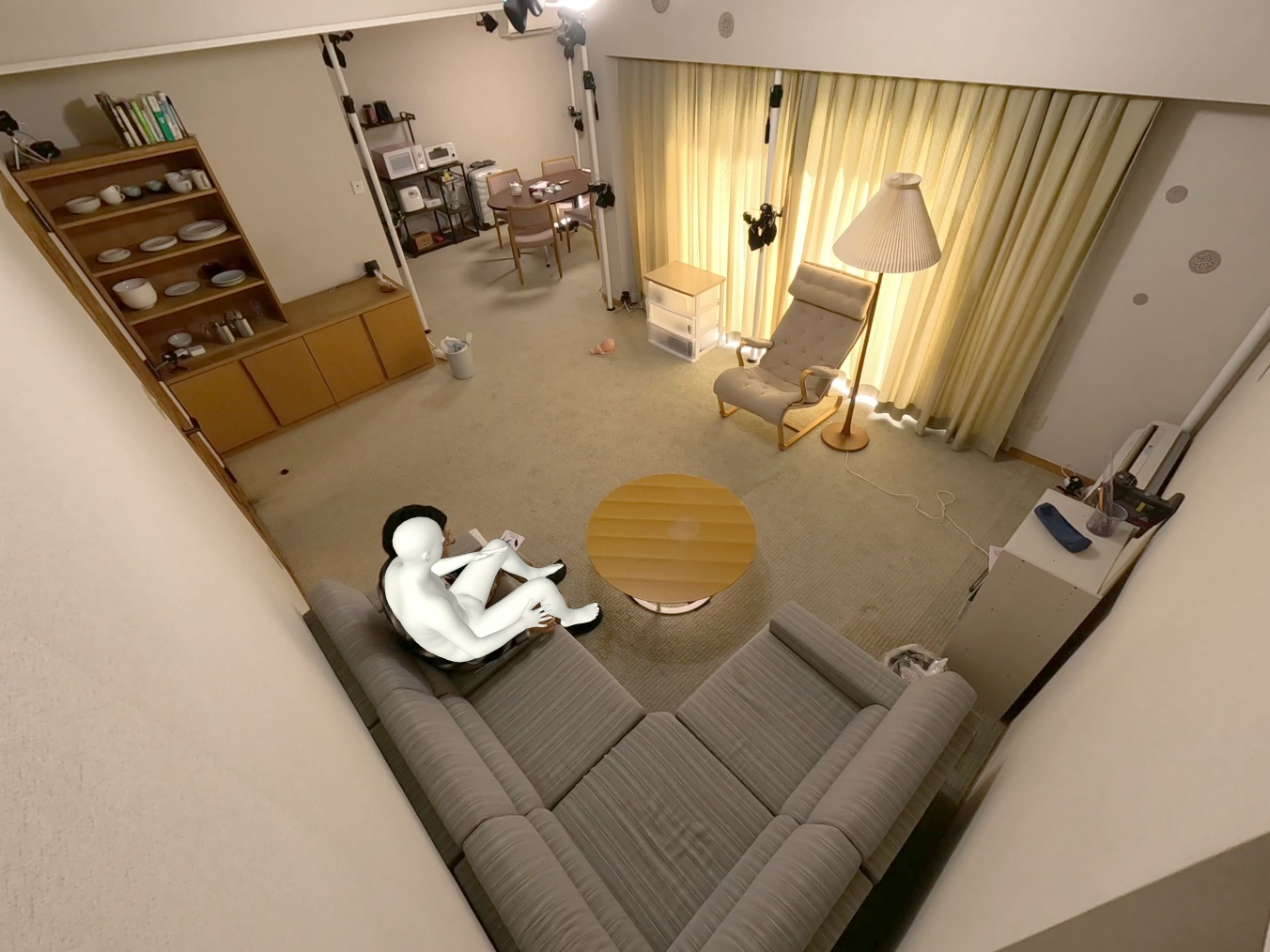}}; \advance \vx by \colwidth
    \node[anchor=north] at (\vx, \vy) {\Pd{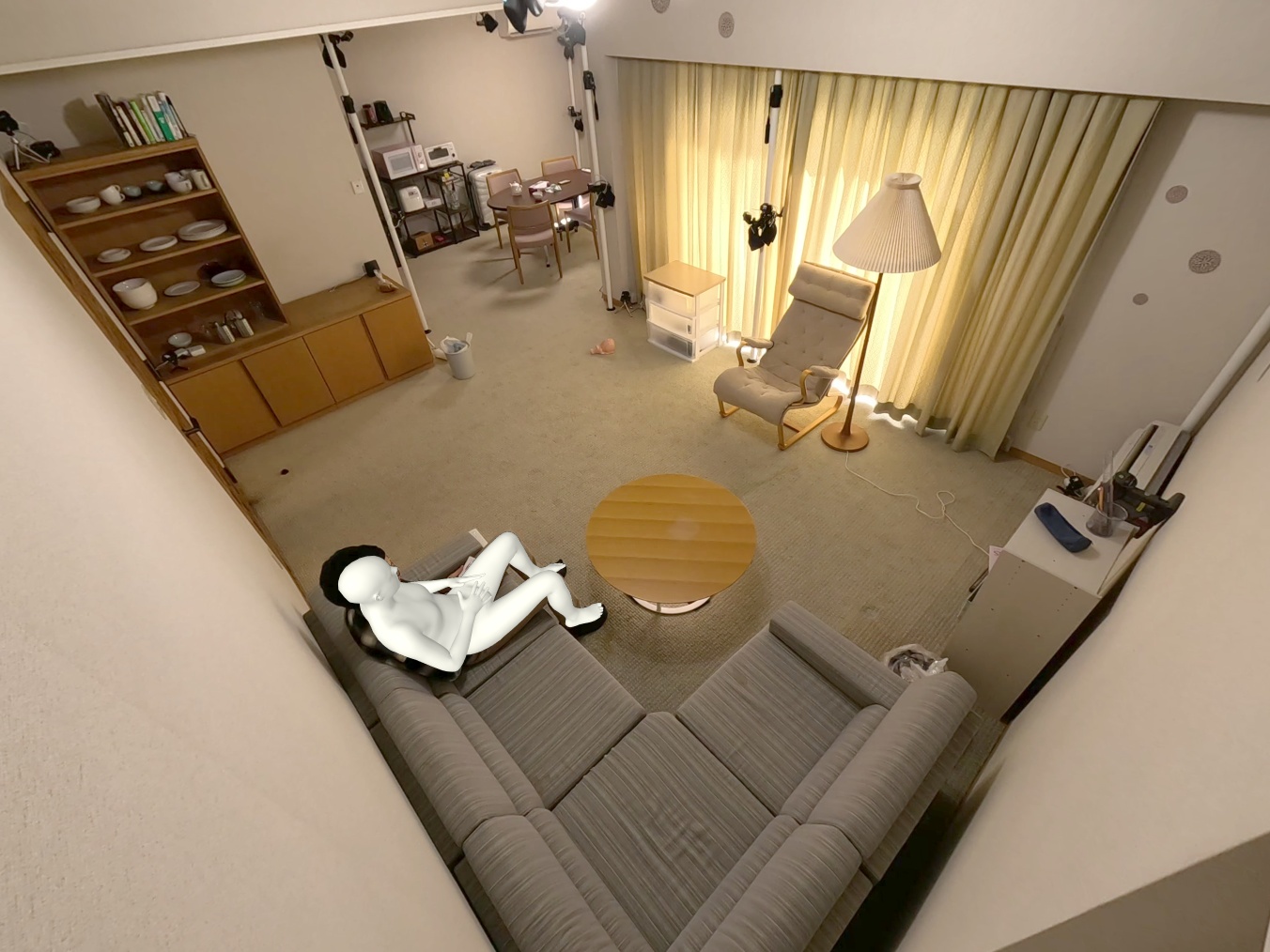}}; \advance \vx by \colwidth
    \node[anchor=north] at (\vx, \vy) {\Pd{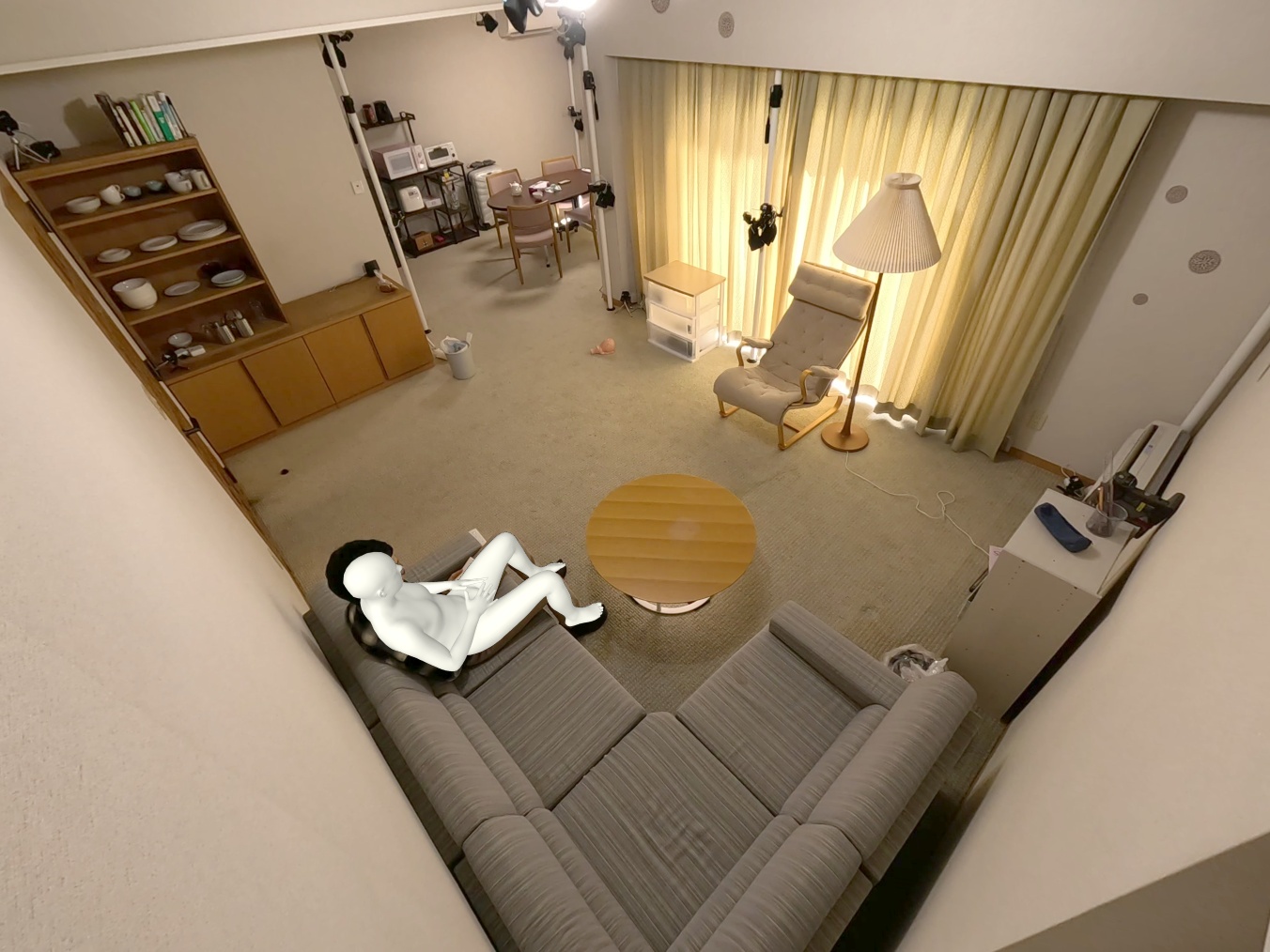}}; \advance \vx by \colwidth

    \vx = 0
    \advance \vy by \rowheight
    \node[anchor=north] at (\vx, \vy) {\Pe{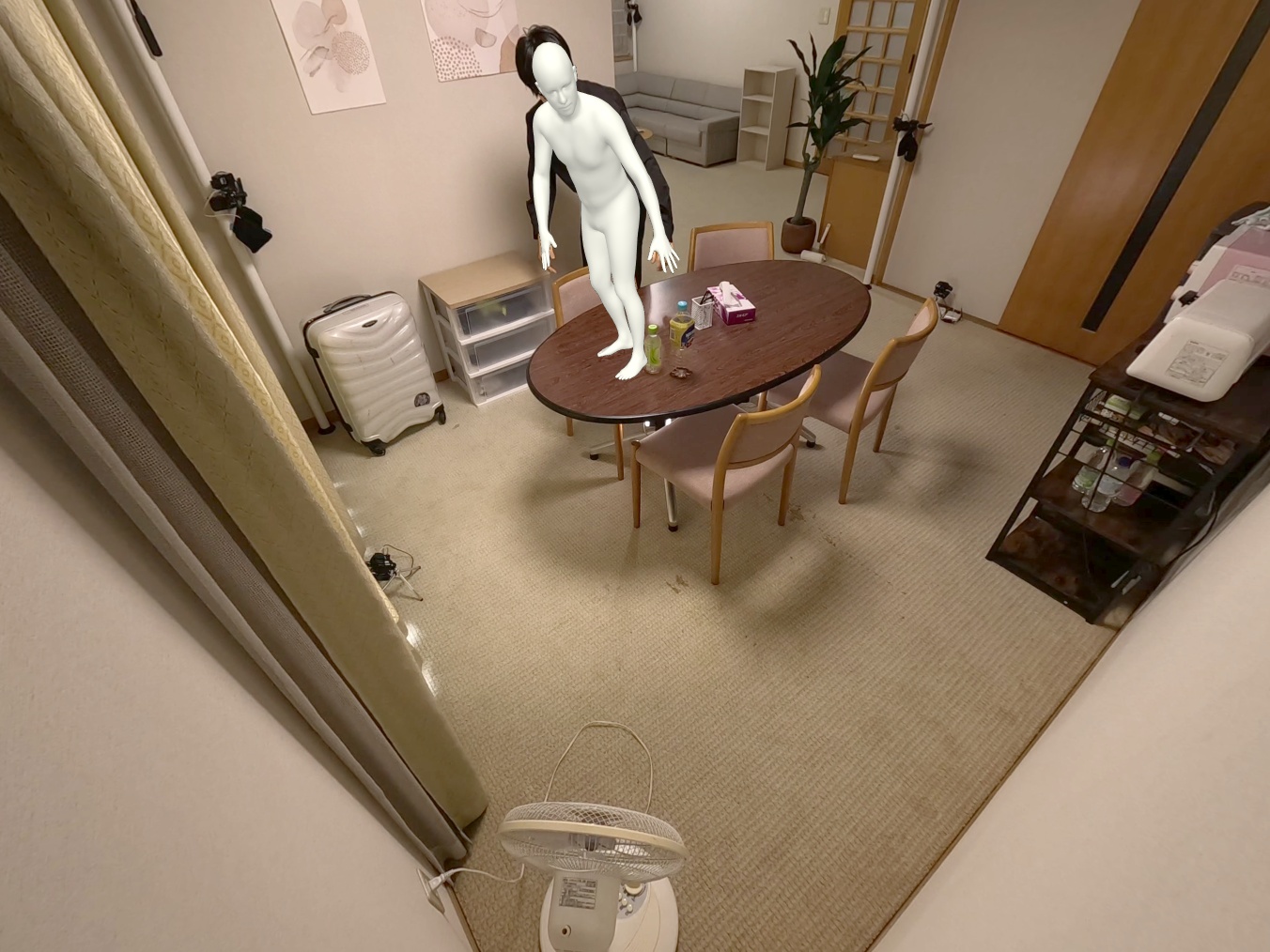}}; \advance \vx by \colwidth
    \node[anchor=north] at (\vx, \vy) {\Pe{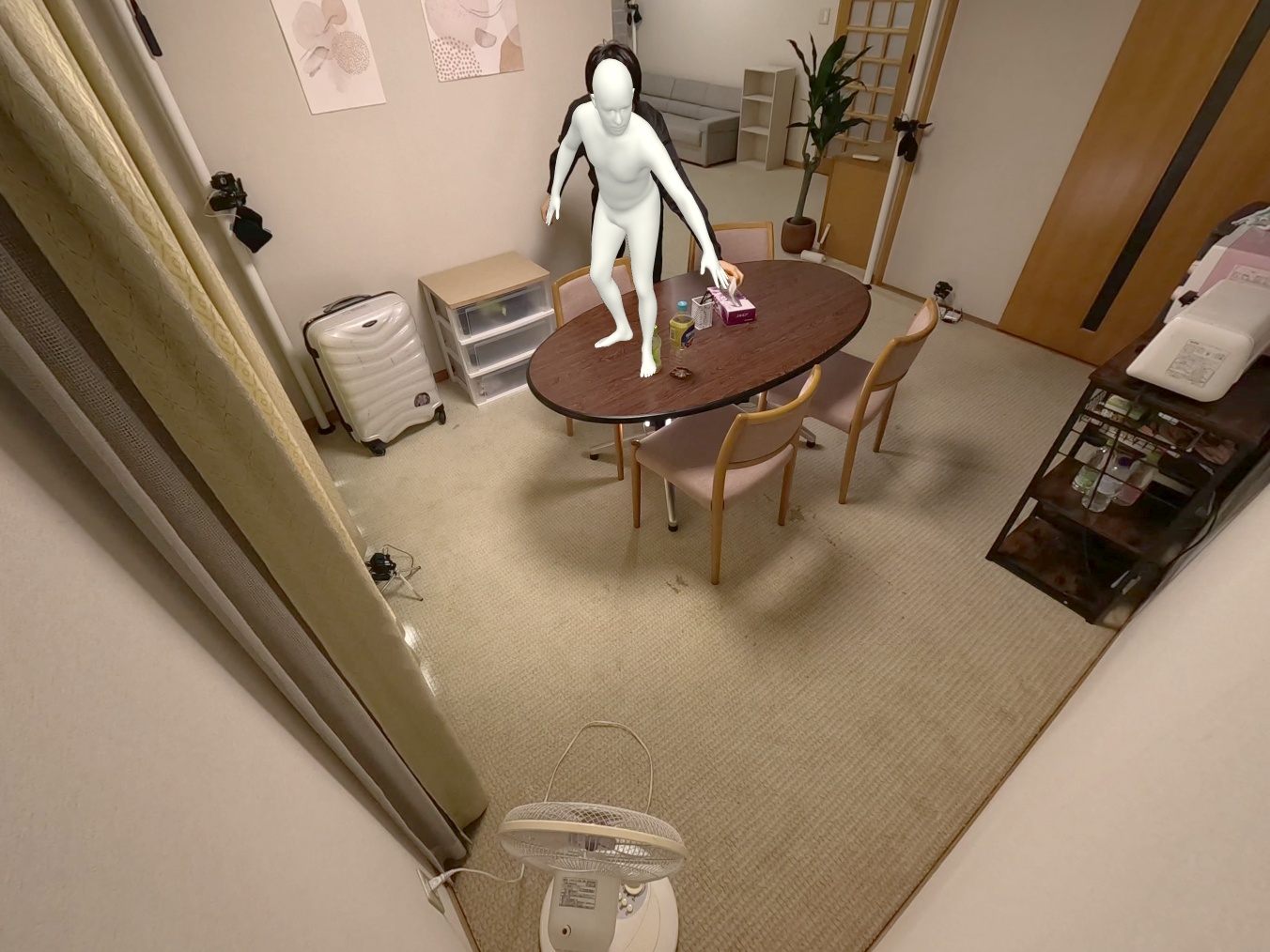}}; \advance \vx by \colwidth
    \node[anchor=north] at (\vx, \vy) {\Pe{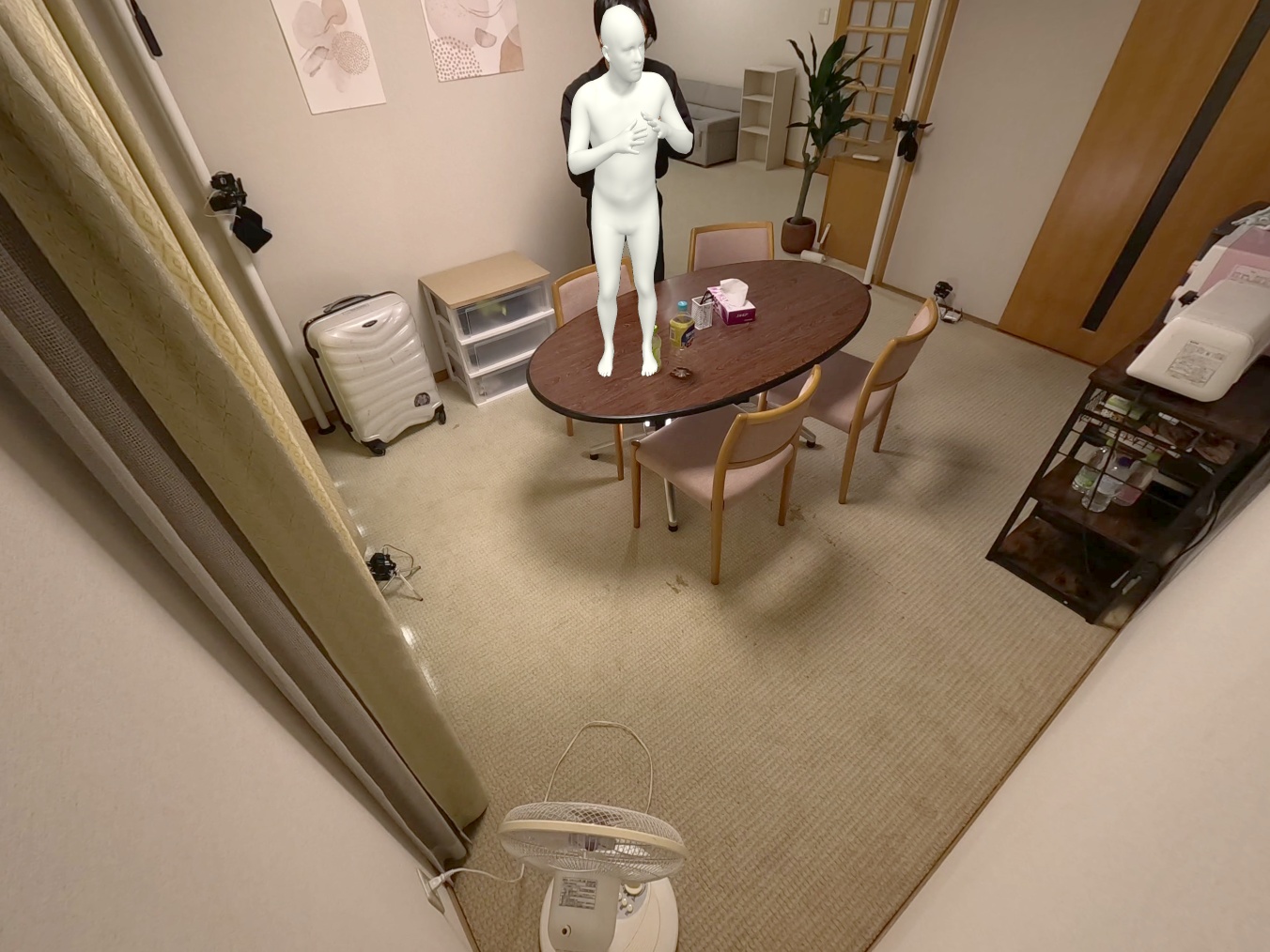}}; \advance \vx by \colwidth
    \node[anchor=north] at (\vx, \vy) {\Pe{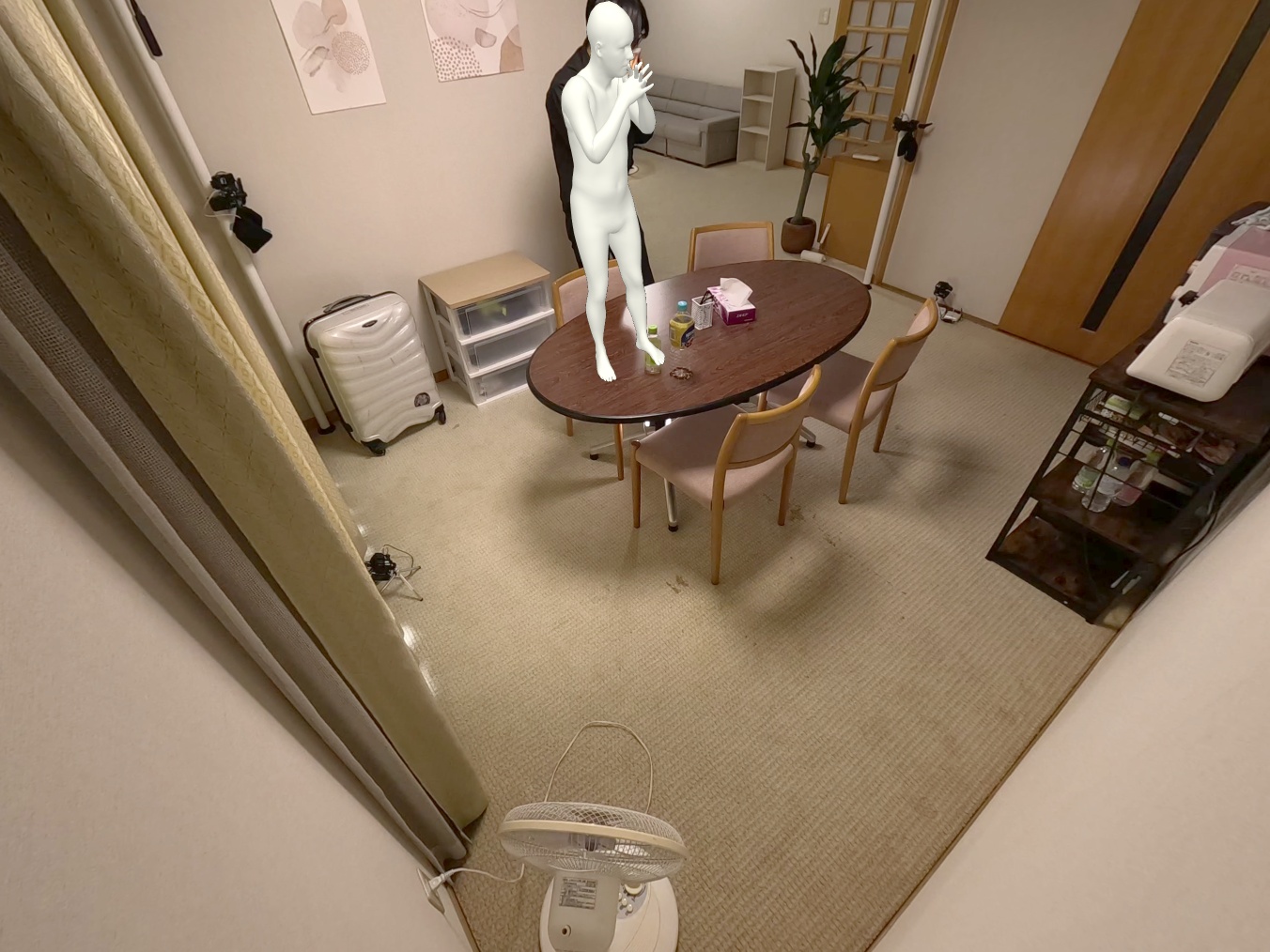}}; \advance \vx by \colwidth
    \node[anchor=north] at (\vx, \vy) {\Pe{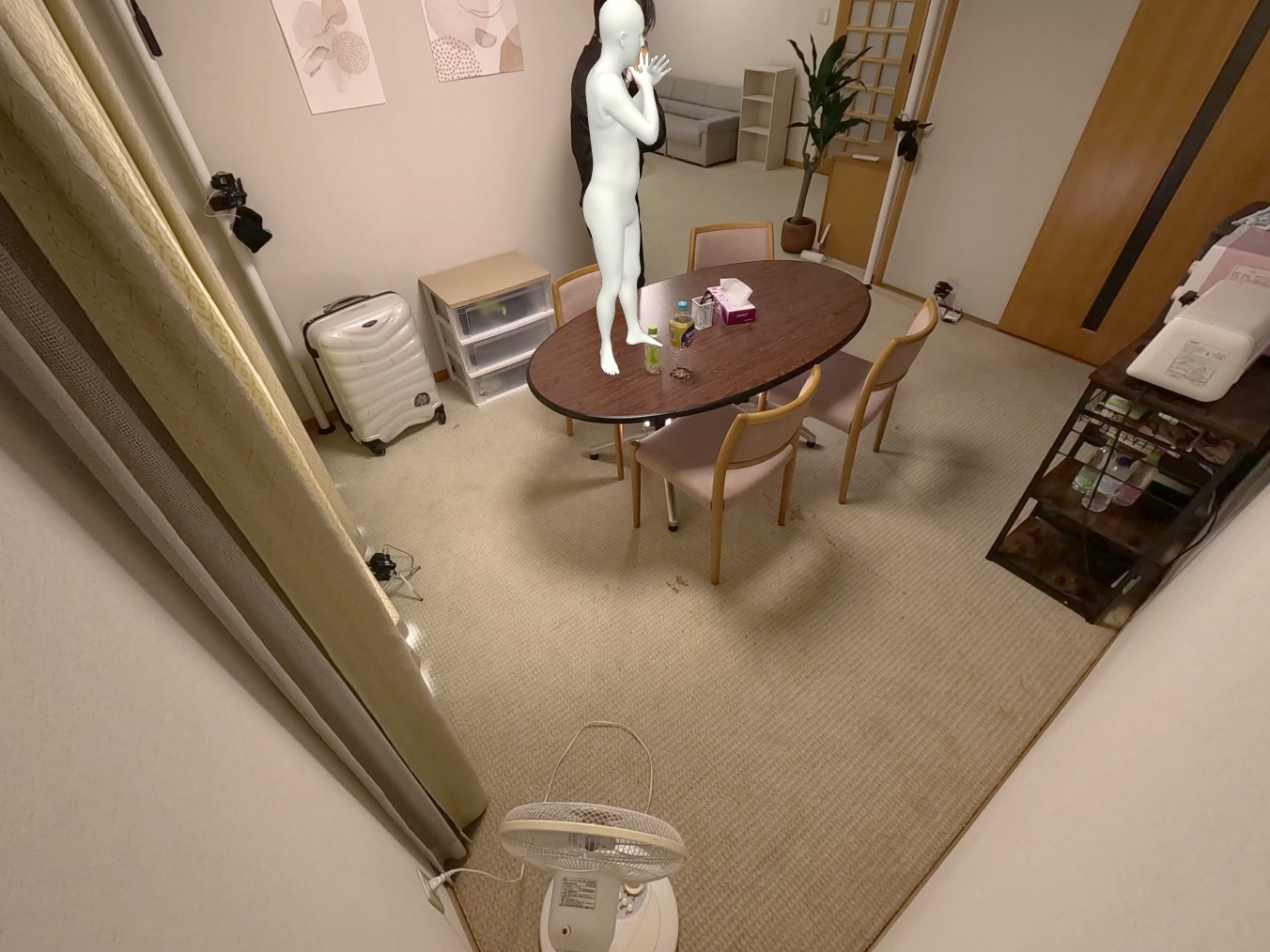}}; \advance \vx by \colwidth

    \vx = 0
    \advance \vy by \rowheight
    \node[anchor=north] at (\vx, \vy) {\Pf{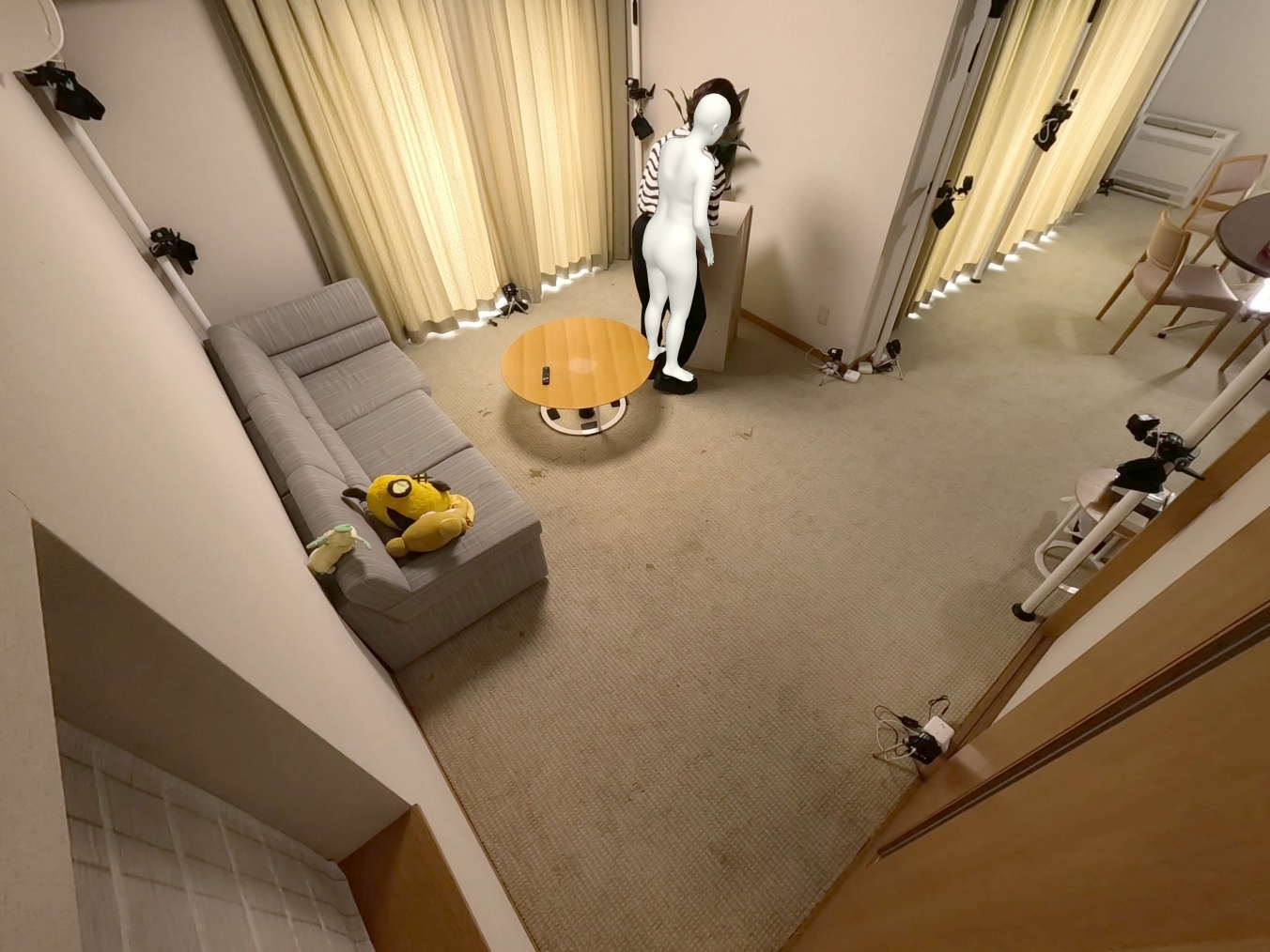}}; \advance \vx by \colwidth
    \node[anchor=north] at (\vx, \vy) {\Pf{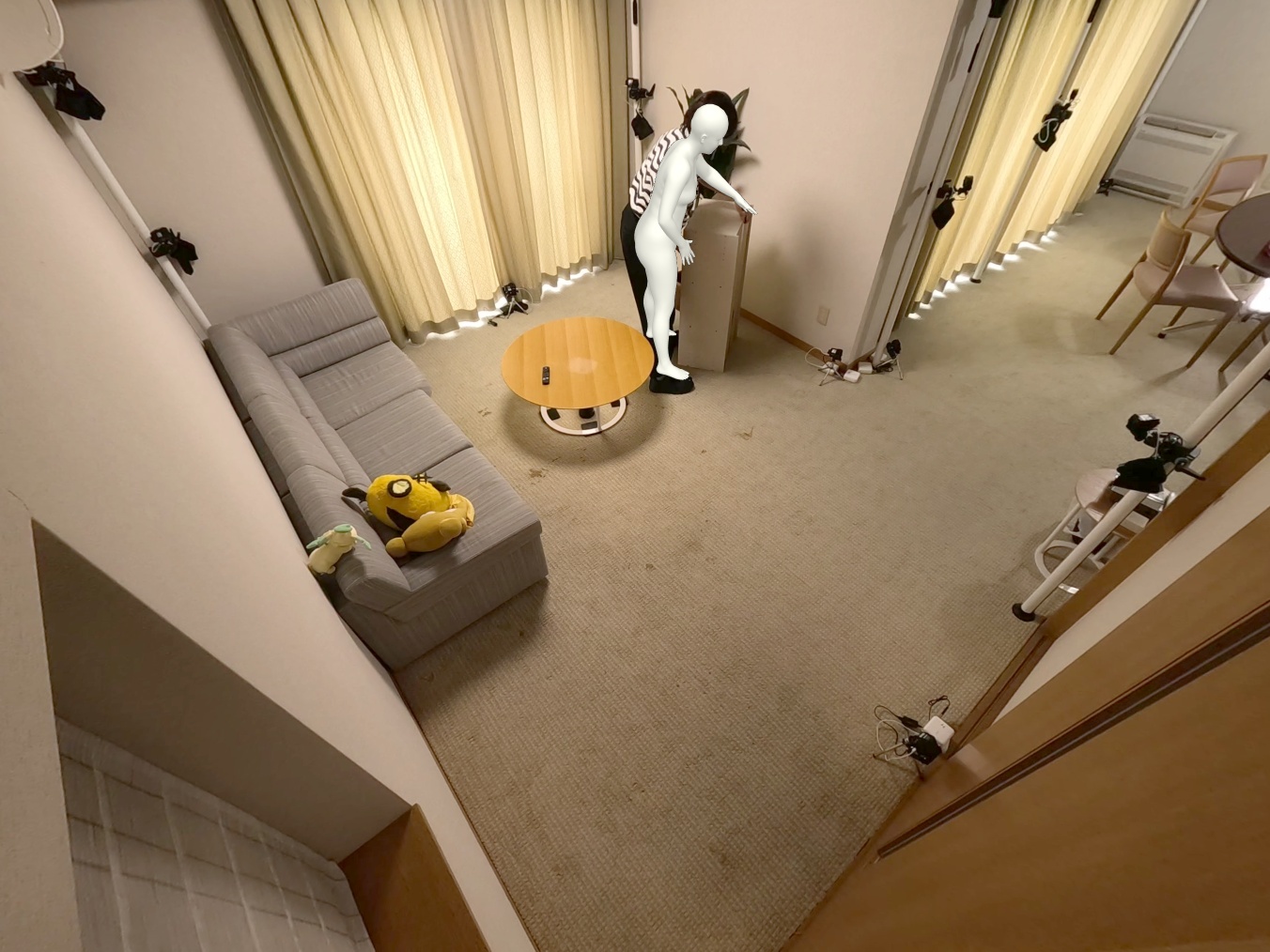}}; \advance \vx by \colwidth
    \node[anchor=north] at (\vx, \vy) {\Pf{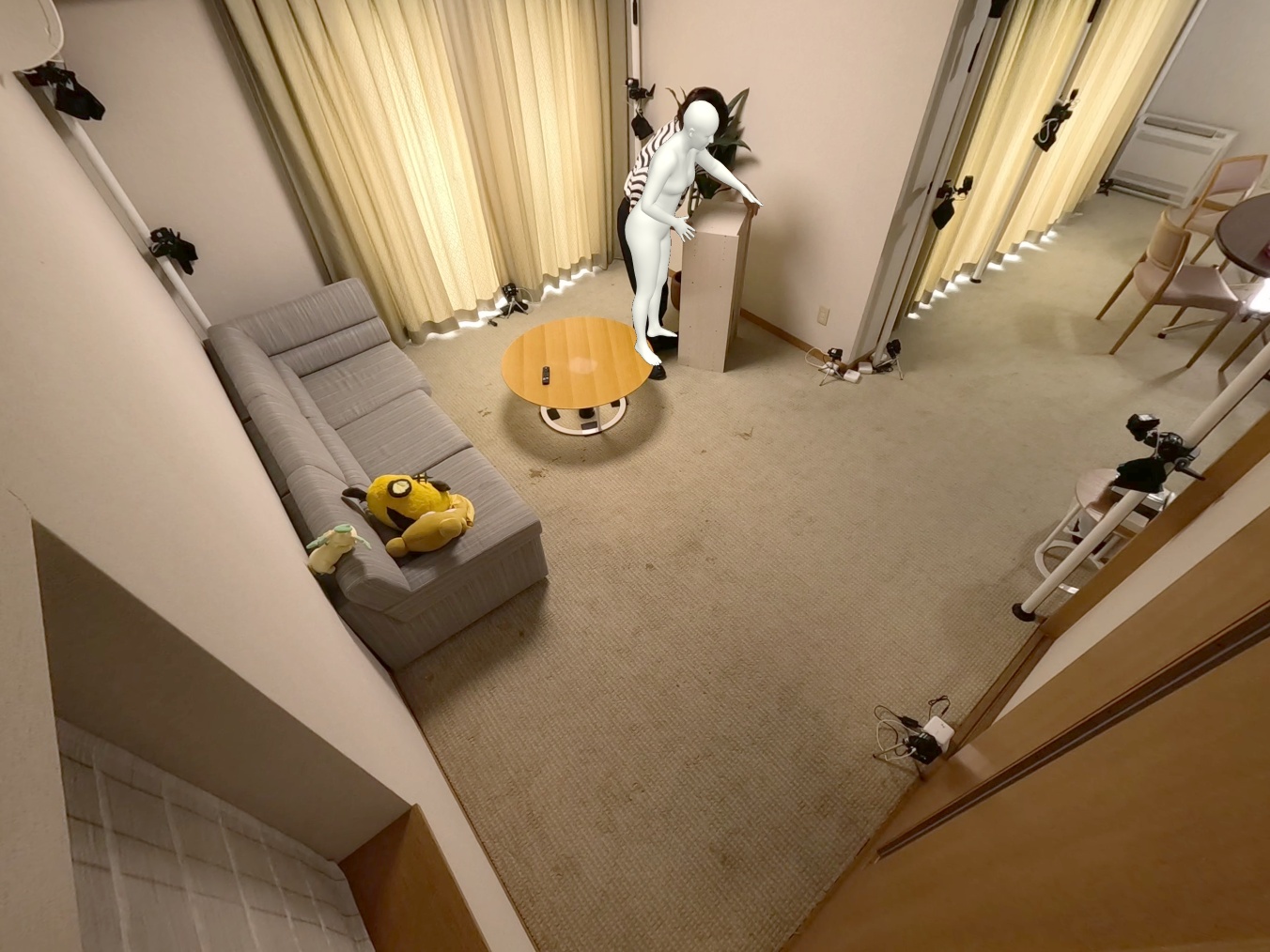}}; \advance \vx by \colwidth
    \node[anchor=north] at (\vx, \vy) {\Pf{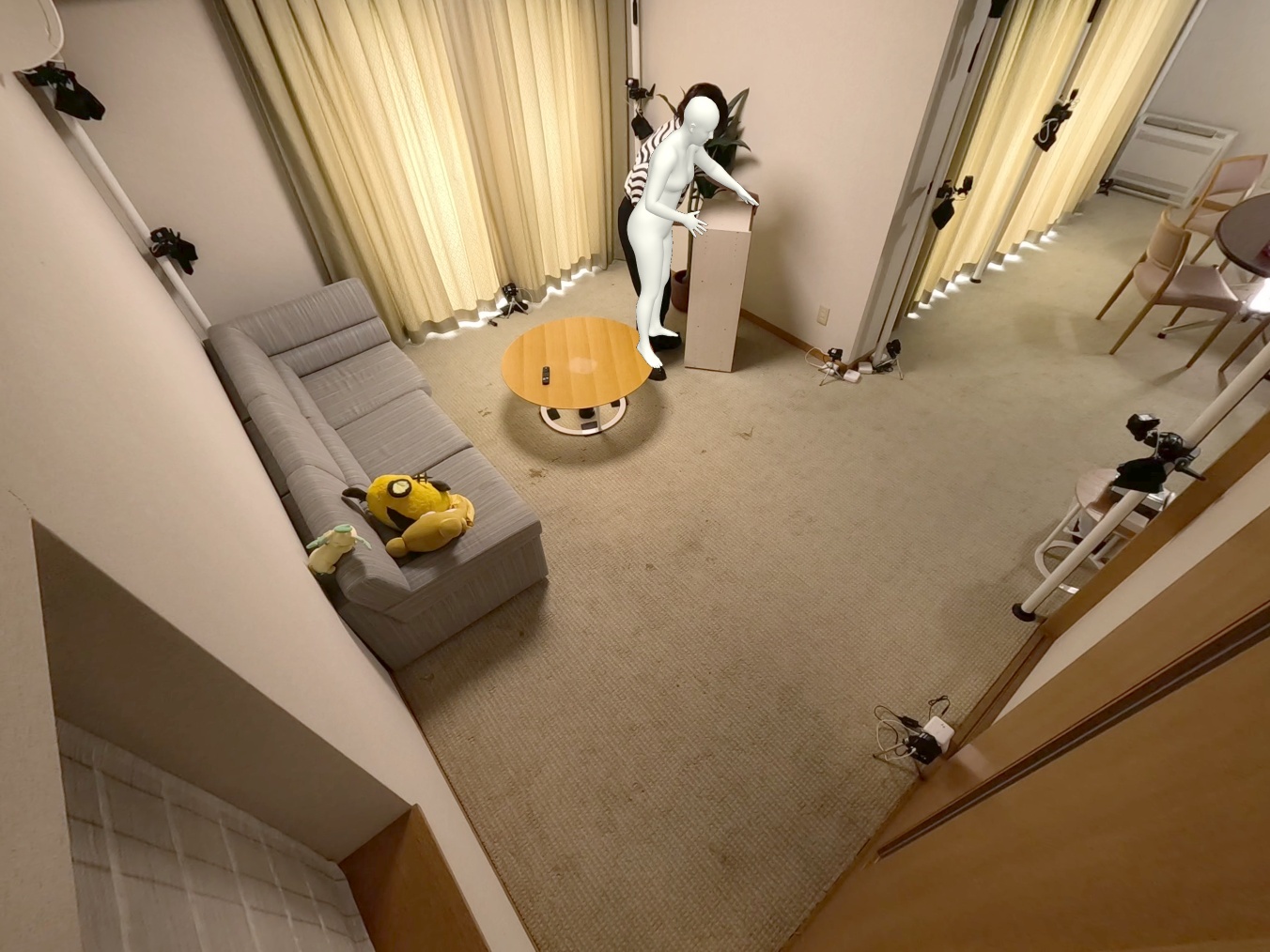}}; \advance \vx by \colwidth
    \node[anchor=north] at (\vx, \vy) {\Pf{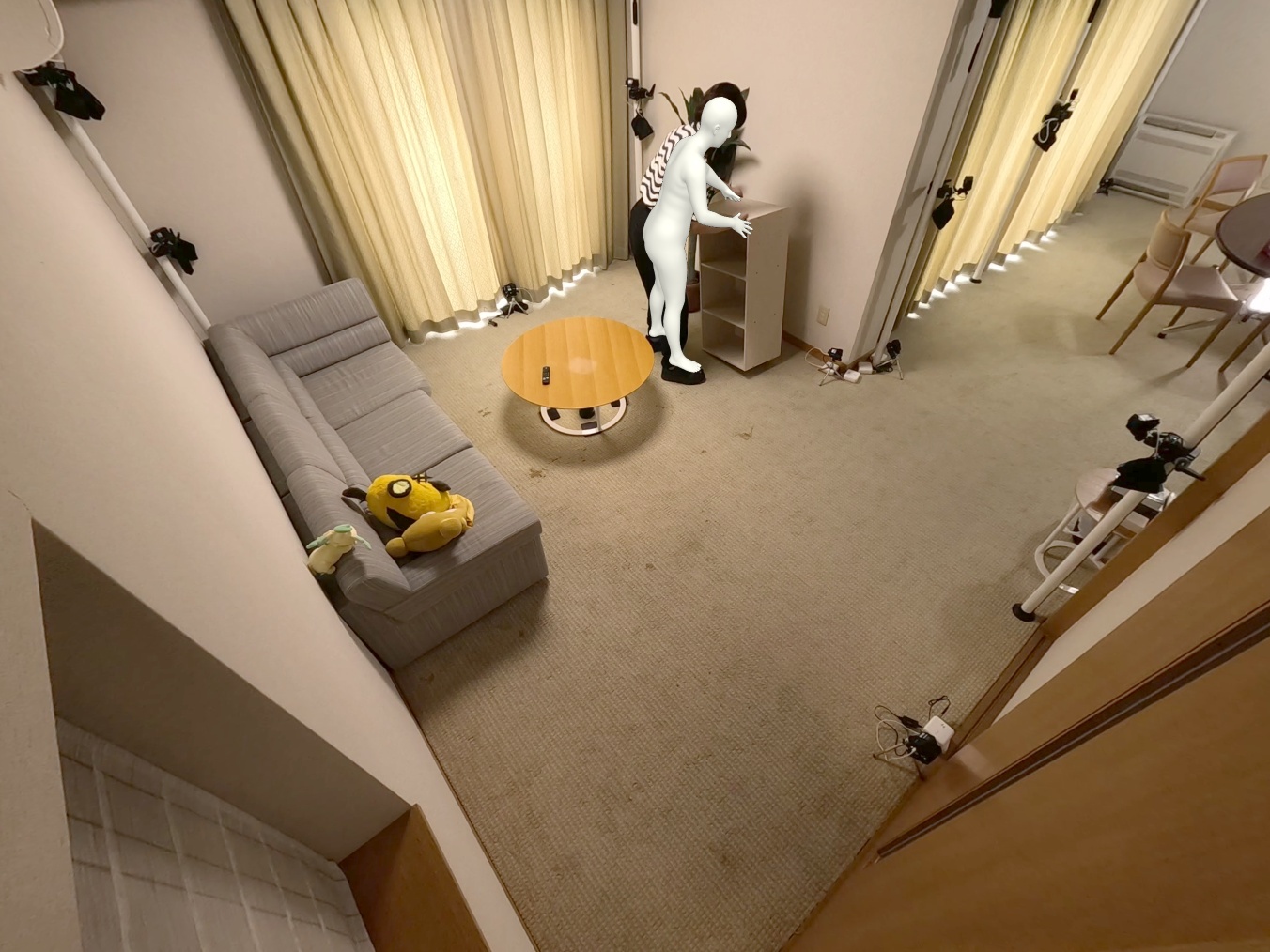}}; \advance \vx by \colwidth

    \vx = 0
    \advance \vy by \rowheight
    \node[anchor=north] at (\vx, \vy) {\Pg{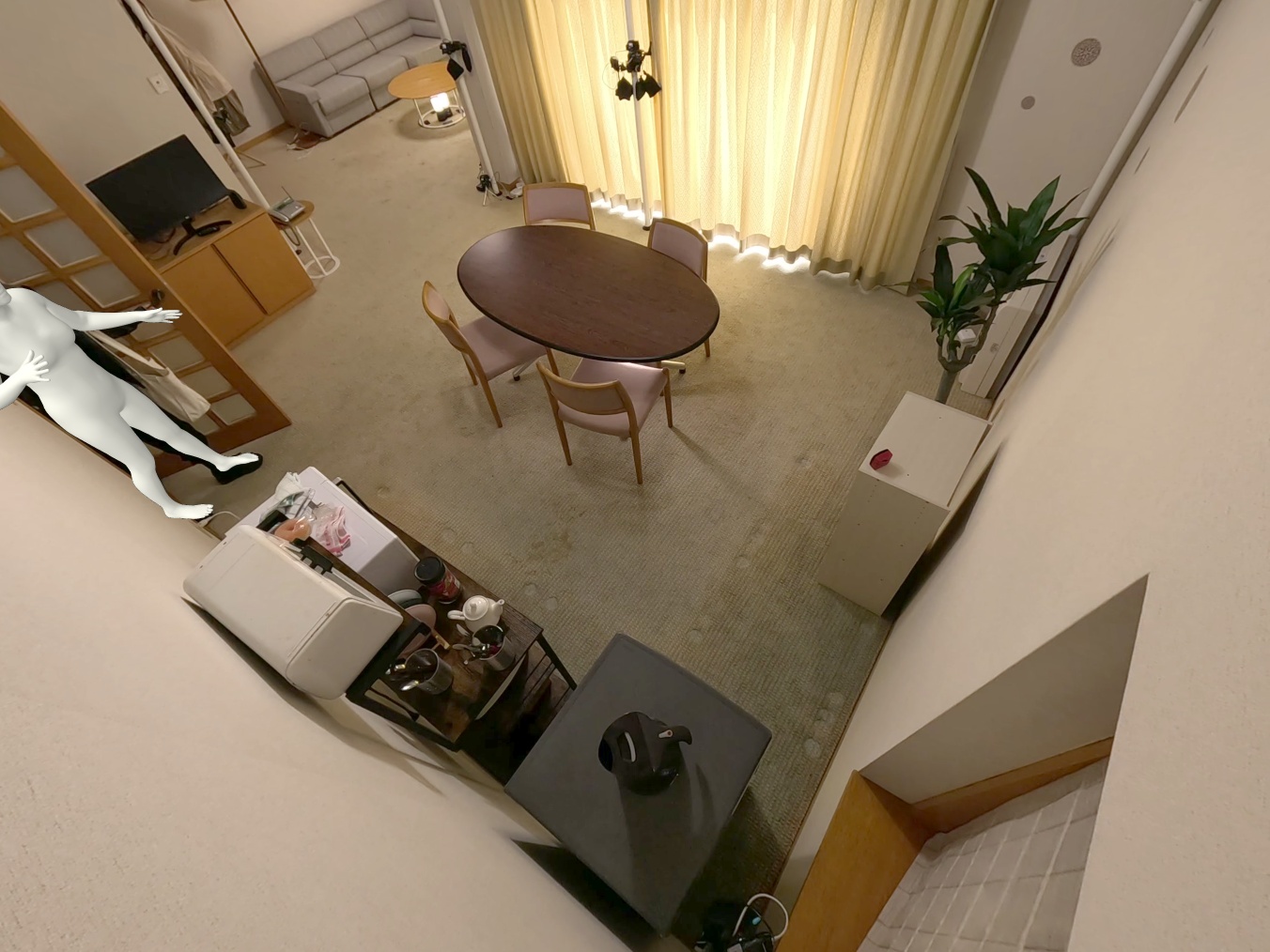}}; \advance \vx by \colwidth
    \node[anchor=north] at (\vx, \vy) {\Pg{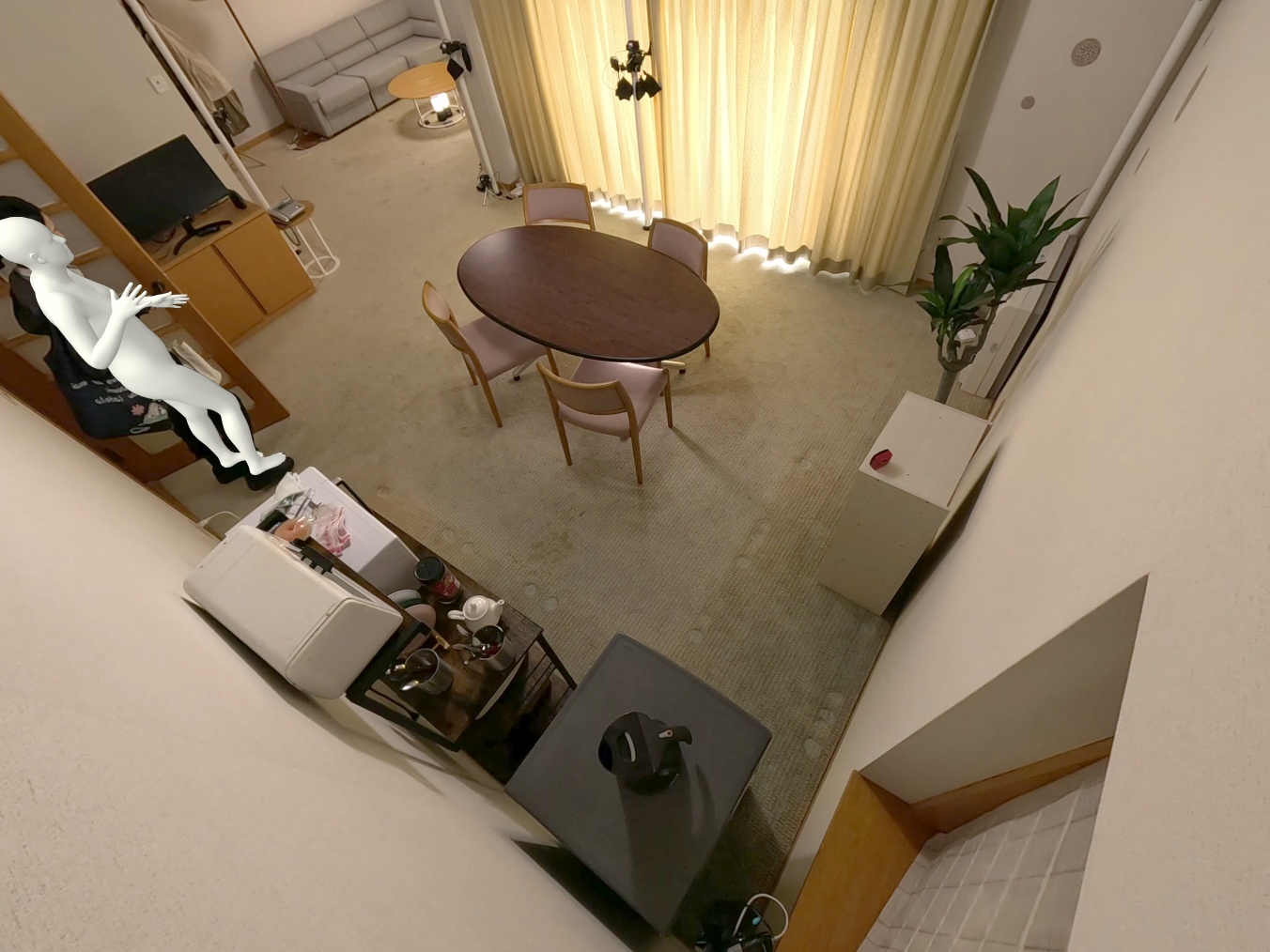}}; \advance \vx by \colwidth
    \node[anchor=north] at (\vx, \vy) {\Pg{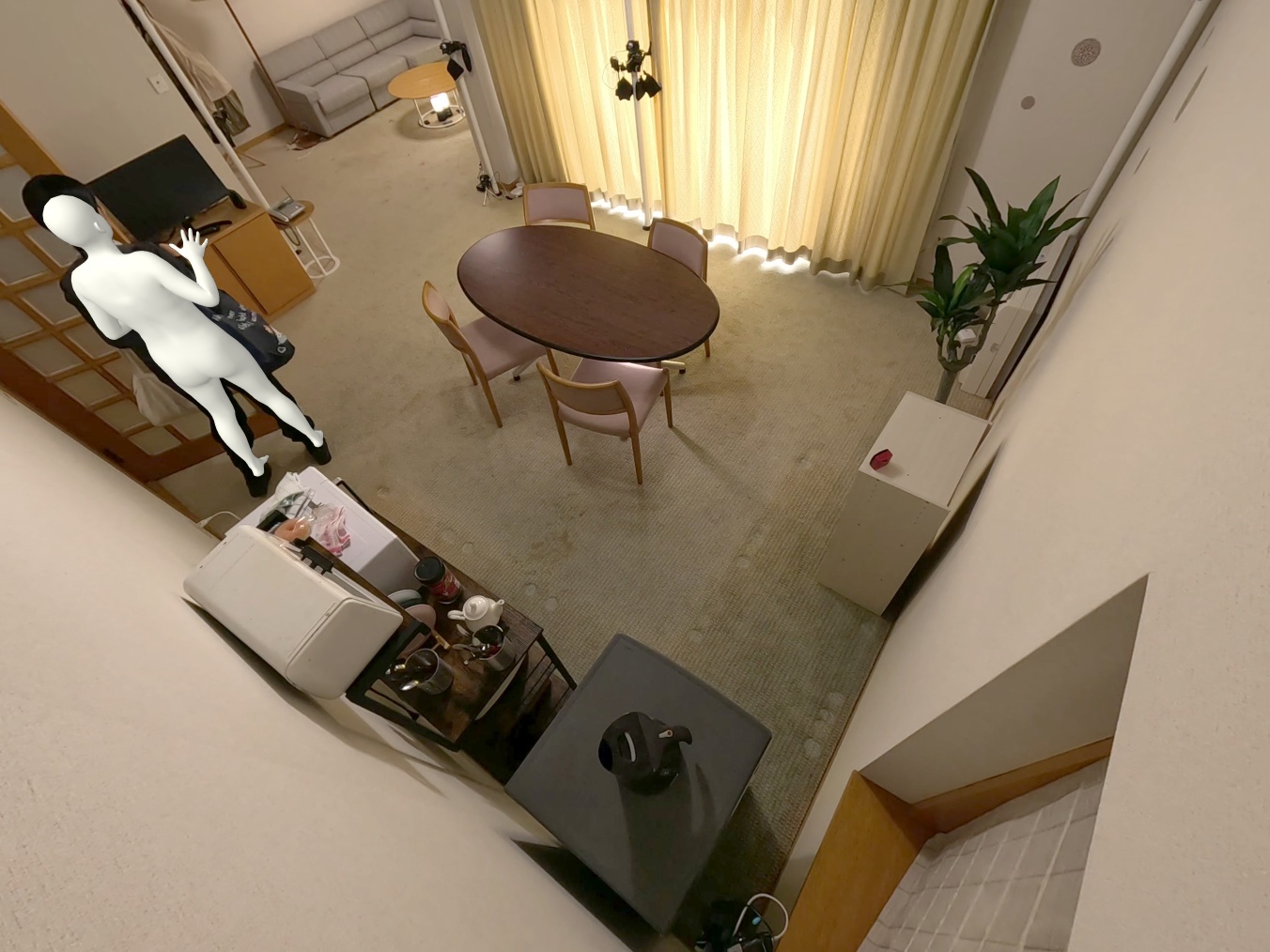}}; \advance \vx by \colwidth
    \node[anchor=north] at (\vx, \vy) {\Pg{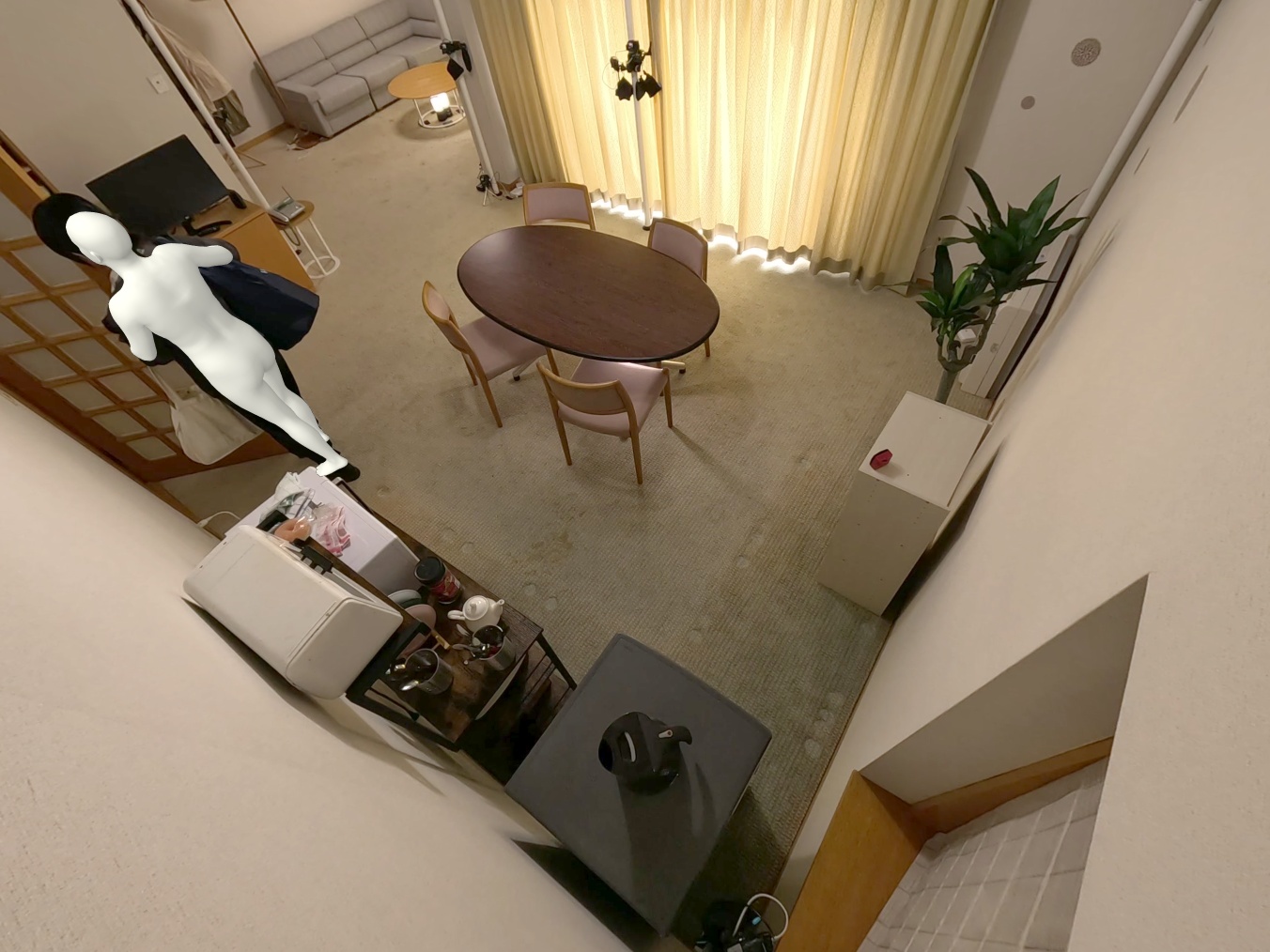}}; \advance \vx by \colwidth
    \node[anchor=north] at (\vx, \vy) {\Pg{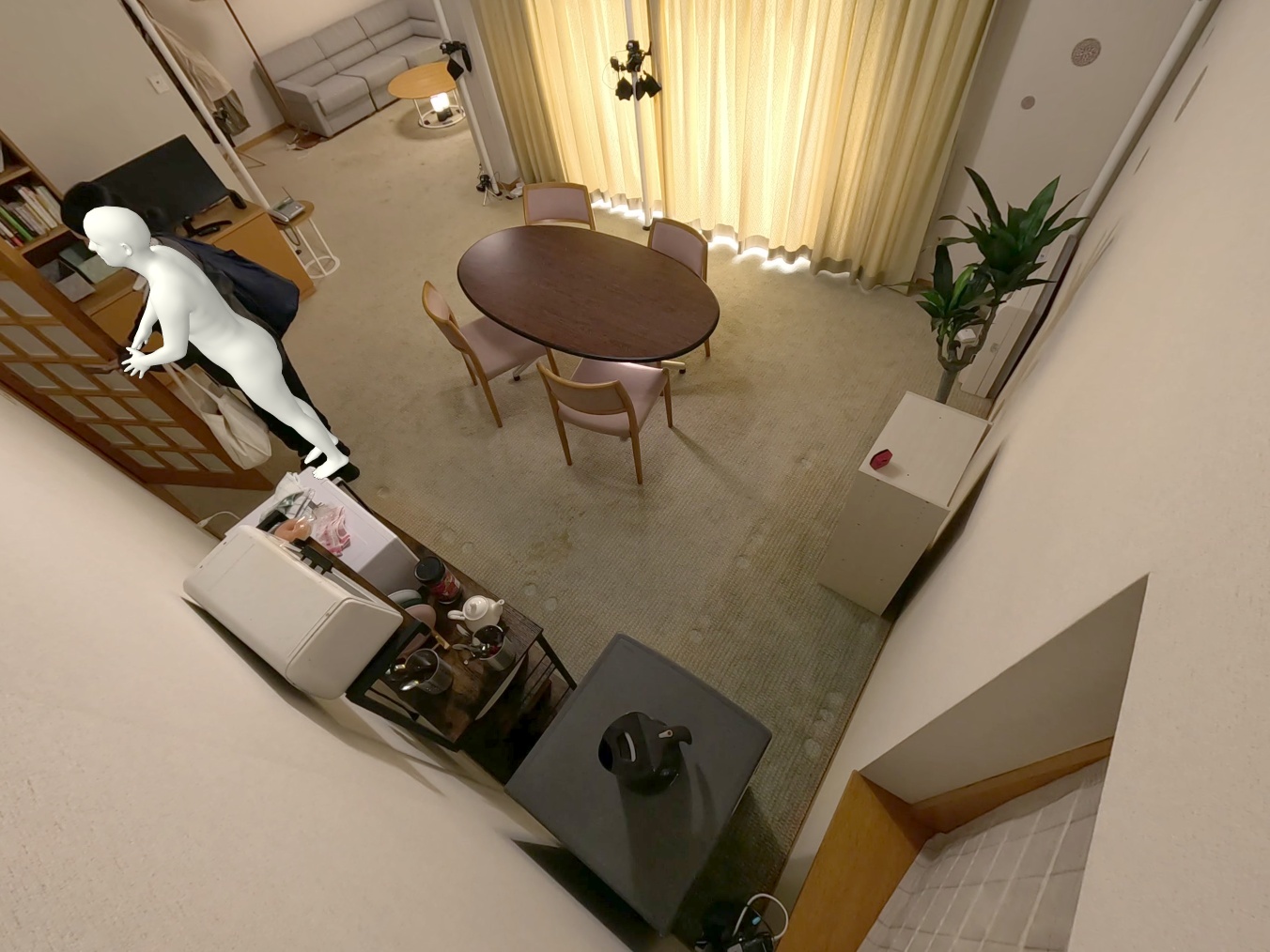}}; \advance \vx by \colwidth

    \vx = 0
    \advance \vy by \rowheight
    \node[anchor=north] at (\vx, \vy) {\Ph{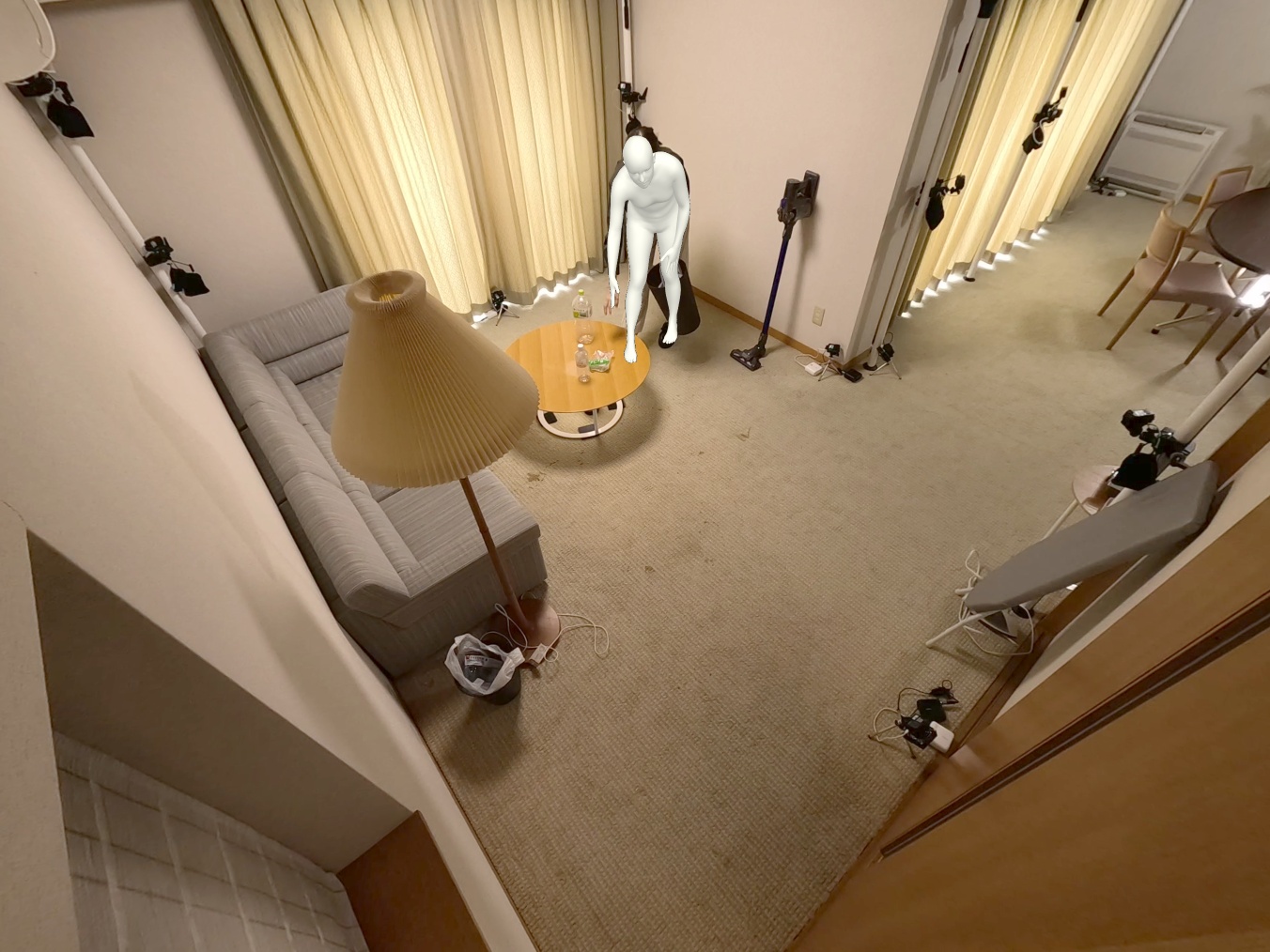}}; \advance \vx by \colwidth
    \node[anchor=north] at (\vx, \vy) {\Ph{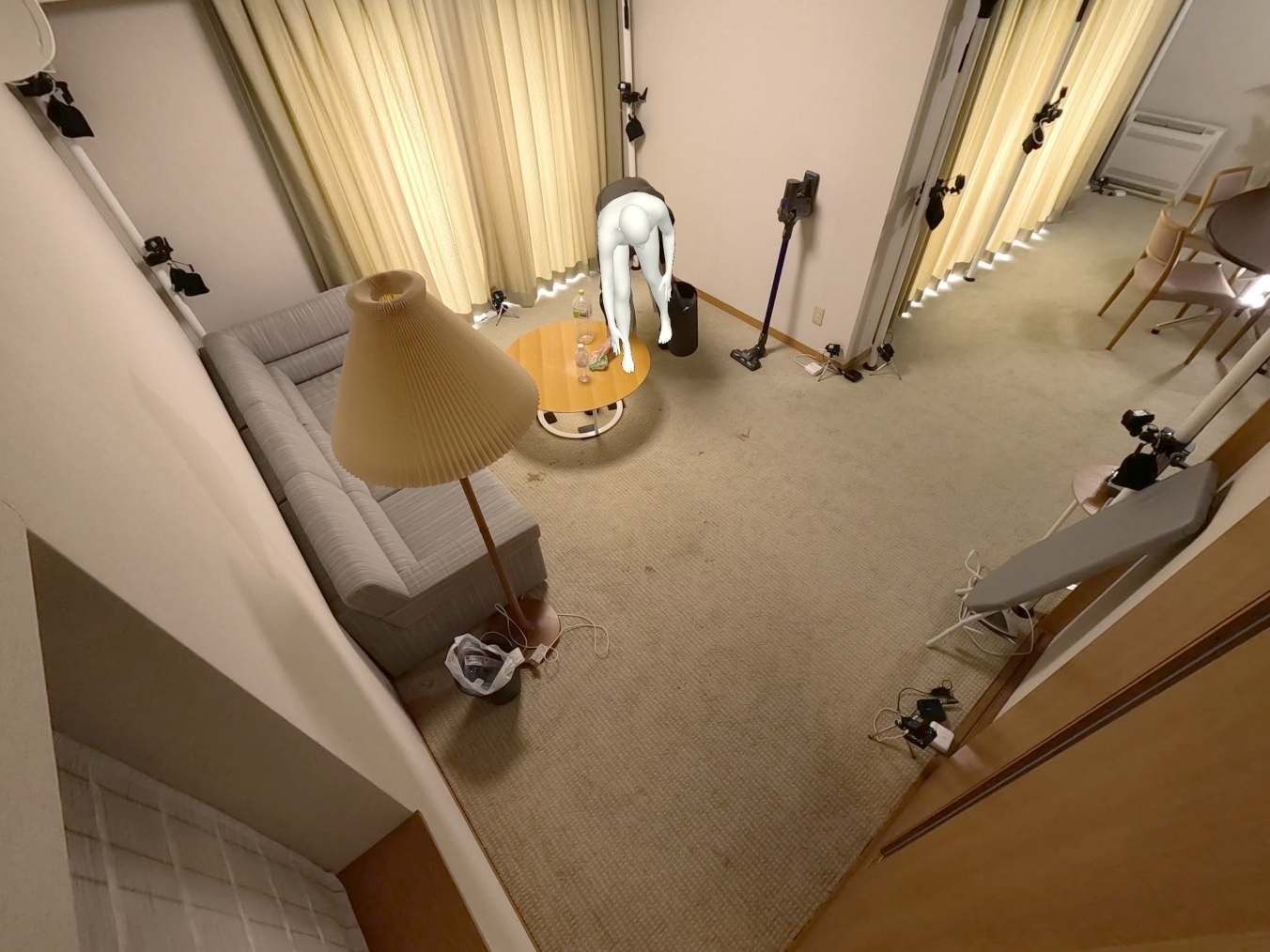}}; \advance \vx by \colwidth
    \node[anchor=north] at (\vx, \vy) {\Ph{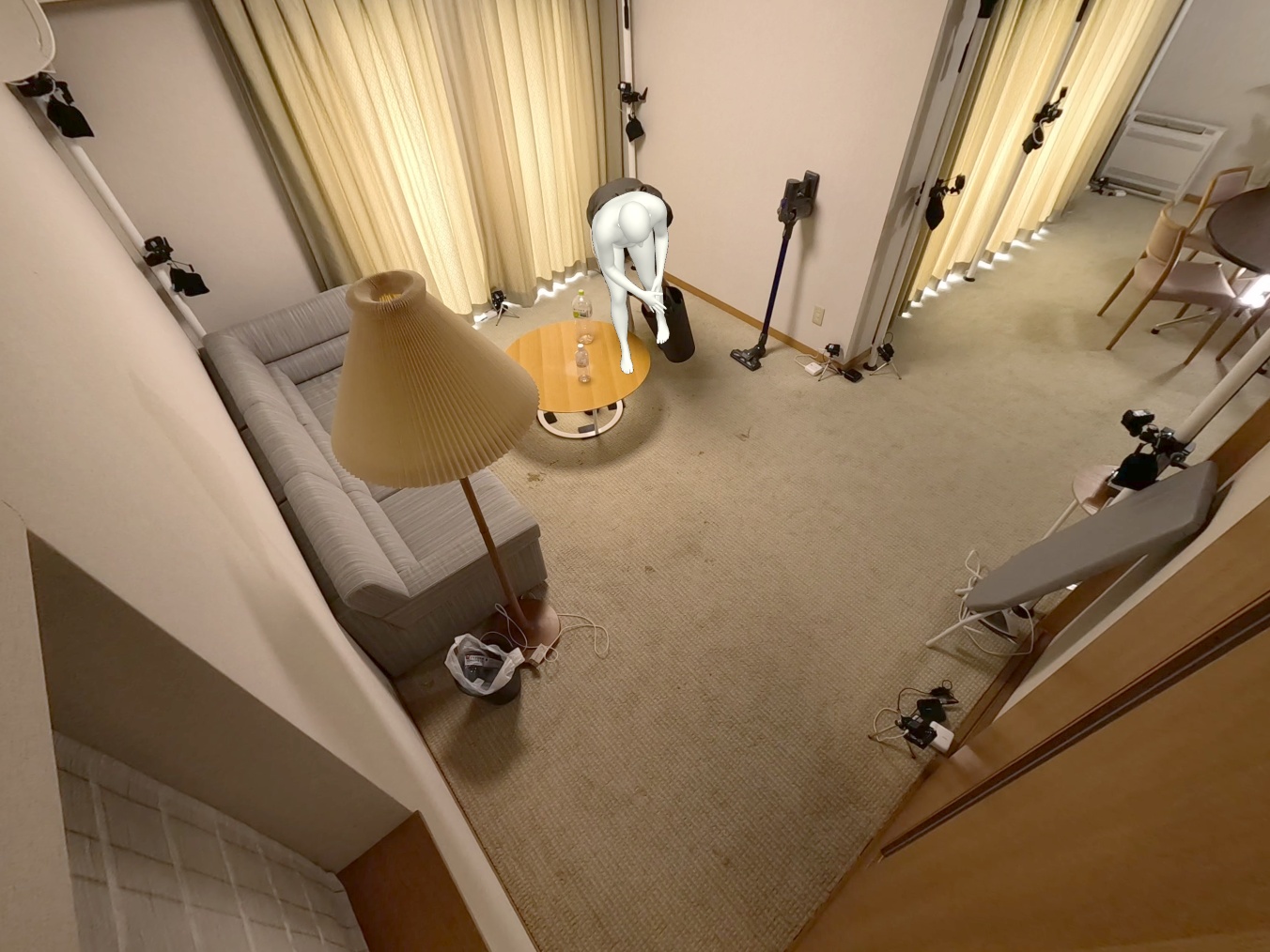}}; \advance \vx by \colwidth
    \node[anchor=north] at (\vx, \vy) {\Ph{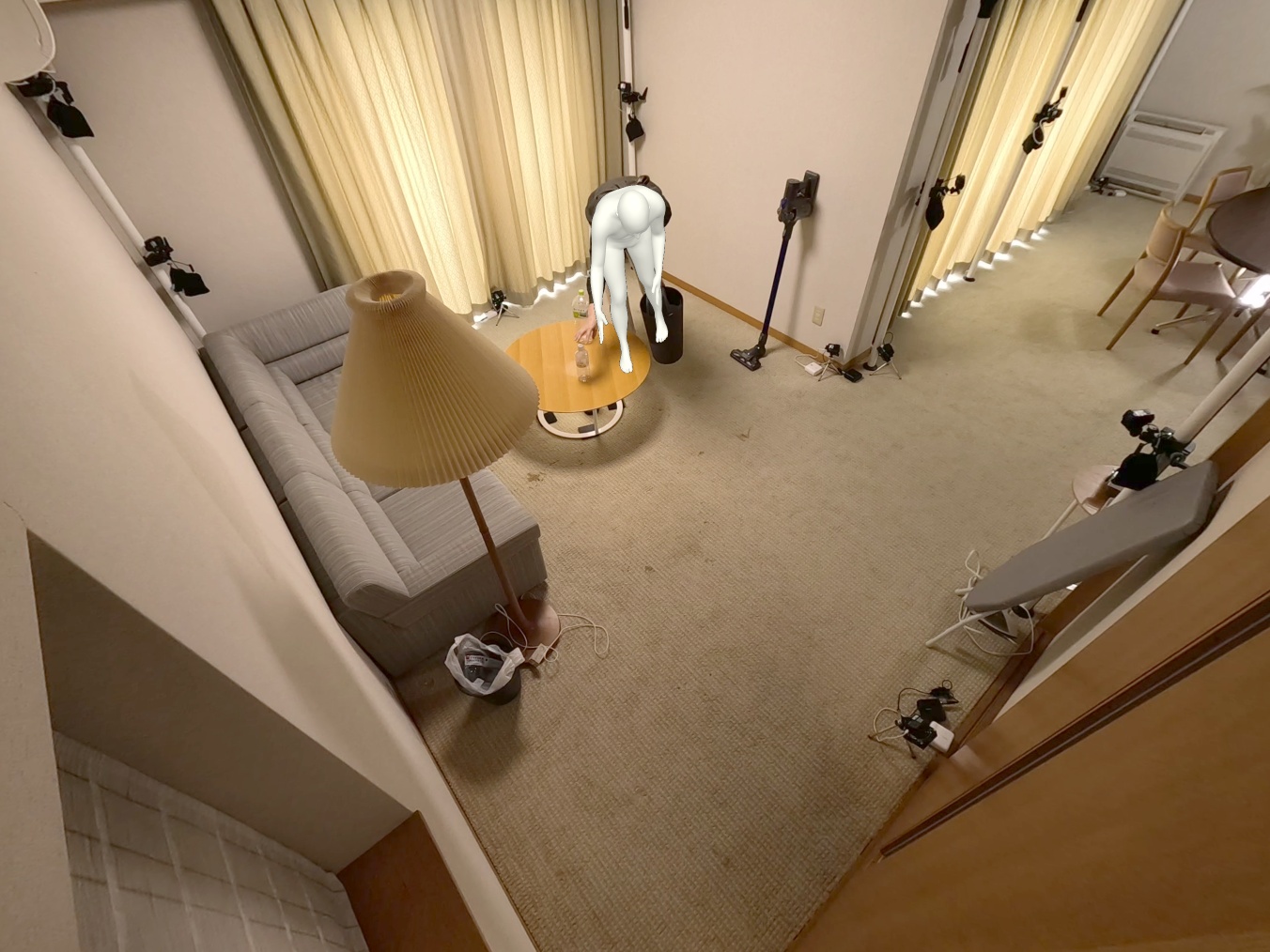}}; \advance \vx by \colwidth
    \node[anchor=north] at (\vx, \vy) {\Ph{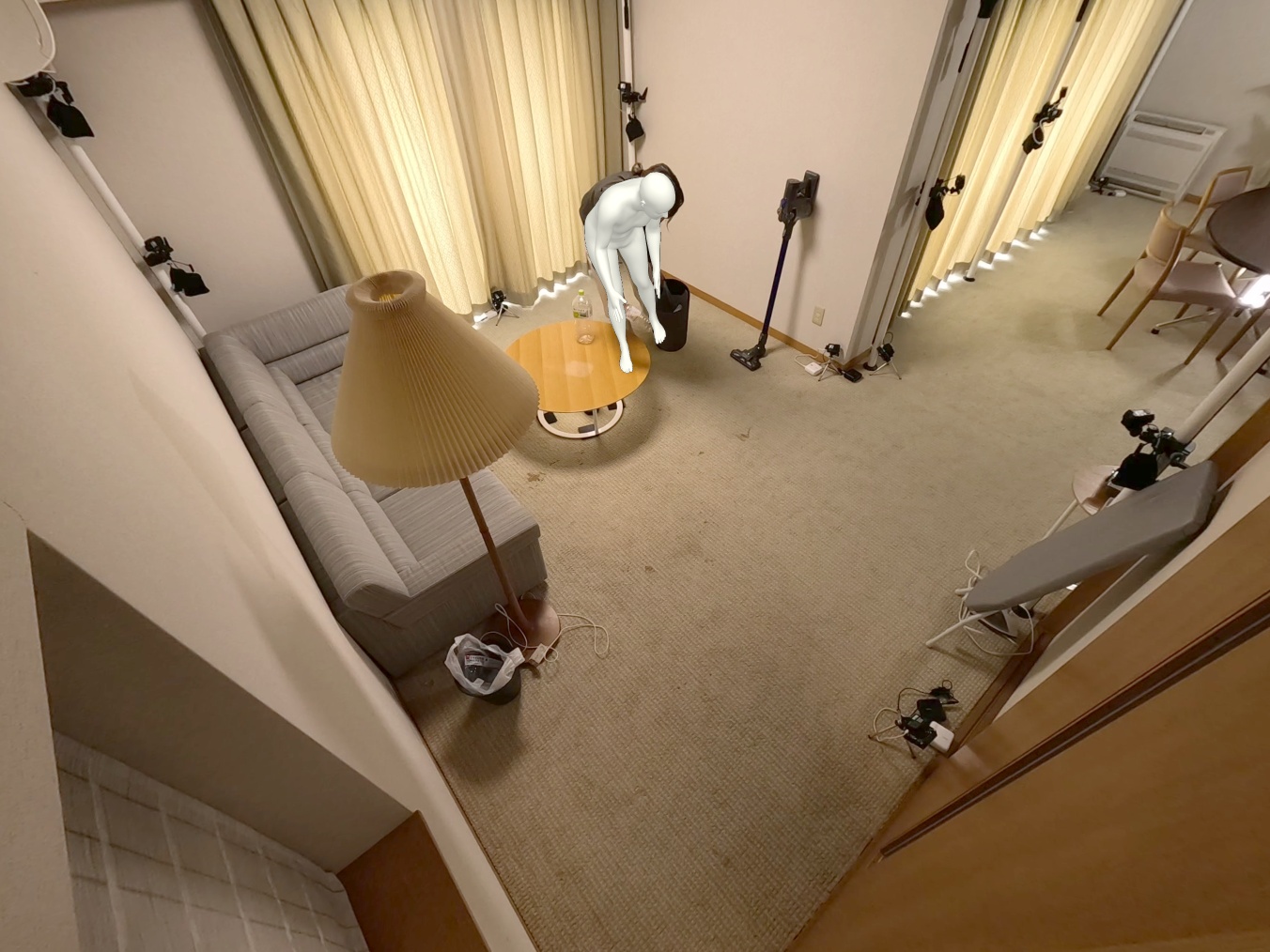}}; \advance \vx by \colwidth

\end{tikzpicture}

%% file: fig/suppl_amd_samples2.tex
\begin{tikzpicture}[x=1mm,y=1mm,every node/.style={inner sep=0pt, text depth=0pt}]
    \newcount\colwidth \colwidth=34  
    \newcount\rowheight \rowheight=-27 

    \newcount\vx \vx=0
    \newcount\vy \vy=0

    \newcommand\imagewidth{0.20\textwidth} 

    \def\RootDir{fig/camera_ready/supp/dataset_examples}
    \def\SceneA{vacuum/2024-09/19/batch2/c-3/gopro11/6}
    \def\SceneB{open_suitcase/2024-09/19/batch3/a-1/gopro11/6}
    \def\SceneC{open_laptop/2024-09/17/batch4/labboy1-1/gopro11/6}
    \def\SceneD{wipe_table/2024-10/10/batch2/a-1/gopro11/1}
    \def\SceneE{use_fan/2024-10/10/batch2/b-3/gopro11/8}
    \def\SceneF{open_fridge/2024-10/16/batch2/a-1/gopro11/6}
    \def\SceneG{start_air_conditioner/2024-10/16/batch2/b-1/gopro11/8}
    \def\SceneH{iron/2024-10/16/batch2/c-2/gopro11/1}

    \def\Pa#1{\includegraphics[width=\imagewidth]{#1}}
    \def\Pb#1{\includegraphics[width=\imagewidth]{#1}}
    \def\Pc#1{\includegraphics[width=\imagewidth]{#1}}
    \def\Pd#1{\includegraphics[width=\imagewidth]{#1}}
    \def\Pe#1{\includegraphics[width=\imagewidth]{#1}}
    \def\Pf#1{\includegraphics[width=\imagewidth]{#1}}
    \def\Pg#1{\includegraphics[width=\imagewidth]{#1}}
    \def\Ph#1{\includegraphics[width=\imagewidth]{#1}}

    \vy = 0
    \node[anchor=north] at (\vx, \vy) {\Pa{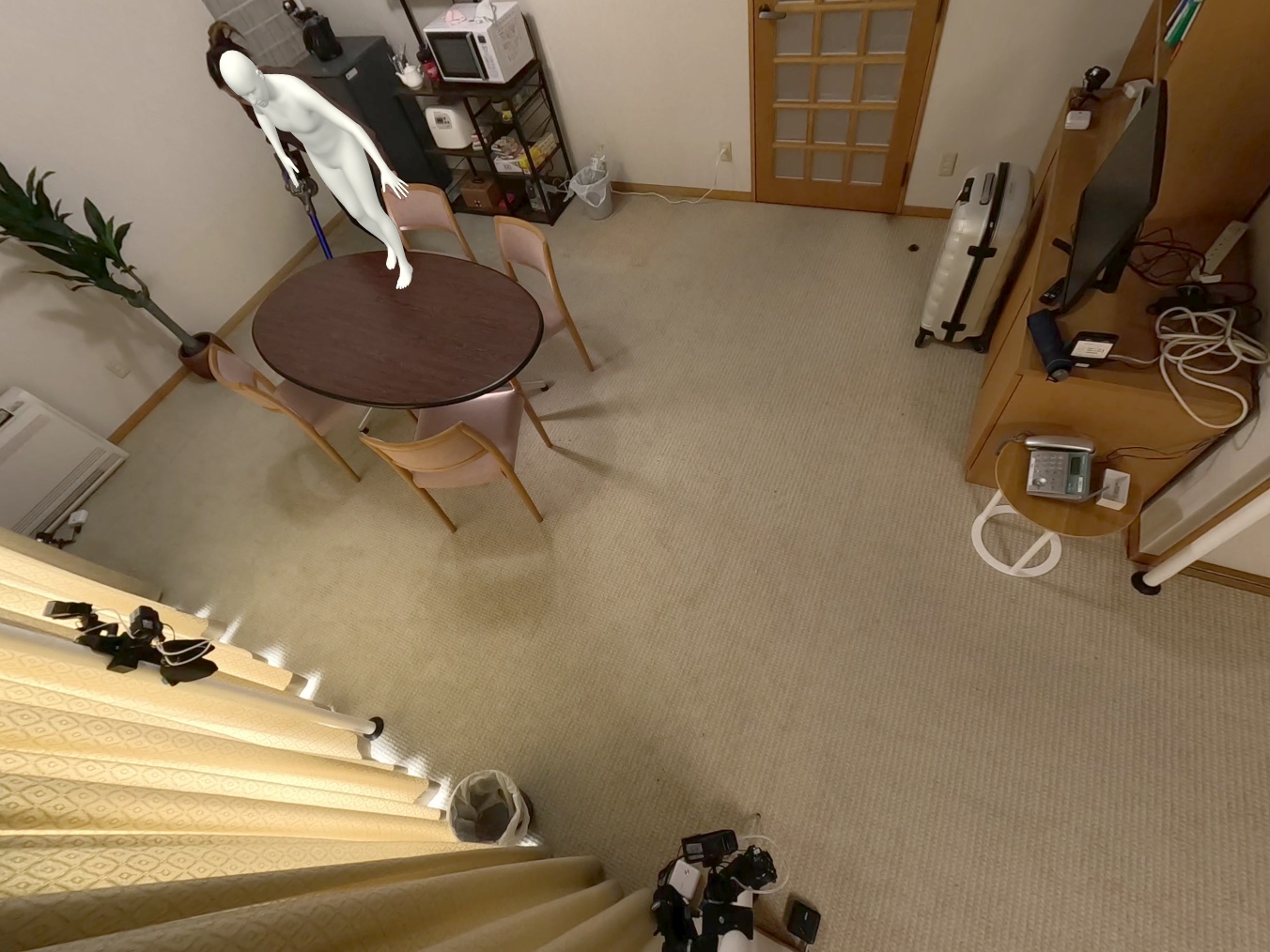}}; \advance \vx by \colwidth
    \node[anchor=north] at (\vx, \vy) {\Pa{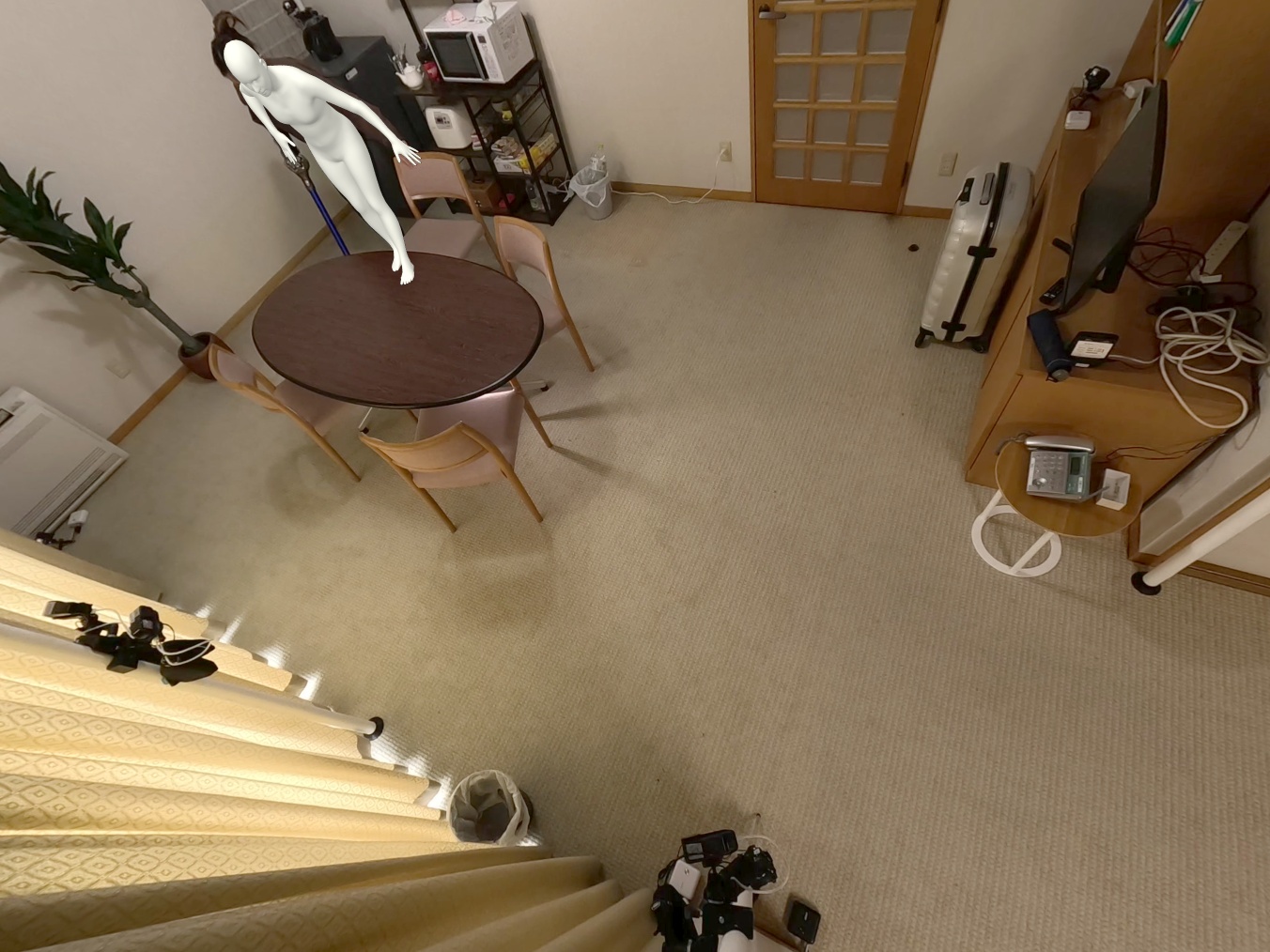}}; \advance \vx by \colwidth
    \node[anchor=north] at (\vx, \vy) {\Pa{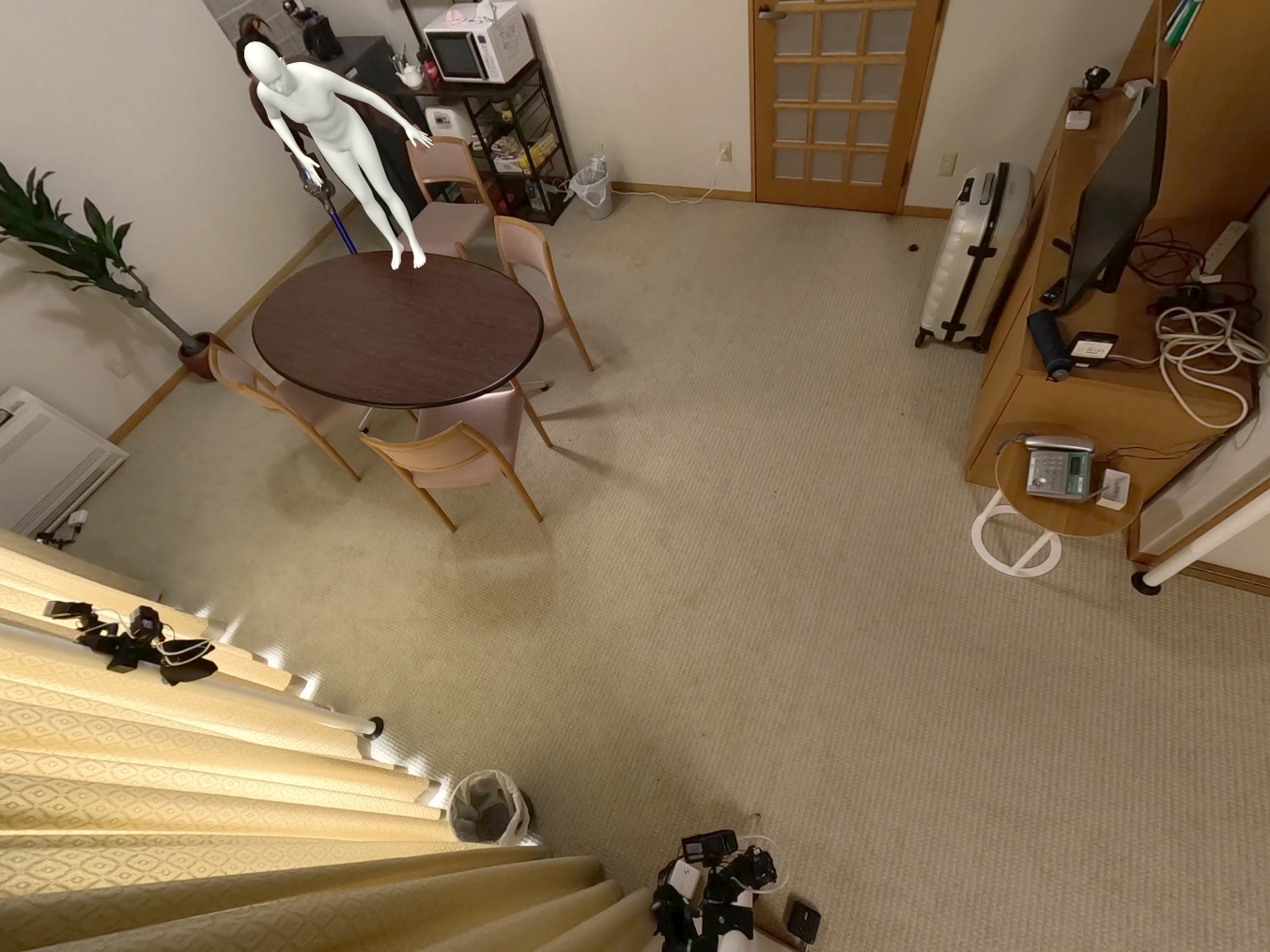}}; \advance \vx by \colwidth
    \node[anchor=north] at (\vx, \vy) {\Pa{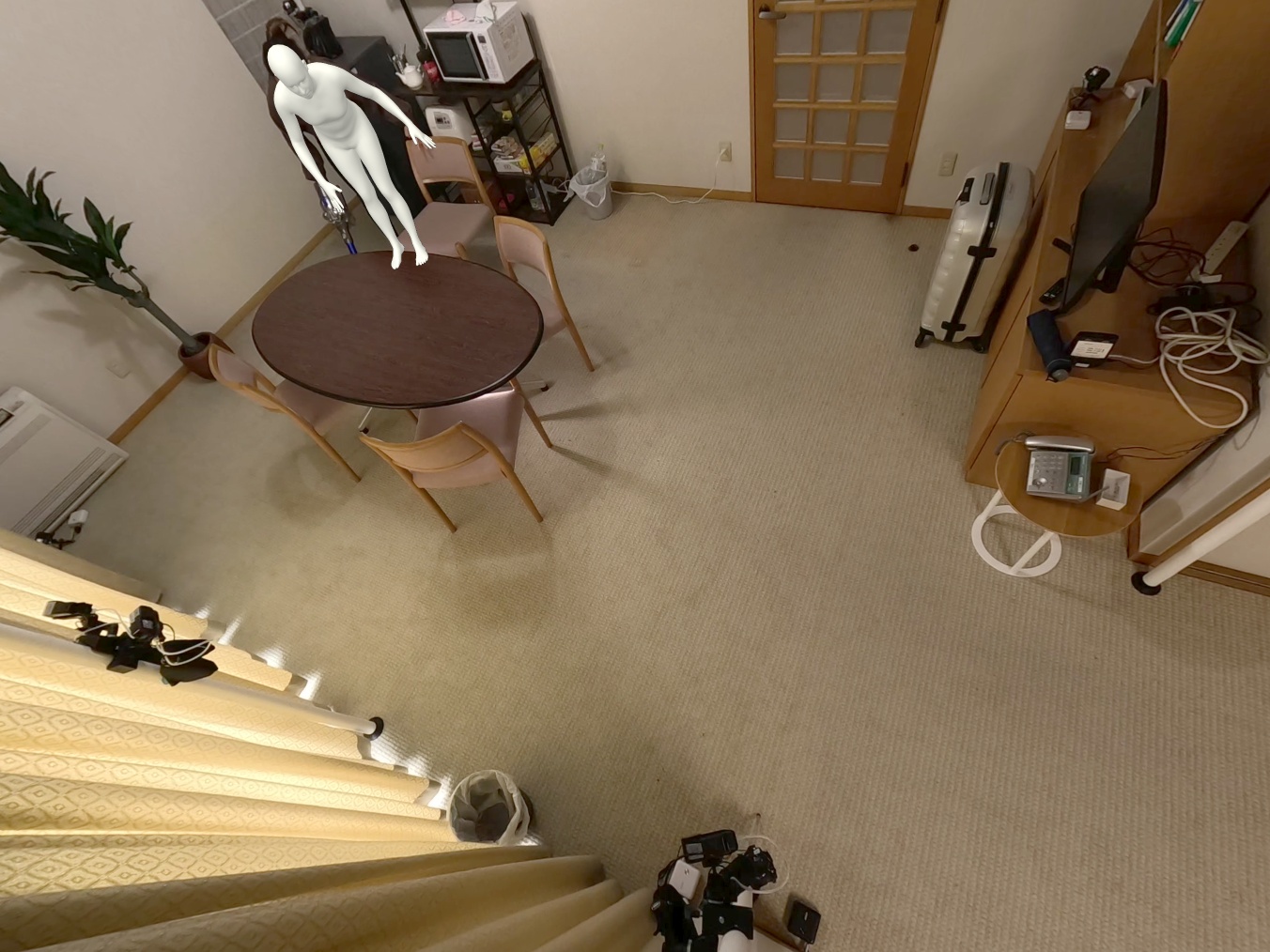}}; \advance \vx by \colwidth
    \node[anchor=north] at (\vx, \vy) {\Pa{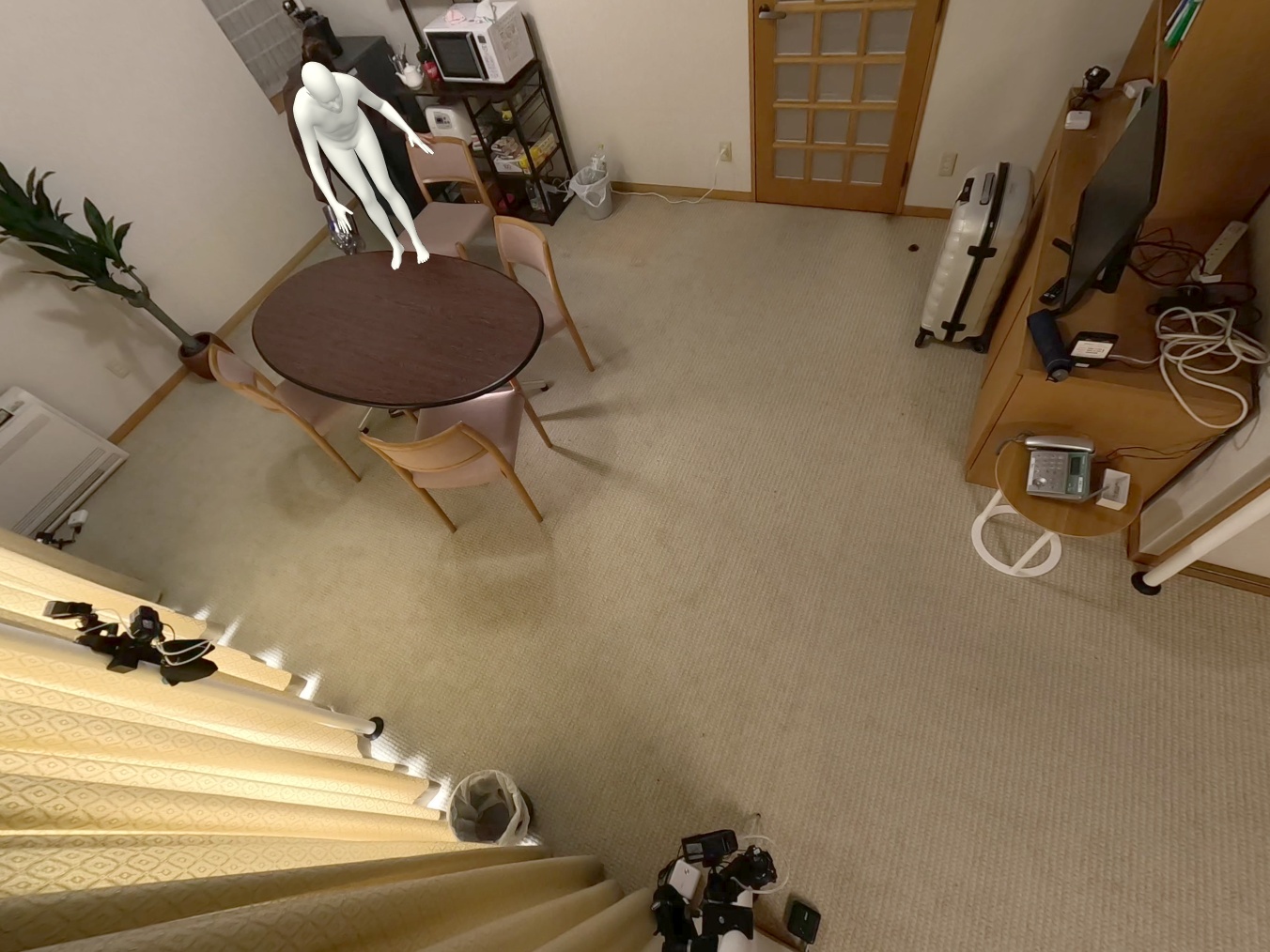}}; \advance \vx by \colwidth

    \vx = 0
    \advance \vy by \rowheight
    \node[anchor=north] at (\vx, \vy) {\Pb{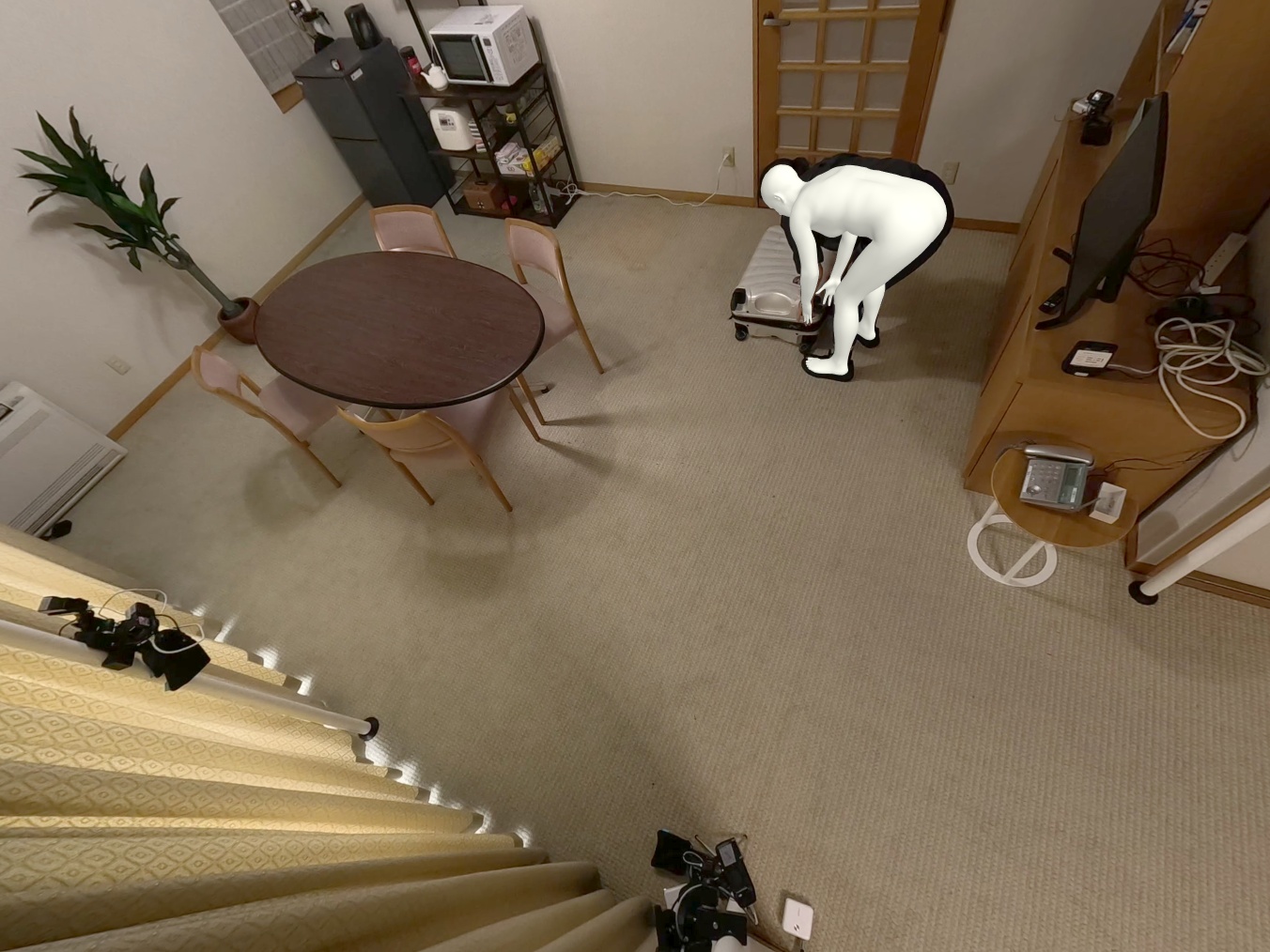}}; \advance \vx by \colwidth
    \node[anchor=north] at (\vx, \vy) {\Pb{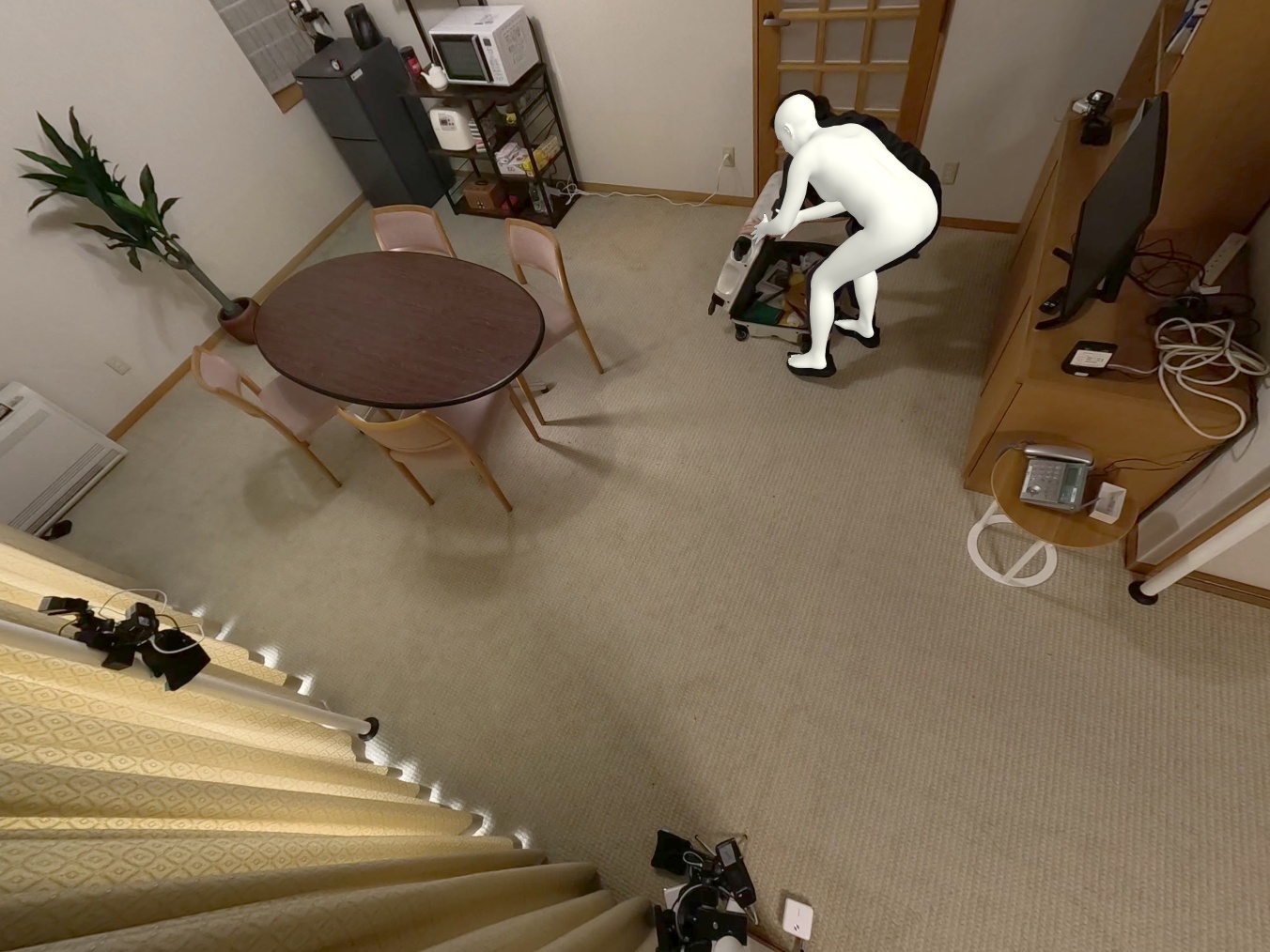}}; \advance \vx by \colwidth
    \node[anchor=north] at (\vx, \vy) {\Pb{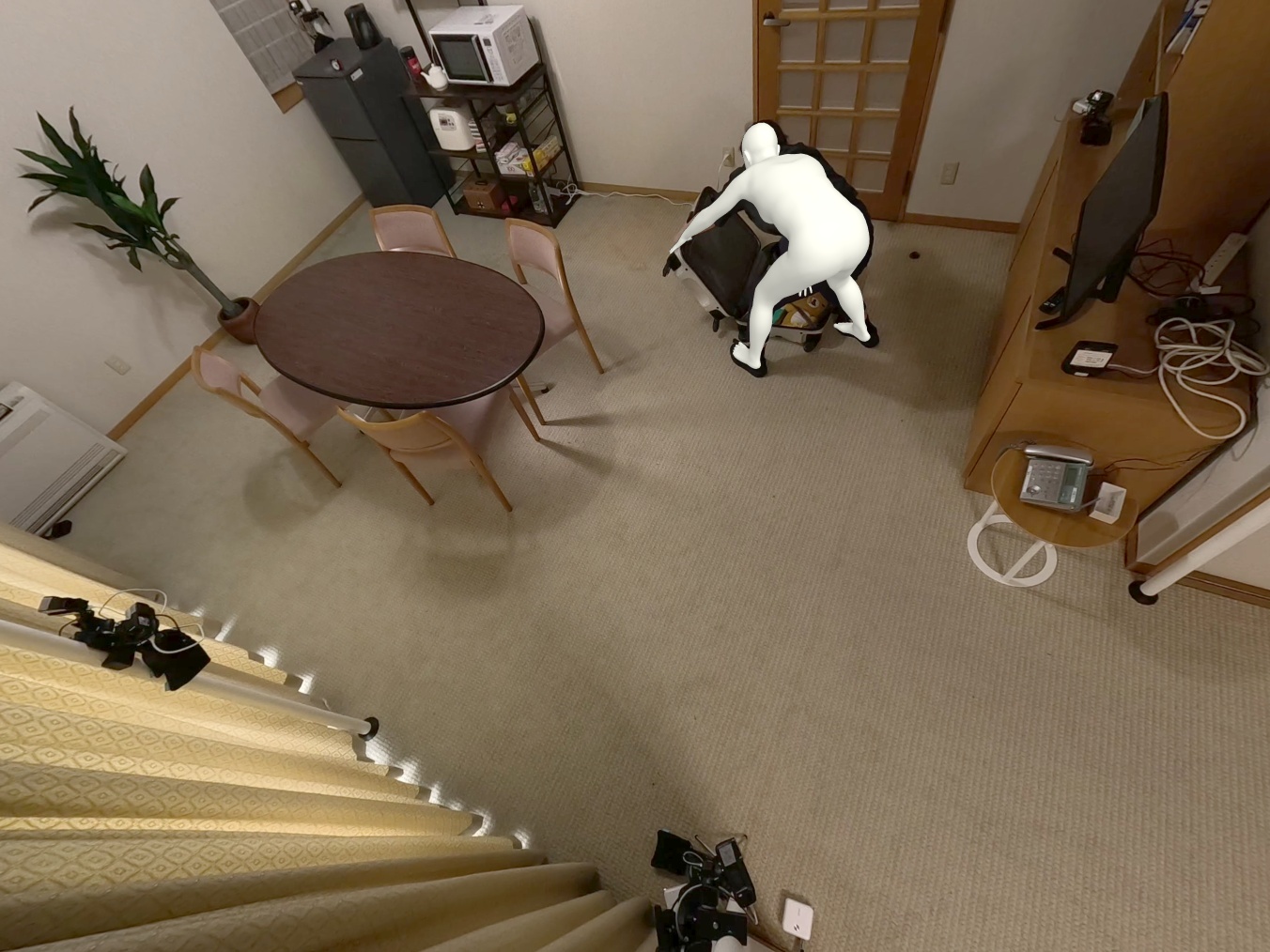}}; \advance \vx by \colwidth
    \node[anchor=north] at (\vx, \vy) {\Pb{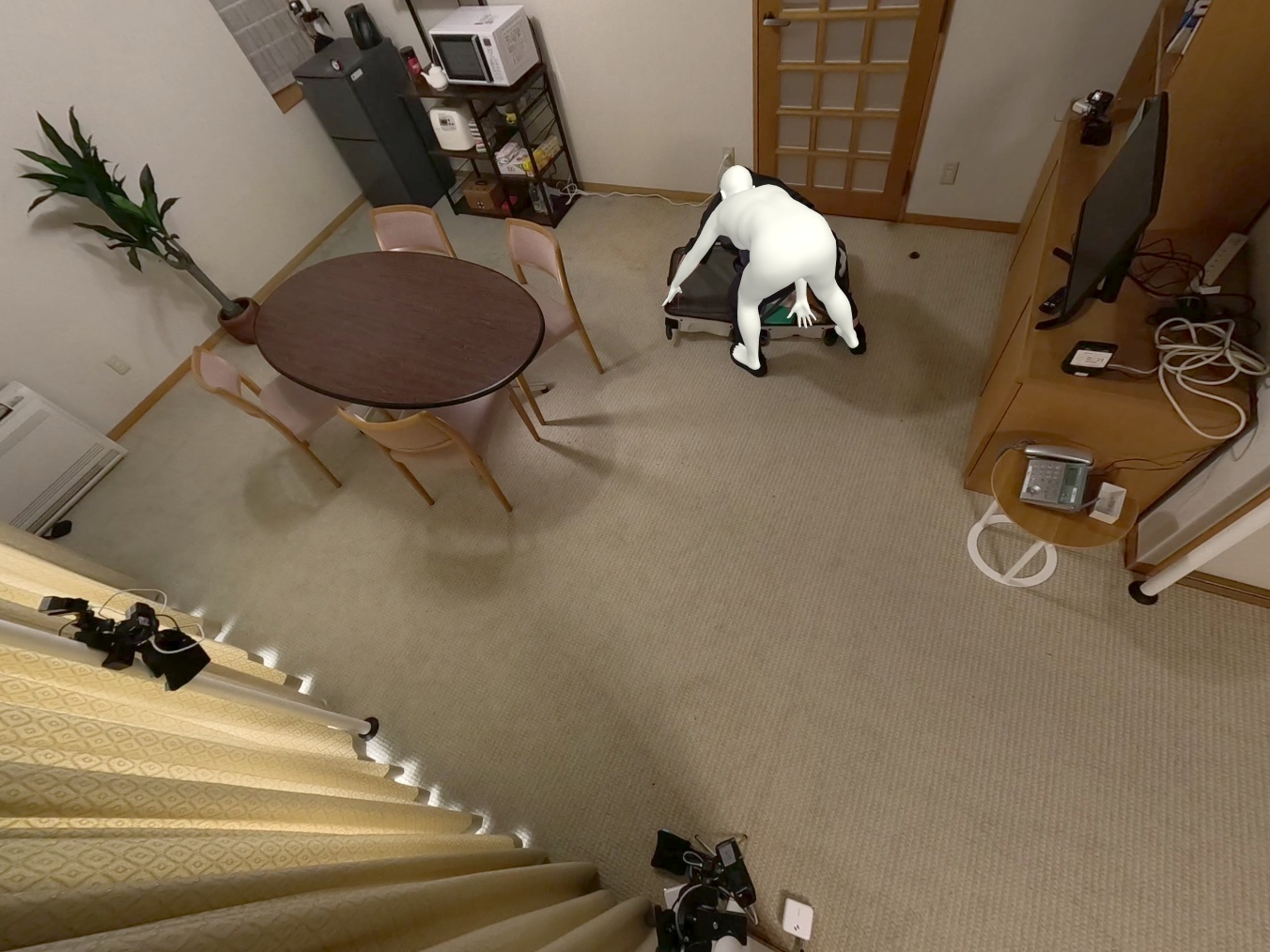}}; \advance \vx by \colwidth
    \node[anchor=north] at (\vx, \vy) {\Pb{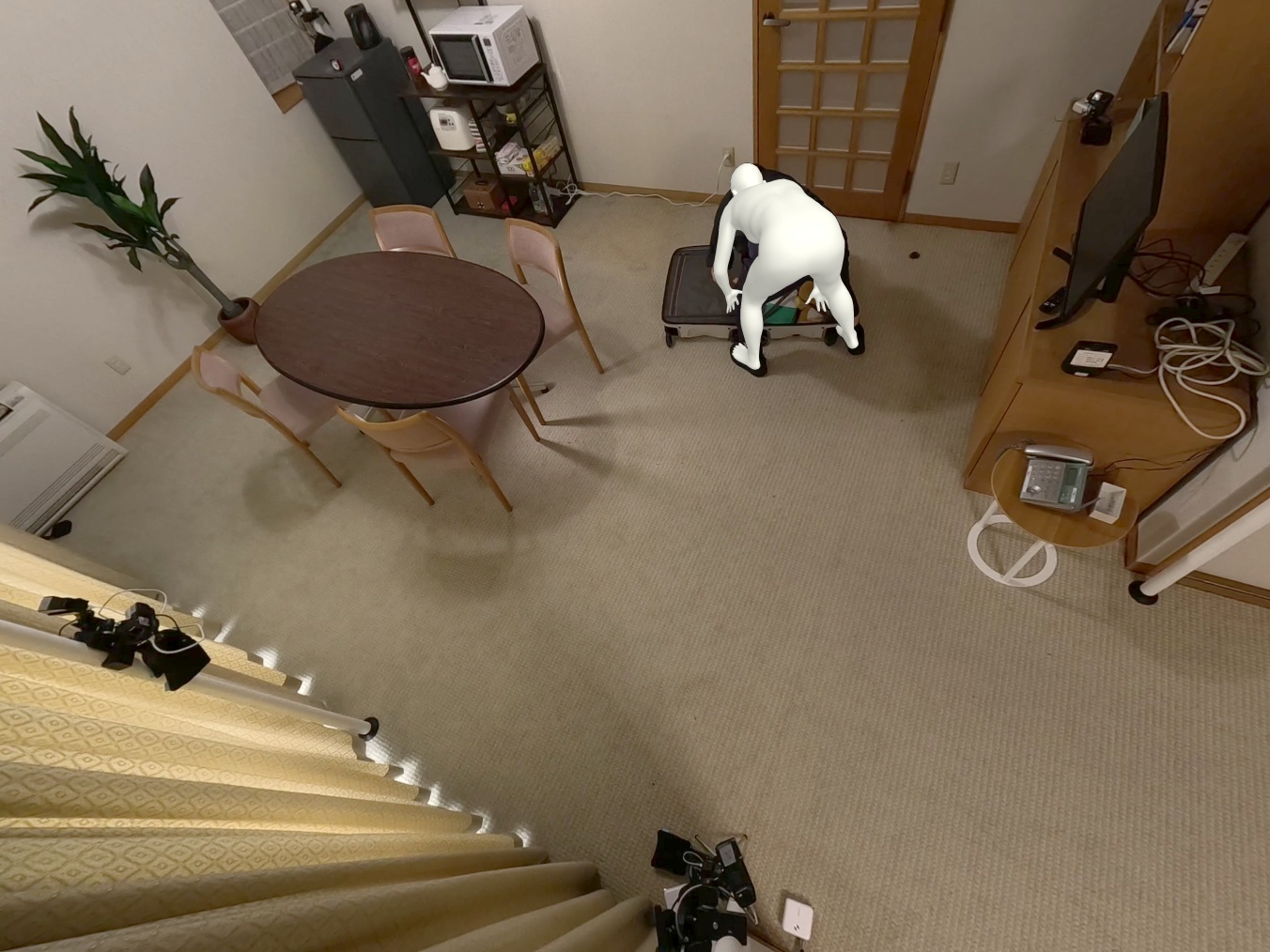}}; \advance \vx by \colwidth

    \vx = 0
    \advance \vy by \rowheight
    \node[anchor=north] at (\vx, \vy) {\Pc{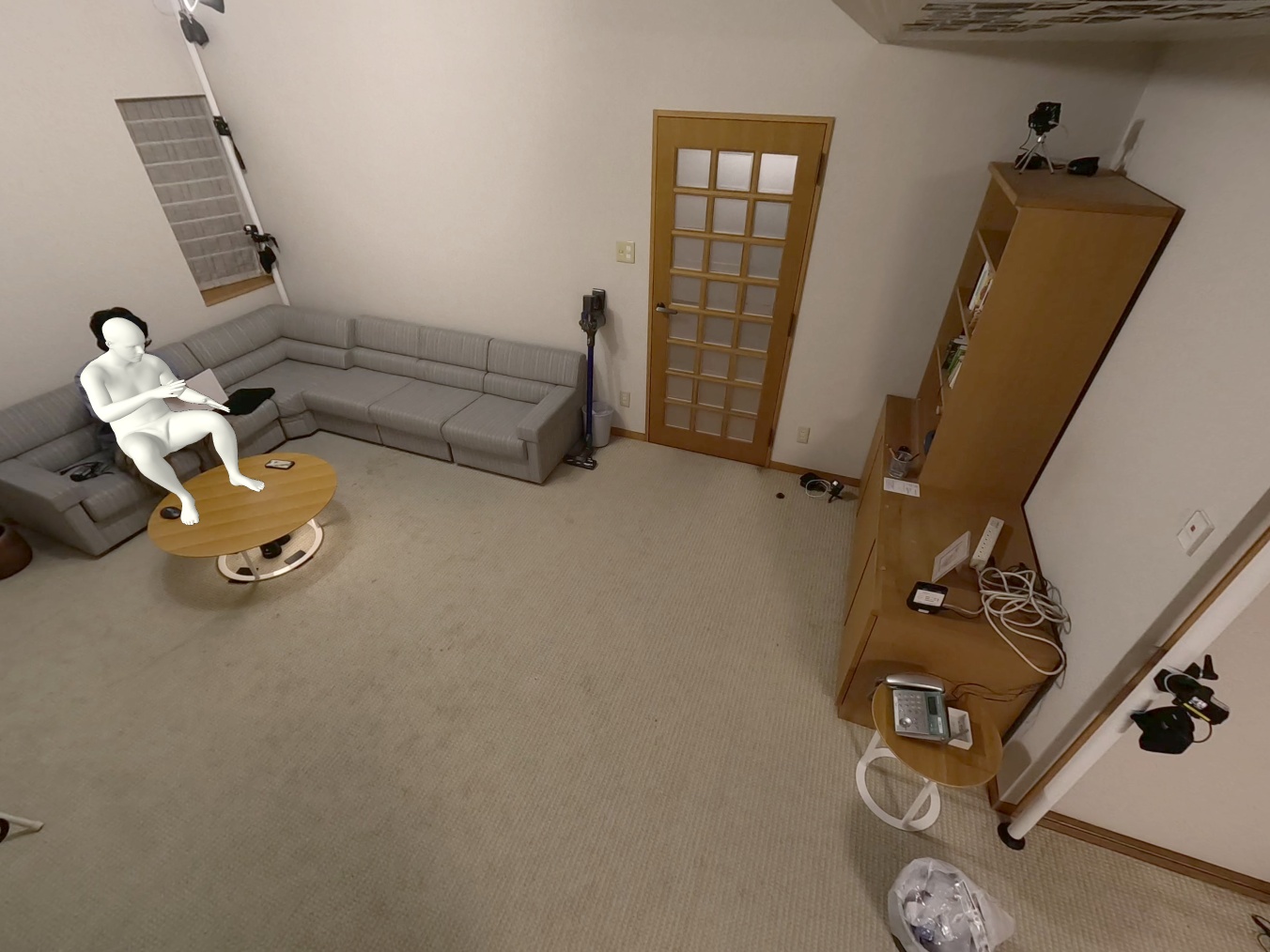}}; \advance \vx by \colwidth
    \node[anchor=north] at (\vx, \vy) {\Pc{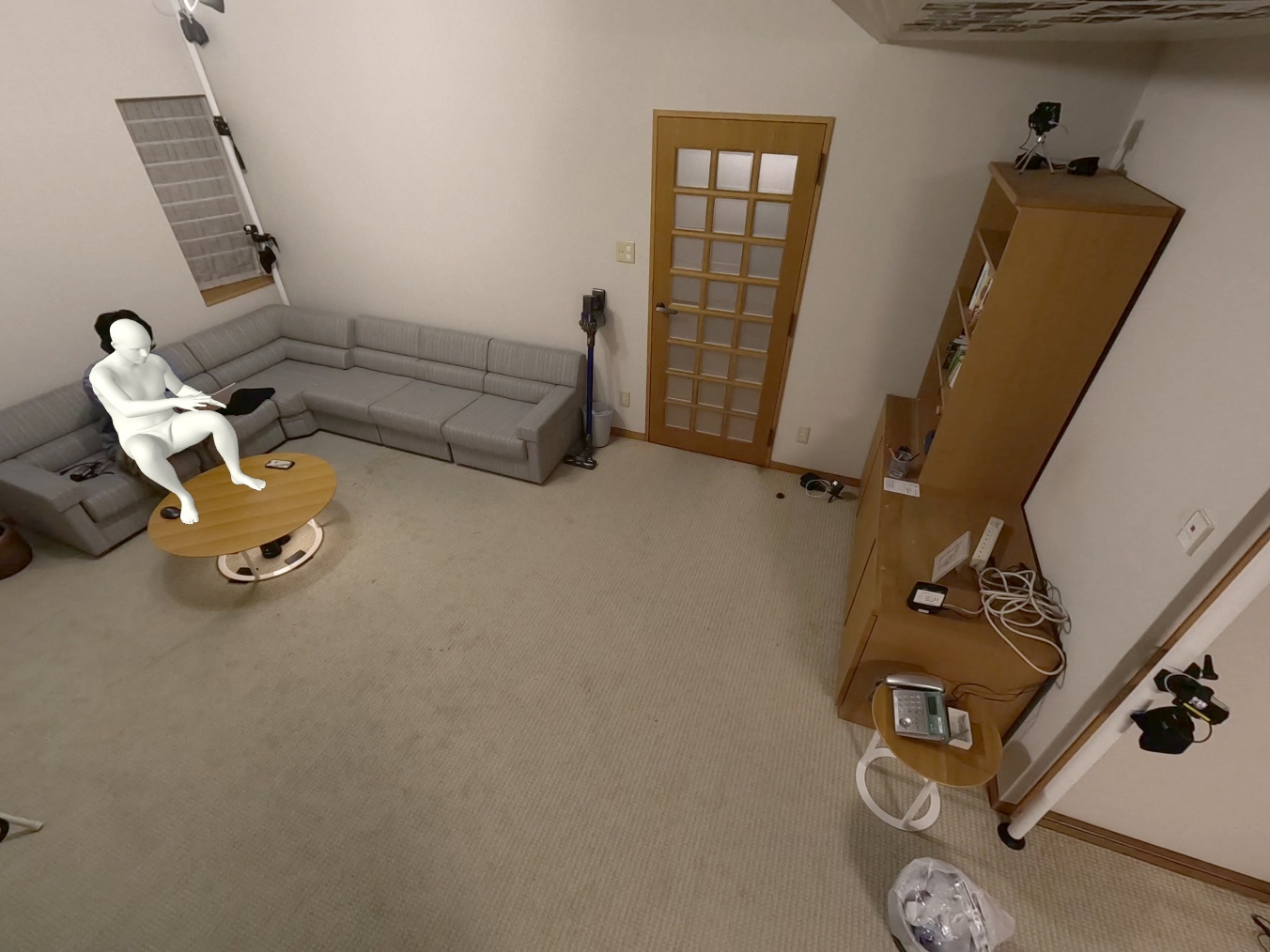}}; \advance \vx by \colwidth
    \node[anchor=north] at (\vx, \vy) {\Pc{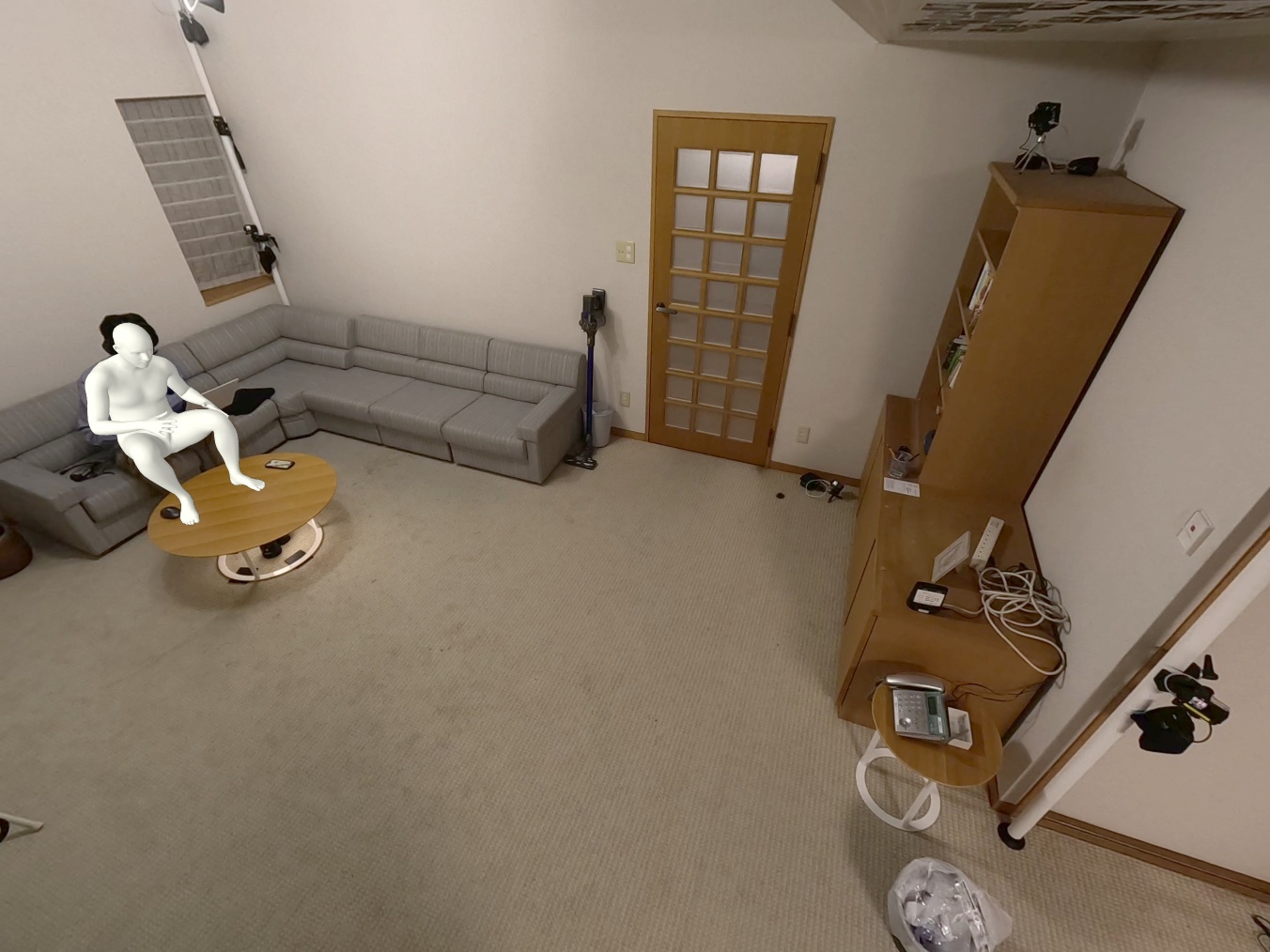}}; \advance \vx by \colwidth
    \node[anchor=north] at (\vx, \vy) {\Pc{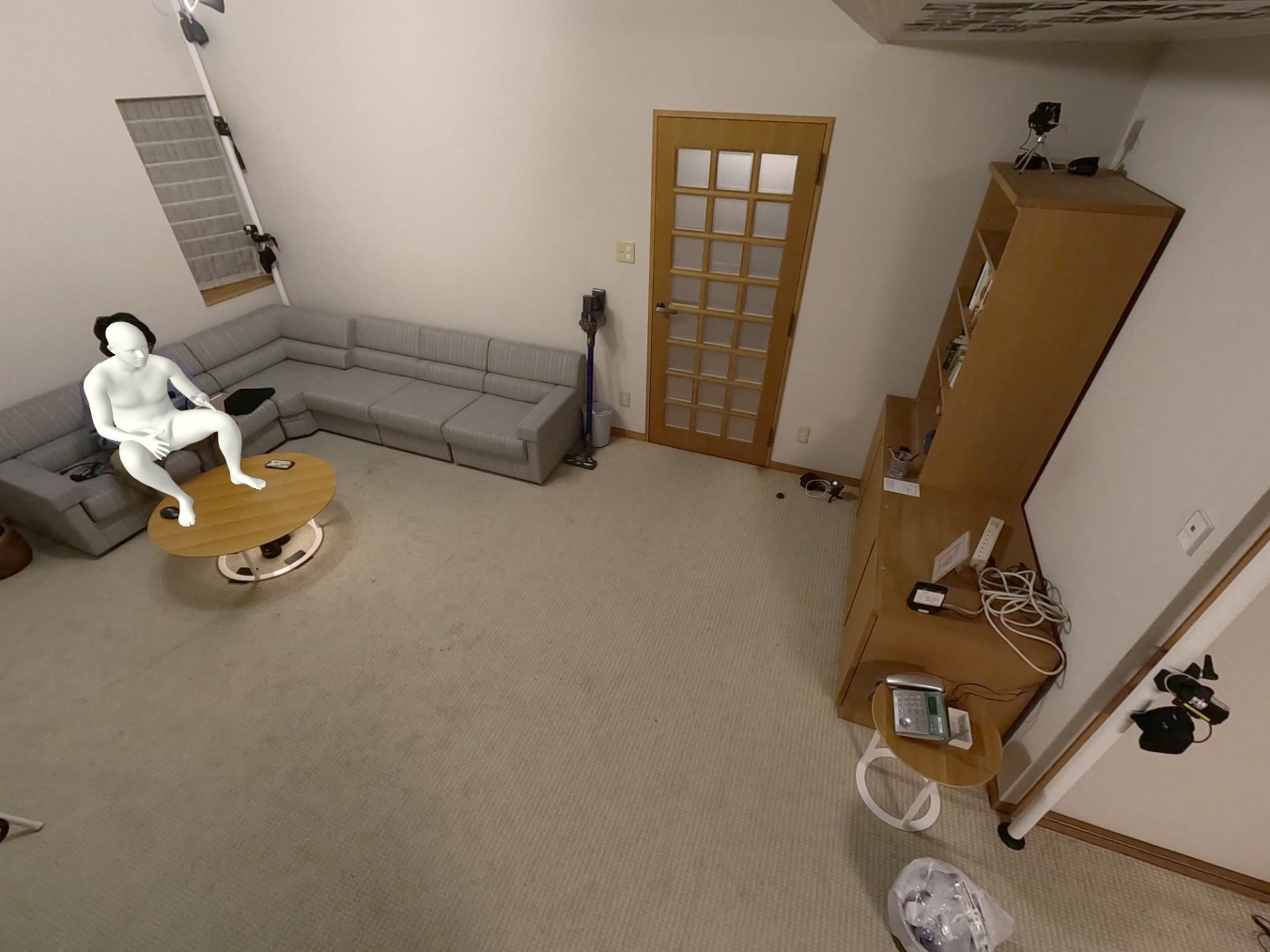}}; \advance \vx by \colwidth
    \node[anchor=north] at (\vx, \vy) {\Pc{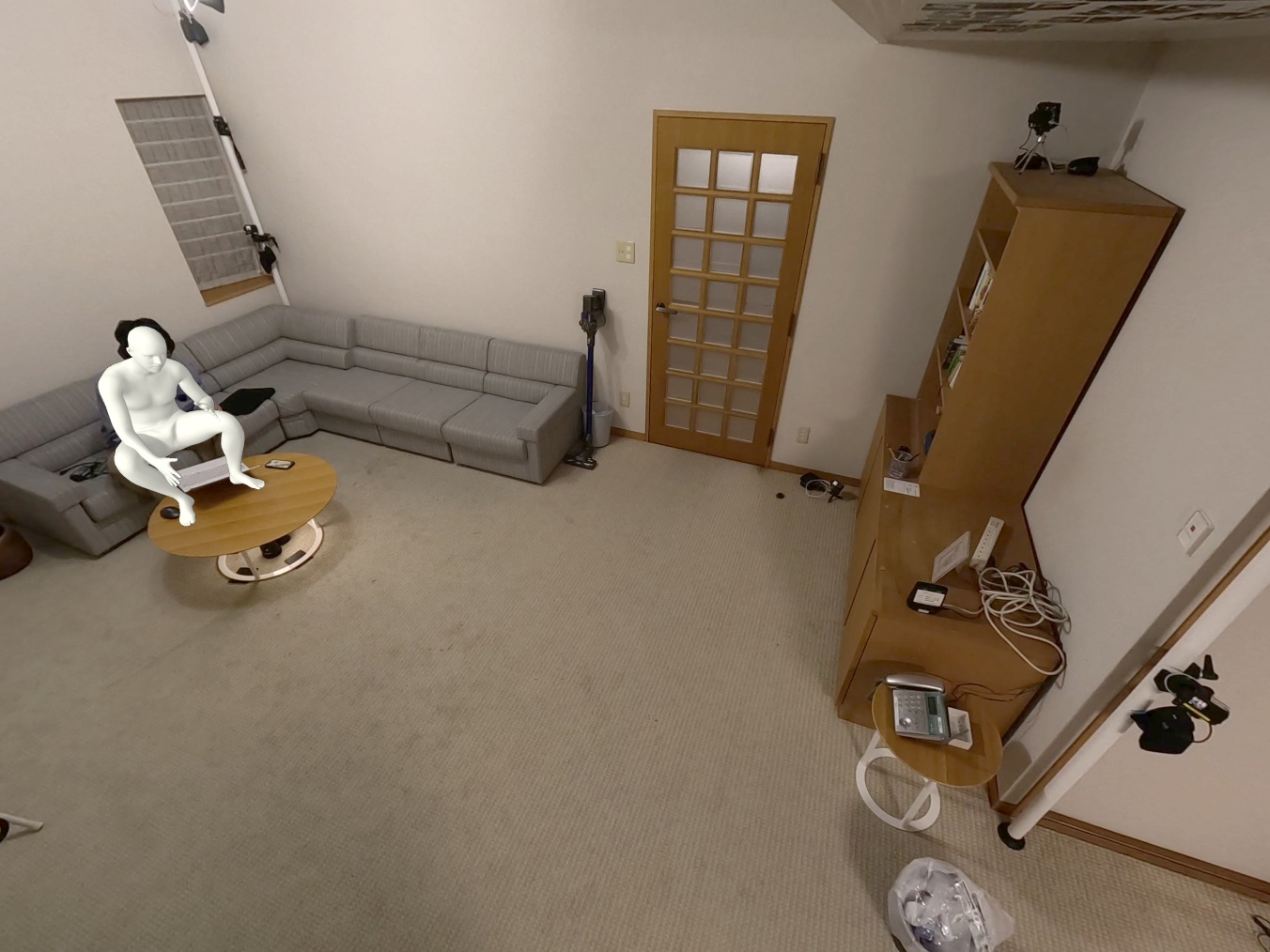}}; \advance \vx by \colwidth

    \vx = 0
    \advance \vy by \rowheight
    \node[anchor=north] at (\vx, \vy) {\Pd{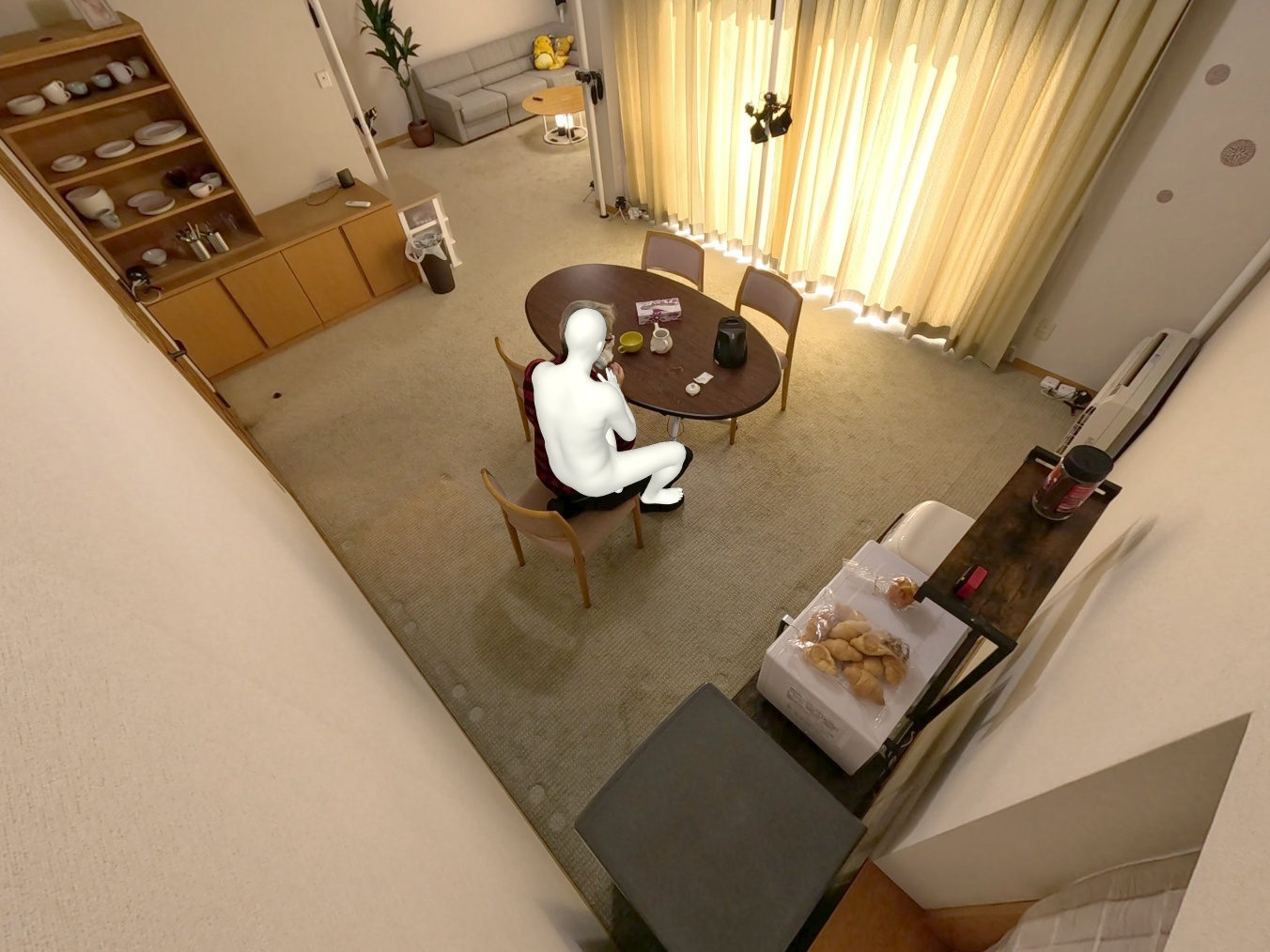}}; \advance \vx by \colwidth
    \node[anchor=north] at (\vx, \vy) {\Pd{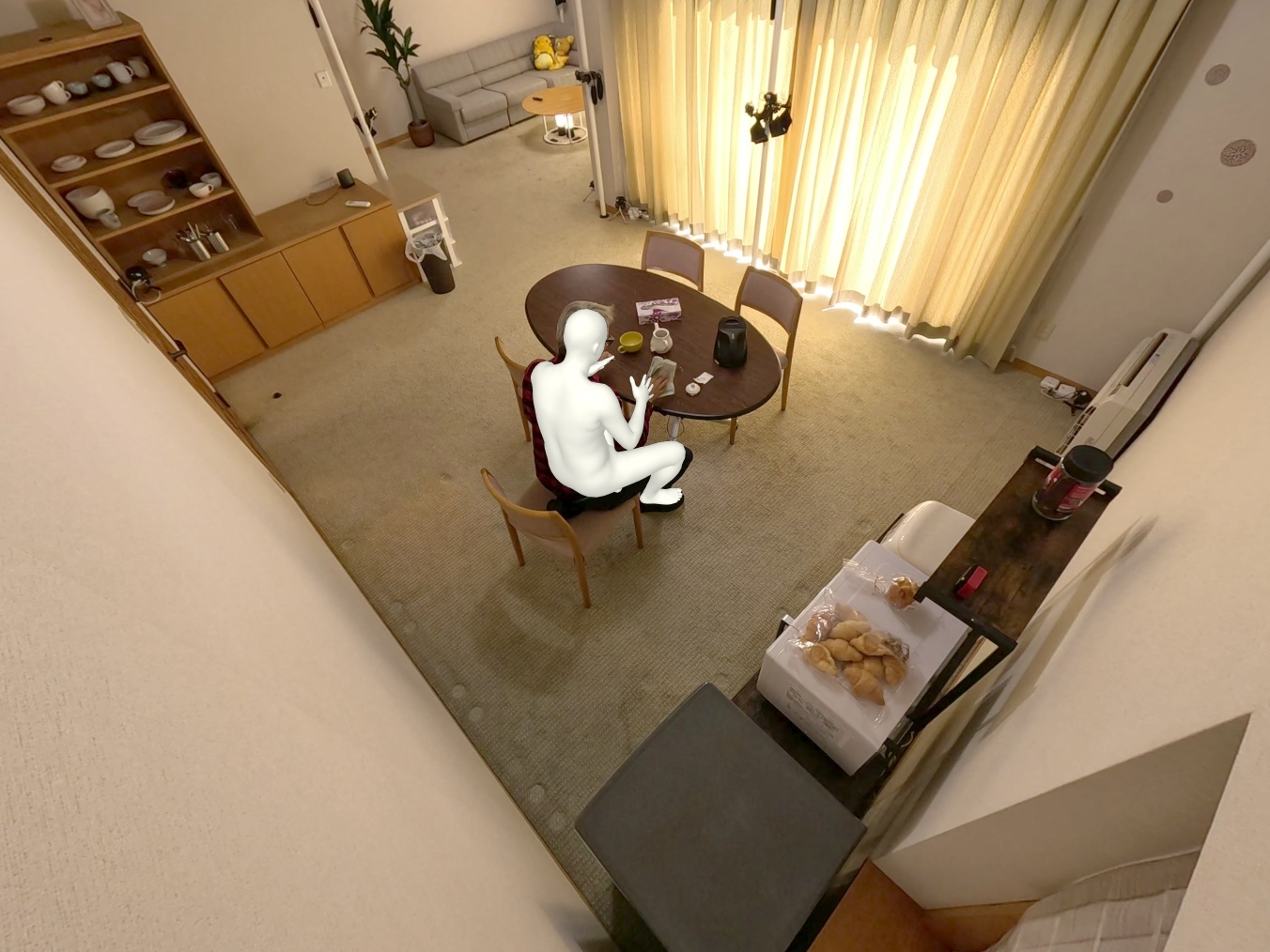}}; \advance \vx by \colwidth
    \node[anchor=north] at (\vx, \vy) {\Pd{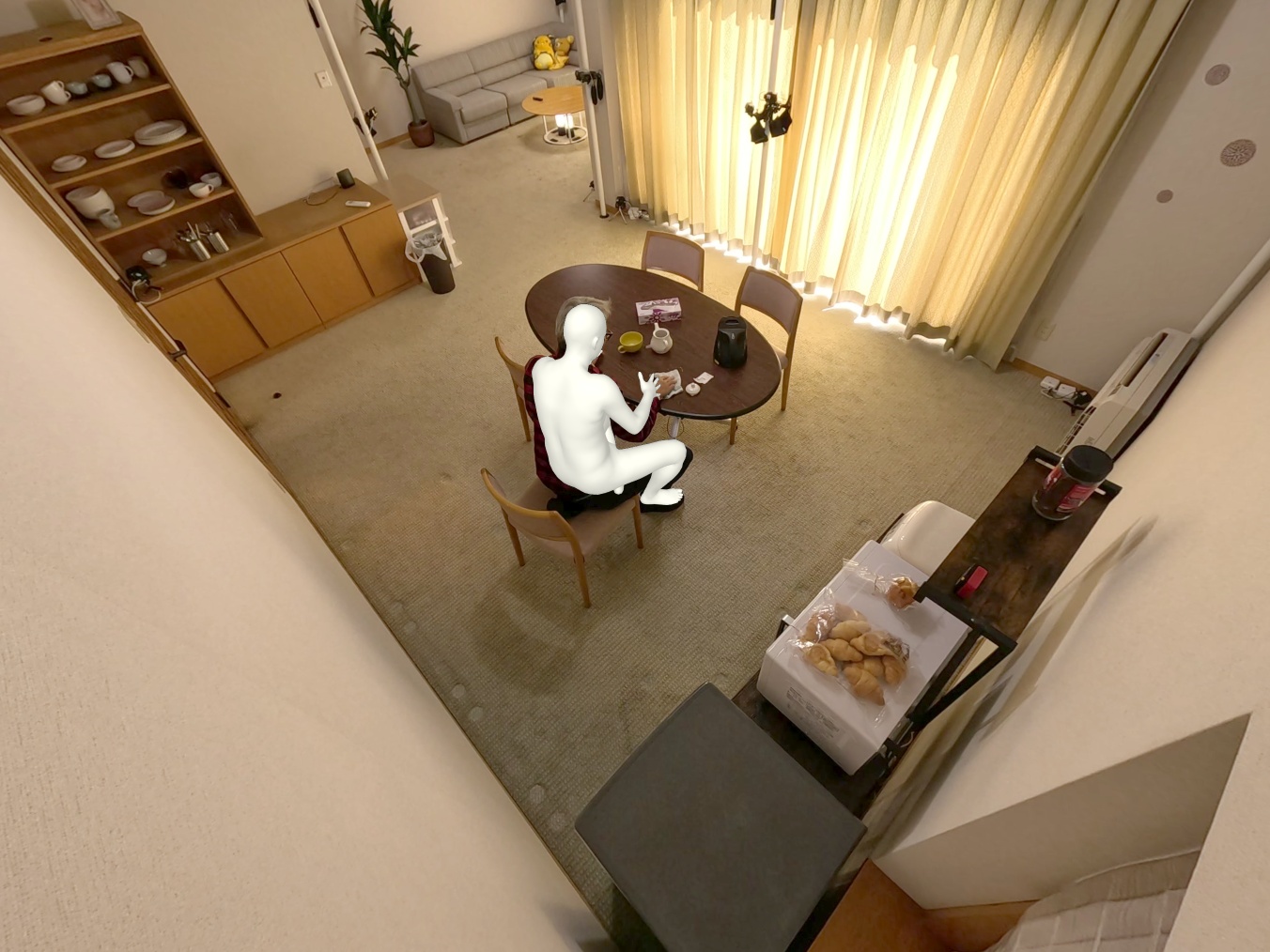}}; \advance \vx by \colwidth
    \node[anchor=north] at (\vx, \vy) {\Pd{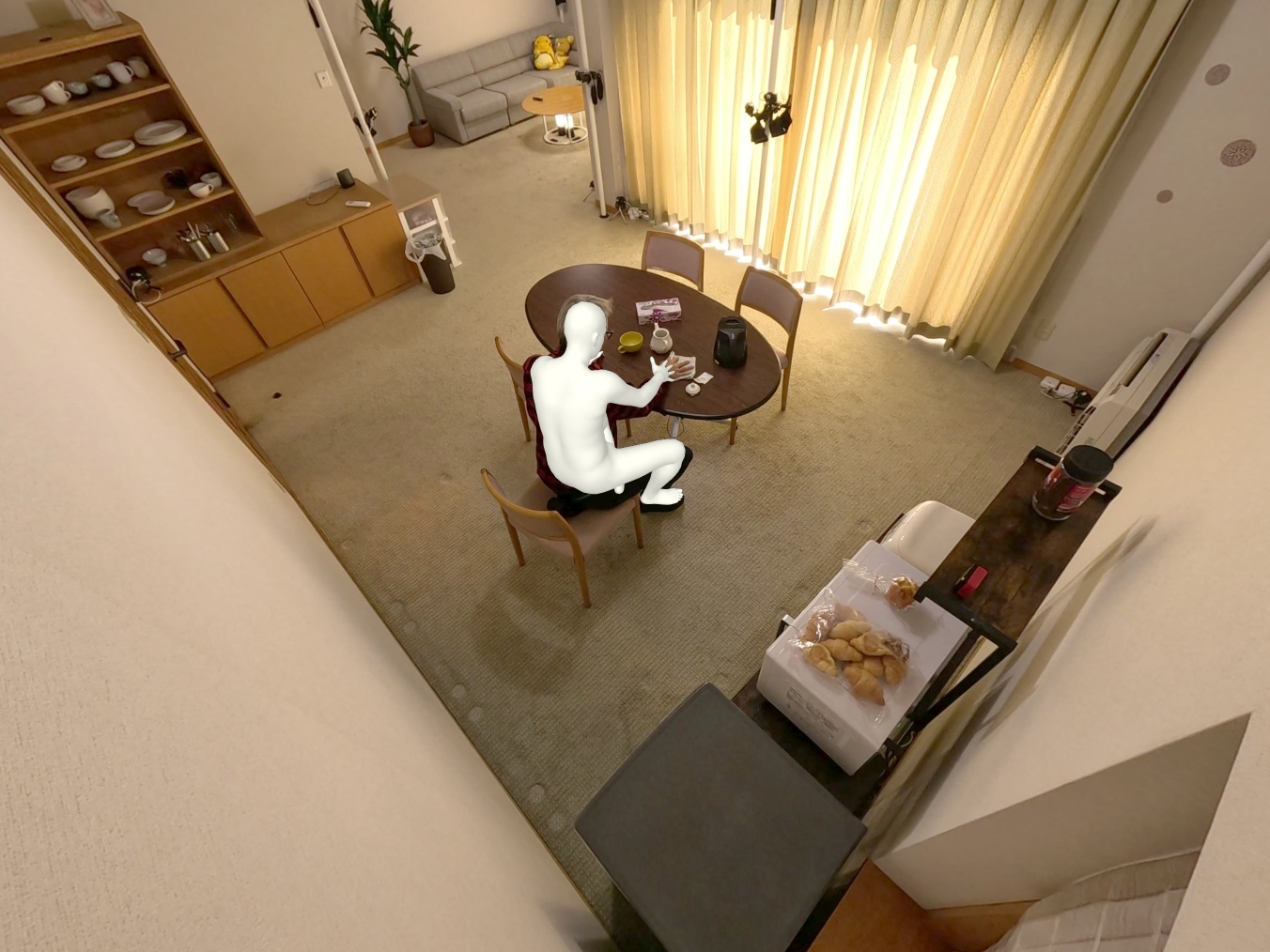}}; \advance \vx by \colwidth
    \node[anchor=north] at (\vx, \vy) {\Pd{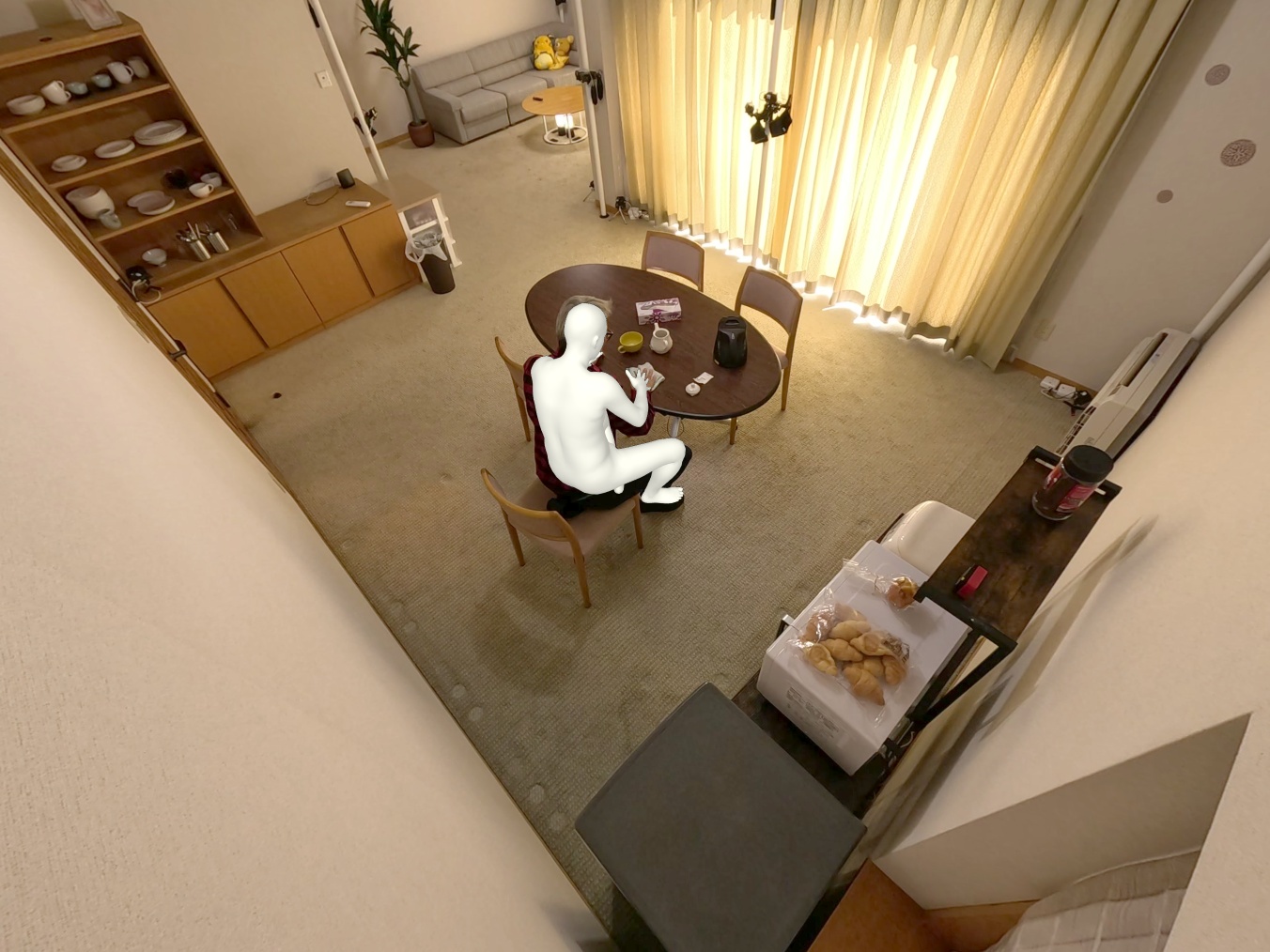}}; \advance \vx by \colwidth

    \vx = 0
    \advance \vy by \rowheight
    \node[anchor=north] at (\vx, \vy) {\Pe{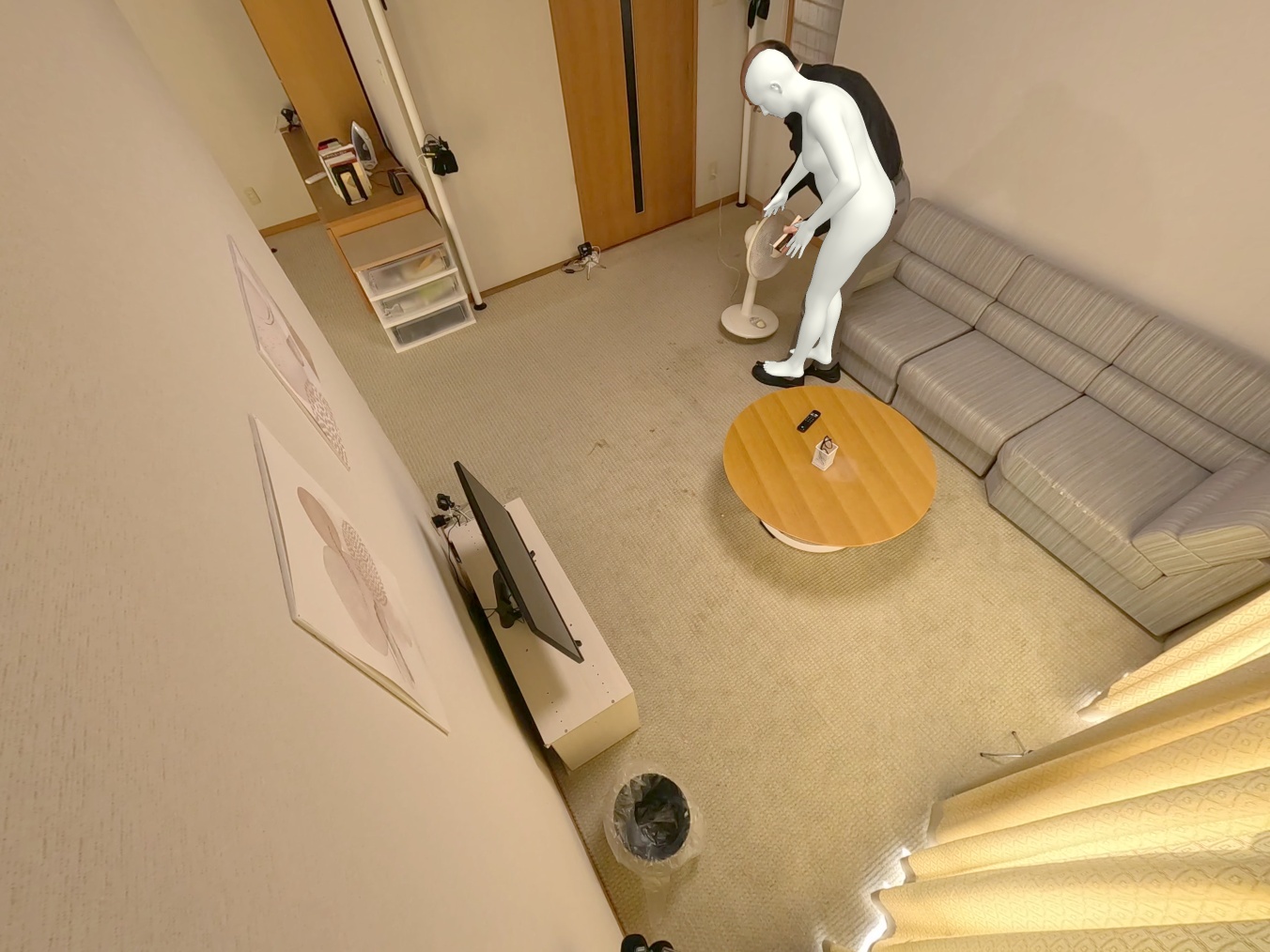}}; \advance \vx by \colwidth
    \node[anchor=north] at (\vx, \vy) {\Pe{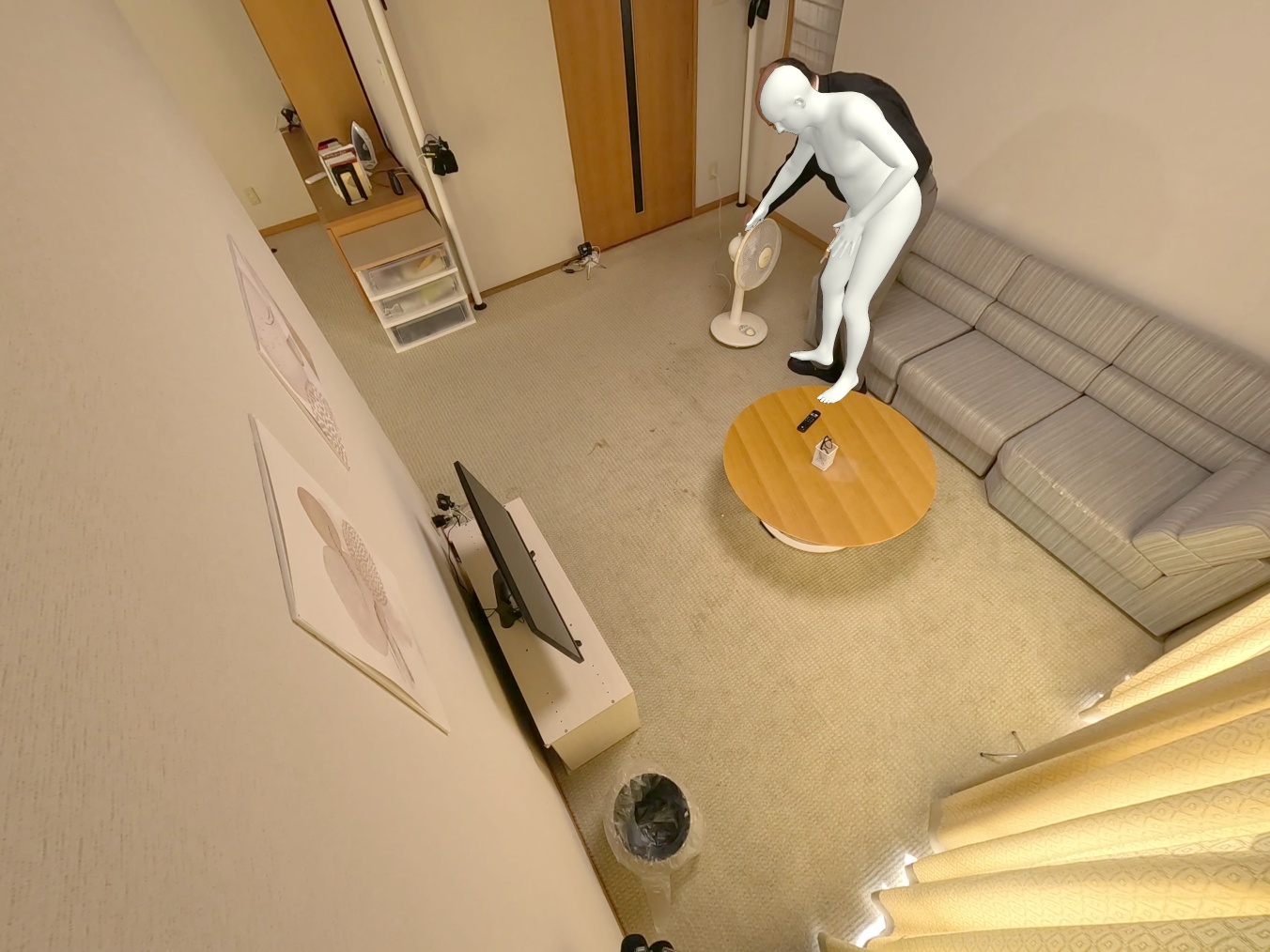}}; \advance \vx by \colwidth
    \node[anchor=north] at (\vx, \vy) {\Pe{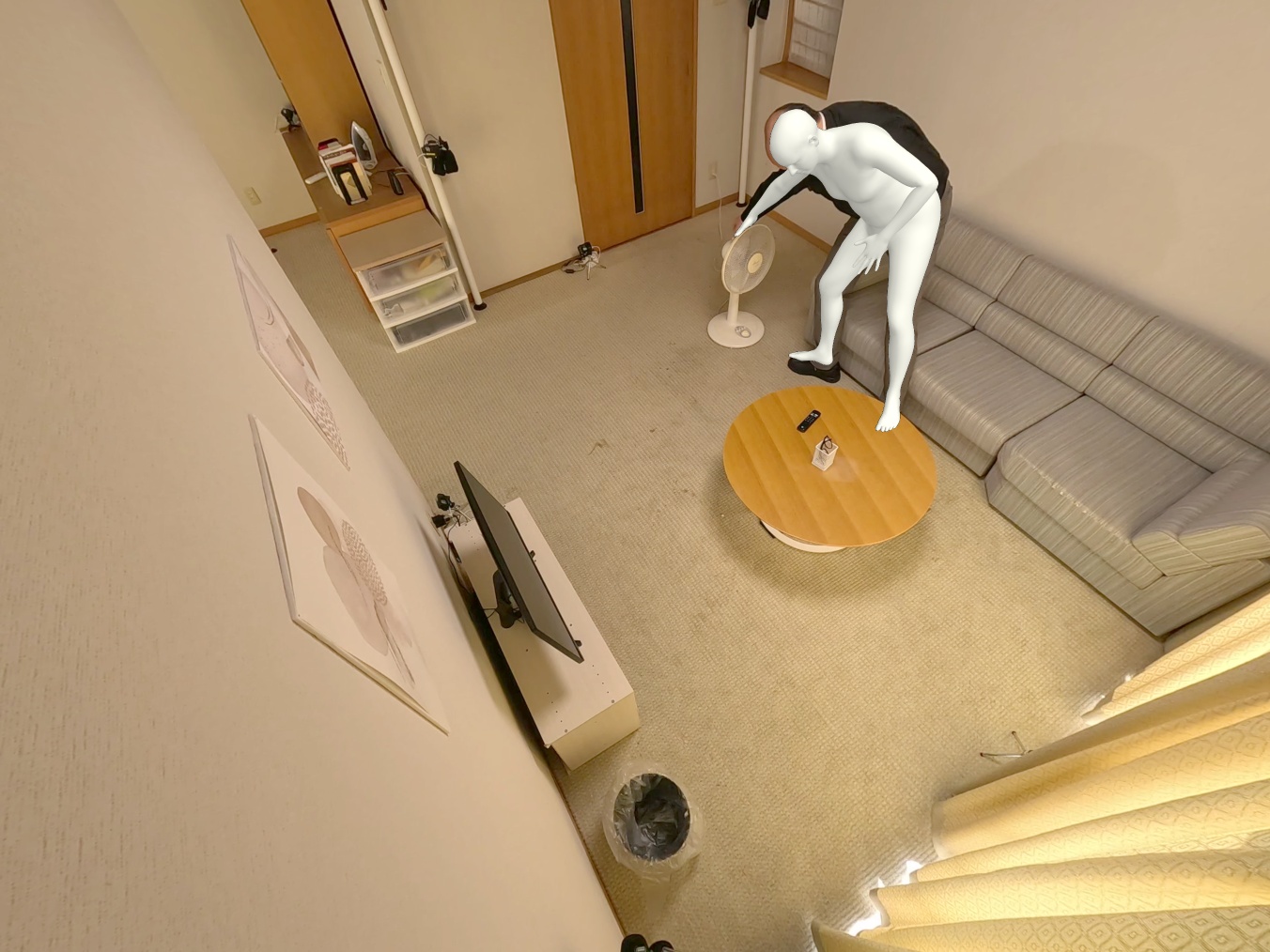}}; \advance \vx by \colwidth
    \node[anchor=north] at (\vx, \vy) {\Pe{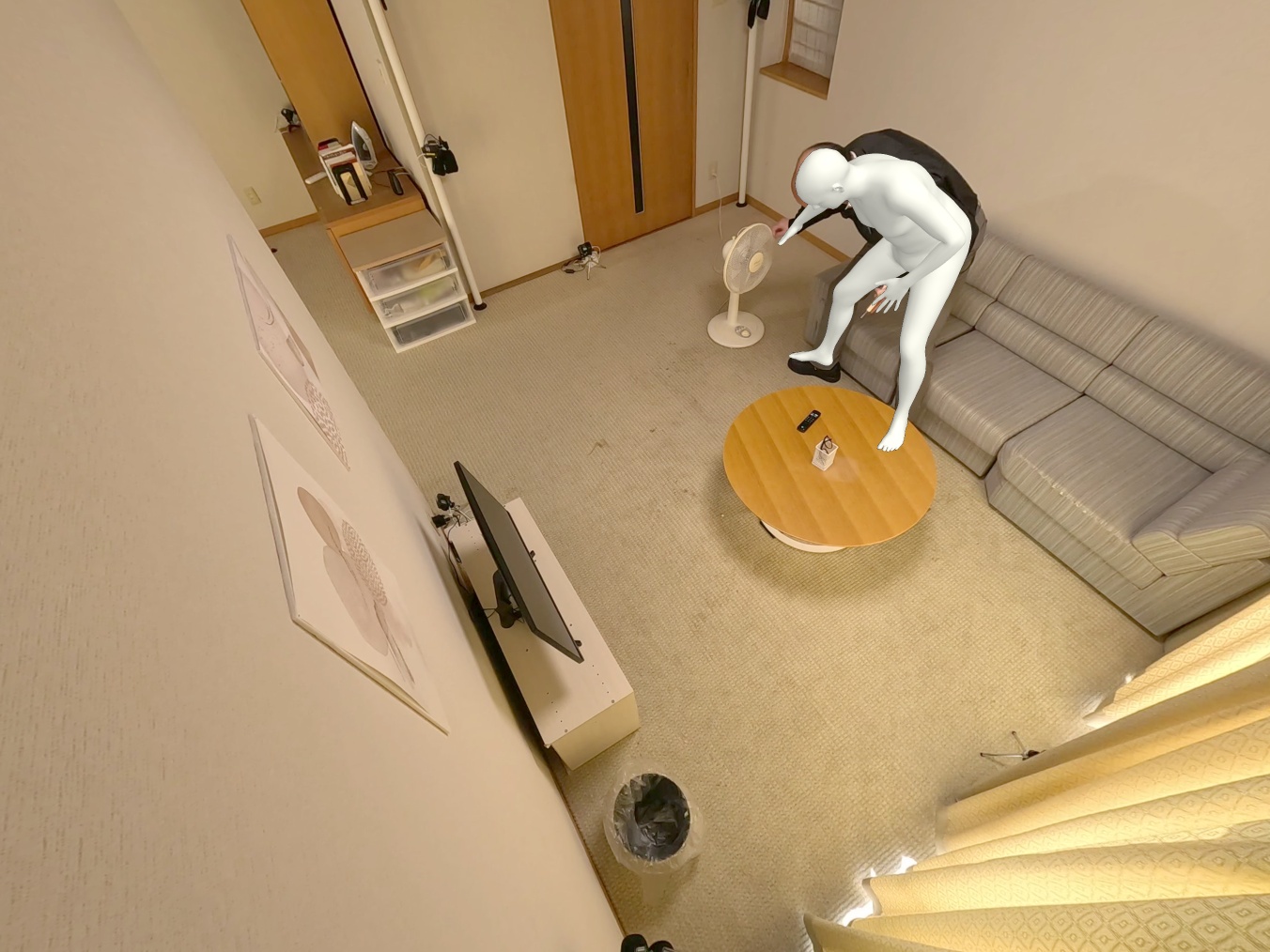}}; \advance \vx by \colwidth
    \node[anchor=north] at (\vx, \vy) {\Pe{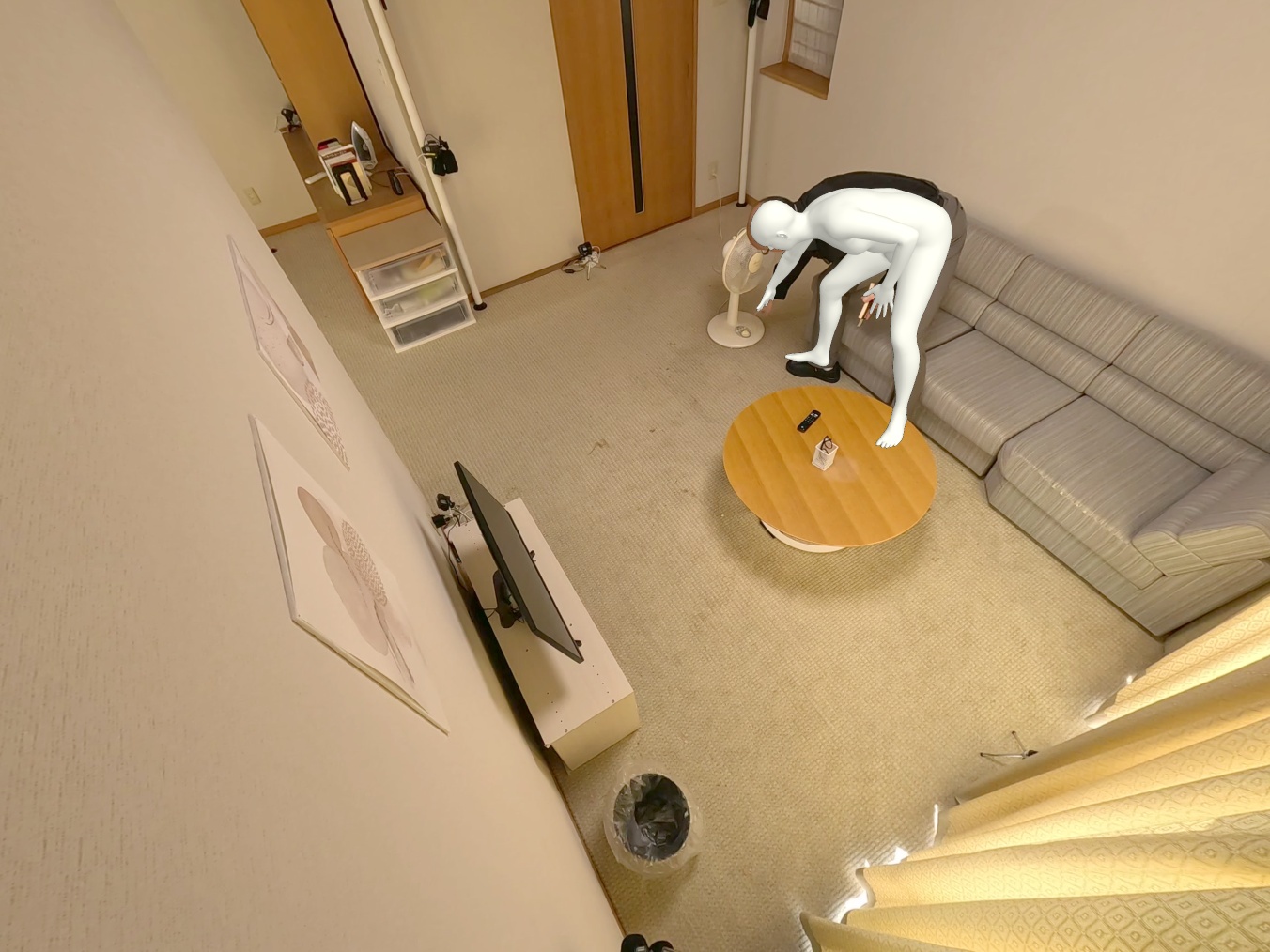}}; \advance \vx by \colwidth

    \vx = 0
    \advance \vy by \rowheight
    \node[anchor=north] at (\vx, \vy) {\Pf{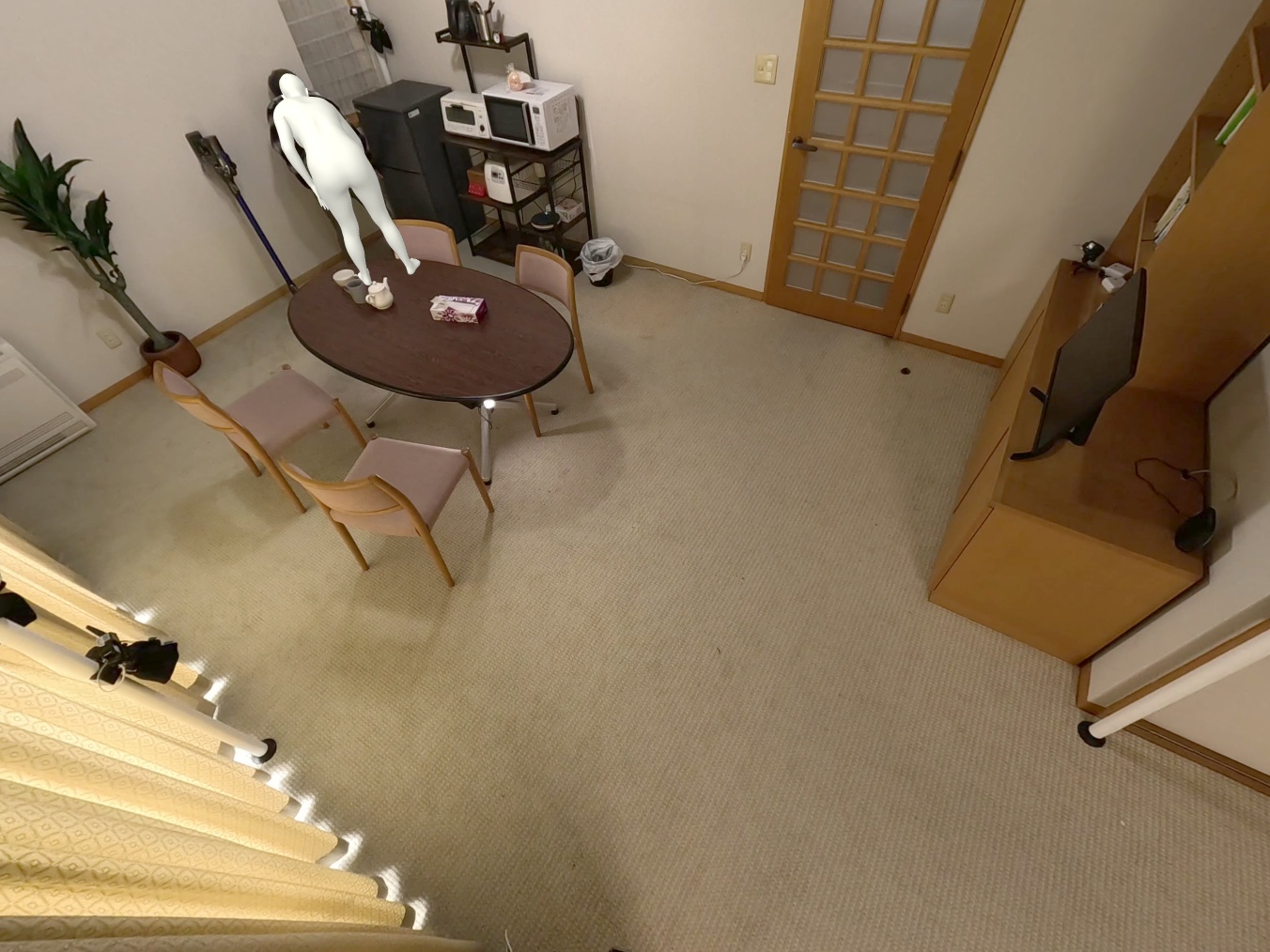}}; \advance \vx by \colwidth
    \node[anchor=north] at (\vx, \vy) {\Pf{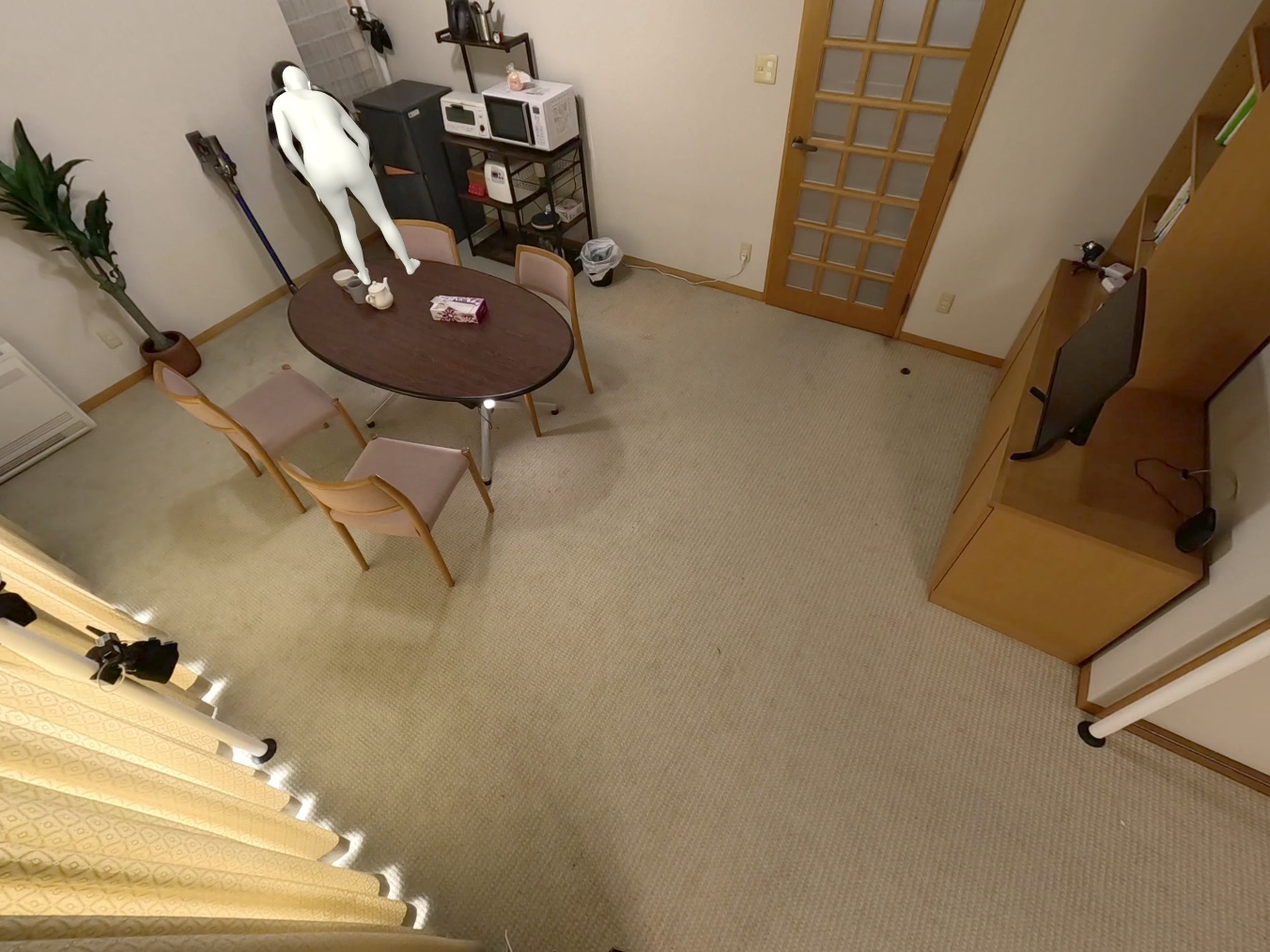}}; \advance \vx by \colwidth
    \node[anchor=north] at (\vx, \vy) {\Pf{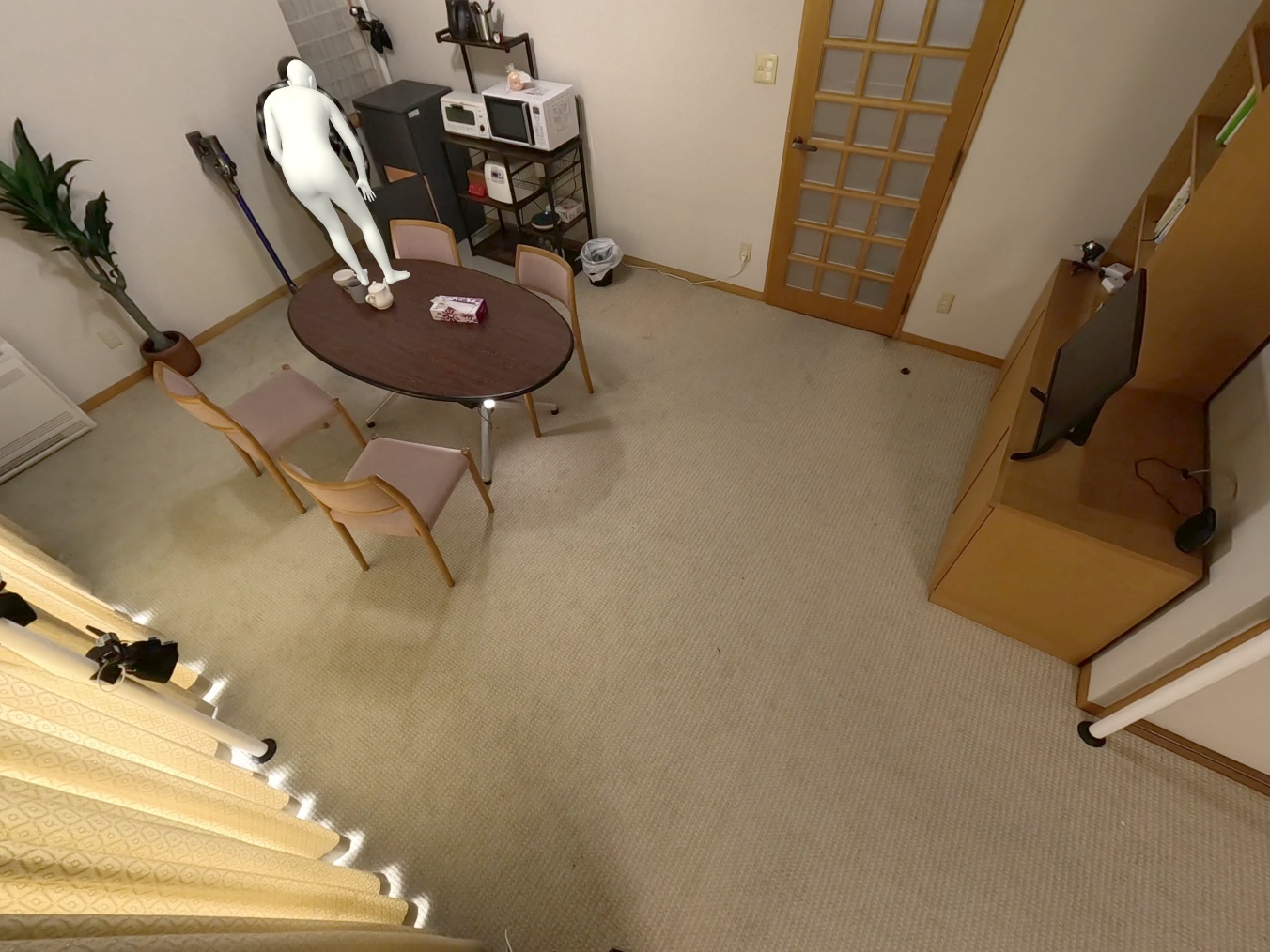}}; \advance \vx by \colwidth
    \node[anchor=north] at (\vx, \vy) {\Pf{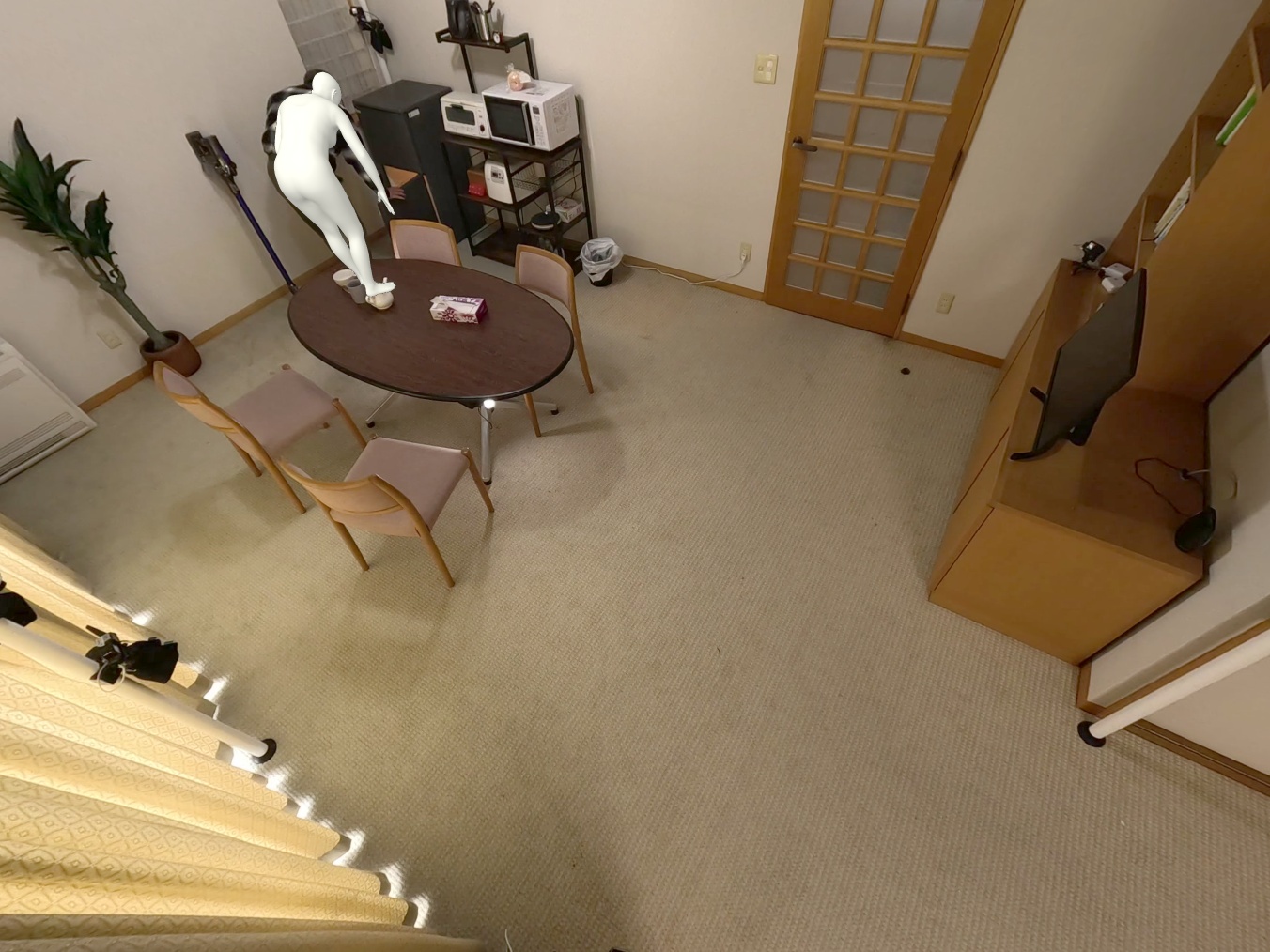}}; \advance \vx by \colwidth
    \node[anchor=north] at (\vx, \vy) {\Pf{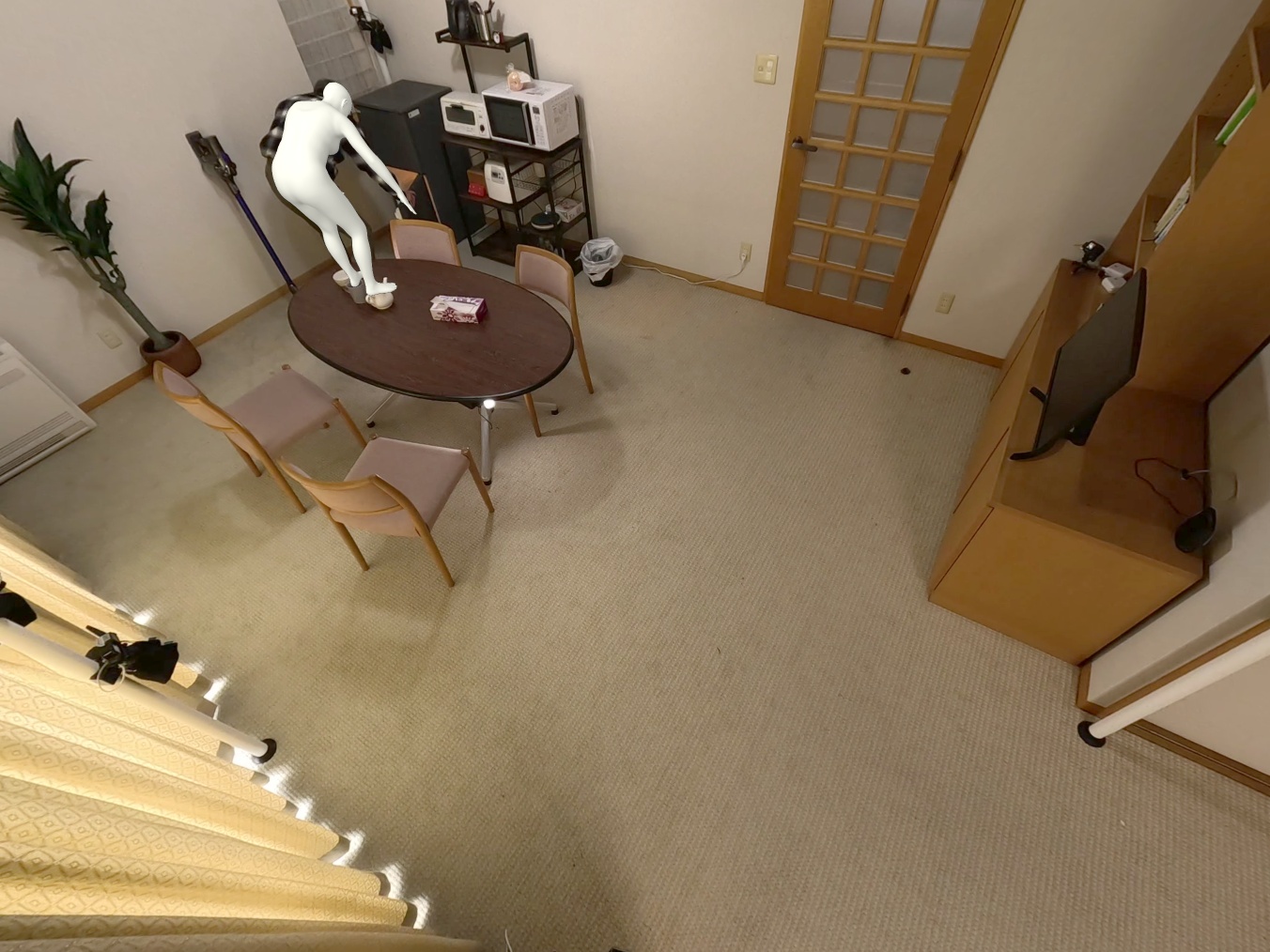}}; \advance \vx by \colwidth

    \vx = 0
    \advance \vy by \rowheight
    \node[anchor=north] at (\vx, \vy) {\Pg{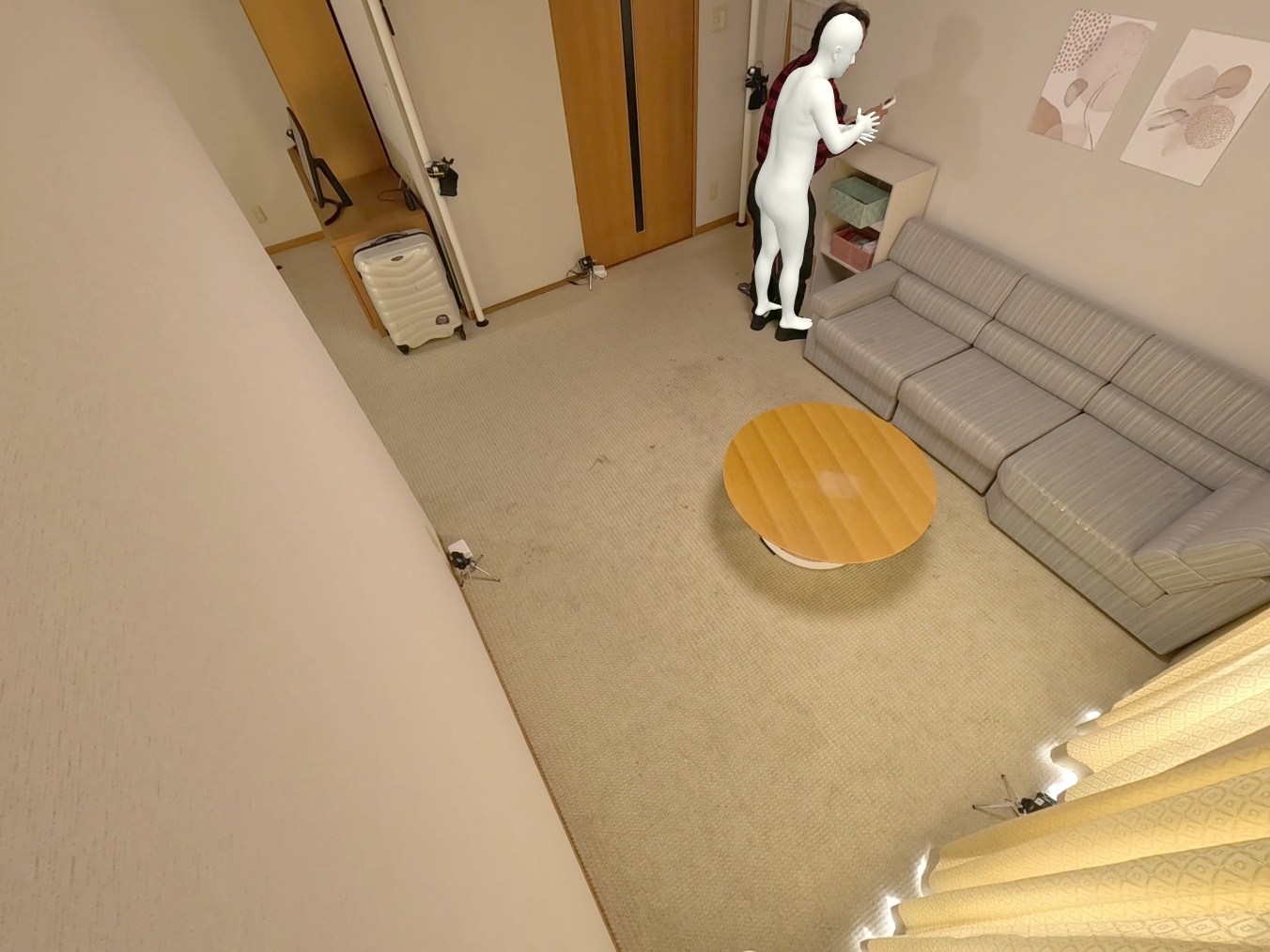}}; \advance \vx by \colwidth
    \node[anchor=north] at (\vx, \vy) {\Pg{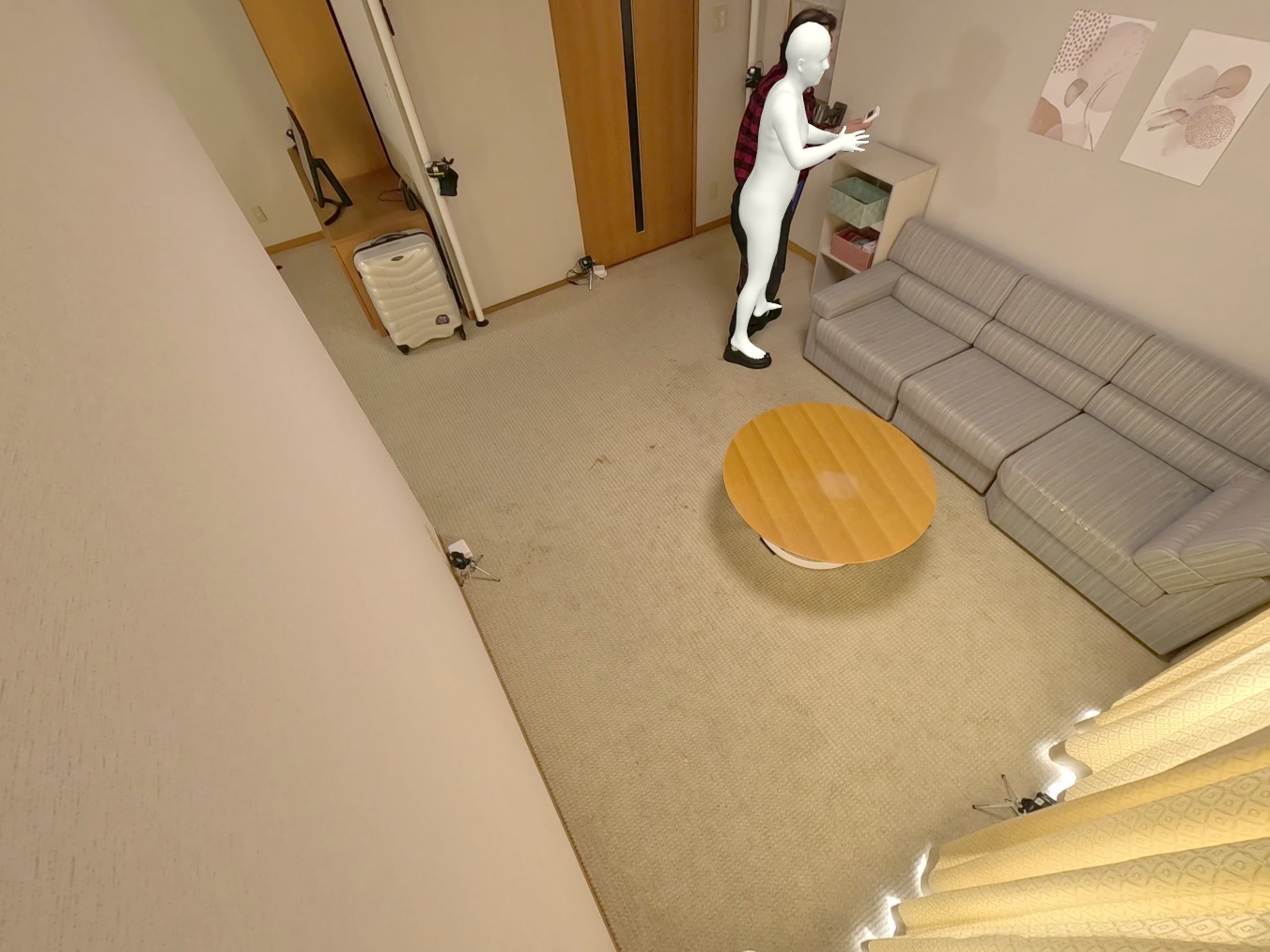}}; \advance \vx by \colwidth
    \node[anchor=north] at (\vx, \vy) {\Pg{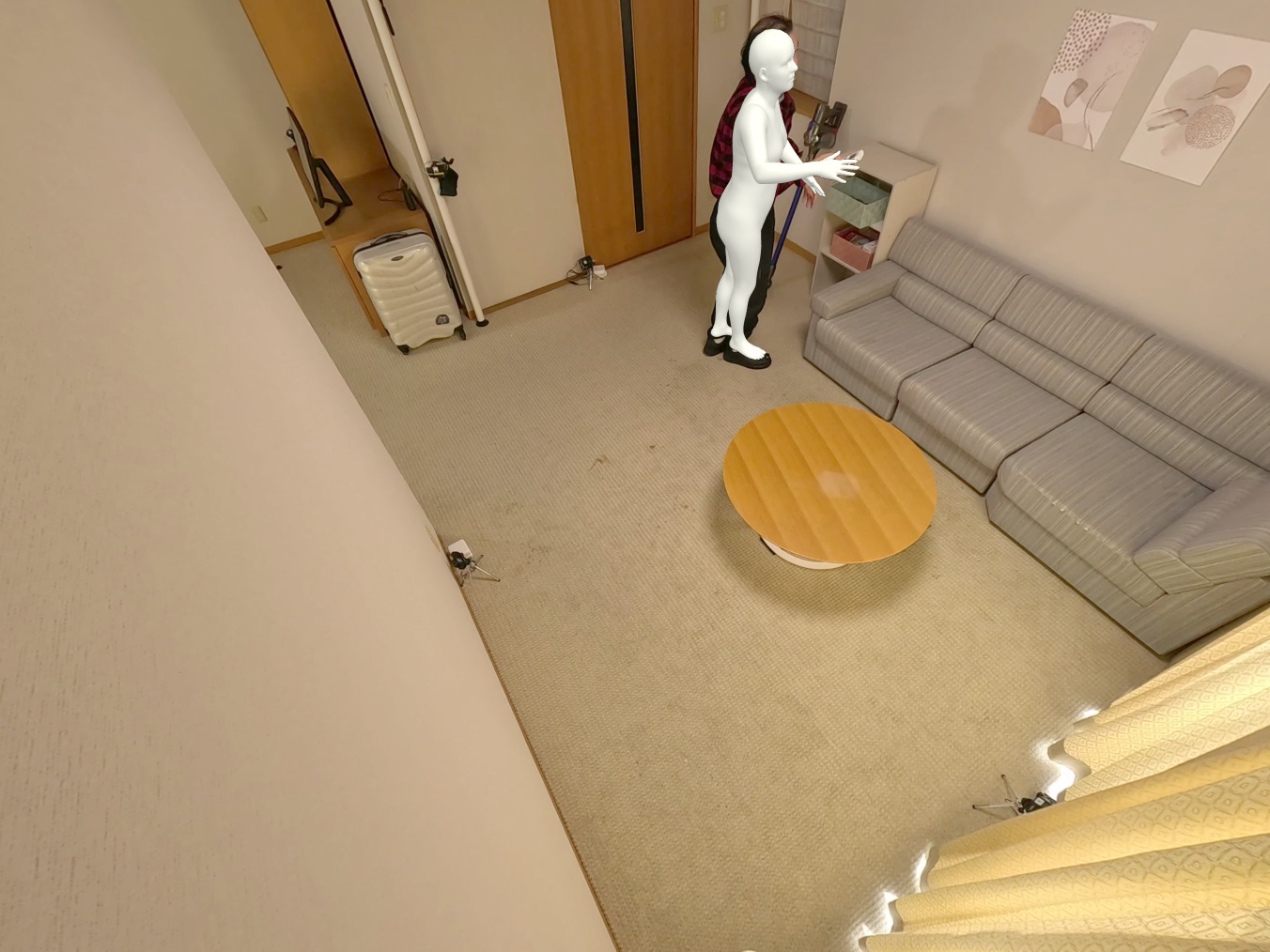}}; \advance \vx by \colwidth
    \node[anchor=north] at (\vx, \vy) {\Pg{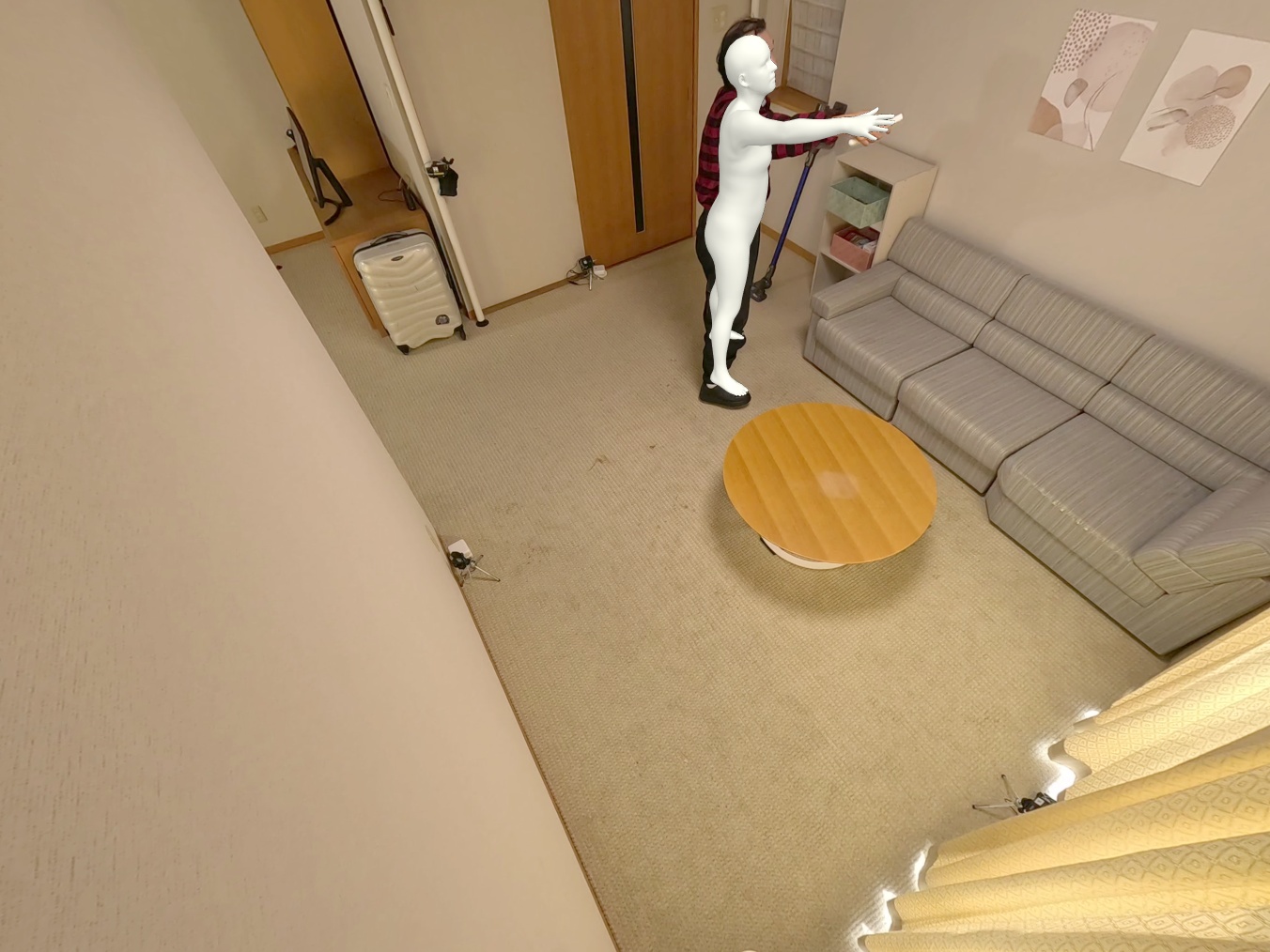}}; \advance \vx by \colwidth
    \node[anchor=north] at (\vx, \vy) {\Pg{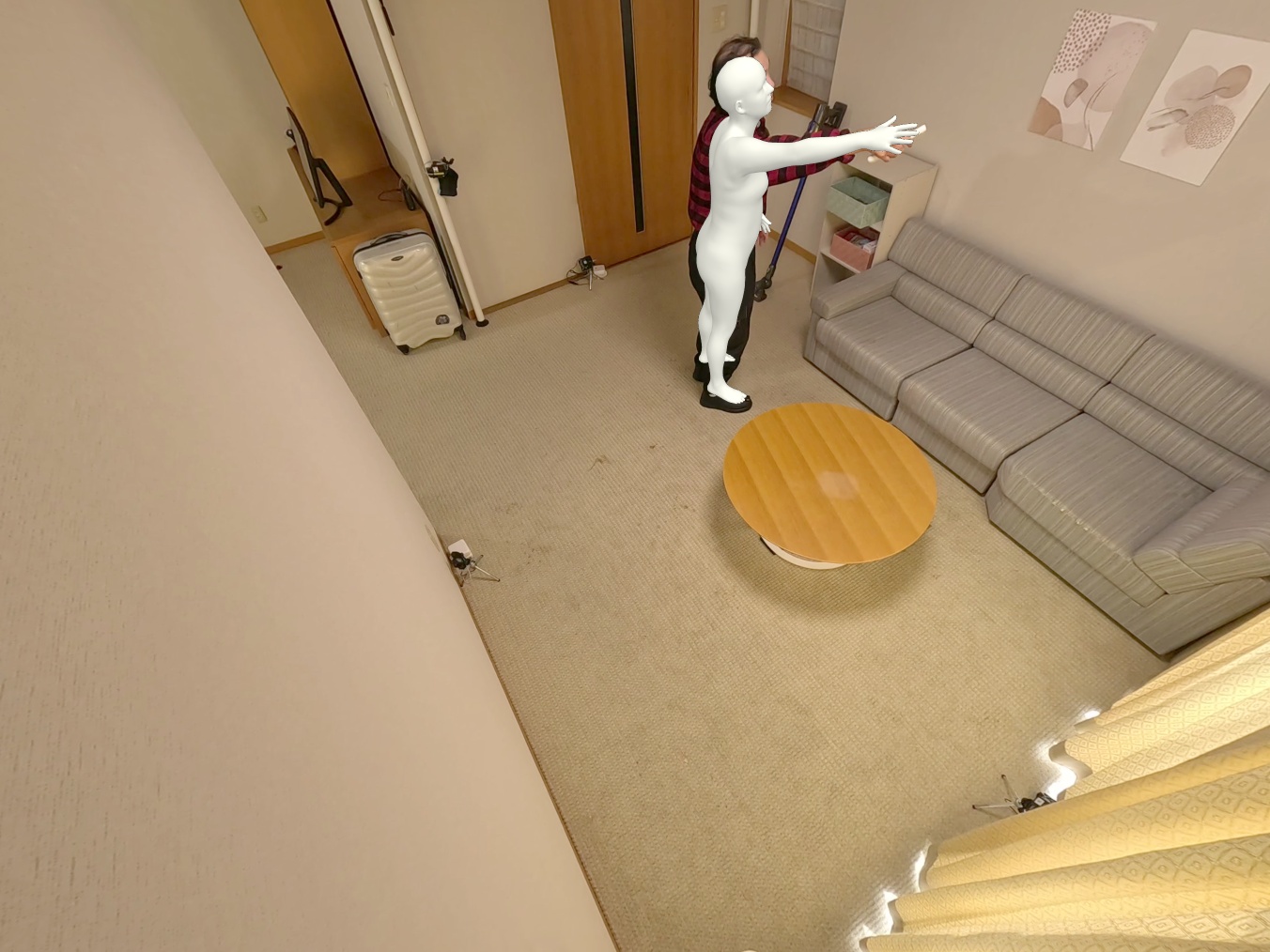}}; \advance \vx by \colwidth

    \vx = 0
    \advance \vy by \rowheight
    \node[anchor=north] at (\vx, \vy) {\Ph{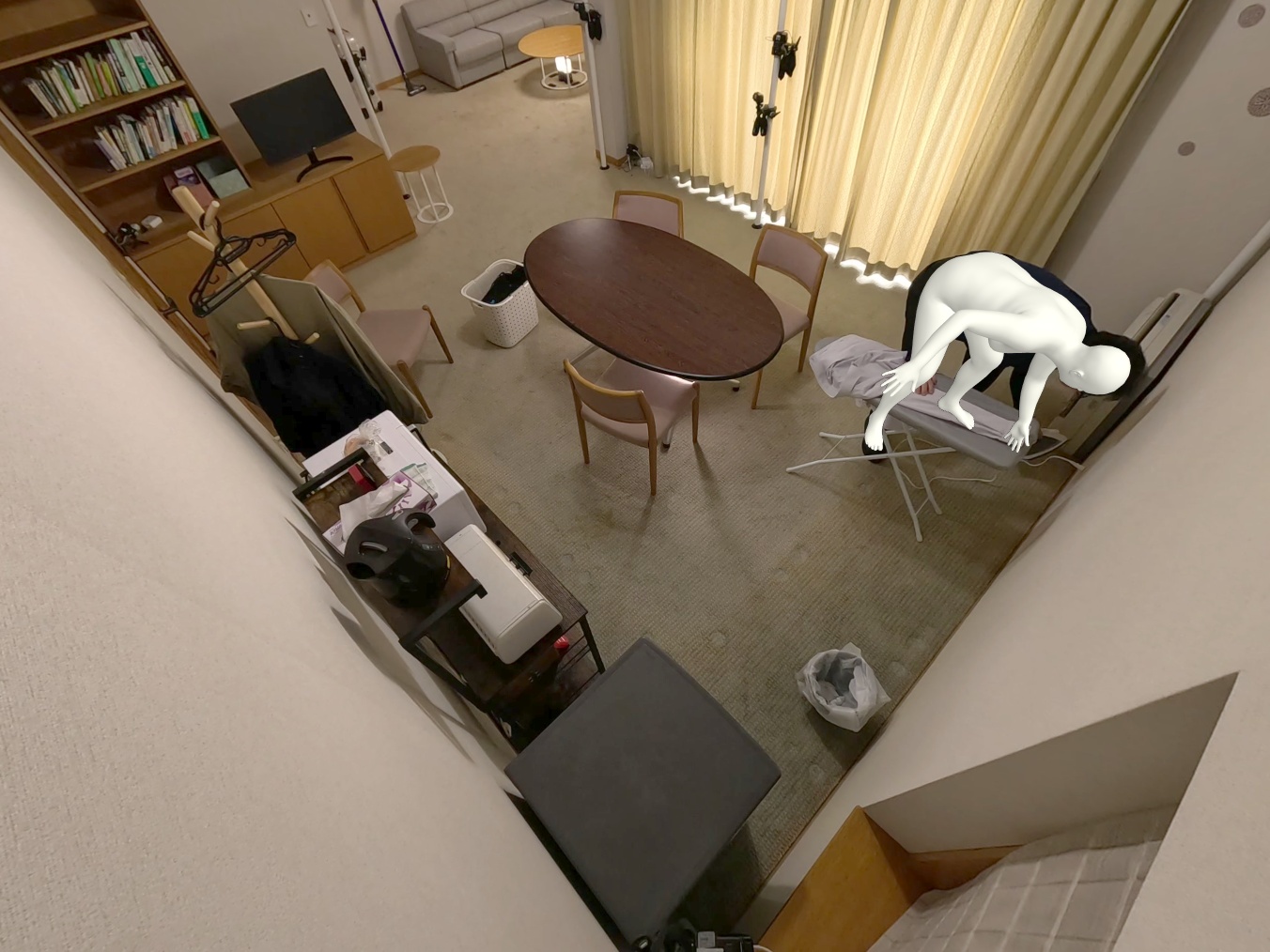}}; \advance \vx by \colwidth
    \node[anchor=north] at (\vx, \vy) {\Ph{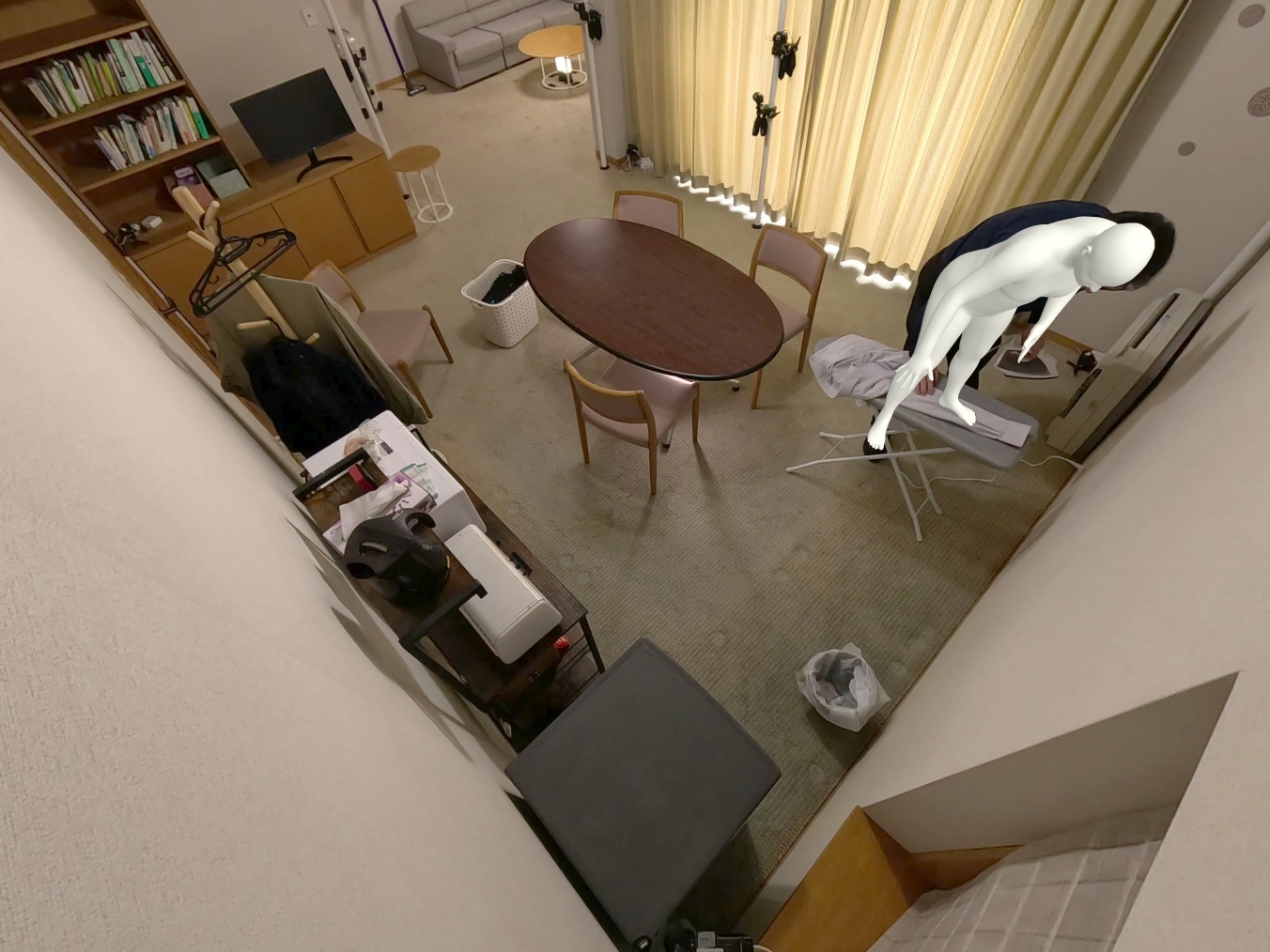}}; \advance \vx by \colwidth
    \node[anchor=north] at (\vx, \vy) {\Ph{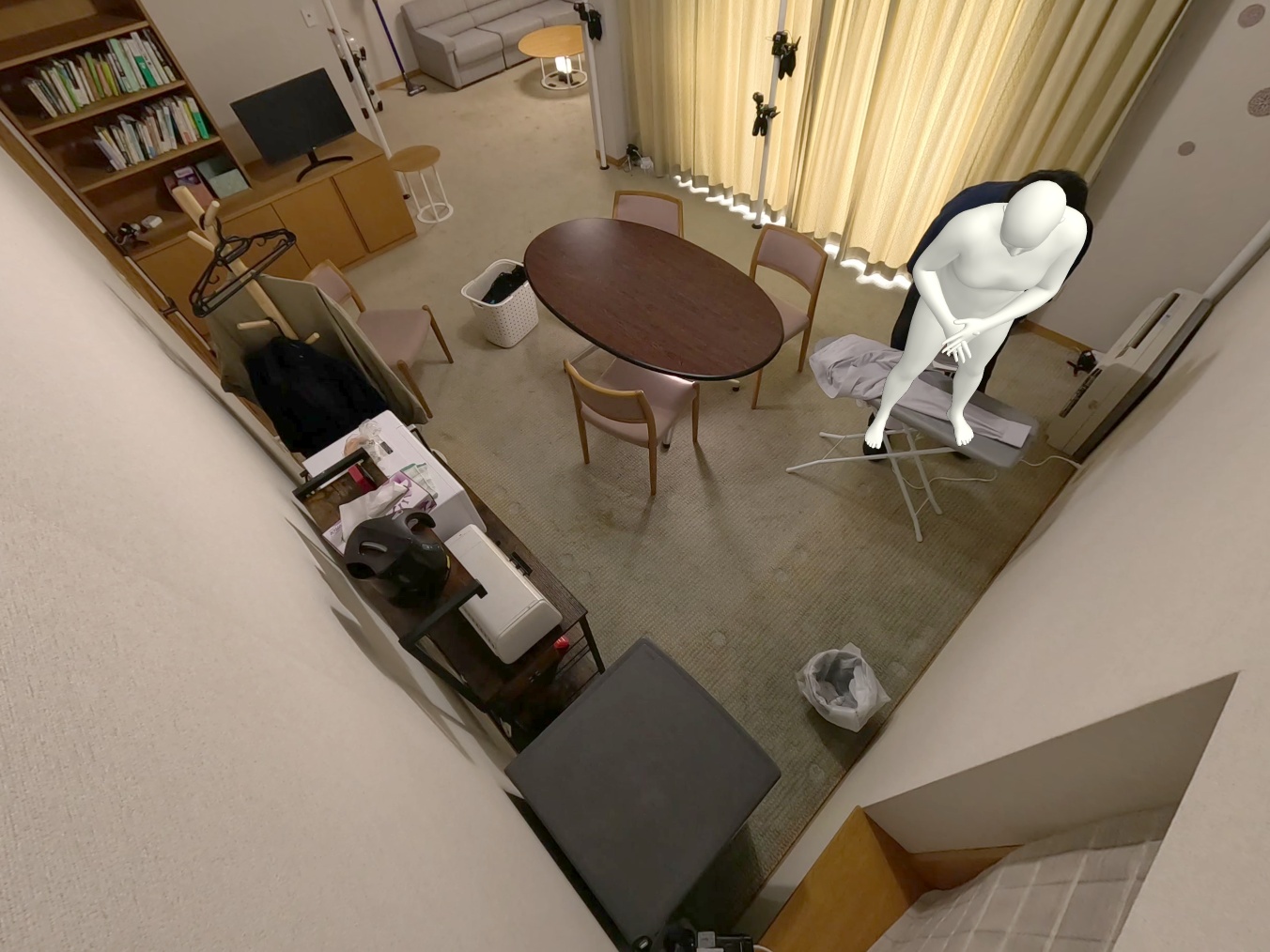}}; \advance \vx by \colwidth
    \node[anchor=north] at (\vx, \vy) {\Ph{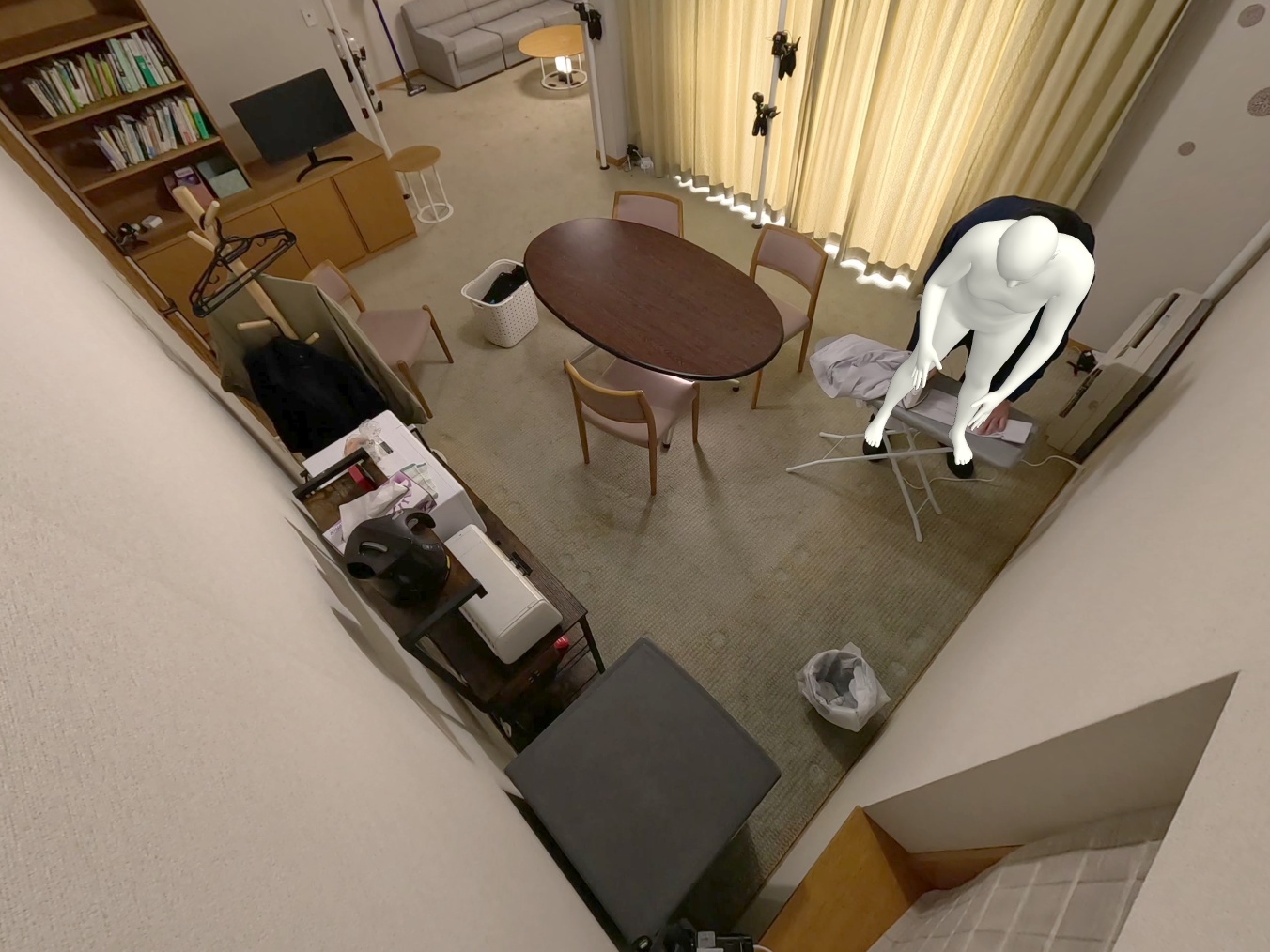}}; \advance \vx by \colwidth
    \node[anchor=north] at (\vx, \vy) {\Ph{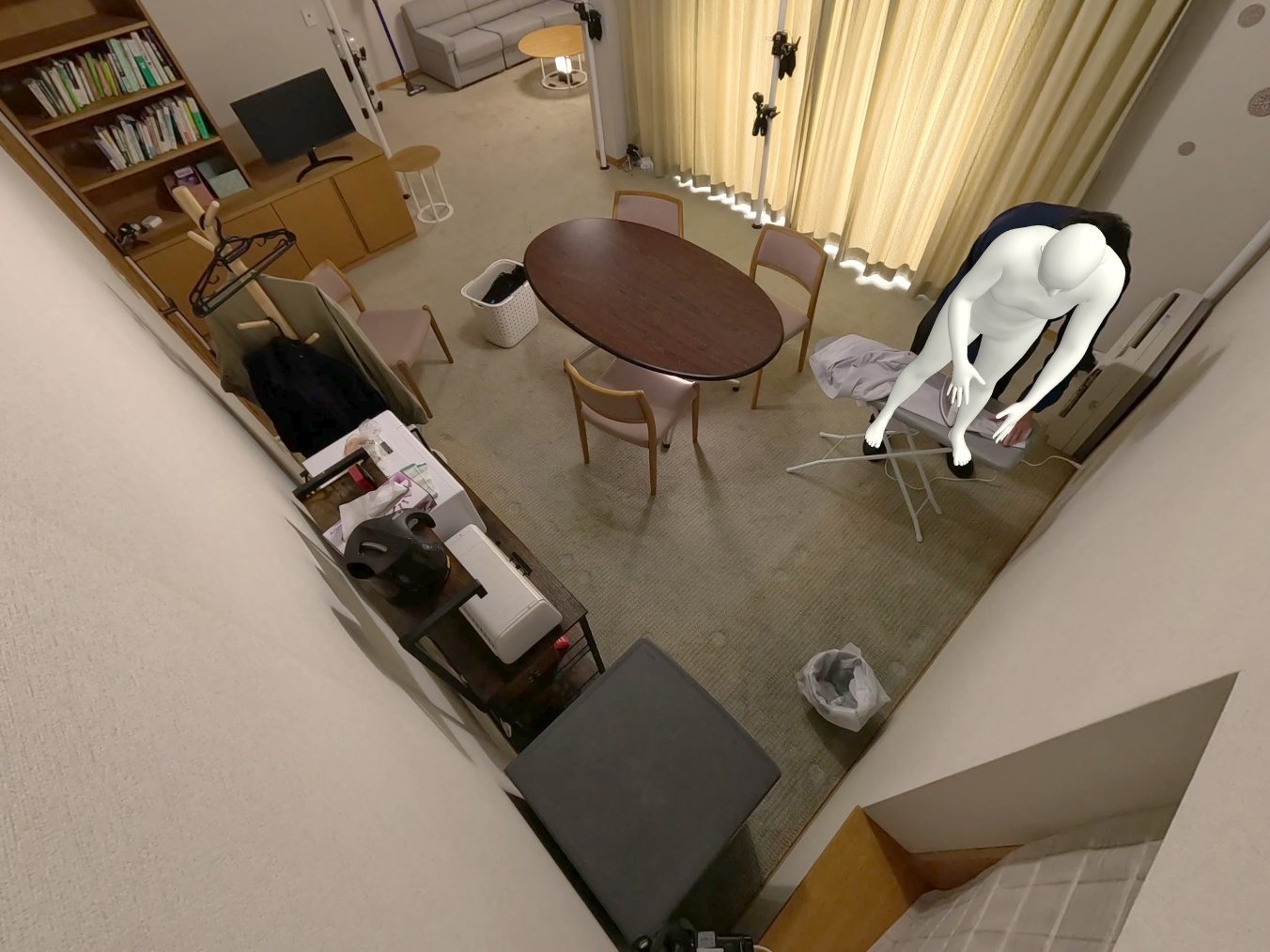}}; \advance \vx by \colwidth

\end{tikzpicture}

%% file: action_tables.tex
\newlength{\strwidth}

\begin{table*}[t!]
    \caption{AMD includes 129 daily actions. Actions performed in nearly identical poses are grouped under labels such as "Do sth." to streamline similar movements.}
    \label{tab:AMDActions}
    \centering
    \begin{tabular}{|c|c|c|}
        \hline
        \begin{minipage}[t]{0.3\linewidth}
            \centering
            \settowidth{\strwidth}{Listen to music with headphones}
            \begin{minipage}[l]{\strwidth}
                \vspace{0.5em} 
                Adjust electric fan direction \\
                Answer the call \\
                Assemble shelf \\
                Blow nose \\
                Carry sth to swh \\
                Change TV channel \\
                Check the time \\
                Chill sth in fridge \\
                Choose book \\
                Choose suitable utensil \\
                Clean \\
                Clip clothes with pegs \\
                Close book \\
                Close door of sth \\
                Close drawer \\
                Close laptop \\
                Close up suitcase \\
                Collect trash \\
                Dispose of trash \\
                Draw line with ruler \\
                Drink sth \\
                Drop sth \\
                Eat food \\
                Eat soup \\
                Empty shelf \\
                End call \\
                Erase with eraser \\
                Fold clothes \\
                Fold drying rack \\
                Fold iron board \\
                Fold newspaper \\
                Hang clothes on hanger \\
                Hang laundry \\
                Heat food with microwave \\
                Insert battery \\
                Insert bookmark \\
                Iron clothes \\
                Lean sth against wall \\
                Lie down \\
                Lie on sofa \\
                Listen to music with headphones \\
                Look for sth \\
                Make coffee
                \vspace{0.5em} 
            \end{minipage}

        \end{minipage}
        &
        \begin{minipage}[t]{0.3\textwidth}
            \centering
            \settowidth{\strwidth}{Remove hanger from clothes}
            \begin{minipage}[l]{\strwidth}
                \vspace{0.5em} 
                Make instant soup \\
                Make tea \\
                Move sth \\
                Notice incorrect time \\
                Notice ringing phone \\
                Open door of sth \\
                Open up laptop \\
                Open up suitcase \\
                Pack suitcase \\
                Pause sth \\
                Pick up sth \\
                Place sth on swh \\
                Play game with smartphone \\
                Plug sth into outlet \\
                Point sth at sth \\
                Pour liquid into cup \\
                Pour powder into cup \\
                Press button \\
                Pull open sth \\
                Pull out drawer \\
                Put away sth \\
                Put lid on sth \\
                Put on bag \\
                Put on coat \\
                Put on glasses \\
                Reach for sth \\
                Read book \\
                Read newspaper \\
                Remove battery \\
                Remove hanger from clothes \\
                Set the time on clock \\
                Set timer \\
                Set up drying rack \\
                Set up iron board \\
                Sit down \\
                Sit on chair \\
                Sit on sofa \\
                Sort laundry \\
                Spill liquid \\
                Spread out newspaper \\
                Spread out washed towel \\
                Stand on chair \\
                Stand sth up
                \vspace{0.5em} 
            \end{minipage}
        \end{minipage}
        &
        \begin{minipage}[t]{0.3\textwidth}
            \centering
            \settowidth{\strwidth}{Take out required dose of medicine}
            \begin{minipage}[l]{\strwidth}
                \vspace{0.5em} 
                Stand up \\
                Stir sth \\
                Take down laundry \\
                Take down wall clock \\
                Take medicine \\
                Take off bag \\
                Take off clothes \\
                Take off glasses \\
                Take off lid \\
                Take out required dose of medicine \\
                Take sth out of swh \\
                Tear open packet \\
                Tidy up \\
                Tie plastic bag \\
                Toast bread \\
                Transfer trash to bag \\
                Turn on TV \\
                Turn on air conditioner \\
                Turn on ceiling light \\
                Turn on electric fan \\
                Turn on reading lamp \\
                Turn off TV \\
                Turn off air conditioner \\
                Turn off ceiling light \\
                Turn off electric fan \\
                Turn off reading lamp \\
                Unclip small clothes from pegs \\
                Unplug sth from outlet \\
                Use lint roller on floor \\
                Use calculator \\
                Use laptop \\
                Use mouse \\
                Use phone \\
                Use fork \\
                Use spoon \\
                Vacuum floor \\
                Wait for timer \\
                Walk \\
                Watch TV \\
                Wipe table \\
                Wipe up liquid \\
                Wrap dish \\
                Write sth
                \vspace{0.5em} 
            \end{minipage}
        \end{minipage} \\
        \hline
    \end{tabular}
\end{table*}